\documentclass[a4paper, UKenglish, cleveref, autoref, thm-restate]{lipics-v2021}
\usepackage{multicol,lipsum}
\usepackage{wrapfig}
\usepackage{footmisc}
\usepackage{algpseudocode}
\pdfoutput=1 
\hideLIPIcs  
\title{Topological Interpretations of GPT-3}
\author{Tianyi Sun and Bradley Nelson}{Department of Statistics, Committee on Computational and Applied Mathematics \\ The University of Chicago}{}{}{}

\authorrunning{T. Sun and B. Nelson} 

\Copyright{Tianyi Sun and Bradley Nelson}


\nolinenumbers 
\EventEditors{John Q. Open and Joan R. Access}
\EventNoEds{2}
\EventLongTitle{42nd Conference on Very Important Topics (CVIT 2016)}
\EventShortTitle{CVIT 2016}
\EventAcronym{CVIT}
\EventYear{2016}
\EventDate{December 24--27, 2016}
\EventLocation{Little Whinging, United Kingdom}
\EventLogo{}
\SeriesVolume{42}
\ArticleNo{23}

\begin{document}
\maketitle
\begin{abstract}
This is an experiential study of investigating a consistent method for deriving the correlation between sentence vector and semantic meaning of a sentence. We first used three state-of-the-art word/sentence embedding methods including GPT-$3$, Word2Vec, and Sentence-BERT, to embed plain text sentence strings into high dimensional spaces. Then we compute the pairwise distance between any possible combination of two sentence vectors in an embedding space and map them into a matrix. Based on each distance matrix, we compute the correlation of distances of a sentence vector with respect to the other sentence vectors in an embedding space. Then we compute the correlation of each pair of the distance matrices. 
We observed correlations of the same sentence in different embedding spaces and correlations of different sentences in the same embedding space. These observations are consistent with our hypothesis and take us to the next stage. 
\end{abstract}
\section{Introduction}
GPT-$3$\cite{brown2020language} is a robust auto-regressive language model developed by OpenAI. It trained with $175$ billion parameters, has $96$ layers and trained on large datasets, including Common Crawl, open clone of OpenAI's unreleased WebText, two internet-based books corpora, and English-language Wikipedia. It achieves the state-of-the-art or human-level results in most of the Natural Language Processing (NLP) tasks, especially in generation task. Because of the robustness of the model, it is not publicly available for the audience to avoid misuse. The goal of this paper is to develop a way of interpret the model in the topological perspective. We start from the output of the model and compare it with the output of models that we have already known/understood.

In GPT-$3$ and Word2Vec embedding space, two word vectors that are close to each other have similar meanings. This inspired us to question about sentence vectors in sentence embedding space: What does sentence vector represent? Does two sentence vectors that are adjacent to each other have similar meaning? If so, are two adjacent sentence vectors semantically similar or emotionally similar?

To tackle this problem, we first embed sentences into a high dimensional space. For each sentence in the sentence scope, we get a numerical sentence vector. Each sentence vector could be either a high dimensional $m\times n$ matrix or a high dimensional $1\times n$ vector depending on the embedding method of our choice. In section \ref{ES}, we are using $100$ sentences for practice. The entire Word2Vec embedding sentence vectors scope has $100$ $243 \times 128$ sentence matrices, where $128$ is the dimension of sentence embedding which is determined by the dimension of embedded word vectors, $243$ is the maximum number of word vectors of a sentence vector in the scope. So each sentence vector is a $243 \times 128$ matrix. Similarly, the entire GPT-$3$ embedding sentence vectors scope has $100$ $\operatorname{len}(sample) \times 768$ sentence vectors. Each sentence vector has a dimension of $768$, which is determined by the initialized model parameter of GPT-$3$ word embedding. $\operatorname{len}(sample)$ represents the number of word vectors in each sentence vector( the length of the sentence). So each sentence vector is a high dimensional matrix in its embedding space. The entire Sentence-BERT embedding vectors scope is a bit different, it has $100$ $1 \times 384$ sentence vectors. Since Sentence-BERT is a direct sentence embedding transformer, each embedded sentence vector is a $1 \times 384$ matrix. 

Second, we use different methods to compute the pairwise distance of any combination of two high dimensional sentence matrices. The methods of our choice for computing the distances of high dimensional sentence matrices are bottleneck distance and cosine distance. Cosine distance is used for computing semantic similarity between sentence vectors in the original paper of Sentence-BERT\cite{DBLP:journals/corr/abs-1908-10084}. The method we used for computing the distance of plain text sentence strings is Levenshtein distance.  

Next we compute the pairwise distances of sentence vectors within a single sentence embedding cloud using multi-dimensional scaling (MDS). MDS is used to visualize the similarity of high-dimensional individual cases of a set in an abstract two-dimensional Cartesian space. We also compute the pairwise distance of sentence embedding clouds using Canonical Correlation Analysis (CCA) and the scaled Hausdorff distance (SHD).

\section{Background}
General ways to extract sentence meanings and look into their similarities are sentiment analysis and topic extraction. 

Sentiment analysis is a supervised learning task. We need to pre-train a classifier using a labeled dataset. Classical labeled dataset for sentiment analysis contains two values \cite{socher-etal-2013-recursive} or three values, those are positive, negative, and neutral. Some recent labeled datasets contain more labels \cite{kleinberg-etal-2020-measuring}. Some datasets with more sentiment labels are task specific. It does not make sense to use task specific dataset to pre-train a classifier and then use the classifier to predict on another domain of task. So finding applicable/appropriate datasets sometimes are a big problem. For sentiment analysis task, some recent state-of-the-art models are BERT\cite{devlin2018bert} and RoBERTa\cite{liu2019roberta}. 

Topic extraction is an unsupervised learning task. LDA is a model for topic extraction task. The result of topic extraction is some clustered groups of sentences with some extracted words ordered by decreasing weights. The weights could be determined by TF-IDF. In order to avoid extracting meaningless words and terms, such as ``where'' and ``which'', we should pre-process those sentences by removing stop words. 

However, those high level summaries of sentences are too general to capture the internal subtle differences between sentences. So in this and in the work that follows, we are going to study/investigate methods and general pipelines to capture subtle differences of sentences structures. 

\section{Data}

The dataset small\_117M\_k40\_test\footnote{\url{https://github.com/openai/gpt-2-output-dataset}} contains $5000$ examples. Each example is either a sentence or a paragraph. For each example, there are three descriptions. Those are the length of the sentence(s), a boolean value of whether the sentence(s) is/are ended or not, and a text string of the example. In consideration of the time complexity, in this work, we take the first $100$ samples as an experimental study. 

\section{Methods}

\subsection{Embedding Methods}
\begin{wrapfigure}{r}{0.5\linewidth}
    \centering
    \includegraphics[width=\linewidth]{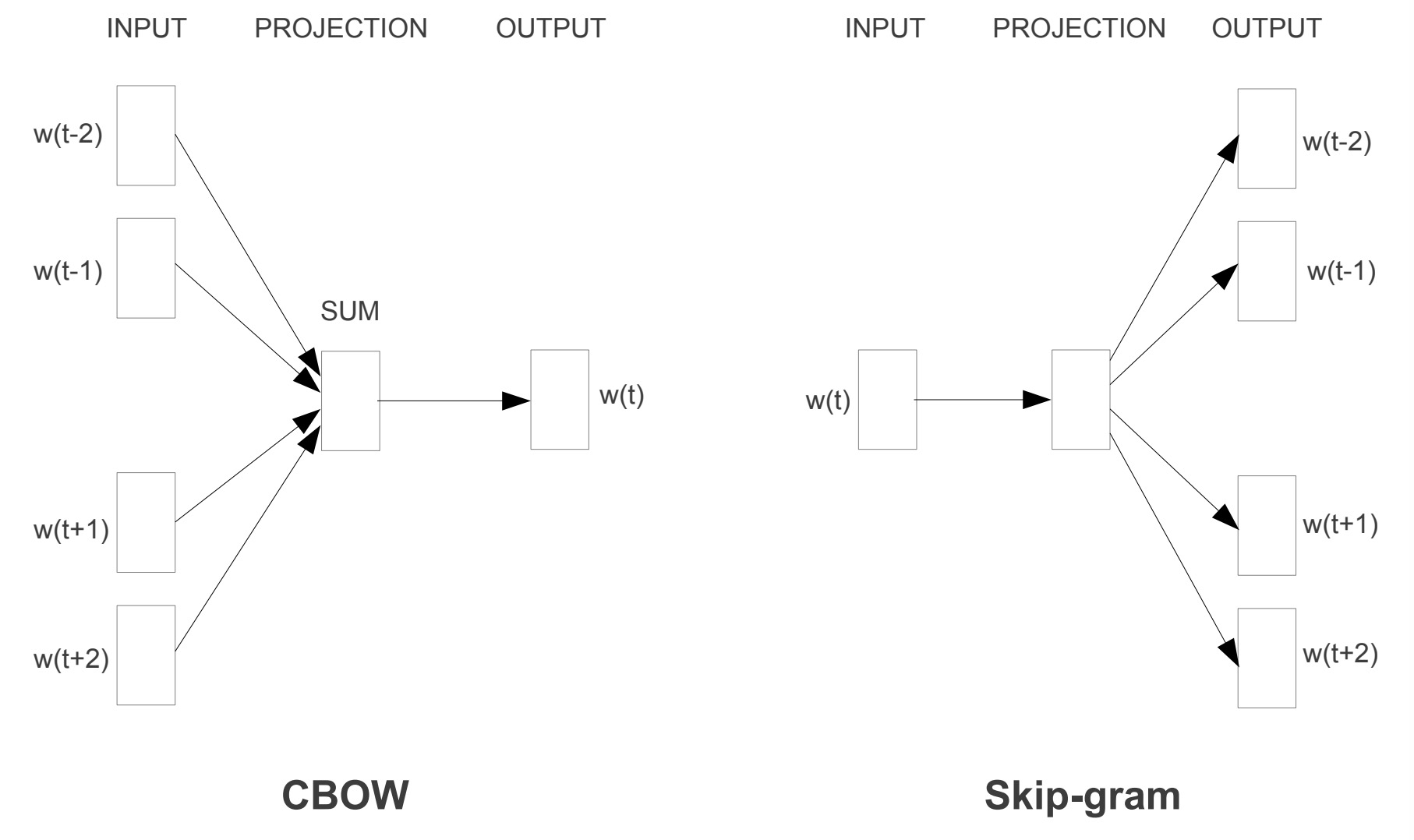}
    \caption{The CBOW architecture on the left predicts the current word based on the random order contextual words. The Skip-gram predicts contextual words given a middle word.}
    \label{fig:word2vec}
\end{wrapfigure}
In order to embed those sentences into a numerical vector space, the embedding methods of our choice are GPT2Tokenizer, Word2Vec, and Sentence-BERT. 
\subsubsection{GPT-3}
GPT-$3$\cite{brown2020language} is trained on datasets including: 
\begin{itemize}
    \item Common Crawl corpus, which contains raw web page data, metadata extracts and text extracts of over $12$ years;
    \item OpenWebText, which contains blogs, papers, codes of over $23$ million URLs and over $10$ million HTML pages; 
    \item Two internet-based books corpora\cite{DBLP:journals/corr/abs-2001-08361}, and Wikipedia.
\end{itemize}
Because GPT-$3$ is trained from many large datasets, we use the pre-trained GPT-$3$ models, such as, GPT2Tokenizer and TFGPT2Model.

\subsubsection{Word2Vec}
Word2Vec\cite{mikolov2013efficient}, a family of model architectures and optimizations used to learn word embedding from large datasets, provided state-of-the-art performance on some datasets of measuring syntactic and semantic word similarities since 2013. The paper proposed two efficient model architectures, continuous Bag-of-Words (CBOW) and continuous Skip-gram (Skip-gram), for learning distributed representation of words. Figure \ref{fig:word2vec} represents the architectures of the two models. This two models save run-time complexity in the way that they avoid of using N-gram neural network language model.

CBOW uses continuous distributed representation of the context, the projection layer is shared for all words and the order of words in the history does not influence the projection. Skip-gram, on the other hand, input a word into a log-linear classifier with continuous projection layer, and predict words within a certain contextual range of that word. The larger the contextual range of input word is, the better the performance of word vectors is. So Word2Vec word embedding is contextually dependent on the dataset.

\subsubsection{Sentence-BERT}
The Sentence-BERT model \cite{DBLP:journals/corr/abs-1908-10084} is a sentence embedding method. It is a structure to derive semantically meaningful sentence embedding that can be compared using cosine-similarity.

\subsection{Distance Computation Methods}
The way of our choice to compute similarity between sentences is to compute the pairwise distance of any combination of two out of one hundred samples. The distance computation methods of our choice are Levenshtein distance, Bottleneck distance\cite{zomorodian2004computing}, and Cosine similarity. Since the embedding methods are different, the shapes of the same sentence in different embedding spaces are different. we use different methods to compute the distance of sentence vectors, according to the embedding methods of our choice. 

\subsubsection{Levenshtein Distance}
The method we used to compute the distances of plain text sentences is Levenshtein distance.
\begin{definition}[Levenshtein distance\cite{LD}]
The Levenshtein distance between two strings, $x$ and $y$, is denoted by $\operatorname{lev}(x, y)$, where $x[n]$ is the $n$th character(s) of the string $x$,
$$\operatorname{lev}(x, y) = \begin{cases}
  |x| & \text{ if } |y| = 0, \\
  |y| & \text{ if } |x| = 0, \\
  \operatorname{lev}\big(x[1::],y[1::] \big) & \text{ if } x[0] = y[0], \\
  1 + \min \begin{cases}
          \operatorname{lev}\big(x[1::], y\big) \\
          \operatorname{lev}\big(x, y[1::]\big) \\
          \operatorname{lev}\big(x[1::], y[1::] \big) \\
       \end{cases} & \text{ otherwise.}
\end{cases}
$$
\end{definition}

Roughly speaking, the Levenshtein distance between two sentences, $x$ and $y$, is to compute how many alphabets/characters needed to be rewritten from sentence $x$ to sentence $y$. In our case, we modify this method to compute how many words we need to rewrite from sentence $x$ to sentence $y$.

\subsubsection{Bottleneck Distance}
The method we used to compute the distances of sentence embedding in the Word2Vec embedding space and the distances of sentence embedding in the GPT-$3$ embedding space is the bottleneck distance. 

\begin{definition}[Partial Matching\cite{kerber2017geometry}]
Given two multi-sets\footnote{We treat undecorated persistence diagrams as plain multi-sets of points in the extended plane $\Bar{\mathbb{R}}^2$.} $P$ and $Q$. A {\bf partial matching} between $P$ and $Q$, denoted as {\bf $M: P\leftrightarrow Q$}, is understood as in graph theory, that is, it is a subset $M$ of $P \times Q$ that satisfies the following constraints: 
\begin{itemize}
    \item every point $p \in P$ is matched with at most one point of $Q$, i.e., there is at most one point $q \in Q$ such that $(p,q) \in M$.
    \item every point $q \in Q$ is matched with at most one point of $P$, i.e., there is at most one point $p \in P$ such that $(p,q) \in M$.
\end{itemize}
\end{definition}

\begin{definition}[Bottleneck cost\cite{892133}]
The chosen cost function for partial matchings $M: P\leftrightarrow Q$ is the {\bf bottleneck cost $c(M)$}: 
\[
c(M)= \max \Bigg\{ \sup_{(p,q)\in M} \|p-q\|_\infty, \quad \sup_{s \in P \sqcup Q\operatorname{unmatched}} \frac{|s_y-s_x|}{2} \Bigg\}.
\]
\end{definition}

\begin{definition}[Bottleneck distance\cite{892133}]
The {\bf bottleneck distance} between the two multi-sets $P, Q$ is the smallest possible bottleneck cost achieved by partial matchings between them: 
$$\operatorname{d_b}(P, Q) = \inf_{M:P\leftrightarrow Q} c(M). $$
\end{definition}

\begin{theorem}{\bf Bottleneck Stability Theorem for Persistence Diagrams\cite{892133}.}
Let $\mathbb{X}$ be a triangulable space with continuous tame functions $f,g: \mathbb{X} \rightarrow \mathbb{R}$. Then the bottleneck distance between the persistence diagrams of $f$ and $g$ in the extended plane $\Bar{\mathbb{R}}^2$ is at most $ \|f - g\|_\infty$.
\end{theorem}

\subsubsection{Cosine Distance}
The method we used to compute the distances of sentence vectors in the Sentence-BERT embedding space is the cosine distance. 
\begin{definition}[Cosine Similarity]
The {\bf cosine similarity}, $S_C (\mathbf{A},\mathbf{B})$, is defined as   
$$S_C (\mathbf{A},\mathbf{B}):= \cos(\theta) = \frac{\mathbf{A} \cdot \mathbf{B}}{\|\mathbf{A}\| \|\mathbf{B}\|} = \frac{\sum\limits_{i=1}^{n}{\mathbf{A}_i\mathbf{B}_i}}{\sqrt{\sum\limits_{i=1}^{n}{\mathbf{A}_i^2}}  \sqrt{\sum\limits_{i=1}^{n}{\mathbf{B}_i^2}}},
$$ where $\mathbf{A}$ and $\mathbf{B}$ represents two sentence vectors in the same embedding space in our case. 
\end{definition}
\begin{definition}[Cosine Distance\cite{cosine}]
The {\bf cosine distance}, $D_C (\mathbf{A},\mathbf{B})$, is used for the complement of cosine similarity in positive space, which is defined as 
$$ D_C (\mathbf{A},\mathbf{B}) = 1- S_C (\mathbf{A},\mathbf{B}).$$
\end{definition}
The cosine distance is not a proper distance metric, as it does not have the Cauchy–Schwarz inequality property, so that it violates the coincidence axiom. To repair the Cauchy–Schwarz inequality property while maintaining the same ordering, it is necessary to convert to angular distance or Euclidean distance. For angular distances, the Cauchy–Schwarz inequality can be expressed directly in terms of the cosines. 

The ordinary Cauchy–Schwarz inequality for angles (i.e., arc lengths on a unit hyper-sphere) gives us that $$|~\operatorname{arc}{AC} - \operatorname{arc}{CB}~| \le ~\operatorname{arc}{AB}~ \le ~\operatorname{arc}{AC}~ + ~\operatorname{arc}{CB}~.$$
Because the cosine function decreases as an angle in $[0,\pi]$ radius increases, these inequalities reversed when we take the cosine of each value$$\cos(\operatorname{arc}{AC} - \operatorname{arc}{CB}) \ge \cos(\operatorname{arc}{AB}) \ge \cos(\operatorname{arc}{AC} + \operatorname{arc}{CB}).$$
Using the cosine addition and subtraction formulas, these two inequalities can be written in terms of the original cosines:
\begin{align*}
\cos(A,C) \cdot \cos(C,B) + \sqrt{\left(1-\cos(A,C)^2\right)\cdot\left(1-\cos(C,B)^2\right)}   
& \geq \cos(A,B)         \\
 \geq \cos(A,C) \cdot \cos(C,B) - \sqrt{\left(1-\cos(A,C)^2\right)\cdot\left(1-\cos(C,B)^2\right)}
\end{align*}
This form of the Cauchy–Schwarz inequality can be used to bound the minimum and maximum similarity of two objects $A$ and $B$ if the similarities to a reference object $C$ are already known. 

\subsection{Correlation Analysis Methods}
The way of our choice to find the correlation of distances of a sentence vector with respect to the other sentence vectors in the same sentence embedding space is Multidimensional Scaling. To find the correlation of two clouds of sentence vectors embedded by different embedding methods, the ways of our choice are Canonical Correlation Analysis and Hausdorff Distance. 

\subsubsection{Multidimensional Scaling} Multidimensional Scaling\cite{mds}(MDS) is used to visualize the level of similarity of individual cases of a dataset. In our case, MDS is used to visualize similarity/information of pairwise distances among a set of $100$ embedded sentence vectors into a configuration of $100$ points mapped into an abstract Cartesian space. Technically, MDS refers to a set of related ordination techniques used in information visualization, in particular to display the information contained in a distance matrix. It is a form of non-linear dimensionality reduction. 

The MDS seeks to approximate the lower-dimensional representation by minimising a loss function $\text{Strain}$. In classical MDS, the $\text{Strain}$ is given by: 
$$ {\text{Strain}}_{D}(x_{1},x_{2},...,x_{n})=
\sqrt{{\frac {\sum _{i,j}{\bigl (}b_{ij}-x_{i}^{T}x_{j}{\bigr )}^{2}}{\sum _{i,j}b_{ij}^{2}}}},$$
where $x_{i}$ denotes vector in $n$-dimensional space, and $b_{ij} \in B$ defined on step 2 of the following classical MDS algorithm.

Classical MDS algorithm uses the fact that the coordinate matrix $X$ can be derived by eigenvalue decomposition from $B=XX'$. And the matrix $B$ can be computed from proximity matrix $D$ by using double centering.
\begin{enumerate}
    \item Set up the squared proximity matrix $D^{(2)}=[d_{ij}^{2}]$.
    \item Apply double centering $B=-{\frac {1}{2}}CD^{(2)}C$ using the centering matrix ${\textstyle C=I-{\frac {1}{n}}J_{n}}$, where $n$ is the number of objects, $I$ is the $n\times n$ identity matrix, and $J_{n}$ is an ${\textstyle n\times n}$ matrix of all ones.
    \item Determine the $m$ largest eigenvalues $\lambda_{1},\lambda_{2},\cdots,\lambda_{m}$ and corresponding eigenvectors $e_{1},e_{2},\cdots,e_{m}$ of $B$, where $m$ is the number of dimensions desired for the output.
    \item $X=E_{m}\Lambda_{m}^{1/2}$, where $E_{m}$ is the matrix of $m$ eigenvectors and $\Lambda_{m}$ is the diagonal matrix of $m$ eigenvalues of $B$.
\end{enumerate}
Classical MDS assumes Euclidean distances, so it is not applicable for direct dissimilarity ratings.

\subsubsection{Canonical Correlation Analysis}
Canonical correlation analysis\cite{cca}(CCA) is a way of inferring information from cross-covariance matrices. Given two vectors $X = (X_1, ..., X_n)$ and $Y= (Y_1, ..., Y_m)$ of random variables, and there are correlations among the variables. Canonical correlation analysis will find linear combinations of $X$ and $Y$ which have maximum correlation with each other.

Given two column vectors $X = (x_1, \cdots, x_n)'$ and $Y = (y_1, \cdots, y_m)'$ of random variables with finite second moments, one may define the cross-covariance $\Sigma_{XY} = cov(X, Y)$ to be the $n \times m$ matrix whose $(i,j)$ entry is the covariance $\operatorname {cov}(x_{i},y_{j})$. Canonical correlation analysis seeks vectors $a \in \mathrm{R}^n$ and $b\in \mathrm{R}^m$ such that the random variables $a^TX$ and $b^TY$ maximize the correlation $\rho = \operatorname{corr}(a^TX, b^TY).$ The random variables $U = a^TX $ and $V = b^TY$ are the first pair of canonical variables. Then one seeks vectors maximizing the same correlation subject to the constraint that they are to be uncorrelated with the first pair of canonical variables. This gives the second pair of canonical variables. This procedure may be continued up to the $\operatorname{min}\{ m,n\}$ times. $$(a', b') = \operatorname{argmax}_{a,b} \operatorname{corr}(a^T, b^TY).$$

\subsubsection{Hausdorff Distance} 
The Hausdorff distance, or Hausdorff metric, measures how far two sets are from each other. Two sets are close in the Hausdorff distance if every point of either set is close to some point of the other set. 
\begin{definition}[Hausdorff Distance\cite{hd}]
Let $X$ and $Y$ be two non-empty subsets of a metric space $(M, d)$. We define their Hausdorff distance $d_H(X,Y)$ by $$d_H(X,Y) = \operatorname{max}\{\operatorname{sup}_{x\in X} d(x,Y), \operatorname{sup}_{y\in Y} d(X, y)\},$$ where $d(a, B) = \operatorname{inf}_{b\in B} d(a, b) $ quantifies the distance from a point $a\in X $ to the subset $B \subseteq X$. 
\end{definition}

In general, $d_{\mathrm H}(X,Y)$ may be infinite. If both $X$ and $Y$ are bounded, then $d_{\mathrm H}(X,Y)$ is guaranteed to be finite.
$d_{\mathrm H}(X,Y)=0$ if and only if $X$ and $Y$ have the same closure.

We modify the above Hausdorff distance into a scaled Hausdorff distance to compute the minimum value of the Hausdorff distance with the corresponding scaled value $\alpha$:
$$ d_{SH}(X, Y) = \operatorname{min}_{\alpha>0} d_H(\alpha X, Y) =\operatorname{min}_{\alpha>0} \operatorname{sup}_{y\in Y} d(\alpha \*X, y).$$

\section{Experimental Study}\label{ES}
\subsection{Distance Matrix Computation}
We first compute six distance matrices based on the plain text sentence strings and sentence vectors in each of the three embedding spaces. The description of how we compute this six distance matrices are in Sections \ref{pt+ld}, \ref{gpt2+ld}, \ref{word2vec+ld}, and \ref{sbert+cd}. The results are in Figure \ref{fig:heat}. 
Then for each of the distance matrix, we compute MDS to visualize the similarity of pairwise distances in each embedding space. The results are in Figure \ref{fig:mds}.
Next we compute the canonical correlation for any possible combinations of pairwise embedding spaces. Based on the results of distance matrices, we are not including the $H_0$ values of pairwise bottleneck distance matrices of sentence vectors embedded by GPT-$3$ and Word2Vec for computing the canonical correlations. The CCA results of all the rest possible combinations of matrices are in Figure \ref{fig:cca}. 
Lastly we compute the scaled Hausdorff Distances for any possible combinations of pairwise embedding spaces excluding the $H_0$ values of pairwise bottleneck distance matrices of sentence vectors embedded by GPT-$3$ and Word2Vec. The optimal Scaled Hausdorff distance results are in Table \ref{tab:hausdorff}. The approximations to the optimal results are visualized in Figure \ref{fig:hausdorff}. 

\subsubsection{Plain text \& Levenshtein distance} \label{pt+ld}
The plain text contains one hundred text string samples. We first split each text string sample into a list of words. Second, we compute the Levenshtein distance between any combination of two lists of words. Third, we map the distance values into a $100 \times 100$ matrix. Each component of $(i, j)$ in the matrix indicates a distance value between sentence $i$ and sentence $j$. The Levenshtein distance matrix is in Figure \ref{fig:heat_levedis_word_100}.

\subsubsection{GPT-3 \& Bottleneck distance} \label{gpt2+ld}
To begin with, we constructed a word cloud based on the one hundred samples using GPT2Tokenizer. The vocabulary size is $50,257$, which is initialized in the model. For each sentence, we label words in that sentence as the index of the sentence it contained and gather those word vectors following the order of the word in the sentence to construct a sentence matrix cloud with one hundred sentence matrices in it. Each sentence is represented as a $\operatorname{len}(sample) \times 768$ matrix, where the length of each sentence matrix, $\operatorname{len}(sample)$, is determined by the number of words in that sentence, and $768$ is the dimension of each word vector. 

\begin{figure}
\begin{subfigure}[b]{0.48\textwidth}
    \includegraphics[width=\textwidth]{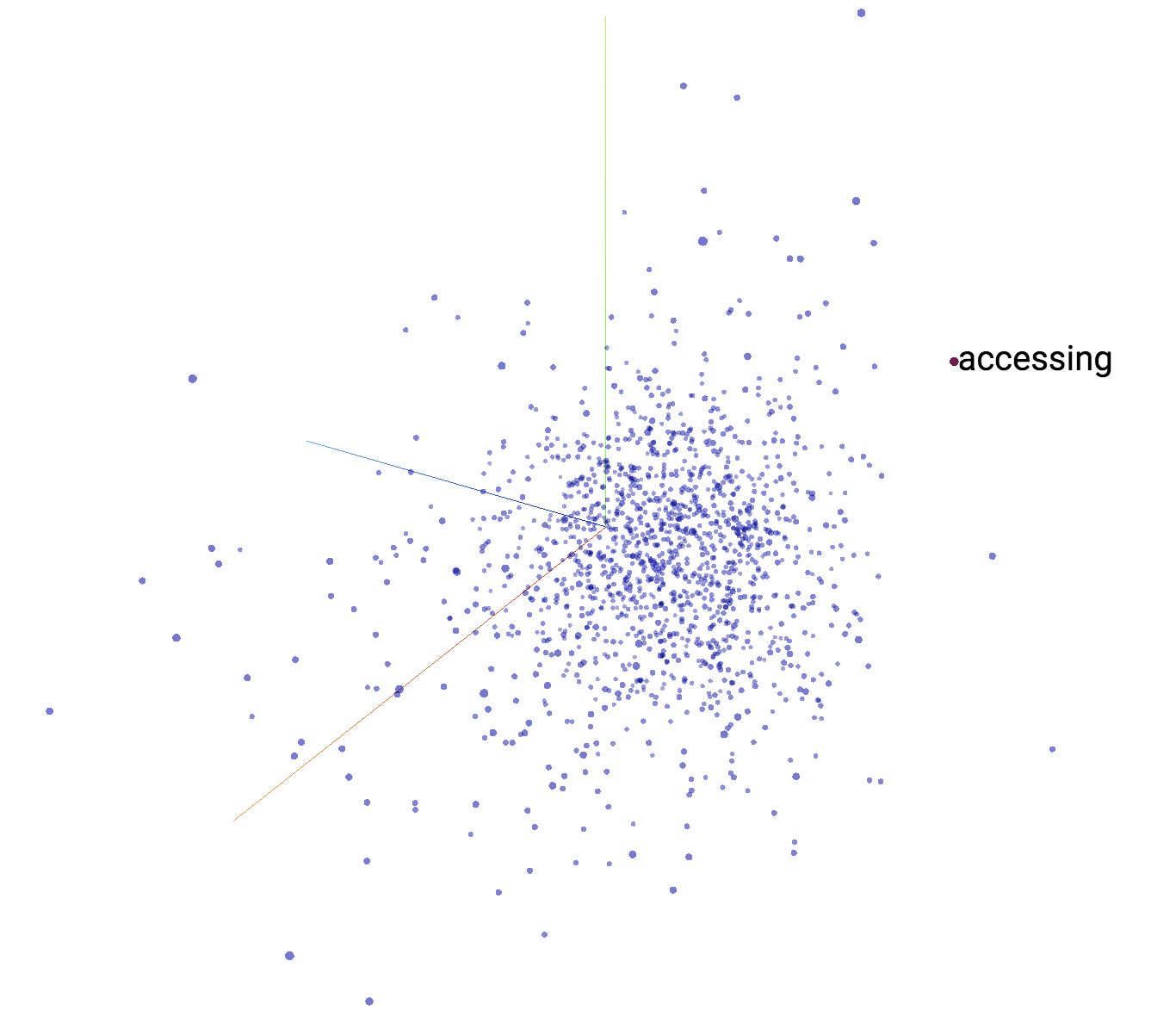}
    \caption{A visualization of Word2Vec word embedding learned from the first 100 text samples of small\_117M\_k40\_test dataset.}
    \label{fig:wordcloud}
\end{subfigure}
\begin{subfigure}[b]{0.48\textwidth}
    \includegraphics[width=\textwidth]{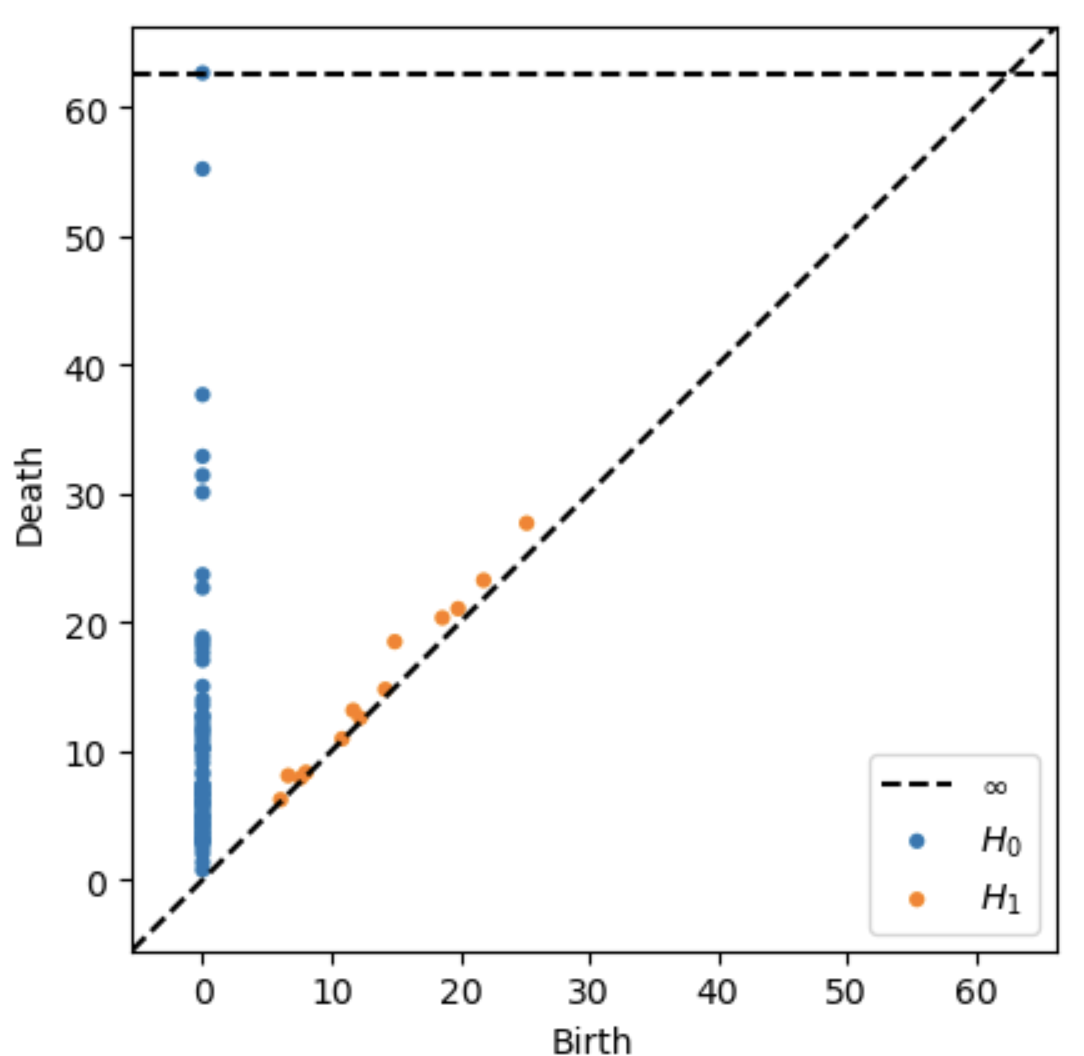}
    \caption{Vietoris-Rips persistence diagram of the third sentence embedding.}    
    \label{fig:rips_sent3}
\end{subfigure}
    \caption{An example of visualization of general Rips persistence diagram for individual sentence matrix sample. The figure is the third sample's GPT-$3$ embedding matrix Rips persistence diagram. The input is a numerical vector with seventy seven $768$-dimensional word vectors following the order of the corresponding words in that sample. The $Birth$ indicates the time that features appeared in the word cloud as adding more points for computation. The $Death$ indicates the time that features disappeared. The $H_1$ points lay on the diagonal are features appeared and disappeared immediately. The $H_0$ points lay at $Birth$ index $0$ are features with a relatively long duration. The $H_0$ point at the top indicates that the word cloud from the third sample has one connected component.}
\end{figure}

Then we compute the bottleneck distance of two sentence matrices. In the first round, shown in Figure \ref{fig:rips_sent3}, we do Principal Component Analysis (PCA) with $2$ components to reduce the dimension of each sentence matrix from $768$ to $2$. This will not lose too much information, since the total variance that PCA with $2$ components can capture is above ninety percent. Then we compute a Rips persistence diagram for each reduced sentence matrix. We take the third sample as an example, the matrix of the third sample is reduced from $77 \times 768$ to $77 \times 2$. In the second round, we directly compute the Rips persistence diagram for each original high-dimensional sentence matrix without doing PCA. The Rips persistence diagram for the third sample is in Figure \ref{fig:rips_sent3}. Lastly, we compute pairwise bottleneck distance for any possible combination of two Rips persistence diagrams computed in the second round. We then map them to a $100 \times 100$ matrix. The distance matrices for GPT-$3$ embedding are in Figure \ref{fig:heat_gpt2_rips_bd_H0_100} and \ref{fig:heat_gpt2_rips_bd_H1_100}, for $H_0$ and $H_1$ values of bottleneck distance respectively.

\subsubsection{Word2Vec \& Bottleneck distance} \label{word2vec+ld}
Similar as what we have done for GPT-$3$ embedding in Section \ref{gpt2+ld}, we first constructed a word cloud for Word2Vec embedded word vectors from the one hundred samples. The vocabulary size, which is also the number of word vectors, is $1192$. Then we construct a dictionary for the word cloud, where the keys are words, and the values are corresponding word vectors. Then we map those word vectors to the list of words of each sample in order, so that we construct a $243 \times 128$ matrix for each sentence. The $243$ is the maximum length among those one hundred samples. Which means that regardless of the actual length, each sentence matrix has the same length. The $128$ is the dimension of word vectors. So we get a sentence cloud with one hundred sentence matrices in it. Each sentence matrix is $243 \times 128$. 

Then we compute the bottleneck distance for each pair of sentence matrices as we have done in Section \ref{gpt2+ld}. The distance matrices for Word2Vec embedding are in Figure \ref{fig:heat_word2vec_rips_bd_H0_100} and \ref{fig:heat_word2vec_rips_bd_H1_100}, for $H_0$ and $H_1$ values of bottleneck distance respectively.

\subsubsection{Sentence-BERT \& Cosine distance} \label{sbert+cd}
\begin{figure}
     \centering
     \begin{subfigure}[b]{0.32\textwidth}
         \centering
         \includegraphics[width=\textwidth]{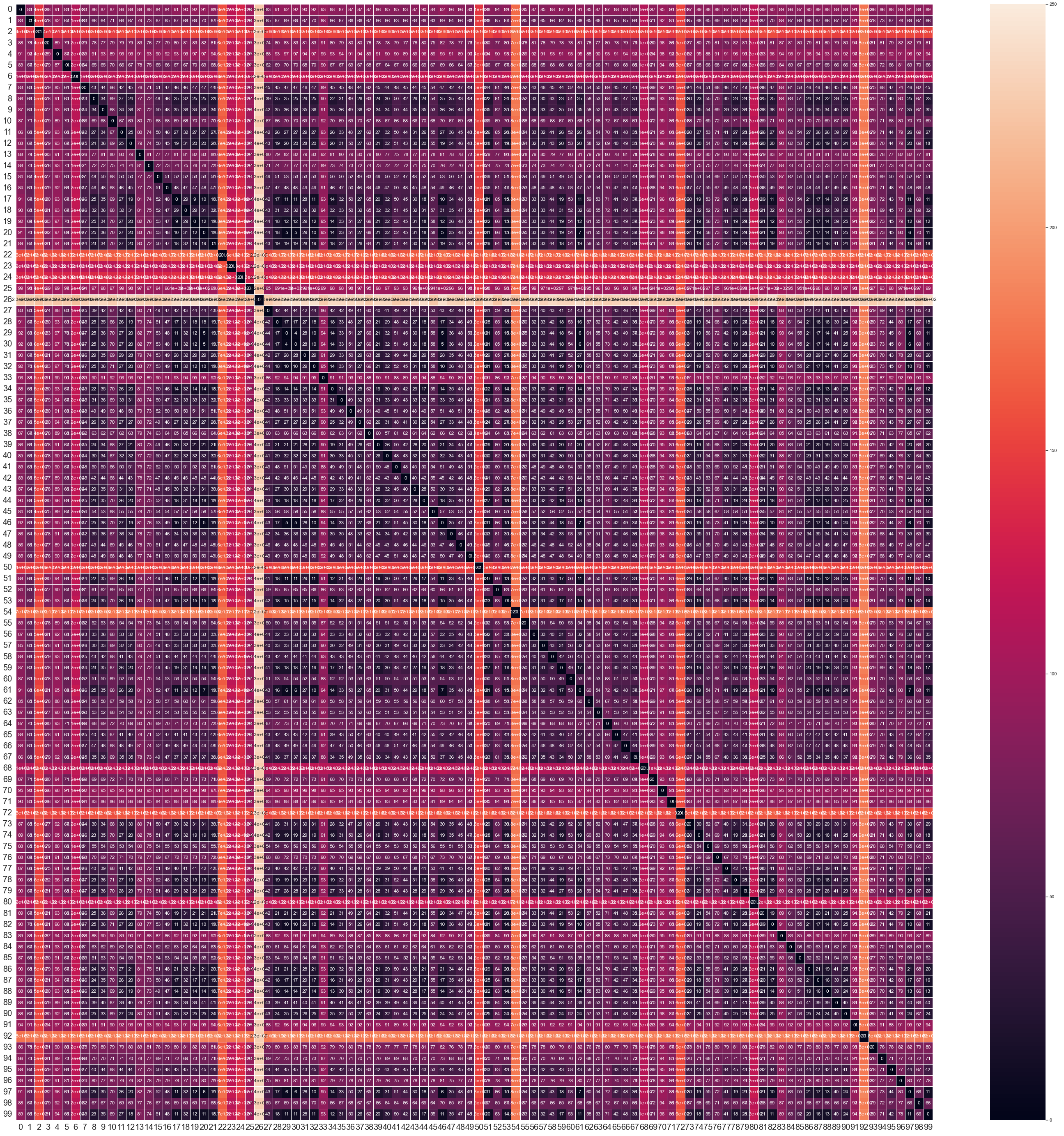}
         \caption{This figure shows the pairwise Levenshtein distance matrix of plain text sentence strings. Since the maximum distance is $243$, the range of the distance is from $0$ to $243$.}
         \label{fig:heat_levedis_word_100}
     \end{subfigure}
     \hfill
     \begin{subfigure}[b]{0.32\textwidth}
         \centering
         \includegraphics[width=\textwidth]{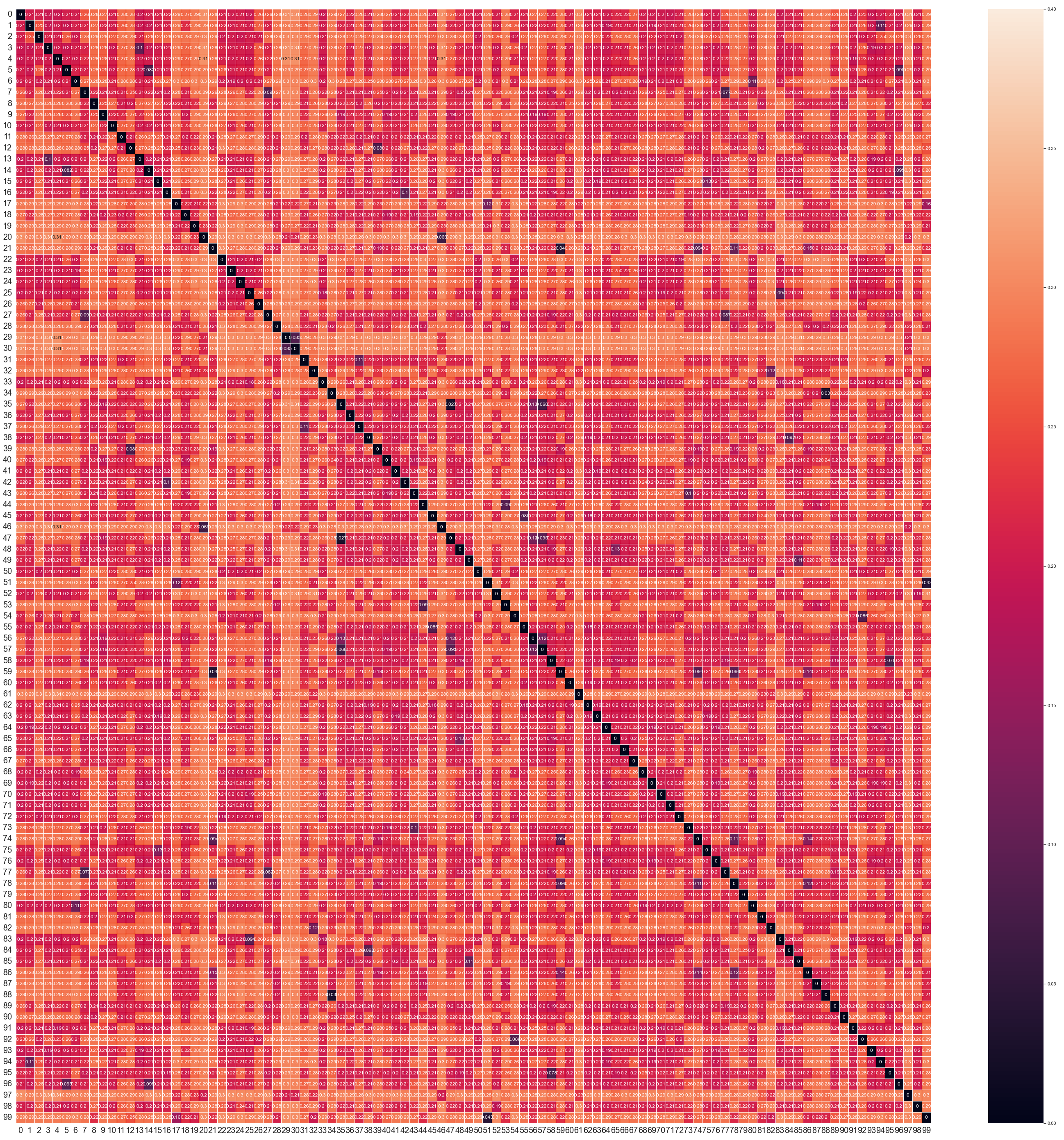}
         \caption{This figure shows the $H_0$ value of the pairwise bottleneck distances matrix of sentences embedded by Word2Vec. Since the maximum distance is about $173.23$, the range of the distance is from $0$ to $180$.}
         \label{fig:heat_gpt2_rips_bd_H0_100}
     \end{subfigure}
     \hfill
     \begin{subfigure}[b]{0.32\textwidth}
         \centering
         \includegraphics[width=\textwidth]{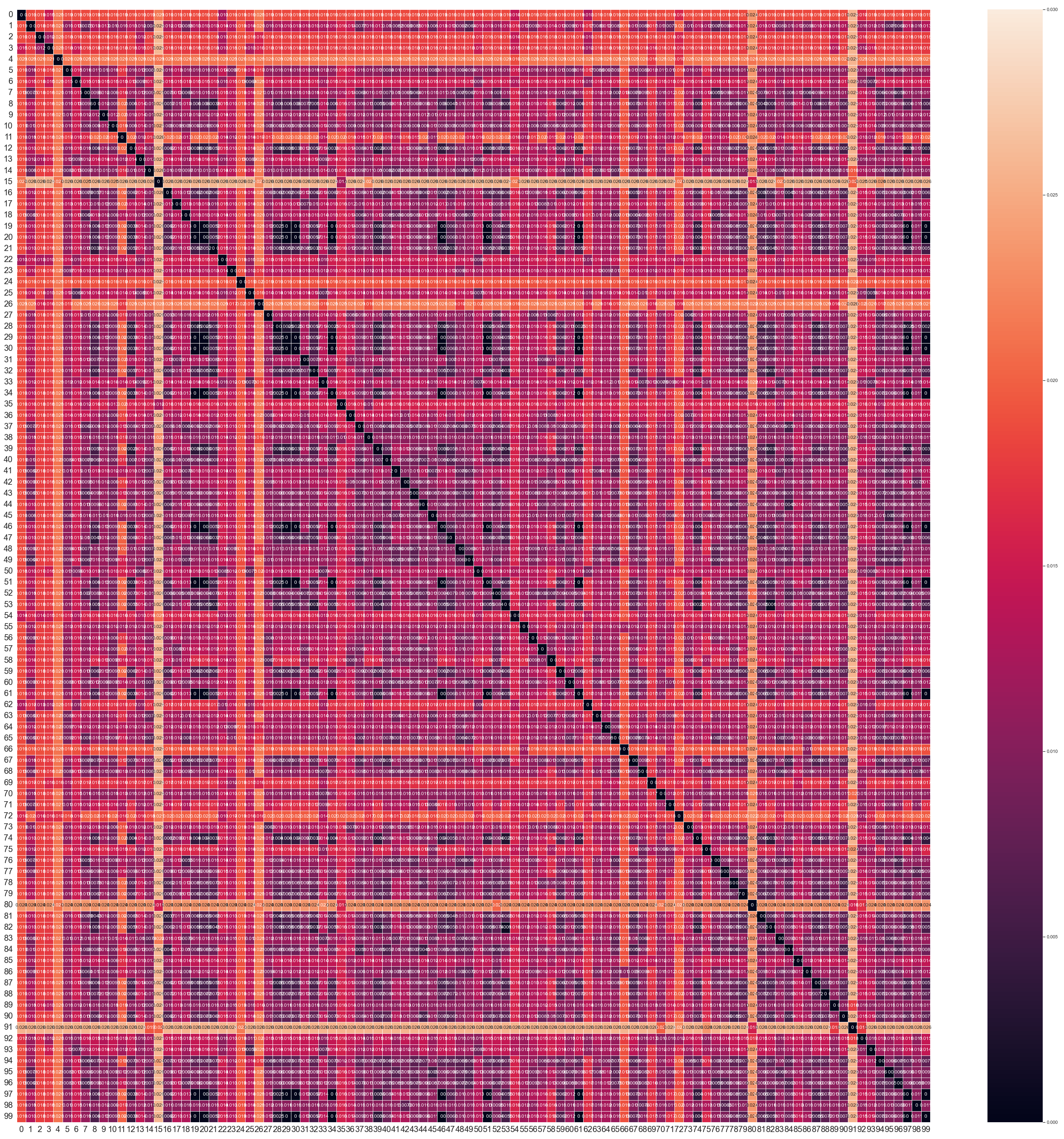}
         \caption{The $H_1$ values of the pairwise bottleneck distances matrix of sentences embedded by Word2Vec.}
         \label{fig:heat_gpt2_rips_bd_H1_100}
     \end{subfigure}
     \hfill
     \begin{subfigure}[b]{0.32\textwidth}
         \centering
         \includegraphics[width=\textwidth]{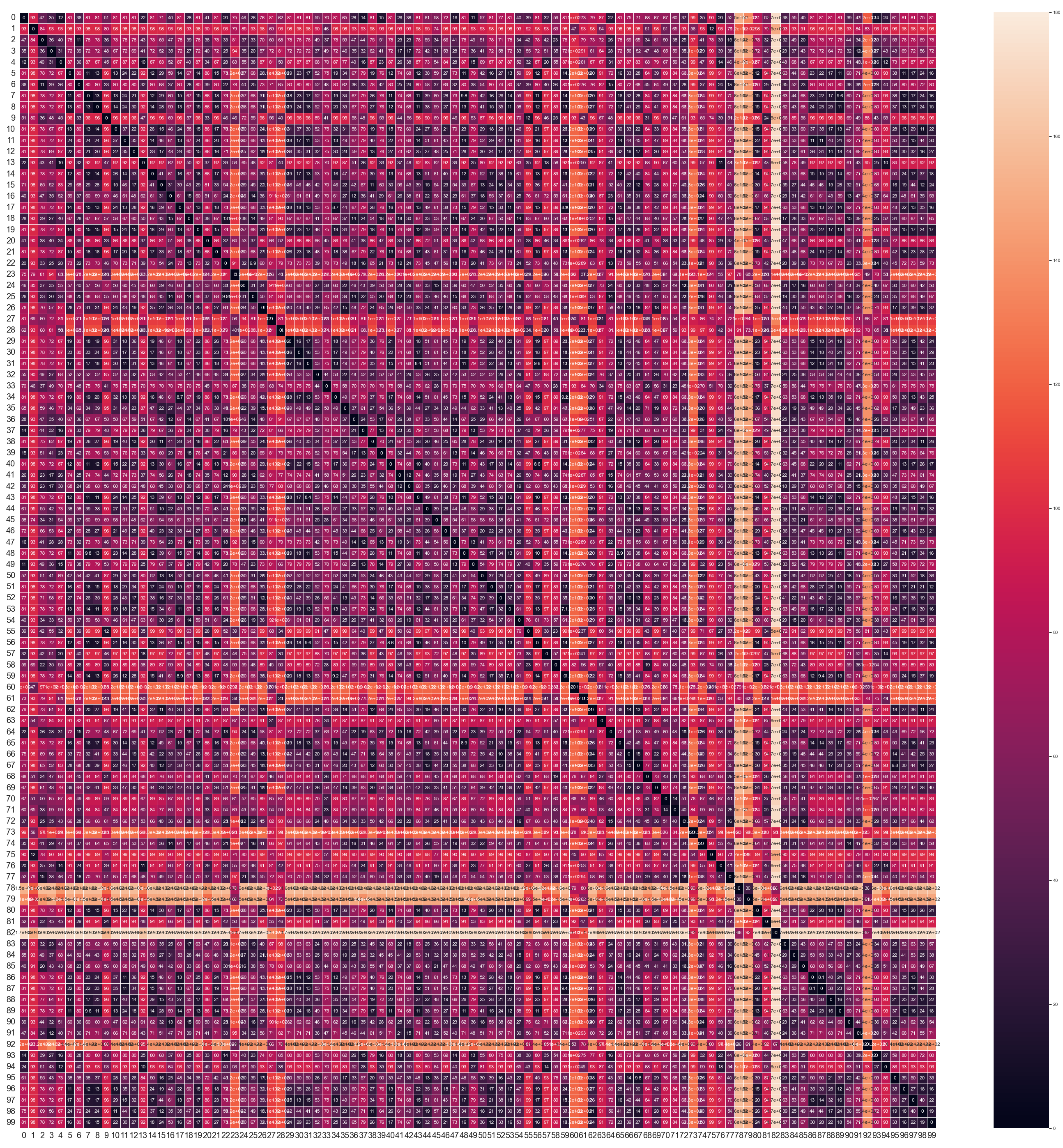}
         \caption{This figure shows the $H_0$ value of the pairwise bottleneck distances matrix of sentences embedded by GPT-$3$. Since the maximum distance is about $0.31$, the range of the distance is from $0$ to $0.4$. }
         \label{fig:heat_word2vec_rips_bd_H0_100}
     \end{subfigure}
     \hfill
     \begin{subfigure}[b]{0.32\textwidth}
         \centering
         \includegraphics[width=\textwidth]{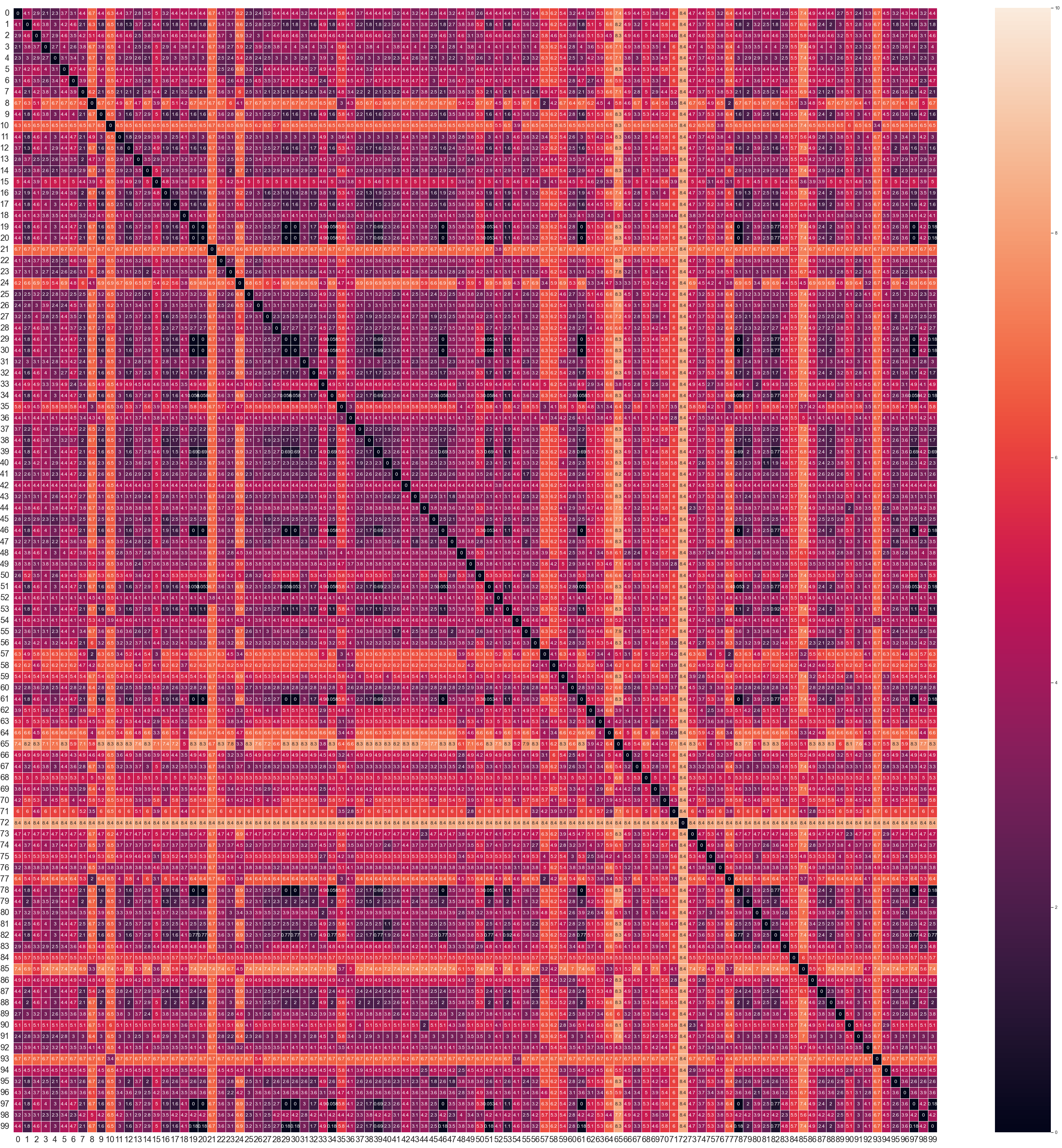}
         \caption{The $H_1$ value of the pairwise bottleneck distances matrix of sentences embedded by GPT-$3$. The range of the distance is from $0$ to $0.03$, since the maximum distance length is about $0.026$.}
         \label{fig:heat_word2vec_rips_bd_H1_100}
     \end{subfigure}
     \hfill
     \begin{subfigure}[b]{0.32\textwidth}
         \centering
         \includegraphics[width=\textwidth]{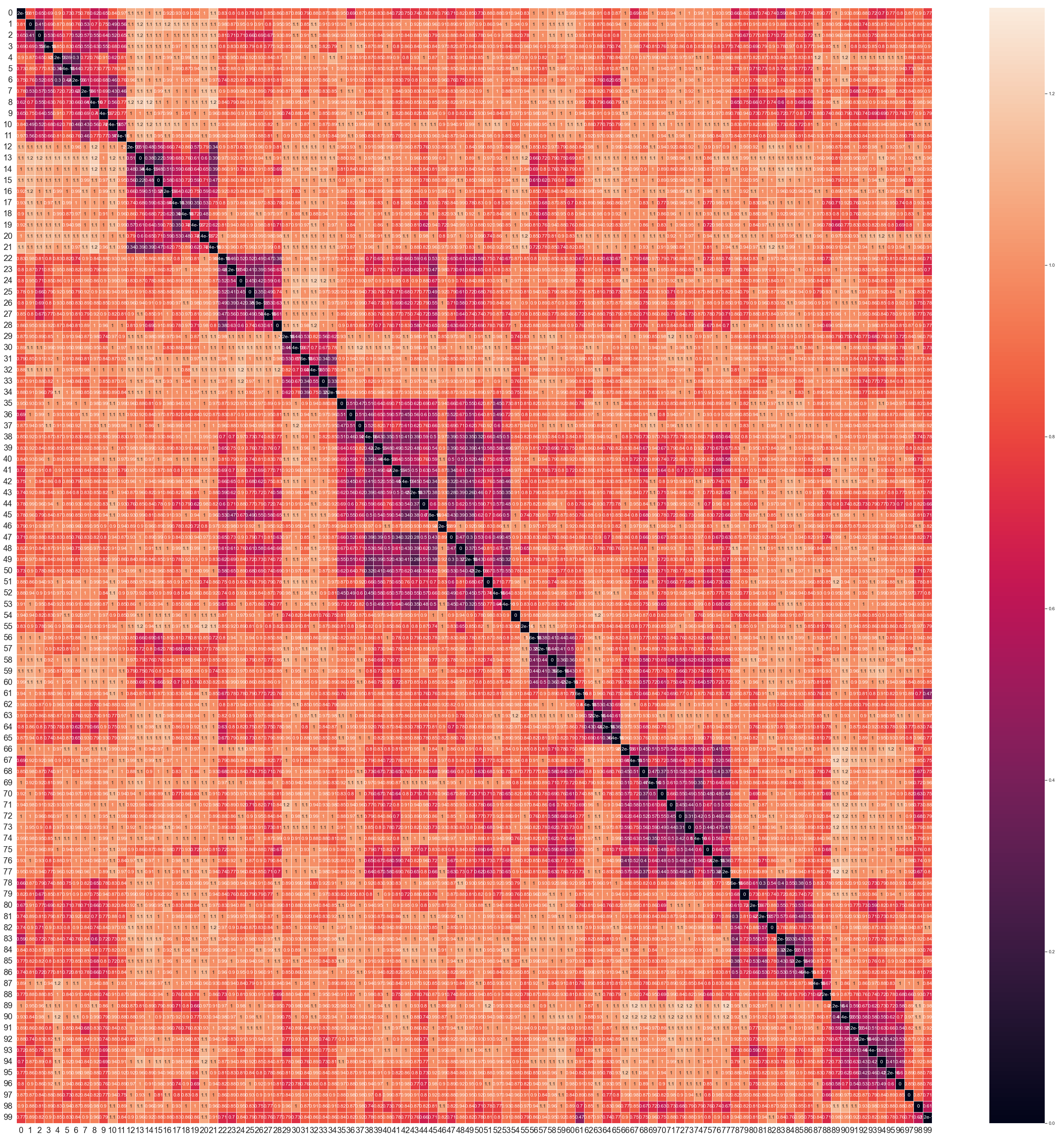}
         \caption{This figure shows the pairwise cosine distance matrix of sentences embedded by Sentence-BERT. Since the maximum cosine distance is $1.224$, the range of the distance is from $0$ to $1.3$. }
         \label{fig:heat_sentbert_cosinedissimilarity_100}
     \end{subfigure}
        \caption{Distance matrices. We observe that the $H_1$ value of bottleneck distances matrix for sentences embedded by GPT-$3$ and the Levenshtein distances matrix for plain text sentences are the most similar.}
        \label{fig:heat}
\end{figure}

We use the pre-trained Sentence-BERT model to generate the sentence embedding vectors. Each sentence vector is a $384$-dimensional vector. Then we construct a sentence vector cloud with one hundred $384$-dimensional sentence vectors in it.  Lastly, we compute the cosine distance for each pair of two sentences vectors and map them into a $100 \times 100$ matrix, which can be observed in Figure \ref{fig:heat_sentbert_cosinedissimilarity_100}.

\subsection{Correlation Analysis} \label{CA}
\begin{figure}
     \centering
     \begin{subfigure}[b]{0.49\textwidth}
         \centering
         \includegraphics[width=\linewidth]{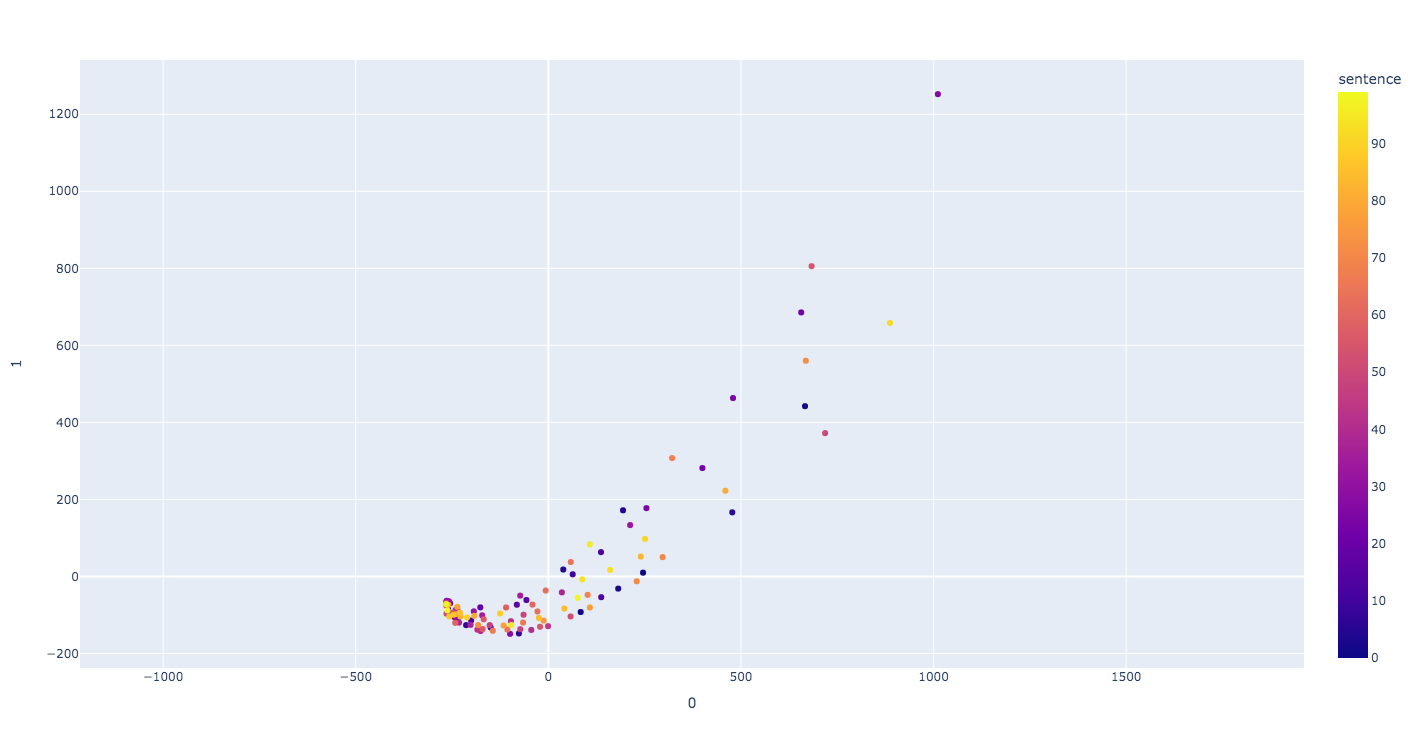}
         \caption{A visualization of the similarity of Levenshtein distances among one hundred plain text sentences in a Cartesian space.}
         \label{fig:levenshtein_distance_word_100_MDS}
     \end{subfigure}
     \hfill
     \begin{subfigure}[b]{0.49\textwidth}
         \centering
         \includegraphics[width=\linewidth]{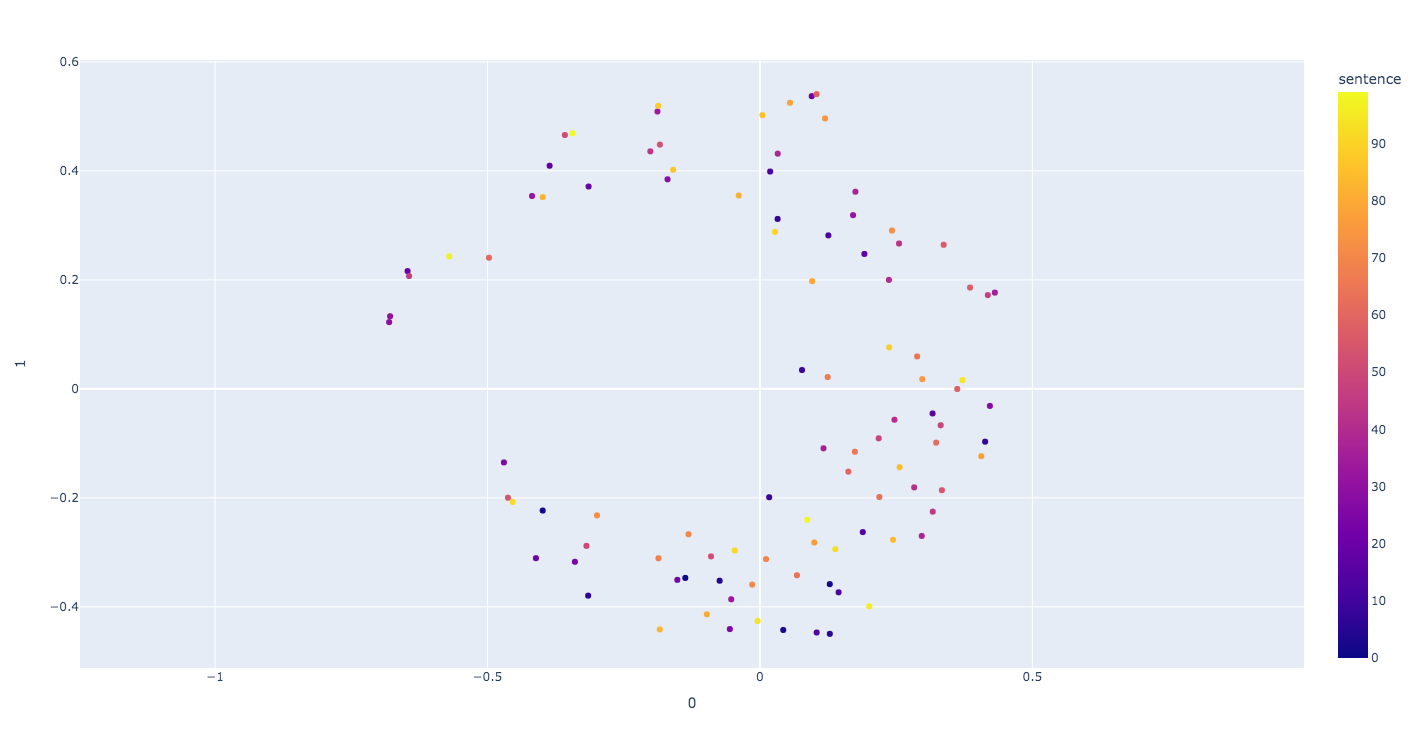}
         \caption{A visualization of similarity of $H_0$ value of the bottleneck distances among one hundred Word2Vec embedded sentence vectors in a Cartesian space.}
         \label{fig:word2vec_rips_bd_H0_100_MDS}
    \end{subfigure}
     \hfill
     \begin{subfigure}[b]{0.49\textwidth}
        \centering
        \includegraphics[width=\linewidth]{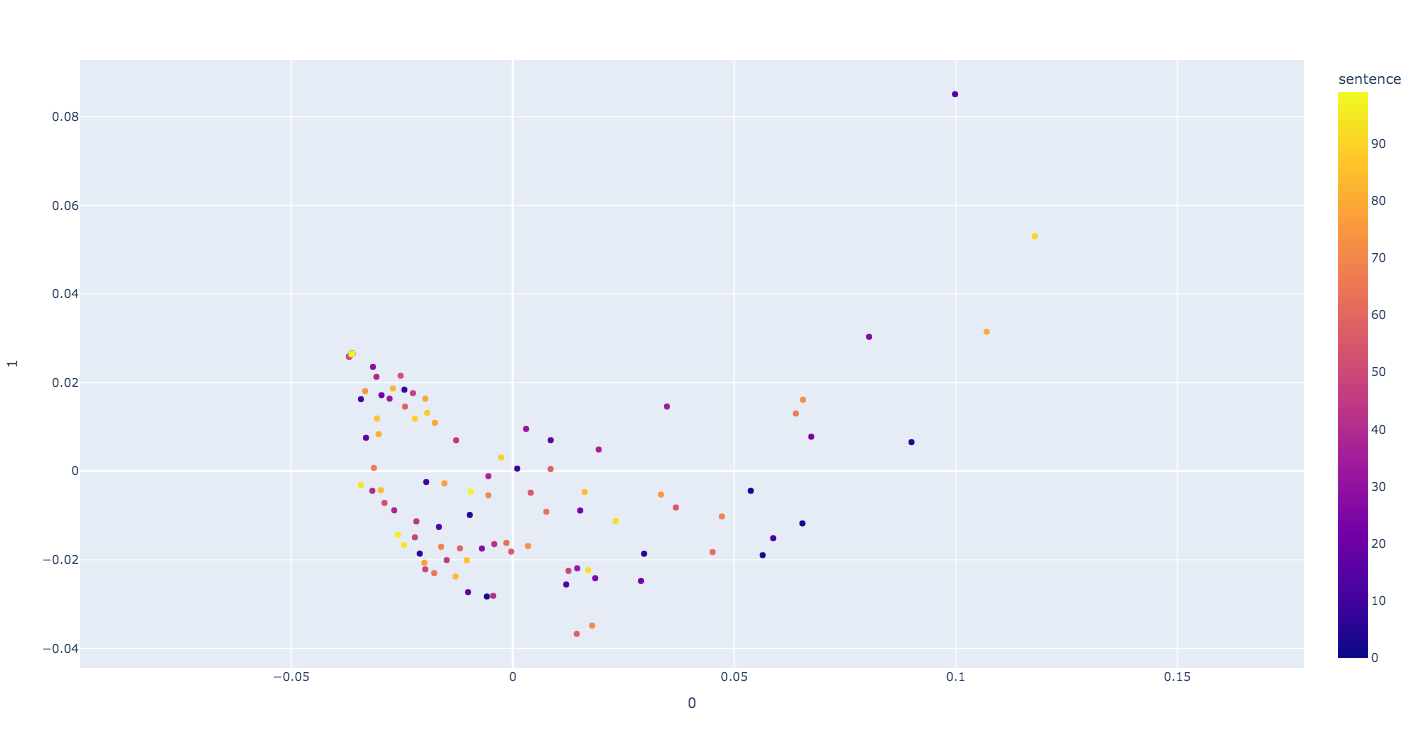}
         \caption{A visualization of similarity of $H_1$ value of the bottleneck distances among one hundred Word2Vec embedded sentence vectors in a Cartesian space.}
         \label{fig:word2vec_rips_bd_H1_100_MDS}
     \end{subfigure}
     \hfill
     \begin{subfigure}[b]{0.49\textwidth}
        \centering
        \includegraphics[width=\linewidth]{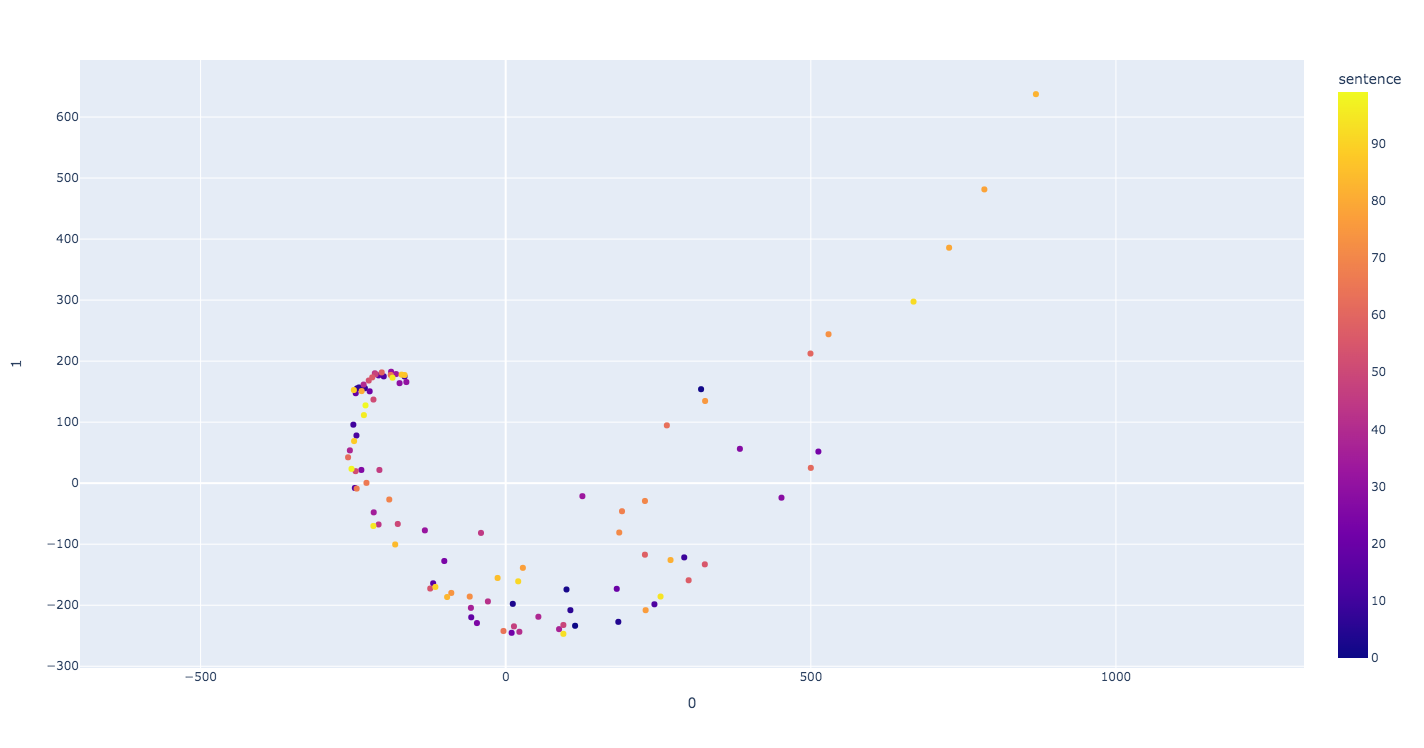}
         \caption{A visualization of the similarity of $H_0$ value of the bottleneck distances among one hundred GPT-$3$ embedded sentence vectors in a Cartesian space. }
         \label{fig:gpt2_rips_bd_H0_100_MDS}  
     \end{subfigure}
     \hfill
     \begin{subfigure}[b]{0.49\textwidth}
        \centering
        \includegraphics[width=\linewidth]{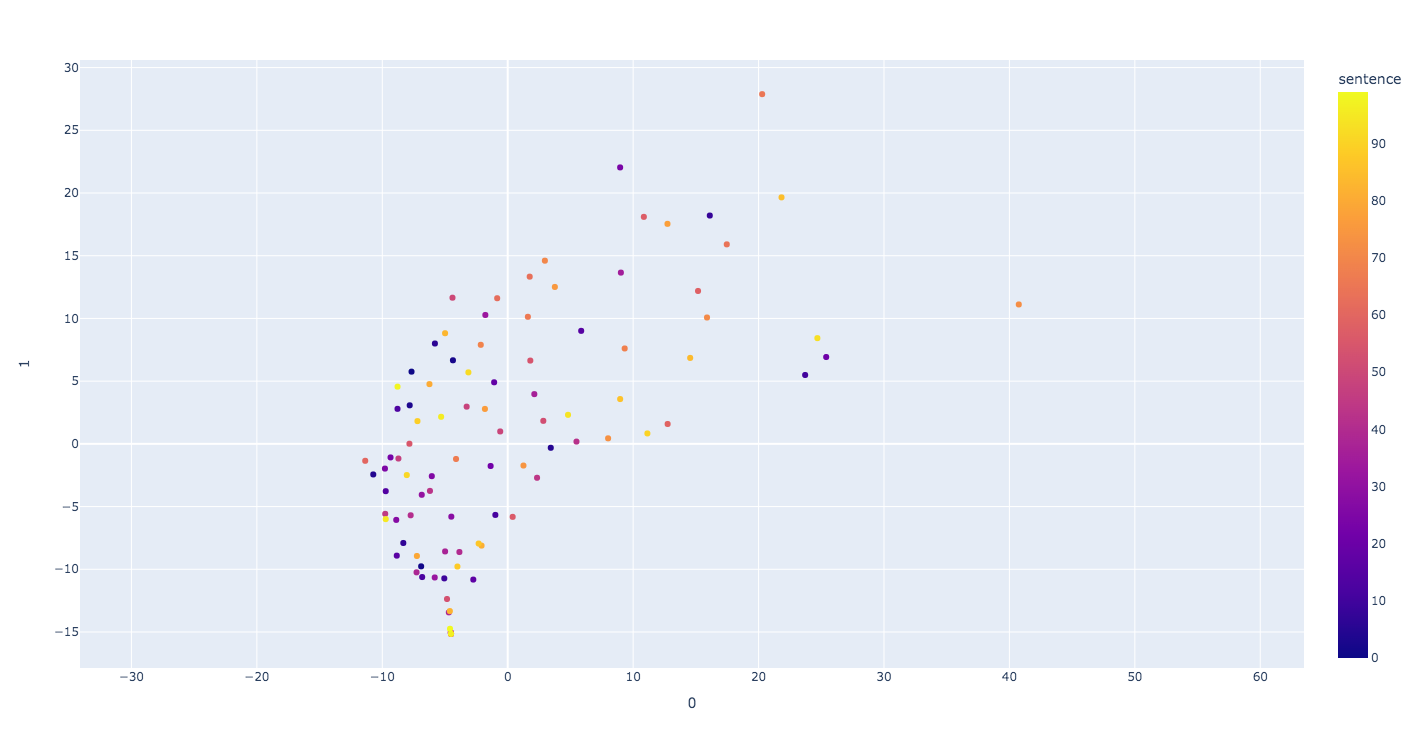}
         \caption{A visualization of similarity of $H_1$ value of the bottleneck distances among one hundred GPT-$3$ embedded sentence vectors in a Cartesian space. }
         \label{fig:gpt2_rips_bd_H1_100_MDS}     
     \end{subfigure}
     \hfill
     \begin{subfigure}[b]{0.49\textwidth}
         \centering
         \includegraphics[width=\linewidth]{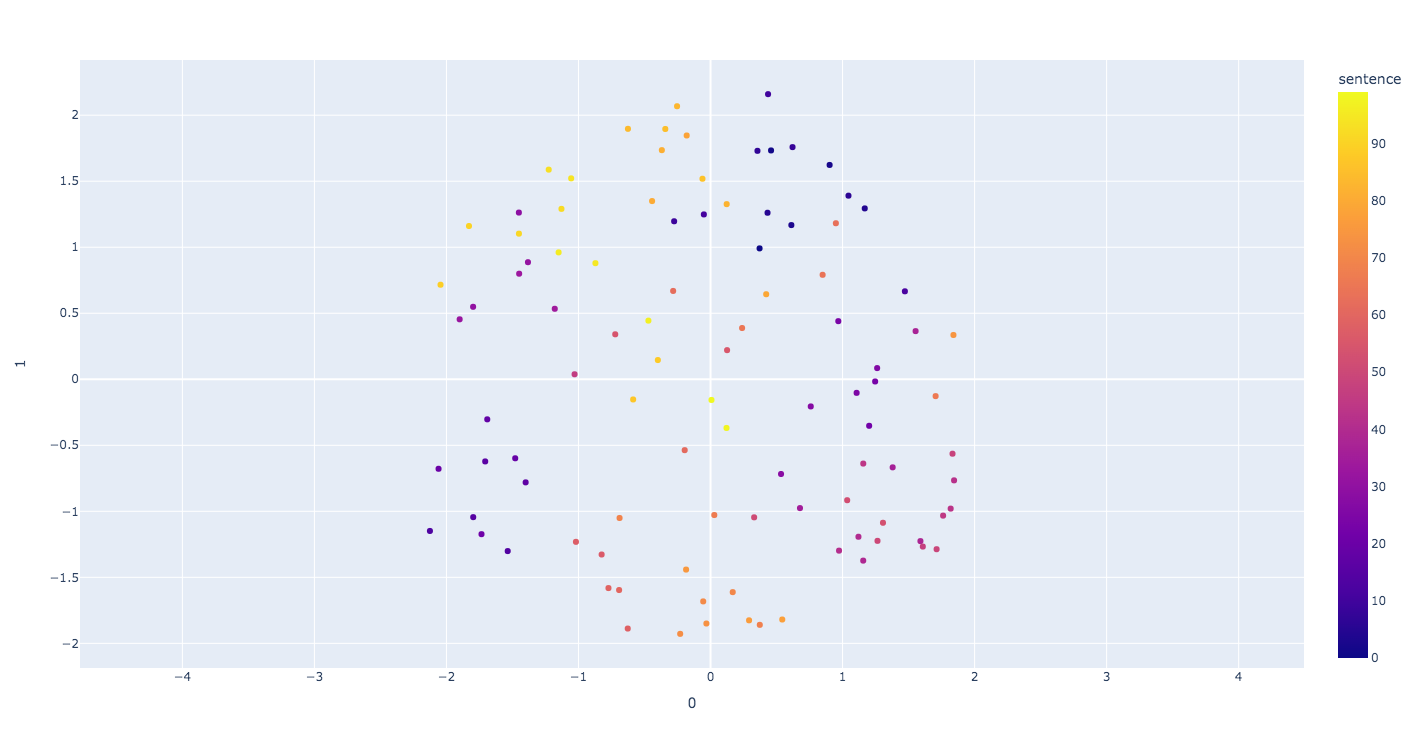}
         \caption{A visualization of similarity of cosine distances among one hundred Sentence-BERT embedded sentence vectors in a Cartesian space.}
         \label{fig:sentbert_cosinedissimilarity_100_MDS}
     \end{subfigure}
\caption{Visualizations of the similarities of pairwise sentence distances in each embedding spaces using MDS.}
\label{fig:mds}
\end{figure}

\begin{figure}
     \centering
     \begin{subfigure}[b]{0.49\textwidth}
         \centering
         \includegraphics[width=\textwidth]{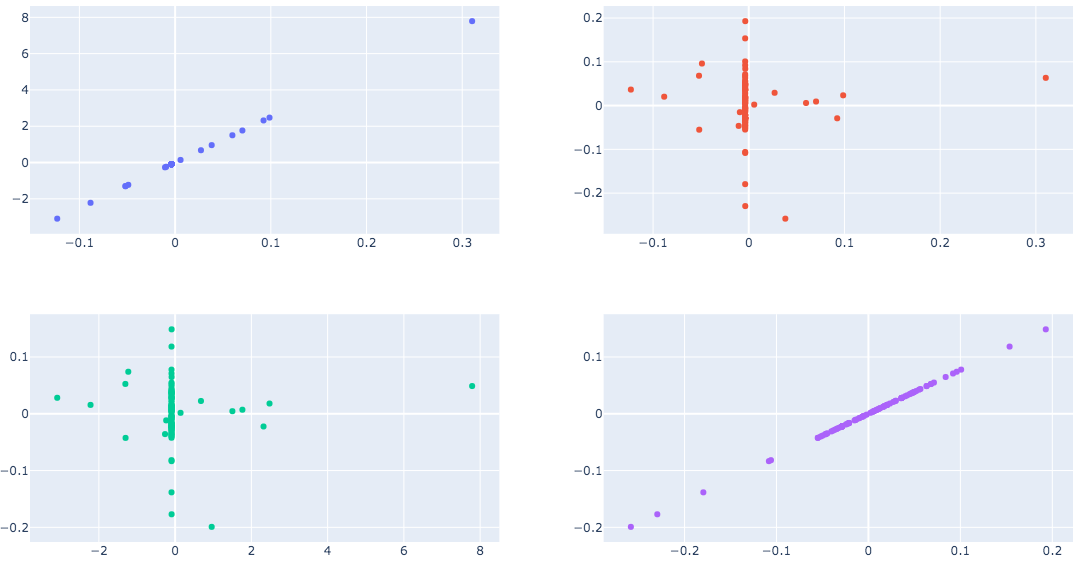}
         \caption{Canonical correlation between $H_1$ values of bottleneck distance matrix for GPT-$3$ embedding and $H_1$ values of bottleneck distance matrix for Word2Vec embedding.}
         \label{fig:CCA_GPT2_h1_word2vec_h1}
     \end{subfigure}
     \hfill
     \begin{subfigure}[b]{0.49\textwidth}
         \centering
         \includegraphics[width=\linewidth]{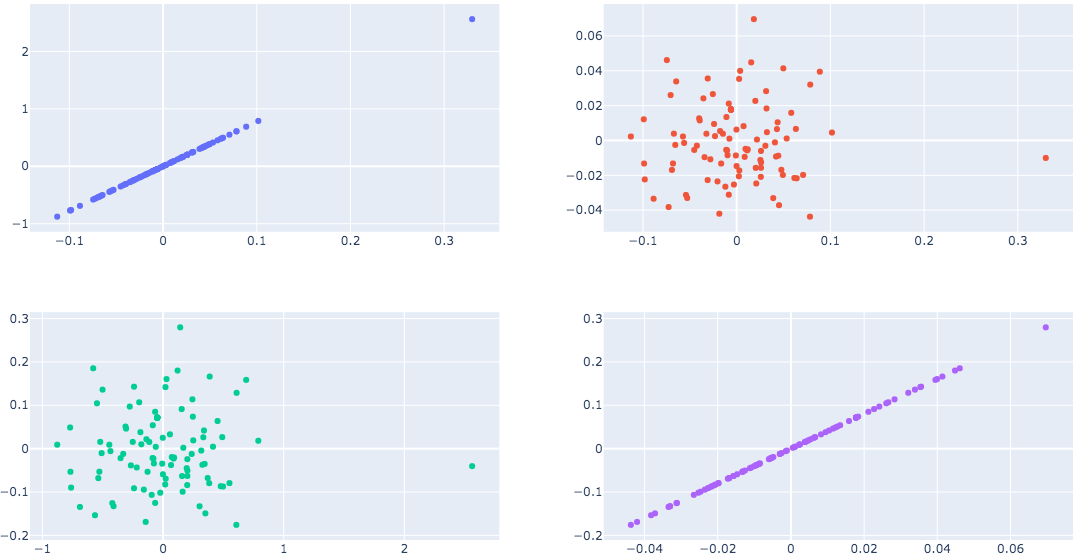}
         \caption{Canonical correlation between $H_1$ values of bottleneck matrix for GPT-$3$ embedding and cosine distance matrix for Sentence-BERT embedding.}
         \label{fig:CCA_GPT2_h1_sentbert}
     \end{subfigure}
     \hfill
     \begin{subfigure}[b]{0.49\textwidth}
         \centering
         \includegraphics[width=\linewidth]{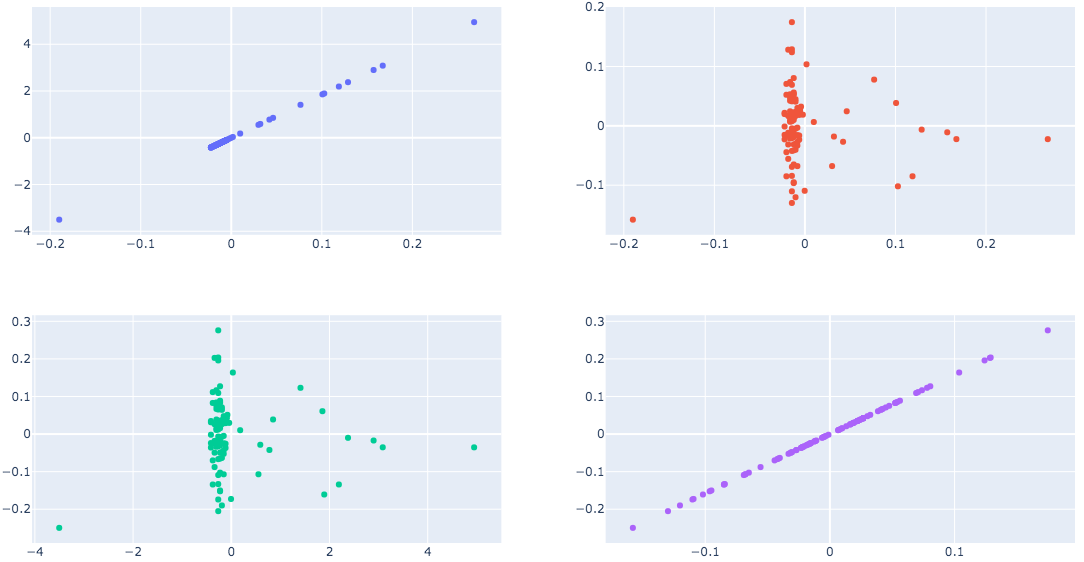}
         \caption{Canonical correlation between $H1$ values of bottleneck distance matrix for GPT-$3$ embedding and levenshtein distance for plain text sentence strings.}
         \label{fig:CCA_GPT2_h1_levenshtein}
     \end{subfigure}
     \hfill
     \begin{subfigure}[b]{0.49\textwidth}
         \centering
         \includegraphics[width=\textwidth]{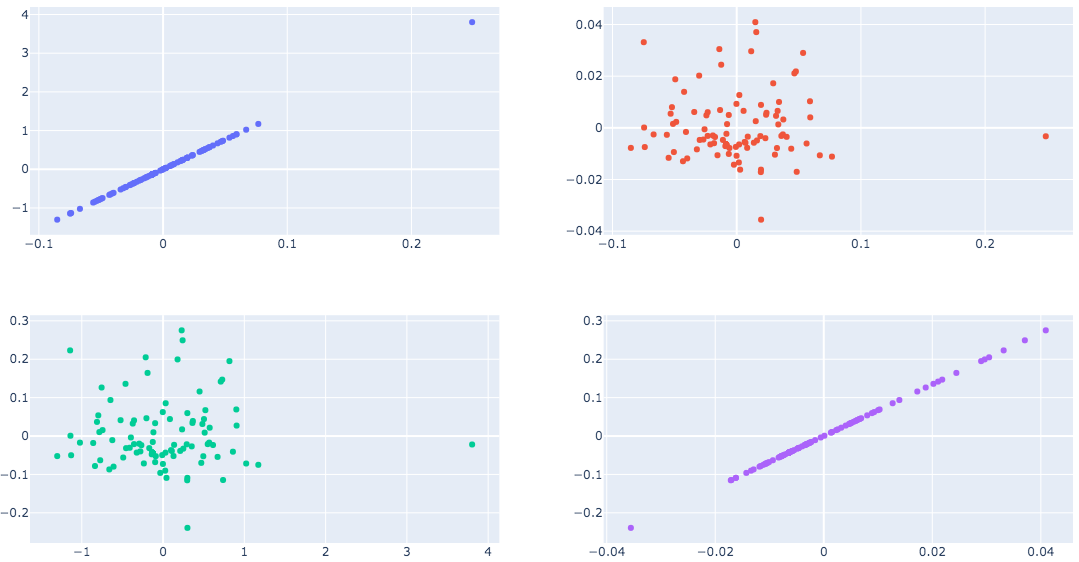}
         \caption{Canonical correlation between $H_1$ value of bottleneck distance matrix for Word2Vec embedding and cosine distance matrix for Sentence-BERT embedding.}
         \label{fig:CCA_word2vec_h1_sentbert}
     \end{subfigure}
     \hfill
     \begin{subfigure}[b]{0.49\textwidth}
         \centering
         \includegraphics[width=\linewidth]{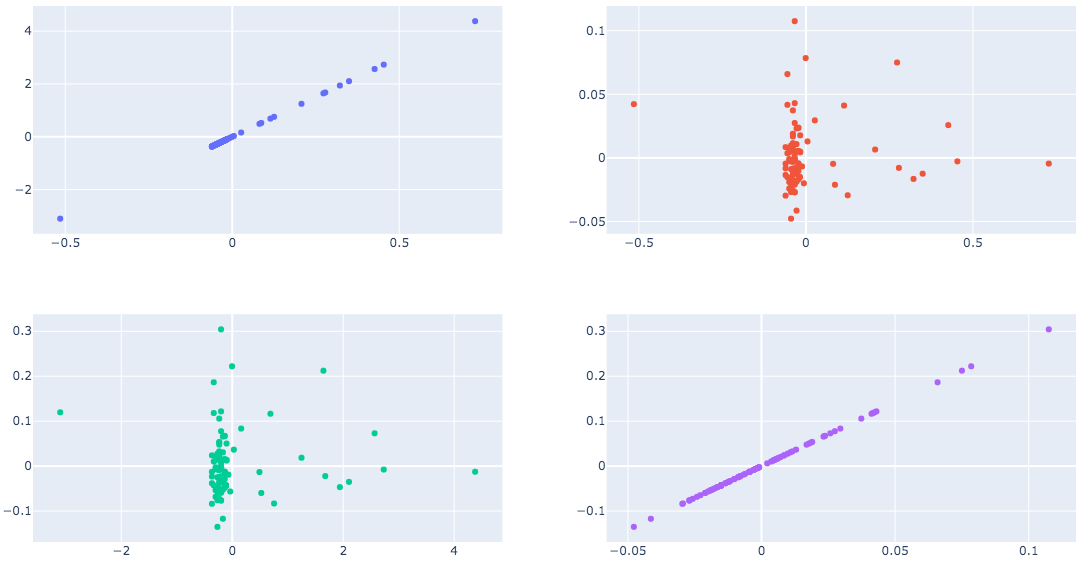}
         \caption{Canonical correlation between $H_1$ value of bottleneck distance matrix for Word2Vec embedding and levenshtein distance matrix for plain text sentence strings.}
         \label{fig:CCA_word2vec_h1_levenshtein}
     \end{subfigure}
     \hfill
     \begin{subfigure}[b]{0.49\textwidth}
         \centering
         \includegraphics[width=\linewidth]{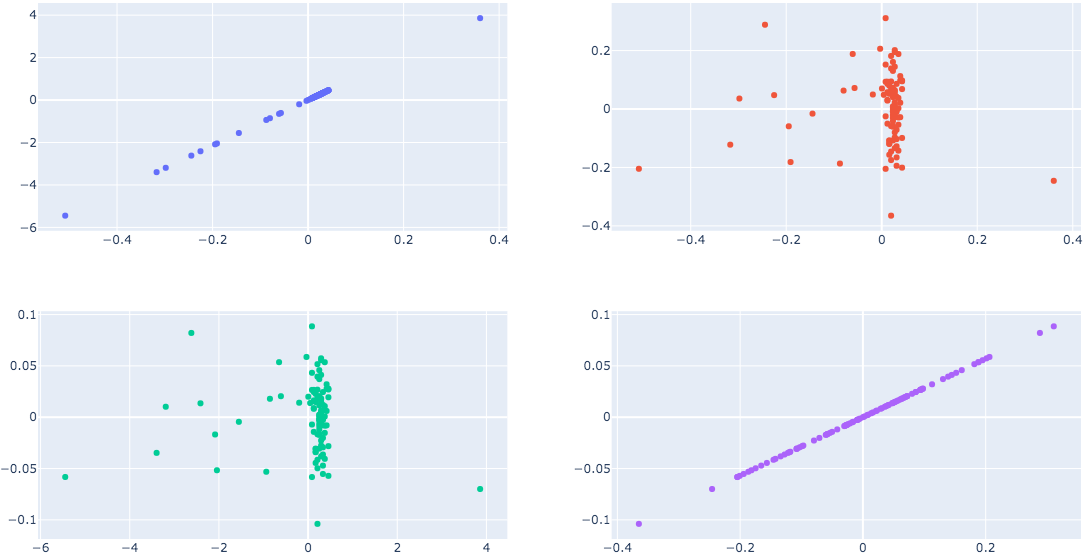}
         \caption{Canonical correlation between cosine distance matrix of Sentence-BERT embedding and Levenshtein distance matrix for plain text sentence strings.}
         \label{fig:CCA_sentbert_levenshtein}
    \end{subfigure}
    \caption{Canonical correlation analysis between any possible combinations of distance matrices. (We omit the two $H_0$ values of bottleneck distance matrices.) CCA computes two scores, $comp.1$ and $comp.2$, for each pair of distance matrices. 
    In each of the above six figures, 
    the upper left figure shows that the correlation between $comp.1$ of matrix $X$ and $comp.1$ of matrix $Y$ is $1$; 
    the upper right figure shows that the correlation between $comp.1$ and $comp.2$ of matrix $X$ is $0$; 
    the lower left figure shows that the correlation between $comp.1$ and $comp.2$ of matrix $Y$ is $0$;
    the lower right figure shows that the correlation between $comp.2$ of matrix $X$ and $comp.2$ matrix $Y$ is $1$. 
    The results show that $comp.2$s and $comp.1$ in each of those matrices have strong correlations but $comp.2$s and $comp.1$ have no correlations at all.}
    \label{fig:cca}
\end{figure}

From the six distance matrices in Figure \ref{fig:heat}, we observe that there are some correlations among sentence clouds in different embedding spaces and correlations among sentence vectors in an embedding space. We further investigate the correlation of sentences within each sentence embedding space by visualizing the similarity of sentence distances in a Cartesian space using MDS. The similarities of the six distance matrices are shown in Figure \ref{fig:mds}. For investigating the correlation of distance matrices across embedding spaces, we compute the CCA and the scaled Hausdorff distances between any possible pair of distance matrices. The results are in Figure \ref{fig:cca} and Table \ref{tab:hausdorff} respectively. 

\begin{table}[!htb]
\footnotesize
\centering
    \begin{tabular*}{\columnwidth}{@{\extracolsep{\fill}}ccc}
         &\textbf{minimum}& \\
        \textbf{Distance Matrices}&\textbf{Hausdorff Distance}&$\alpha$\\
        \hline
        GPT-$3$ embedding bottleneck distance $H_1$ values \& & & \\ Levenshtein distance & $6.7496$ & $0.0391$ \\
        \hline
        GPT-$3$ embedding bottleneck distance $H_1$ values \& & & \\
        Word2Vec embedding bottleneck distance $H_1$ values & $6.0985$ & $323.7458$ \\
        \hline
        GPT-$3$ embedding bottleneck distance $H_1$ values \& & & \\
        Sentence-BERT embedding cosine distance & $6.5734$ & $5.6899$ \\
        \hline
        Word2Vec embedding bottleneck distance $H_1$ values \& & & \\
        Levenshtein distance & $0.0193$ &$0.0001$ \\
        \hline
        Word2Vec embedding bottleneck distance $H_1$ values \& & & \\
        Sentence-BERT embedding cosine distance & $0.0219$ & $0.0184$ \\
        \hline
        Levenshtein distance \& & & \\ 
        Sentence-BERT embedding cosine distance & $1.0565$ & $0.0063$ \\
        \end{tabular*}
        \caption{The optimal values of scaled Hausdorff distances(SHD) for each pair of distance matrices with the corresponding scaled $\alpha$ values. Figure \ref{fig:hausdorff} shows how we approximate the values.}
        \label{tab:hausdorff}
\end{table}

\newpage
\section{Conclusion}
Figure \ref{fig:heat} shows the six distance matrices we computed in Section \ref{ES}. The darker the color is, the closer the two sentence strings are. Since the ranges of matrices are different, we cannot compare two matrices directly by colors, but we can compare the patterns of differences in color. We observe that the pattern of $H_1$ value of bottleneck distances matrix for sentences embedded by GPT-$3$, Figure \ref{fig:heat_gpt2_rips_bd_H1_100}, and the pattern of Levenshtein distances matrix for plain text sentences, Figure \ref{fig:heat_levedis_word_100}, are the most similar, so these two distance matrices might be highly correlated. Similarly, by observation, the $H_1$ value of bottleneck distances matrix for sentences embedded by GPT-$3$ and the $H_1$ value of bottleneck distances matrix for sentences embedded by Word2Vec are similar, so the distance matrices, Figure \ref{fig:heat_gpt2_rips_bd_H1_100} and Figure \ref{fig:heat_word2vec_rips_bd_H1_100}, seem to be highly correlated. In Figure \ref{fig:heat_sentbert_cosinedissimilarity_100}, the squared blocks lay on the diagonal of the distance matrix indicate that these sentences share same topics. 

Figure \ref{fig:mds} shows the MDS of six distance matrices in Figure \ref{fig:heat}. By observation, the MDS of $H_1$ value of bottleneck distances matrix for sentences embedded by GPT-$3$, Figure \ref{fig:gpt2_rips_bd_H1_100_MDS}, and the MDS of $H_1$ value of bottleneck distances matrix for sentences embedded by Word2Vec, Figure \ref{fig:word2vec_rips_bd_H1_100_MDS}, are similar. And each pairwise distance lay in similar coordinates in the two Cartesian spaces Figure \ref{fig:gpt2_rips_bd_H1_100_MDS} and Figure \ref{fig:word2vec_rips_bd_H1_100_MDS}. This reflects that for the same sentence, the sentence vectors embedded by GPT-$3$ and Word2Vec share similar scaled distances with respect to the other sentence vectors in the same embedding space. Moreover, the MDS of Levenshtein distance matrix of pain text sentence string, Figure \ref{fig:levenshtein_distance_word_100_MDS}, and the MDS of $H_0$ value of bottleneck distances matrix for sentences embedded by GPT-$3$, Figure \ref{fig:gpt2_rips_bd_H0_100_MDS}, are similar to some extent. The MDS of cosine distance matrix for sentences embedded by Sentence-BERT, Figure \ref{fig:sentbert_cosinedissimilarity_100_MDS}, and the MDS of $H_0$ value of bottleneck distances matrix for sentences embedded by Word2Vec, Figure \ref{fig:word2vec_rips_bd_H0_100_MDS}, are similar to some extent.  We are surprised to see that the Figure \ref{fig:cca} shows perfect canonical correlations for any pair of distance matrices. 

\begin{figure}
    \centering
    \begin{subfigure}[b]{0.49\textwidth}
         \centering
         \includegraphics[width=\textwidth]{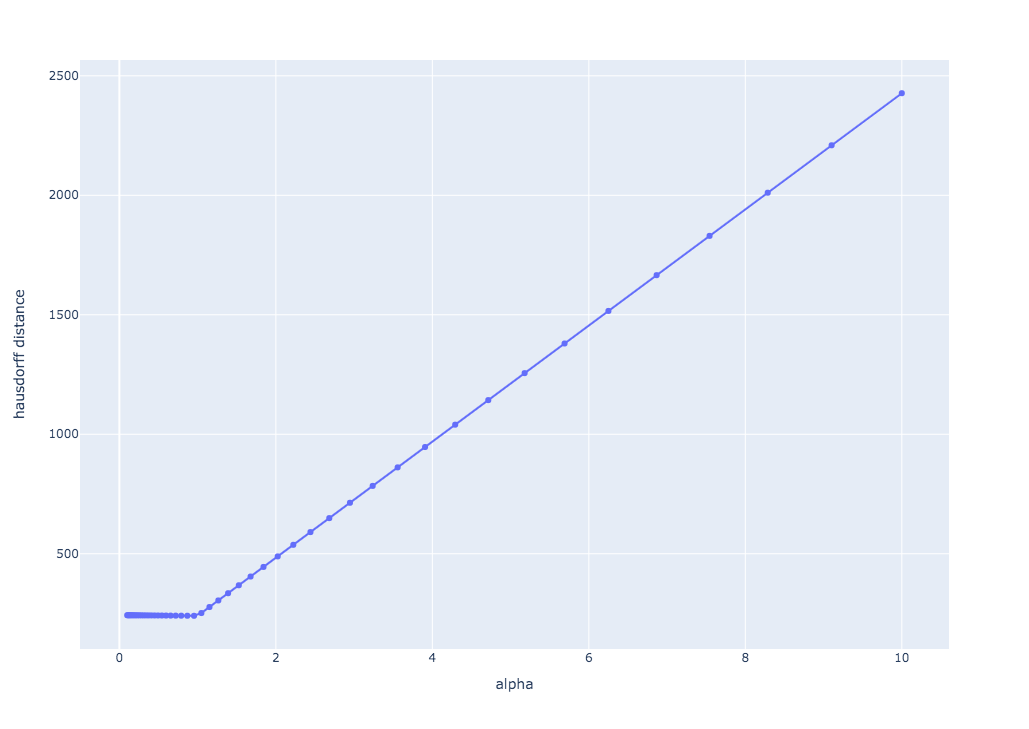}
         \caption{The SHD between bottleneck distances matrix of GPT-$3$ embedding vectors and Levenshtein distance matrix. The minimum Hausdorff distance is $6.7411260909223545$ the corresponding alpha is $0.03906939937054615$.}
         \label{fig:hd_gpt2_levenshtein}
    \end{subfigure}
    \begin{subfigure}[b]{0.49\textwidth}
         \centering
         \includegraphics[width=\textwidth]{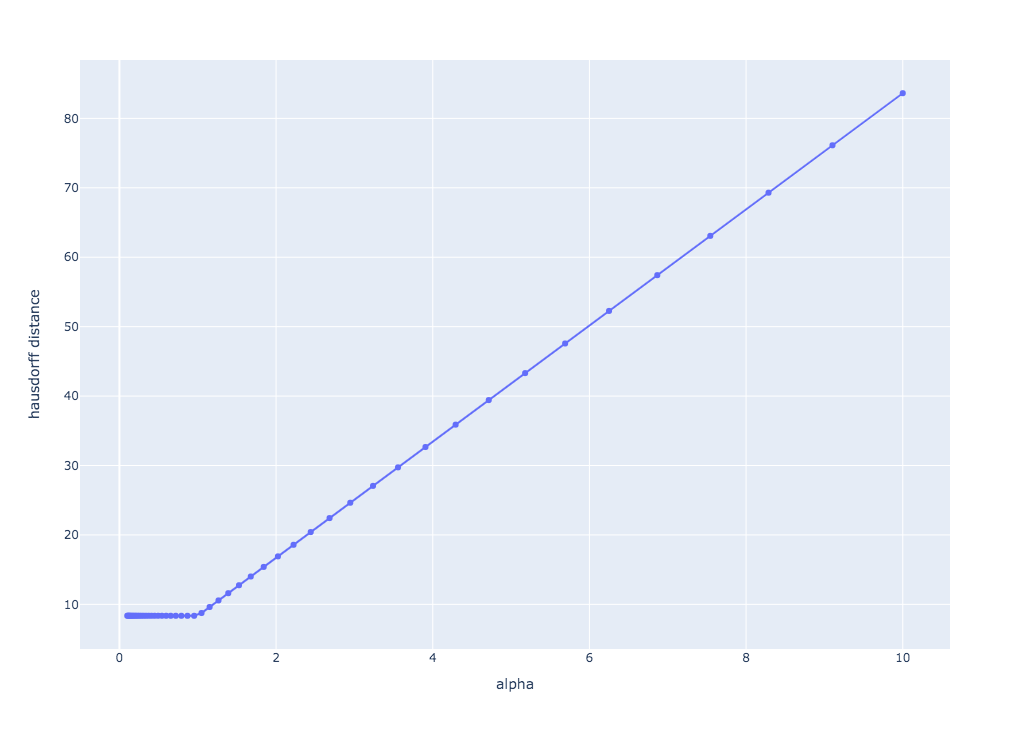}
         \caption{The SHD between bottleneck distances matrix of GPT-$3$ embedding vectors and bottleneck distances matrix of Word2Vec embedding vectors. The minimum Hausdorff distance is $6.098477954565762$ the corresponding alpha is  $323.74575428176433$.}
         \label{fig:hd_gpt2_word2vec}
    \end{subfigure}
    \begin{subfigure}[b]{0.49\textwidth}
         \centering
         \includegraphics[width=\textwidth]{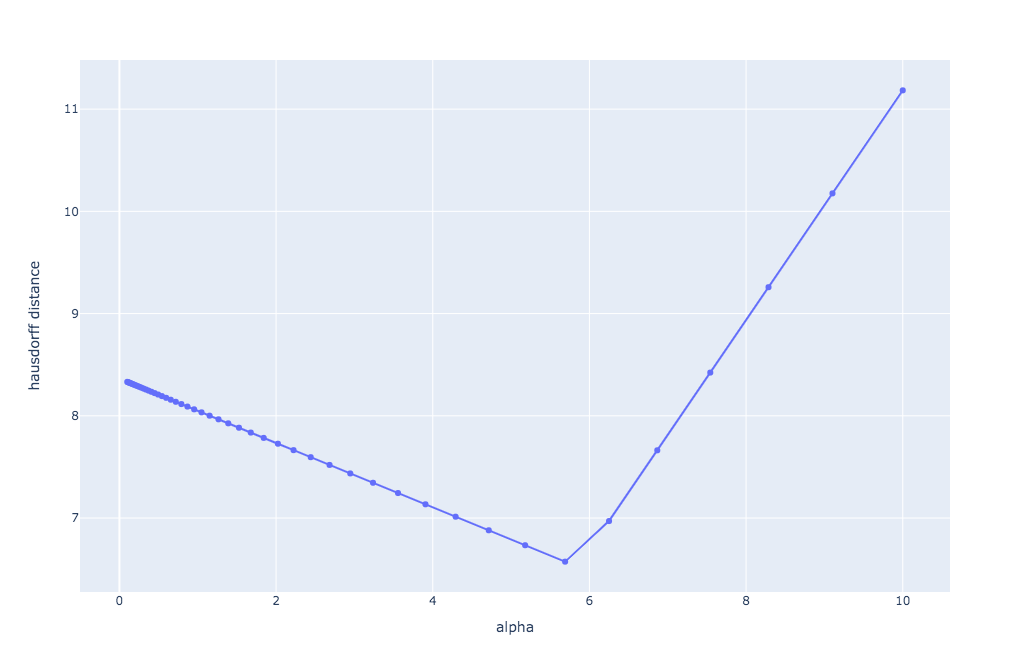}
         \caption{The SHD between bottleneck distances matrix of GPT-$3$ embedding vectors and bottleneck distances matrix of Sentence-BERT embedding vectors. The minimum Hausdorff distance is $6.573426558673857$ the corresponding alpha is $5.689866029018296$.}
         \label{fig:hd_gpt2_sentbert}
    \end{subfigure}
    \begin{subfigure}[b]{0.49\textwidth}
         \centering
         \includegraphics[width=\textwidth]{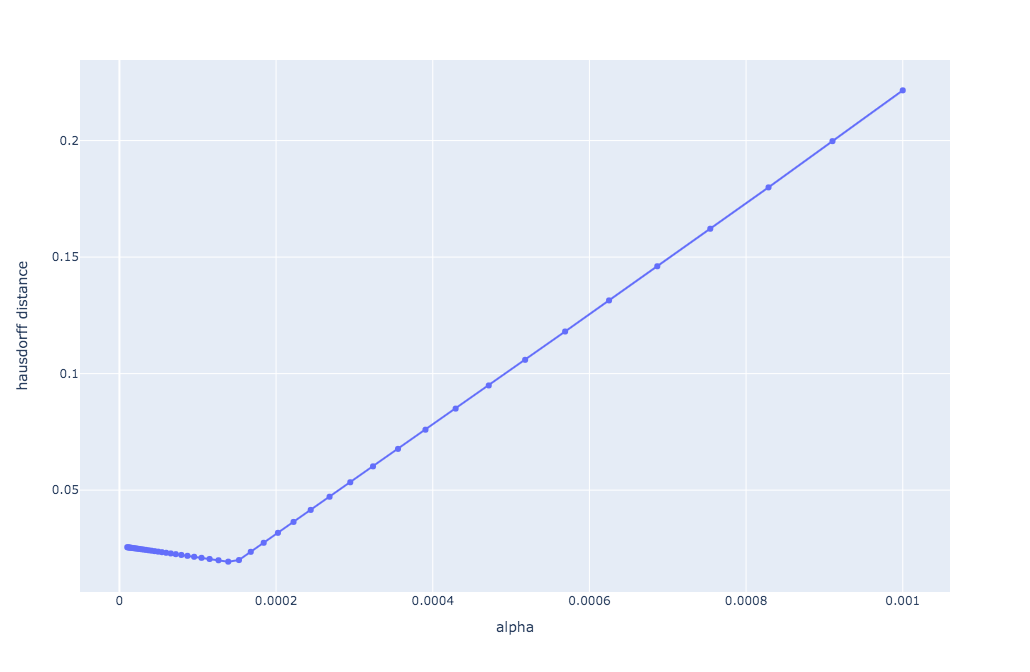}
         \caption{The SHD between bottleneck distances matrix of Word2Vec embedding vectors and Levenshtein distance matrix. The minimum Hausdorff distance is $0.019304307878458445$ the corresponding alpha is $0.00013894954943731373$.}
         \label{fig:hd_word2vec_levenshtein}
    \end{subfigure}
    \begin{subfigure}[b]{0.49\textwidth}
         \centering
         \includegraphics[width=\textwidth]{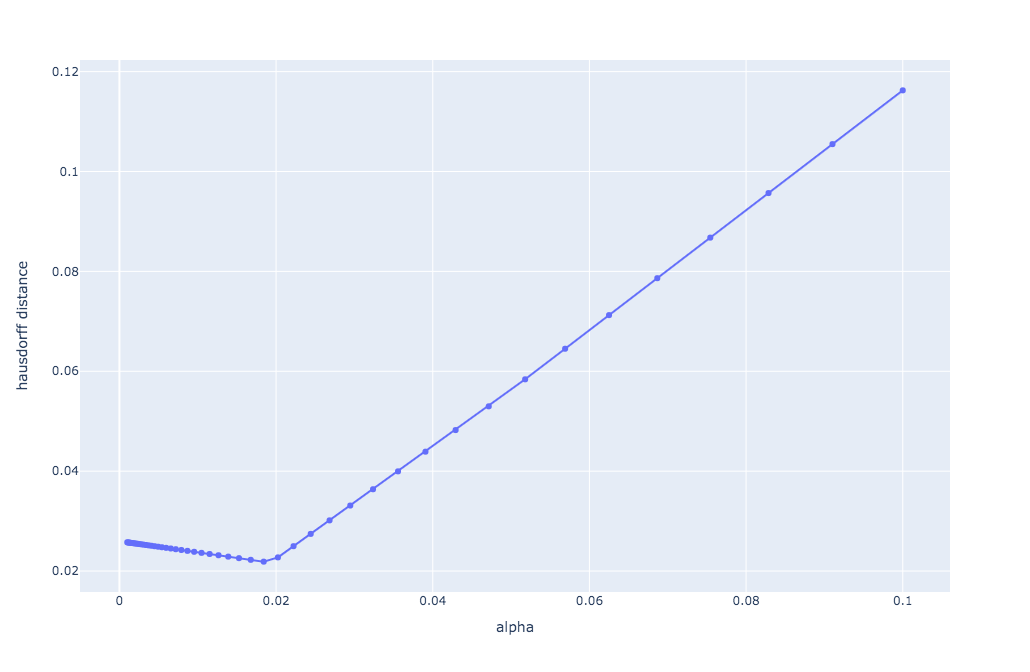}
         \caption{The SHD between bottleneck distances matrix of Word2Vec embedding vectors and bottleneck distances matrix of Sentence-BERT embedding vectors. The minimum Hausdorff distance is $0.02188291702897182$ the corresponding alpha is $0.018420699693267154$.}
         \label{fig:hd_word2vec_sentbert}
    \end{subfigure}
    \begin{subfigure}[b]{0.49\textwidth}
         \centering
         \includegraphics[width=\textwidth]{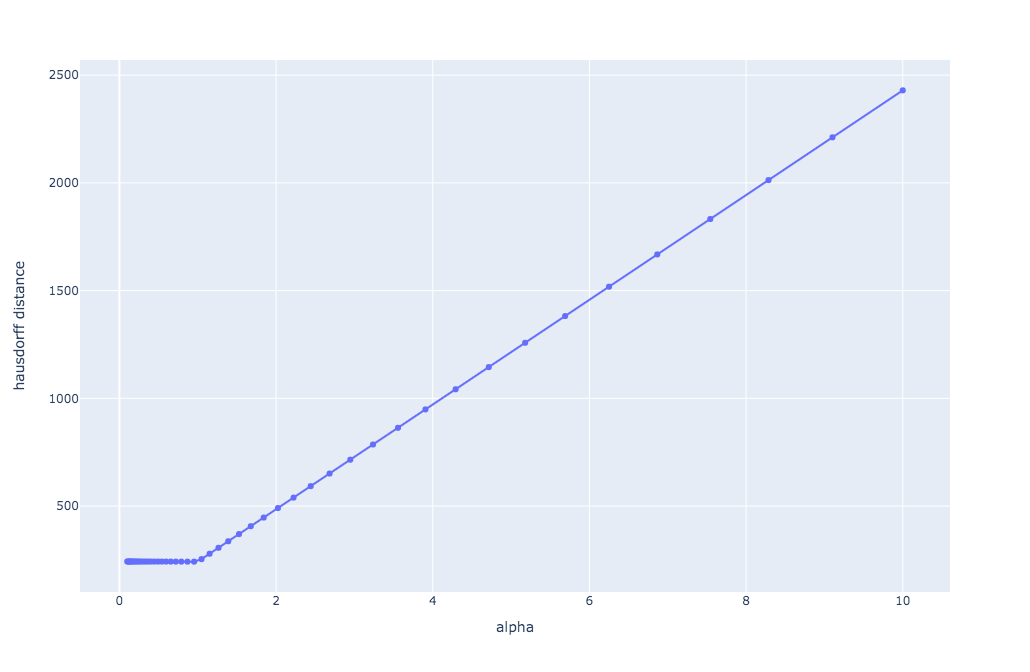}
         \caption{The SHD between bottleneck distances matrix of Sentence-BERT embedding vectors and Levenshtein distance matrix. The minimum Hausdorff distance is $1.0564626656578358$ the corresponding alpha is $0.0062505519252739694$.}
         \label{fig:hd_sentbert_levenshtein}
    \end{subfigure}
\caption{This shows how we approximate the optimal values of scaled Hausdorff distances (SHD) for each pair of distance matrices with the corresponding alpha values.}
\label{fig:hausdorff}
\end{figure}

Table \ref{tab:hausdorff} is the results of optimal values of scaled Hausdorff distances for any possible pair of distance matrices and the corresponding $\alpha$ values. We observe that the minimum Hausdorff distance between $H_1$ value of GPT-$3$ embedding bottleneck distance matrix and any other distance matrices, including Levenshtein distance matrix, $H_1$ value of Word2Vec embedding bottleneck distance matrix, and Sentence-BERT embedding cosine distance matrix, are all in the range from $6$ to $7$ with the corresponding scaled values $\alpha$. The minimum Hausdorff distance between $H_1$ value of Word2Vec embedding bottleneck distance matrix and, either Levenshtein distance matrix, or Sentence-BERT embedding cosine distance matrix, are about $0.02$ with the corresponding scaled values, $\alpha$. Figure \ref{fig:hausdorff} shows how we approximate the optimal scaled Hausdorff distances with the corresponding scaled values $\alpha$. 

\section{Future}
Based on observations and conclusion, we come up with the following directions of research that we will explore in the future. 

\subsection{Interpretation}
Natural language understanding\cite{sun2023truth, sun2005study} is an important task in NLP. This work shows a topological way to explore information inside a model, i.e., GPT-$3$, from outputs. 
The work\cite{naitzat2020topology} shows topological changes in each layer while messages passing through a network. 
Once computed persistence homology of outputs and internal layers we are interested in ways of explaining these homology\cite{sun2023greedy}. 
We will further test our model interpretation pipeline in generative models. Some pre-trained models and preliminary works are available at \url{https://huggingface.co/tianyisun}. 
We will develop topological pipeline for model error discovery and repair\cite{ning23es}. 

If a consistent approach is used to interpret the model, will we see similar results across different models?   
This question will also lead to the whole matrix thing in Figure \ref{fig:heat}. A follow-up question is that what is the correlation between distance of sentence embedding and meaning of sentences?

\subsection{Generation} 
\begin{figure}
    \centering
    \begin{subfigure}[b]{\textwidth}
         \includegraphics[width=0.14\textwidth]{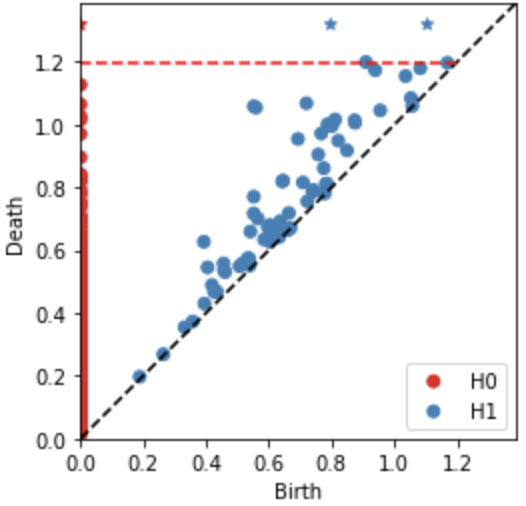}
         \includegraphics[width=0.21\textwidth]{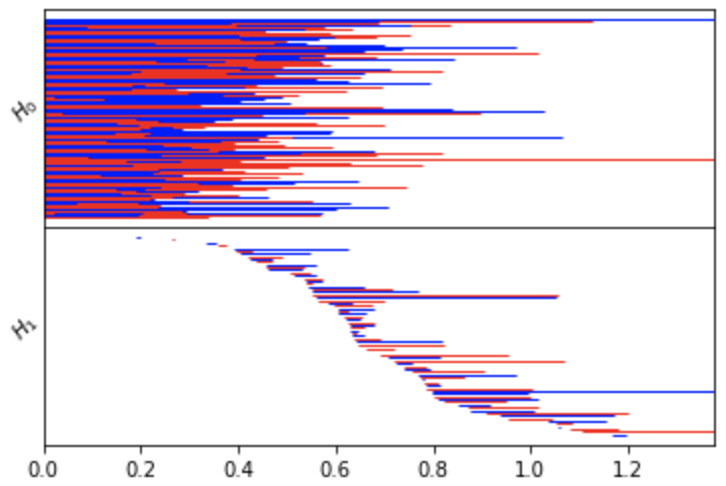}
         \includegraphics[width=0.14\textwidth]{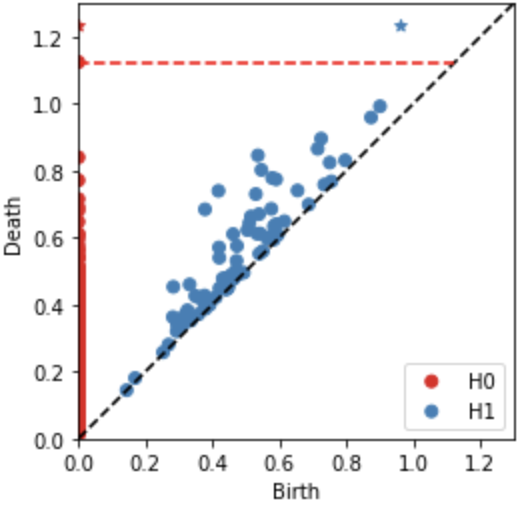}
         \includegraphics[width=0.21\textwidth]{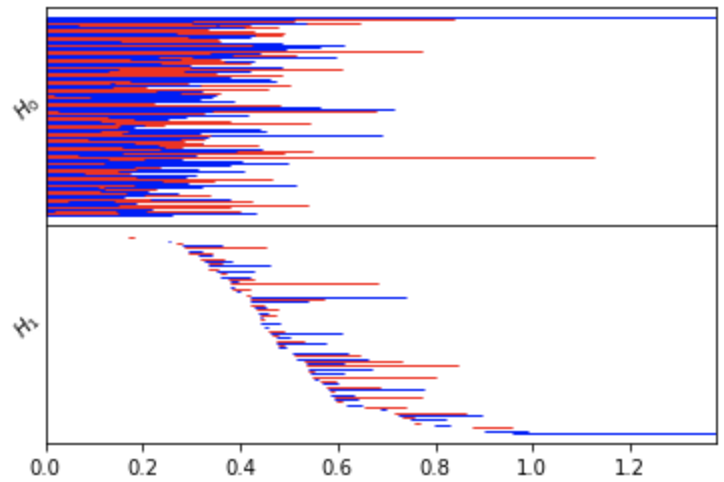}
         \caption{Epoch: $2000$; Loss: $0.115287$; Train Accuracy: $0.964$; Test Accuracy: $0.960$.}
    \end{subfigure}
    \begin{subfigure}[b]{\textwidth}
         \includegraphics[width=0.14\textwidth]{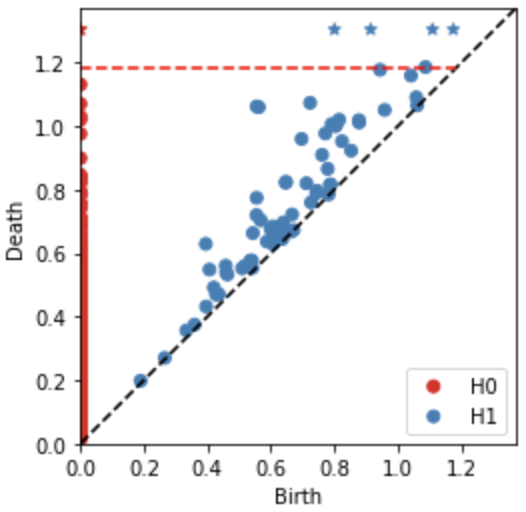}
         \includegraphics[width=0.21\textwidth]{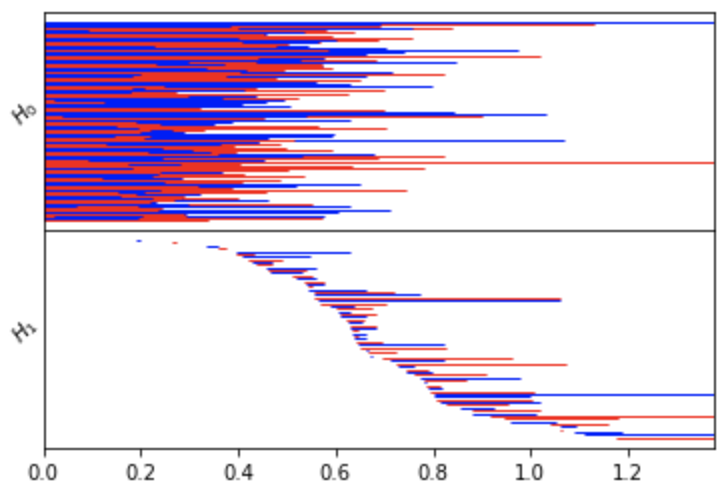}
         \includegraphics[width=0.14\textwidth]{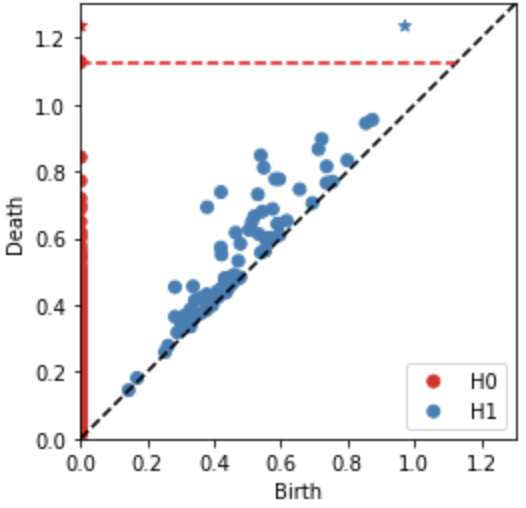}
         \includegraphics[width=0.21\textwidth]{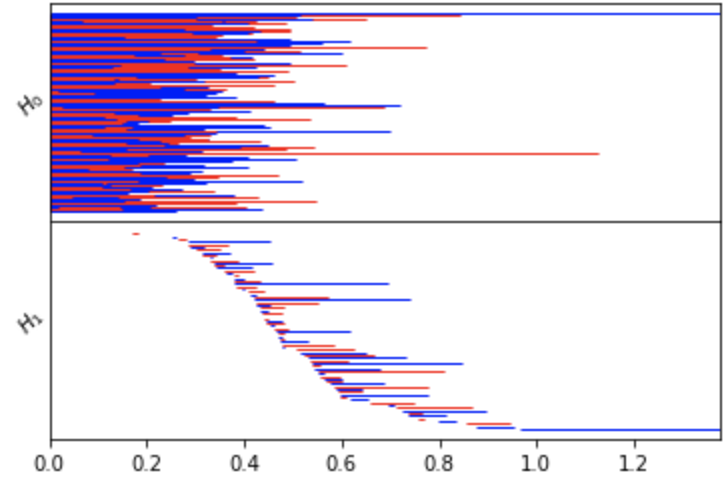}
         \caption{Epoch: $3000$; Loss: $0.089349$; Train Accuracy: $0.980$; Test Accuracy: $0.980$.}
    \end{subfigure}
    \begin{subfigure}[b]{\textwidth}
         \includegraphics[width=0.14\textwidth]{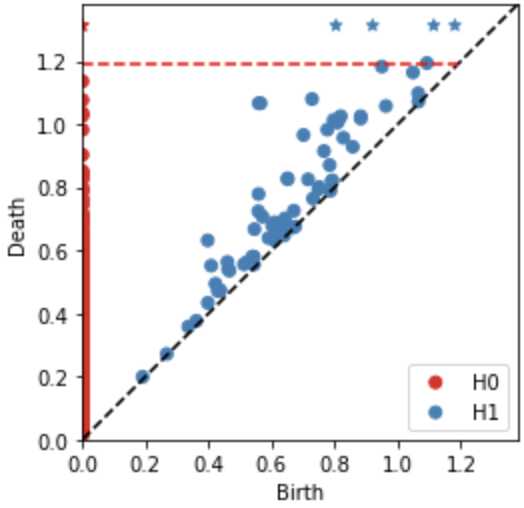}
         \includegraphics[width=0.21\textwidth]{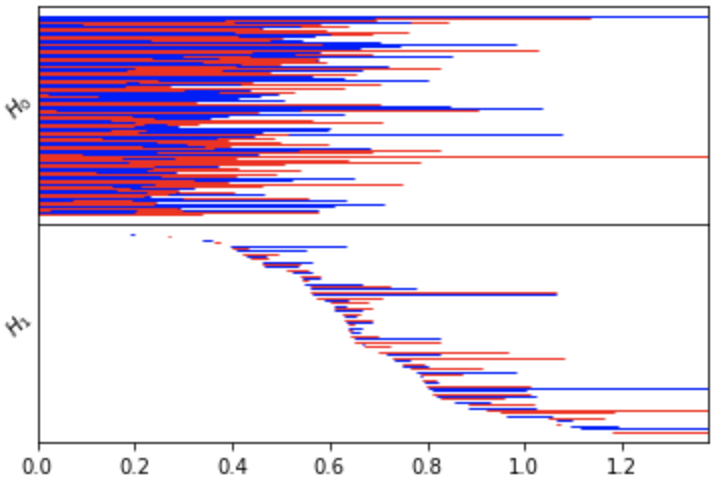}
         \includegraphics[width=0.14\textwidth]{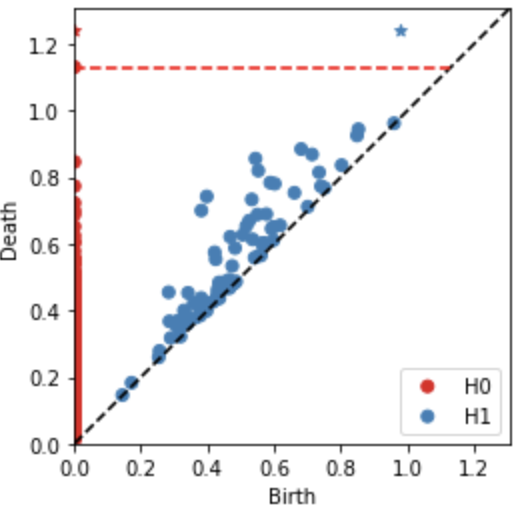}
         \includegraphics[width=0.21\textwidth]{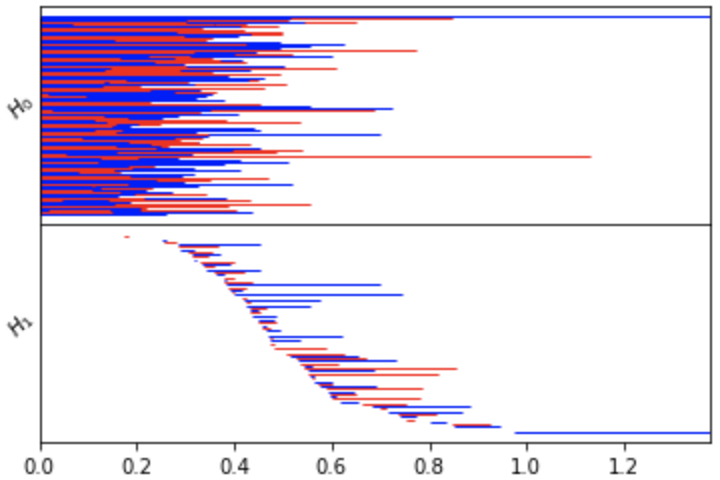}
         \caption{Epoch: $4000$; Loss: $0.072310$; Train Accuracy: $0.984$; Test Accuracy: $0.980$.}
    \end{subfigure}
    \begin{subfigure}[b]{\textwidth}
         \includegraphics[width=0.14\textwidth]{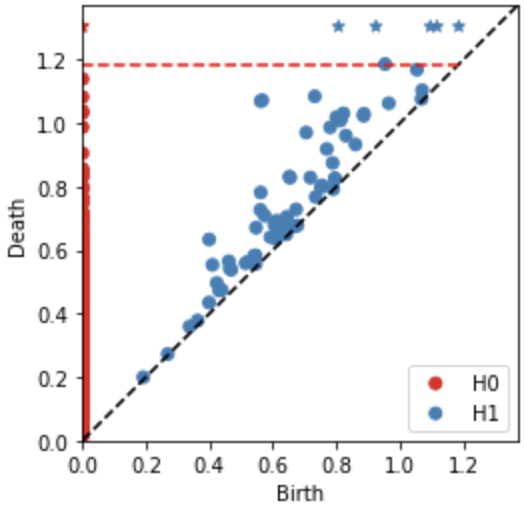}
         \includegraphics[width=0.21\textwidth]{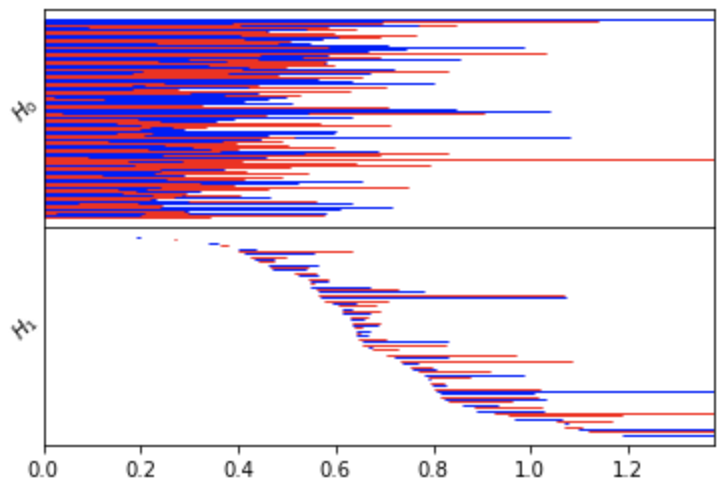}
         \includegraphics[width=0.14\textwidth]{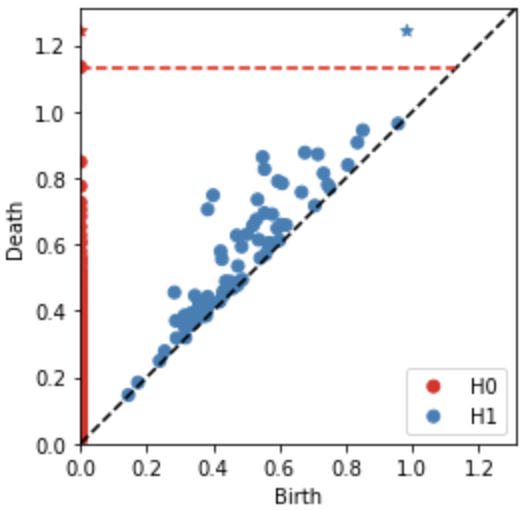}
         \includegraphics[width=0.21\textwidth]{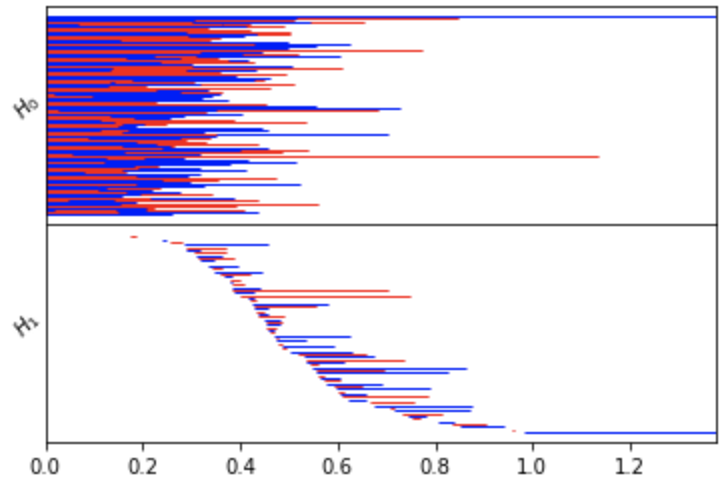}
         \caption{Epoch: $5000$; Loss: $0.060382$; Train Accuracy: $0.991$; Test Accuracy: $0.980$.}
    \end{subfigure}
    \begin{subfigure}[b]{\textwidth}
         \includegraphics[width=0.14\textwidth]{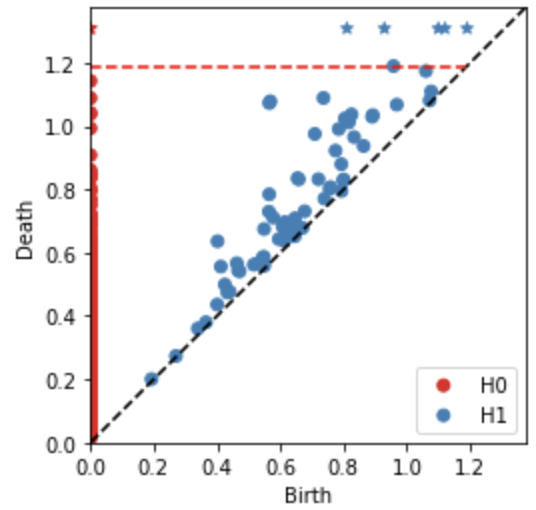}
         \includegraphics[width=0.21\textwidth]{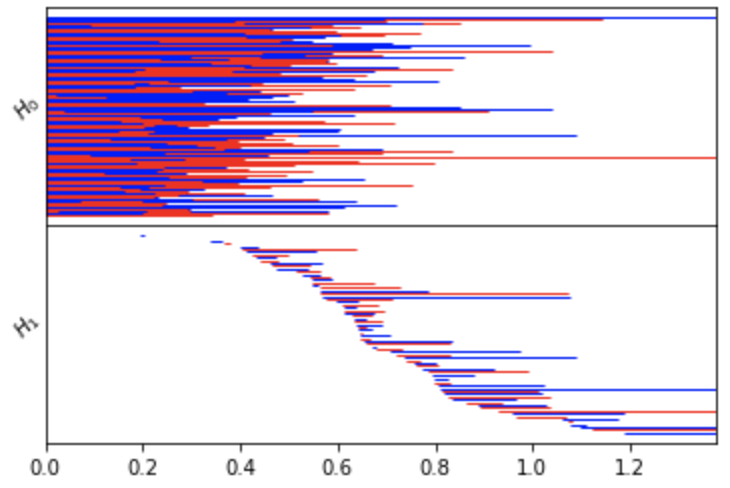}
         \includegraphics[width=0.14\textwidth]{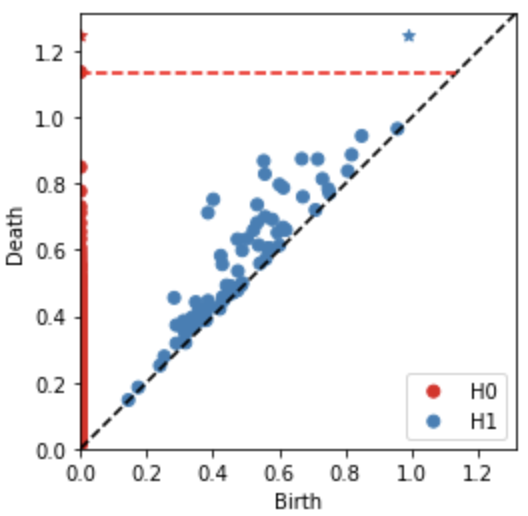}
         \includegraphics[width=0.21\textwidth]{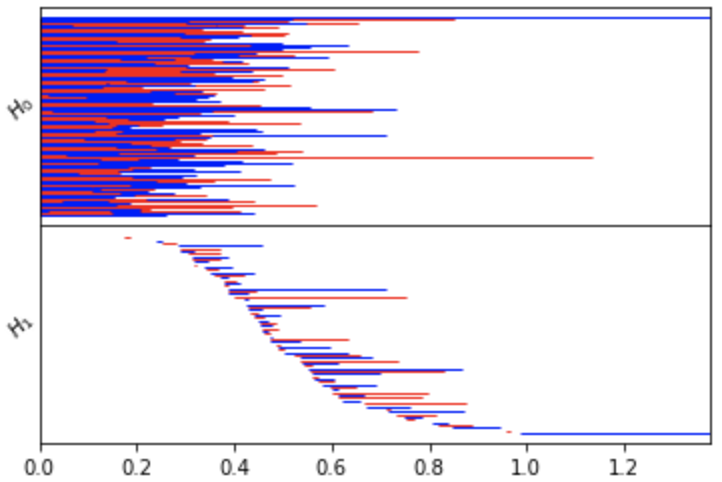}
         \caption{Epoch: $6000$; Loss: $0.051985$; Train Accuracy: $0.991$; Test Accuracy: $0.980$.}
    \end{subfigure}
    \begin{subfigure}[b]{\textwidth}
         \includegraphics[width=0.14\textwidth]{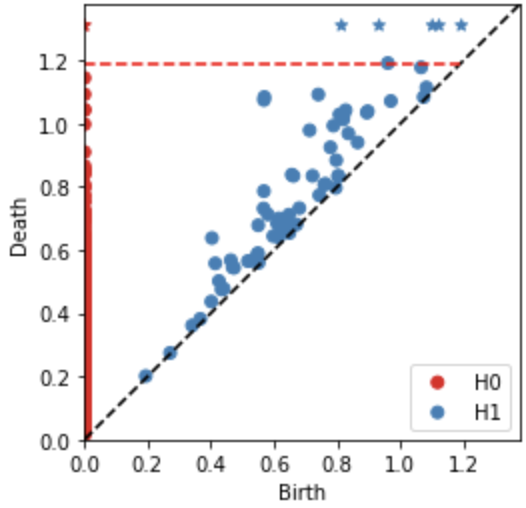}
         \includegraphics[width=0.21\textwidth]{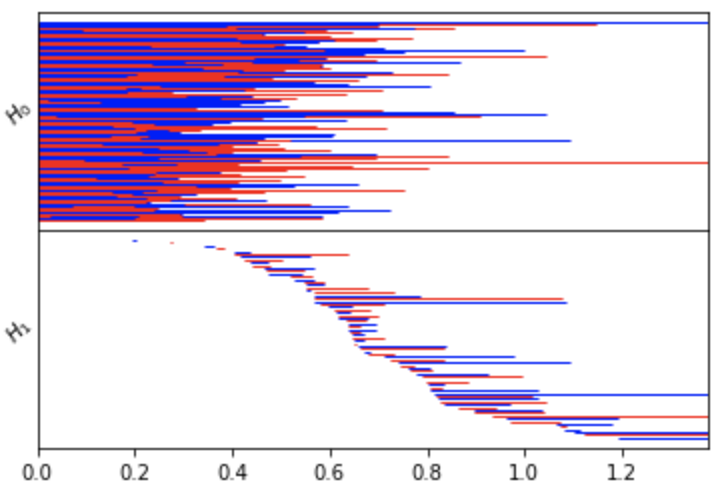}
         \includegraphics[width=0.14\textwidth]{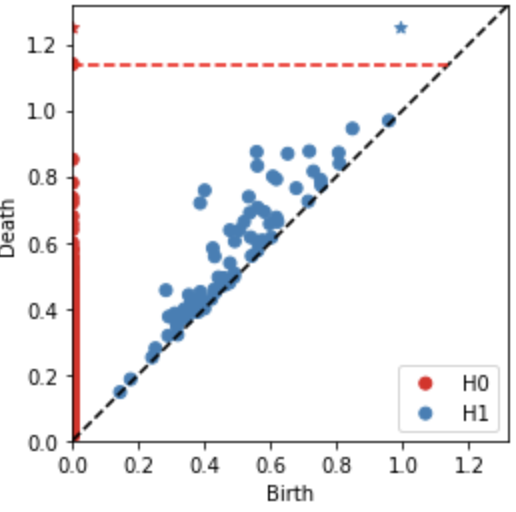}
         \includegraphics[width=0.21\textwidth]{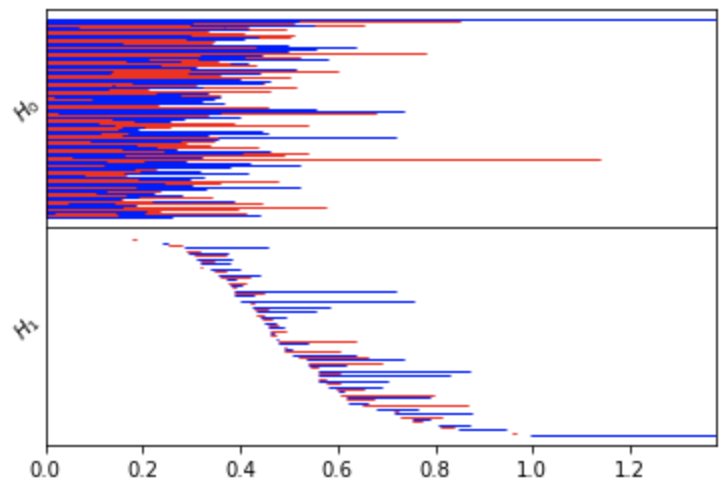}
         \caption{Epoch: $7000$; Loss: $0.045680$; Train Accuracy: $0.996$; Test Accuracy: $0.980$.}
    \end{subfigure}
    \begin{subfigure}[b]{\textwidth}
         \includegraphics[width=0.14\textwidth]{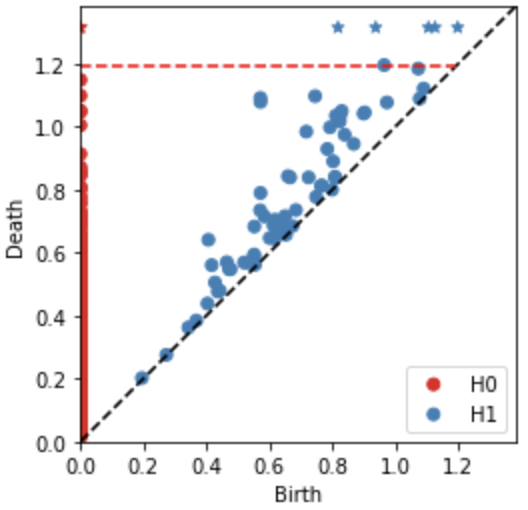}
         \includegraphics[width=0.21\textwidth]{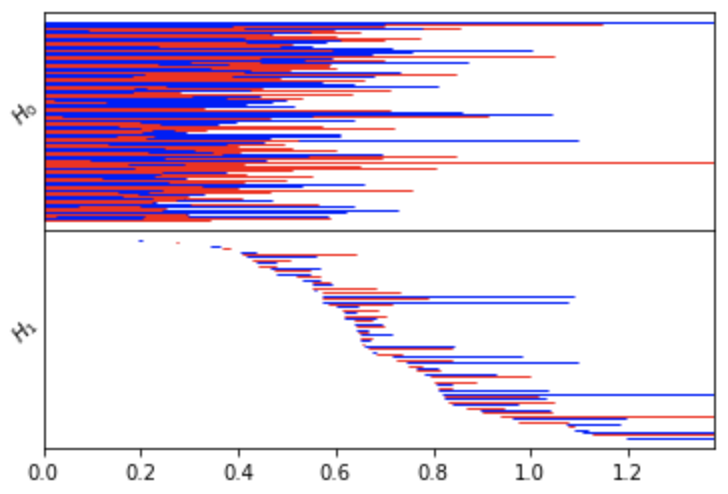}
         \includegraphics[width=0.14\textwidth]{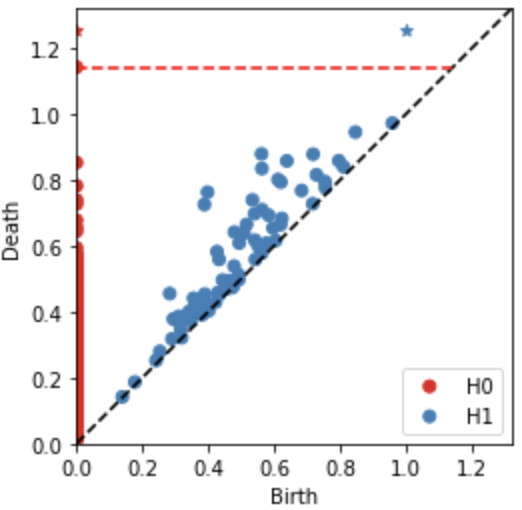}
         \includegraphics[width=0.21\textwidth]{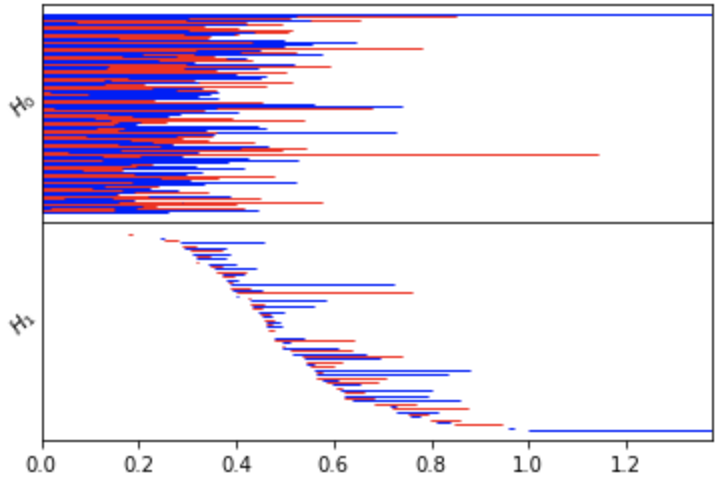}
         \caption{Epoch: $8000$; Loss: $0.040629$; Train Accuracy: $0.998$; Test Accuracy: $0.983$.}
    \end{subfigure} 
    \begin{subfigure}[b]{\textwidth}
         \includegraphics[width=0.14\textwidth]{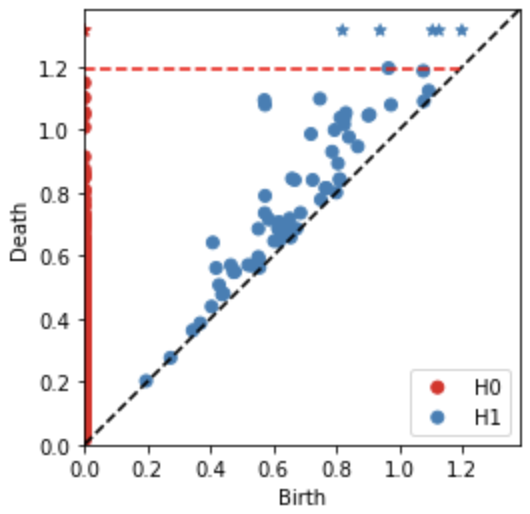}
         \includegraphics[width=0.21\textwidth]{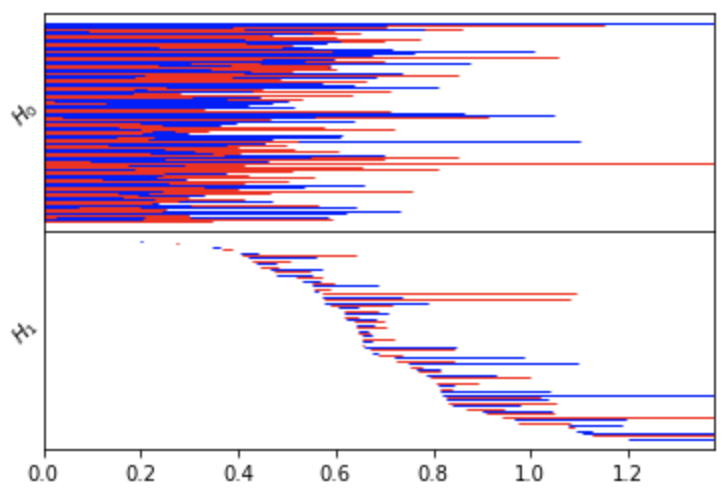}
         \includegraphics[width=0.14\textwidth]{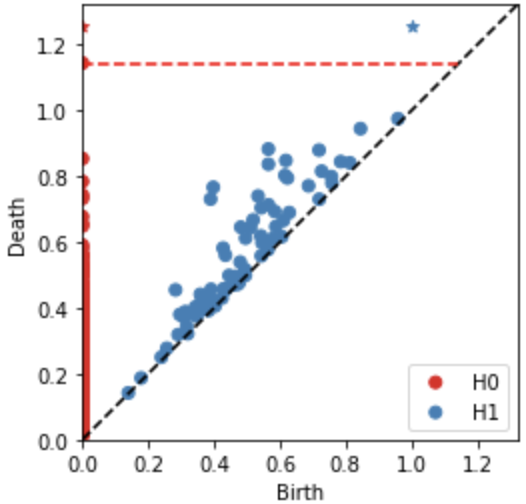}
         \includegraphics[width=0.21\textwidth]{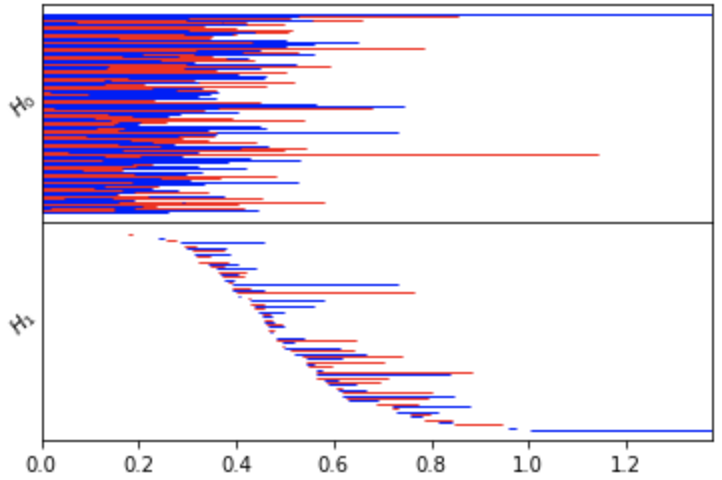}
         \caption{Epoch: $9000$; Loss: $0.036546$; Train Accuracy: $0.998$; Test Accuracy: $0.983$.}
    \end{subfigure} 
    \begin{subfigure}[b]{\textwidth}
         \includegraphics[width=0.14\textwidth]{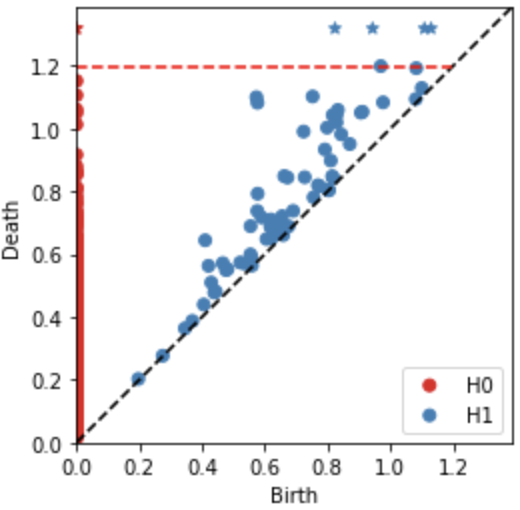}
         \includegraphics[width=0.21\textwidth]{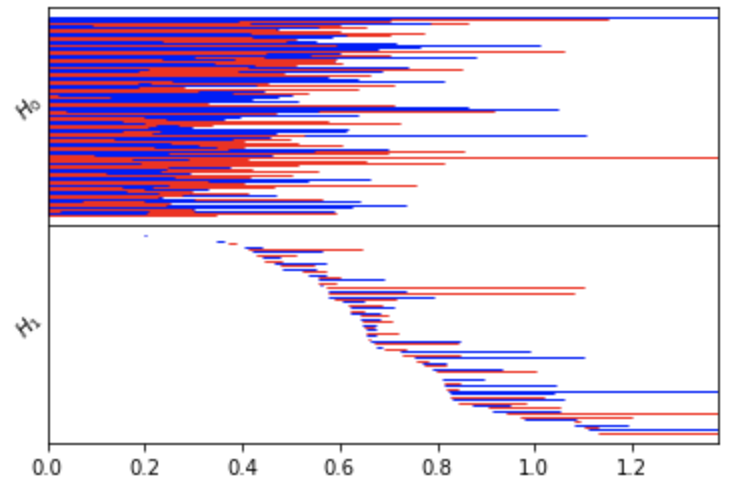}
         \includegraphics[width=0.14\textwidth]{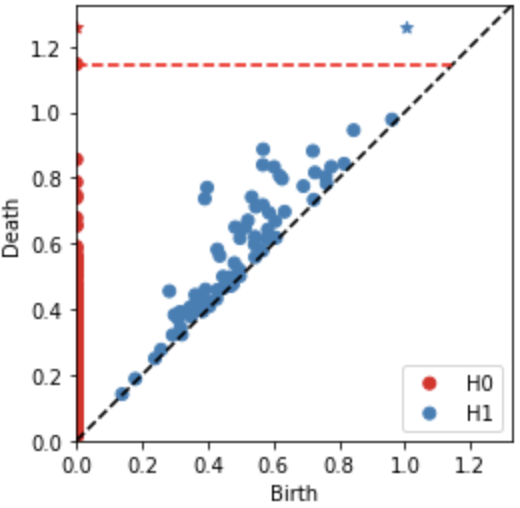}
         \includegraphics[width=0.21\textwidth]{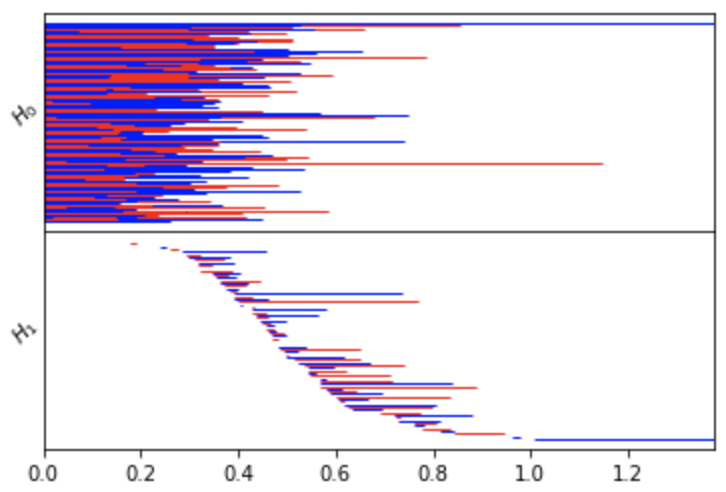}
         \caption{Epoch: $10000$; Loss: $0.033150$; Train Accuracy: $0.998$; Test Accuracy: $0.983$.}
    \end{subfigure}  
\caption{Persistence homology of forward layer before and after apply activation function.}
\label{fig:rips_barcode}
\end{figure}

A question is how is the connected component deciding what the next word should be? We built a two layers neural network with ``ReLU'', ``Sigmoid'', and ``tanh'' activation functions applied to each layer.  We used 
$$ \operatorname{accuracy}(y,\hat{y}) = \frac{1}{n_{samples}} \Sigma^{n_{samples}-1} 1(\hat{y}_i = y_i)$$ 
to evaluate the accuracy of our model. Once the train and test accuracy went higher than $0.9$, we computed Rips persistence diagram and persistence barcode of hidden layer per $1000$ epochs starting from $2000$ epochs. The results are in Figure \ref{fig:rips_barcode}.

In Figure \ref{fig:graph}, we have shown that text strings can be represented as a directed graph. A follow up question is how is the directed graph deciding what the next word should be? There would be a latent space including a set of possibilities. 
We will study this through path homology, isomorphism of directed graph, and graphical neural networks. We will build graph generative models through higher order interactions\cite{sun2023ptensors}.

\section*{Acknowledgement} 
The authors thank Lek-Heng Lim, who provided idea for this work.  The authors thank people who provided helpful discussions.  The authors thank the support of DARPA research.

\newpage
\bibliographystyle{plainurl}
\bibliography{citation}

\newpage
\section{Appendix}
\begin{table}[!htb]
\footnotesize
\centering
    \begin{tabular*}{\columnwidth}{@{\extracolsep{\fill}}cccc}
        \textbf{sentence label} & \textbf{one component} & \textbf{two components} & \textbf{three components}\\
        \hline
        1 & 0.85 & 0.91 & 0.95 \\
        \hline
        2 & 0.9 & 0.94 & 0.97 \\
        \hline
        3 & 0.76 & 0.85 & 0.93 \\ 
        \hline
        4 & 0.86 & 0.92 & 0.94 \\
         \hline 
        5 & 0.82 & 0.91 & 0.94 \\
         \hline 
        6 & 0.82 & 0.89 & 0.92 \\
         \hline 
        7 & 0.83 & 0.9 & 0.94 \\
                  \hline 
         8 & 0.8 & 0.88 & 0.93 \\
                  \hline 
         9 & 0.79 & 0.89 & 0.93 \\
                  \hline 
         10 & 0.71 & 0.9 & 0.94\\
                  \hline 
         11 & 0.85 & 0.92 & 0.94 \\
                  \hline 
         12 & 0.78 & 0.88 & 0.92 \\
                  \hline 
         13 & 0.73 & 0.86 & 0.91 \\
                  \hline 
         14 & 0.71 & 0.81 & 0.91 \\
                  \hline 
         15 & 0.79 & 0.87 & 0.9 \\
                  \hline 
         16 & 0.83 & 0.9 & 0.93 \\
                  \hline 
         17 & 0.84 & 0.9 & 0.93 \\
                  \hline 
         18 & 0.75 & 0.89 & 0.93 \\
                  \hline 
         19 & 0.78 & 0.92 & 0.97 \\
                  \hline 
         20 & 0.83 & 0.92 & 0.95 \\
                  \hline 
         21 & 0.74 & 0.93 & 0.99 \\
                  \hline 
         22 & 0.78 & 0.86 & 0.9 \\
                  \hline 
         23 & 0.86 & 0.91 & 0.95 \\
                  \hline 
         24 & 0.79 & 0.93 & 0.96 \\
                  \hline 
         25 & 0.75 & 0.91 & 0.95 \\
                  \hline 
         26 & 0.88 & 0.93 & 0.96 \\
                  \hline 
         27 & 0.86 & 0.94 & 0.97 \\
                  \hline 
         28 & 0.76 & 0.89 & 0.94 \\
         \hline 
         29 & 0.87 & 0.92 & 0.95 \\
                  \hline 
         30 & 0.79 & 0.91 & 0.96 \\
                  \hline 
         31 & 0.77 & 0.87 & 0.94 \\
                  \hline 
         32 & 0.76 & 0.88 & 0.91 \\
                  \hline 
         33 & 0.9  & 0.94 & 0.96 \\
                  \hline 
         34 & 0.79 & 0.88 & 0.93 \\
                  \hline 
         35 & 0.77 & 0.9  & 0.93 \\
                  \hline 
         36 & 0.76 & 0.84 & 0.9 \\
                  \hline 
         37 & 0.73 & 0.82 & 0.91 \\
                  \hline 
         38 & 0.79 & 0.86 & 0.91 \\
                  \hline 
         39 & 0.75 & 0.84 & 0.9 \\
                  \hline 
         40 & 0.87 & 0.93 & 0.95 \\
                  \hline 
         41 & 0.81 & 0.89 & 0.92 \\
                  \hline 
         42 & 0.7 & 0.89 & 0.93 \\
                  \hline 
         43 & 0.67 & 0.86 & 0.91 \\
                  \hline 
         44 & 0.78 & 0.87 & 0.91 \\
                  \hline 
         45 & 0.82 & 0.89 & 0.93 \\
                  \hline 
         46 & 0.77 & 0.89 & 0.93 \\
                  \hline 
         47 & 0.87 & 0.94 & 0.98 \\
                  \hline 
         48 & 0.68 & 0.83 & 0.89 \\
                  \hline 
         49 & 0.78 & 0.86 & 0.9 \\
                  \hline 
         50 & 0.8  & 0.88 & 0.92 \\
         \end{tabular*}
        \caption{Variances captured by doing PCA with one, two, and three components on sentence vectors.}
        \label{tab:pca_vc1}
\end{table}
\clearpage
\begin{table}[!htb]
\footnotesize
\centering
    \begin{tabular*}{\columnwidth}{@{\extracolsep{\fill}}cccc}
        \textbf{sentence label} & \textbf{one component} & \textbf{two components} & \textbf{three components} \\
        \hline
         51 & 0.8 & 0.88 & 0.91 \\
                  \hline 
         52 & 0.8 & 0.92 & 0.95 \\
                  \hline 
         53 & 0.68 & 0.82 & 0.87 \\
                  \hline 
         54 & 0.74 & 0.87 & 0.92 \\
                  \hline 
         55 & 0.86 & 0.94 & 0.97 \\
                  \hline 
         56 & 0.68 & 0.83 & 0.92 \\
                  \hline 
         57 & 0.77 & 0.85 & 0.89\\
                  \hline 
         58 & 0.66 & 0.84 & 0.92 \\
                  \hline 
         59 & 0.79 & 0.91 & 0.95 \\
                  \hline 
         60 & 0.79 & 0.87 & 0.91 \\
                  \hline 
         61 & 0.86 & 0.92 & 0.96 \\
                  \hline 
         62 & 0.93 & 0.97 & 0.99\\
                  \hline 
         63 & 0.87 & 0.92 & 0.95 \\
                  \hline 
         64 & 0.82 & 0.91 & 0.96 \\
                  \hline 
         65 & 0.86 & 0.91 & 0.95 \\
                  \hline 
         66 & 0.74 & 0.85 & 0.89 \\
                  \hline 
         67 & 0.75 & 0.85 & 0.9 \\
                  \hline 
         68 & 0.79 & 0.88 & 0.91 \\
                  \hline 
         69 & 0.78 & 0.86 & 0.92 \\         
                  \hline 
         70 & 0.88 & 0.93 & 0.95 \\
                  \hline 
         71 & 0.87 & 0.93 & 0.96 \\       
                  \hline 
         72 & 0.84 & 0.9 & 0.95 \\         
                  \hline 
         73 & 0.75 & 0.83 & 0.91\\         
                  \hline 
         74 & 0.87 & 0.94 & 0.97 \\         
                  \hline 
         75 & 0.85 & 0.92 & 0.94 \\
                  \hline 
         76 & 0.87 & 0.92 & 0.96 \\         
                  \hline 
         77 & 0.78 & 0.86 & 0.93 \\         
                  \hline 
         78 & 0.71 & 0.82 & 0.89 \\        
                  \hline 
         79 & 0.87 & 0.94 & 0.97 \\         
                  \hline 
         80 & 0.79 & 0.89 & 0.93 \\         
                  \hline 
         81 & 0.81 & 0.88 & 0.91 \\         
                  \hline 
         82 & 0.8 & 0.92 & 0.96 \\         
                  \hline 
         83 & 0.87 & 0.97 & 0.99 \\        
                  \hline 
         84 & 0.79 & 0.88 & 0.94 \\         
                  \hline 
         85 & 0.84 & 0.94 & 0.97 \\         
                  \hline 
         86 & 0.78 & 0.88 & 0.94 \\         
                  \hline 
         87 & 0.68 & 0.86 & 0.9\\         
                  \hline 
         88 & 0.87 & 0.94 & 0.96 \\        
                  \hline 
         89 & 0.8 & 0.91 & 0.94 \\        
                  \hline 
         90 & 0.78 & 0.89 & 0.93 \\        
                  \hline 
         91 & 0.77 & 0.87 & 0.94 \\         
                   \hline 
         92 & 0.83 & 0.91 & 0.93 \\        
                  \hline 
         93 & 0.75 & 0.88 & 0.95 \\         
                  \hline 
         94 & 0.77 & 0.86 & 0.91 \\          
                  \hline 
         95 & 0.74 & 0.84 & 0.92 \\          
                  \hline 
         96 & 0.76 & 0.86 & 0.92 \\   
                  \hline 
         97 & 0.85 & 0.93 & 0.95 \\   
                  \hline 
         98 & 0.71 & 0.86 & 0.92 \\            
                  \hline 
         99 & 0.75 & 0.86 & 0.91 \\   
                  \hline 
         100 & 0.88 & 0.94 & 0.96 \\   
         \end{tabular*}
        \caption{Variances captured by doing PCA with one, two, and three components on sentence vectors. }
        \label{tab:pca_vc2}
\end{table}

\clearpage
\begin{figure}
    \begin{subfigure}[b]{0.49\textwidth}
         \includegraphics[width=\textwidth]{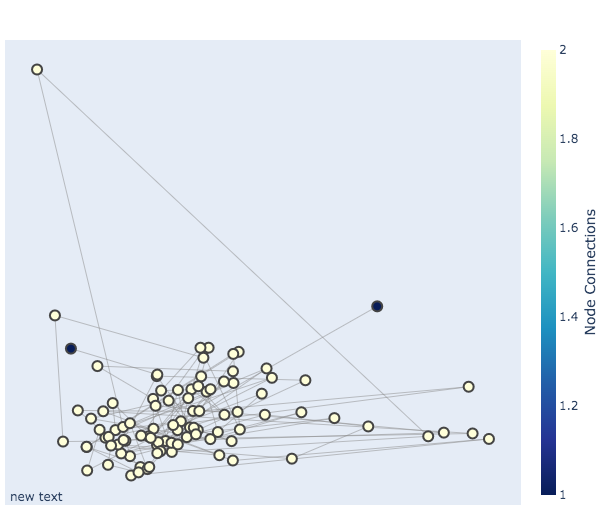}
         \includegraphics[width=\textwidth]{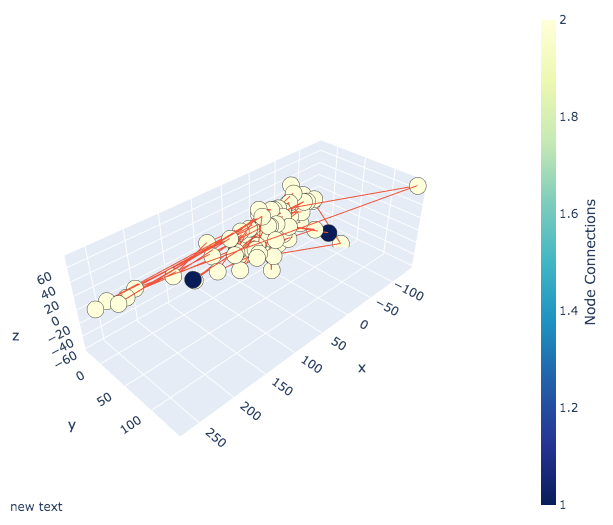}
         \includegraphics[width=\textwidth]{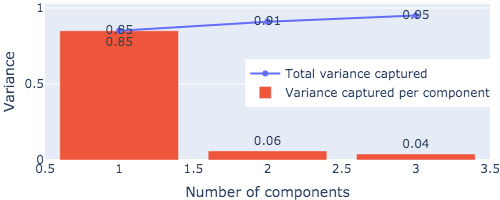}
         \subcaption{``victory: but where does it end? It's been about the last five days. On the last day of October I thought there were going to be a lot of announcements from the game makers over at Eurogamer. And the fact that there will only be one announcement at the close was great but was also the beginning of a really long day of speculation and discussion about how it's going to end and what's to be done about it and that we'd need a lot of help.''}
         \label{fig:1}
    \end{subfigure}
    \begin{subfigure}[b]{0.49\textwidth}
         \includegraphics[width=\textwidth]{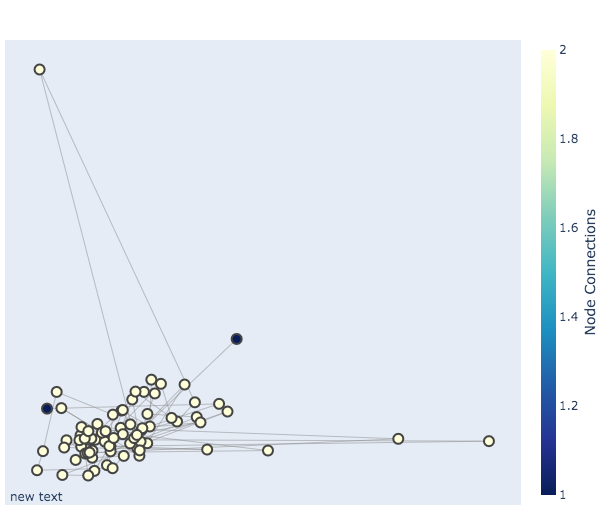}
         \includegraphics[width=\textwidth]{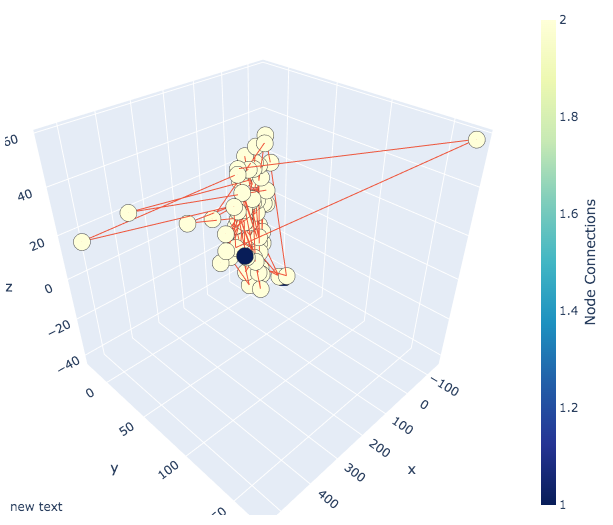}
         \includegraphics[width=\textwidth]{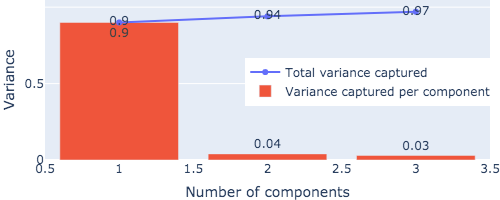}
         \subcaption{``The main reason there aren't any DLC for Dragon Age 2 so far is the studio has a ton of different things to do from what was expected of them to DLC. We also have the biggest development and development team to help us out. From game designer to writer they have a lot of ideas about how to work around some of the more complicated DLCs in the game.''}
         \label{fig:2}
    \end{subfigure}
\end{figure}
\begin{figure}\ContinuedFloat
    \begin{subfigure}[b]{0.49\textwidth}
         \includegraphics[width=\textwidth]{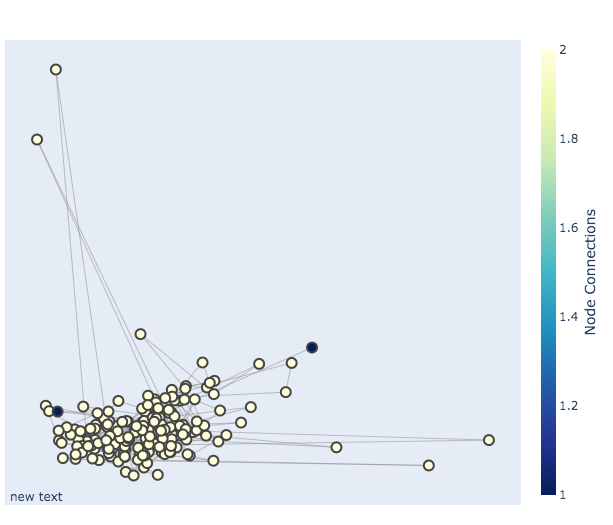}
         \includegraphics[width=\textwidth]{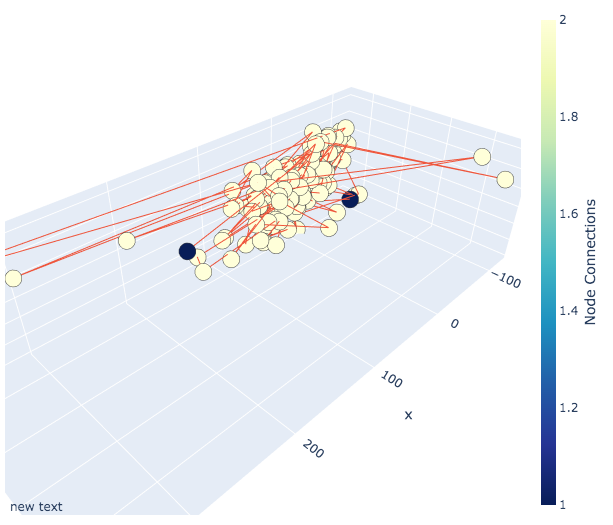}
         \includegraphics[width=\textwidth]{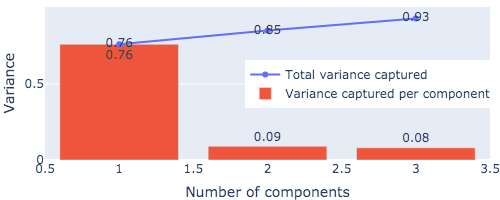}
         \subcaption{``They've had some really interesting discussions with us in the studio including talking about whether to bring DLC like Death Knight: Dragon Age 2: or even Shadow of Mordor through as they're just doing the DLC which you see in the trailer: whether or not to do them in the alpha then the first step is looking into the details for which DLC will be included which is going to be a huge priority. For those of you who don't know: what the game is is essentially a turn-based tactical combat game where you are the commander of an army and you fight against another player and there is the whole premise of 'this is why we're here'. And then each turn you lose: you are the one who loses because you lost and: therefore not necessarily in the game: but where you are able to find things to do. That is where their plans will fall.''}
         \label{fig:3}
    \end{subfigure}
    \begin{subfigure}[b]{0.49\textwidth}
         \includegraphics[width=\textwidth]{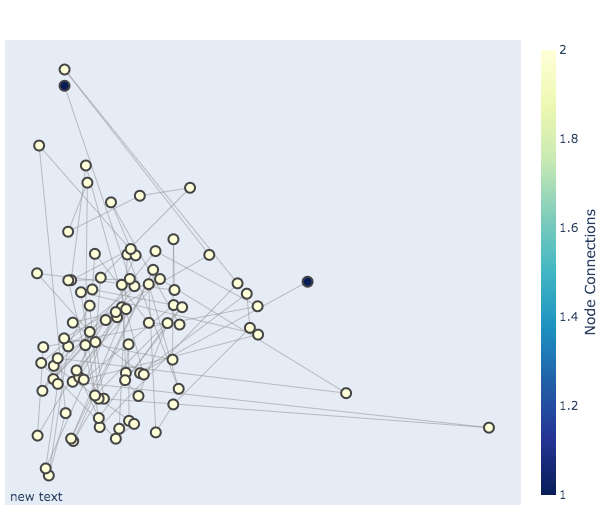}
         \includegraphics[width=\textwidth]{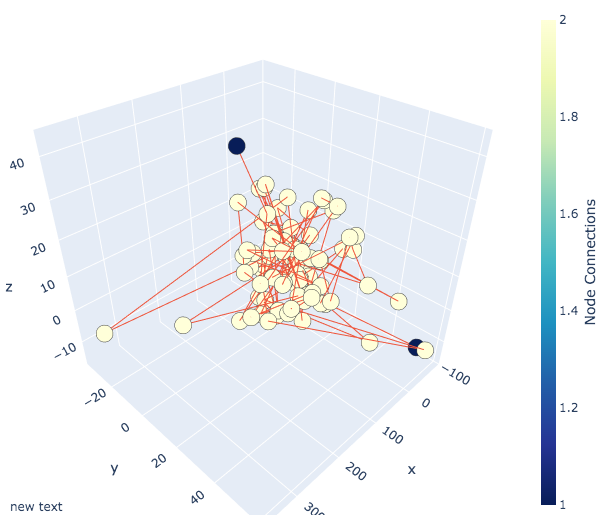}
         \includegraphics[width=\textwidth]{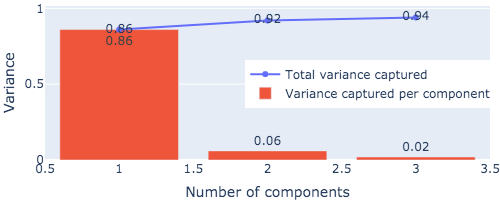}
         \caption{``And then there will be more development with us and even their own development team. They'll get involved but it will be in the form of the first-person shooter game which is something we are excited about and how it will continue to grow as this game is developed further into the franchise. There is a lot more on the creative side of the game: but that is something we are really excited about because it opens up more opportunities for future titles.''}
         \label{fig:4}
    \end{subfigure}
\end{figure}
\begin{figure}\ContinuedFloat
        \begin{subfigure}[b]{0.49\textwidth}
         \includegraphics[width=\textwidth]{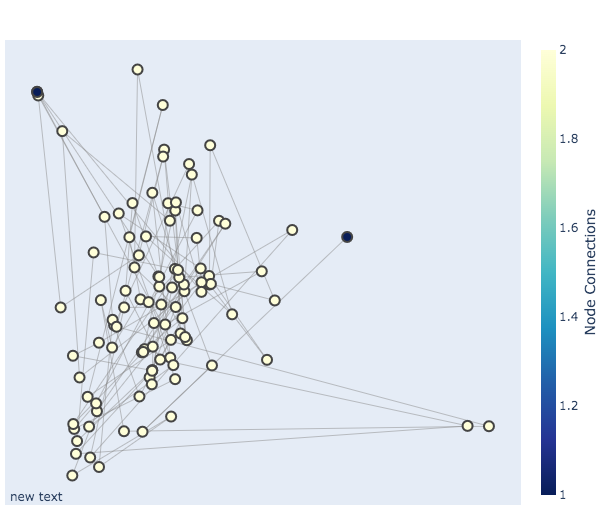}
         \includegraphics[width=\textwidth]{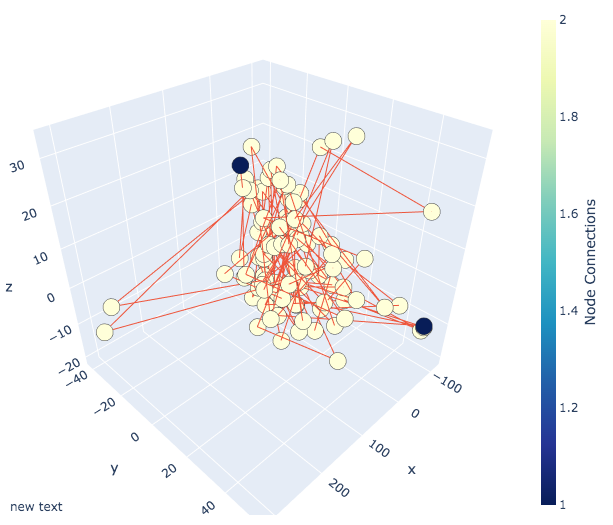}
         \includegraphics[width=\textwidth]{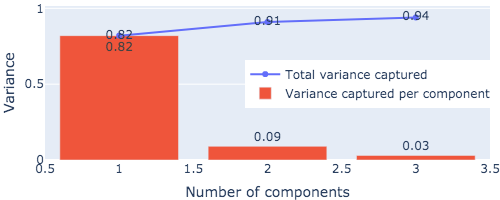}      
         \caption{``It also has a story that we are going to tell in a sort of fantasy world. You play as the Inquisitor of the Inquisition: and you know your family and this is how a little bit about the family and how you deal with various evil people. It will be set in a fantasy world where you have no control over them but you have to fight them. And the Inquisitor is going to get to the first man to rescue them from those evil gods and to use their powers to defeat these evil gods.''}
         \label{fig:5}
    \end{subfigure}
        \begin{subfigure}[b]{0.49\textwidth}
         \includegraphics[width=\textwidth]{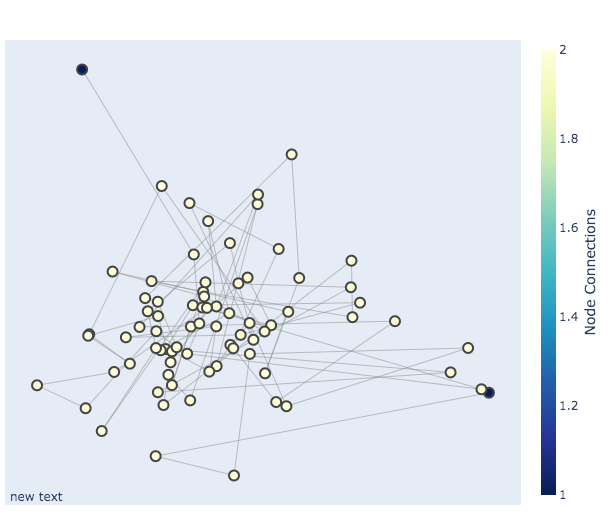}
         \includegraphics[width=\textwidth]{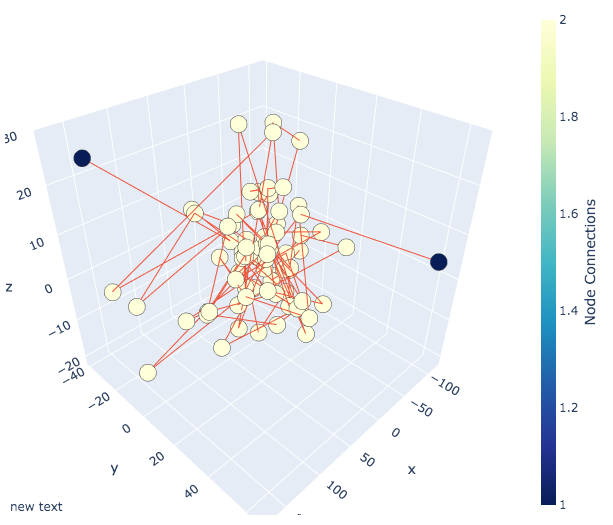}
         \includegraphics[width=\textwidth]{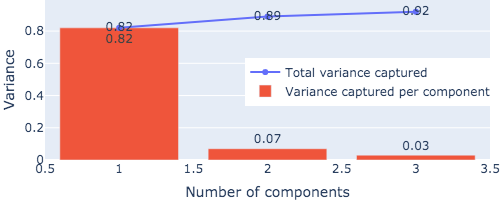}
         \caption{``So what that is going to be is a story that is very story based: and there are lots of things going on but with some really amazing writing: the story really revolves around the Inquisition: the Church: and how they go about their missions through a series of events which you can imagine would become the 'world building missions' coming into play and that is exactly what it is.''}
         \label{fig:6}
    \end{subfigure}
\end{figure}
\begin{figure}\ContinuedFloat
    \begin{subfigure}[b]{0.49\textwidth}
         \includegraphics[width=\textwidth]{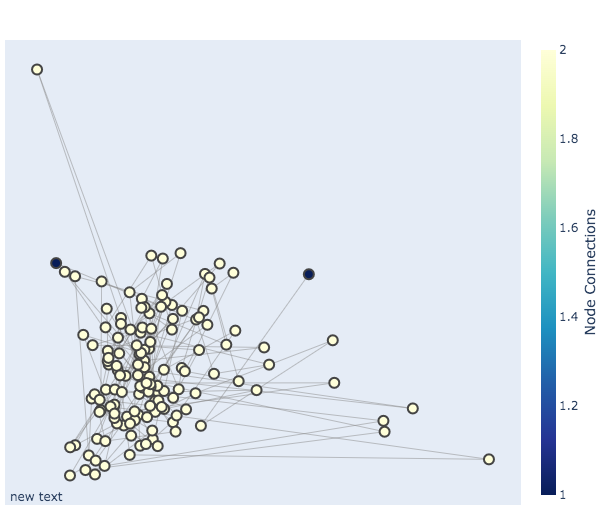}
         \includegraphics[width=\textwidth]{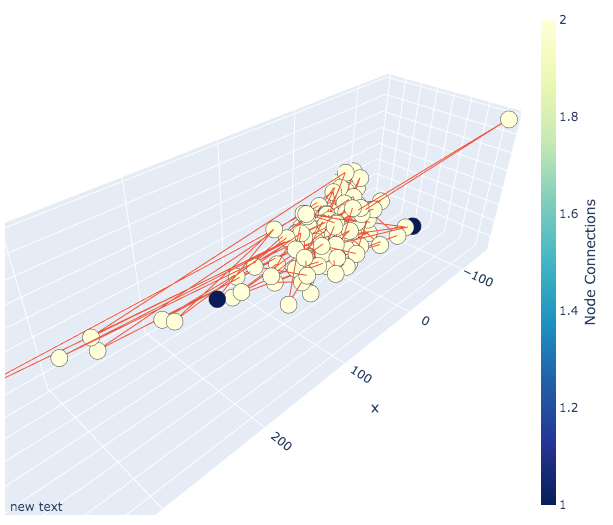}
         \includegraphics[width=\textwidth]{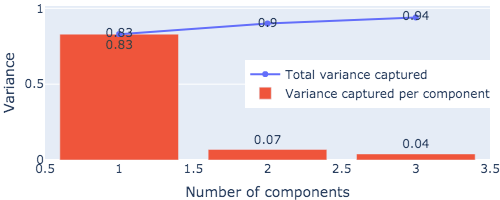}
         \caption{``At the end of the day: it's an exploration game but it is much deeper. And it is not about being the 'game in progress' because you are not going to be the Inquisitor: or the Bishop or the 'bad guy' because the Inquisitor is going to be there and he has to do that for them. You're fighting against demons: battling mages: that sort of thing: the Inquisition is a completely different kind of game: it's just a more unique way of trying to build a larger group and to create new forces to fight alongside that. It will be like an MMORPG with a world that has to be built around the Inquisition as well as some different rules of the game.''}
         \label{fig:7}
    \end{subfigure}
        \begin{subfigure}[b]{0.49\textwidth}
         \includegraphics[width=\textwidth]{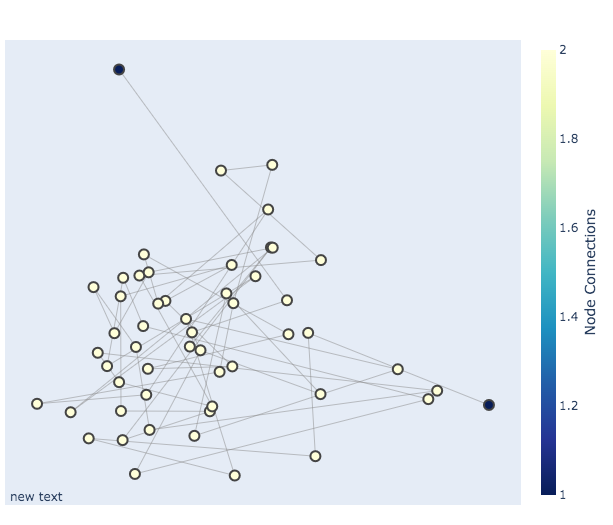}
         \includegraphics[width=\textwidth]{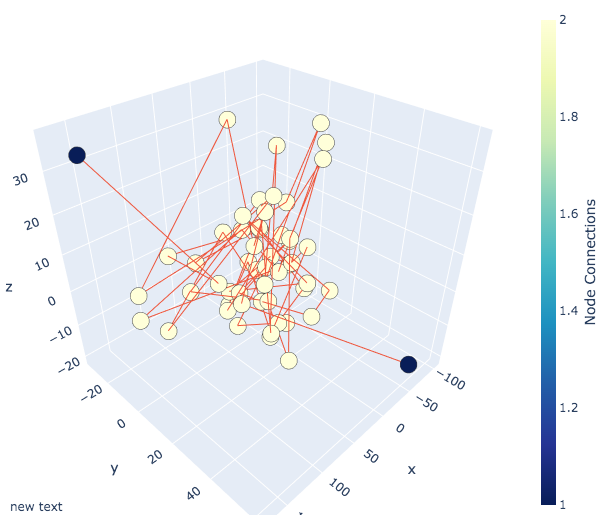}
         \includegraphics[width=\textwidth]{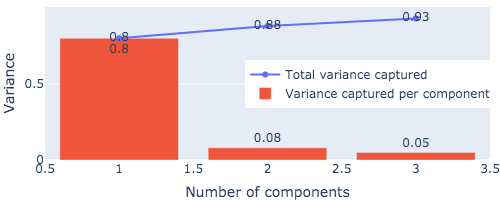}
         \caption{``But of course we have a lot of ideas in the works and it's all great: it's such a good story and we love playing Dragon Age: and it's definitely going to be a very fun game to play and a great place to put the world.''}
         \label{fig:8}
    \end{subfigure}
\end{figure}
\begin{figure}\ContinuedFloat
        \begin{subfigure}[b]{0.49\textwidth}
         \includegraphics[width=\textwidth]{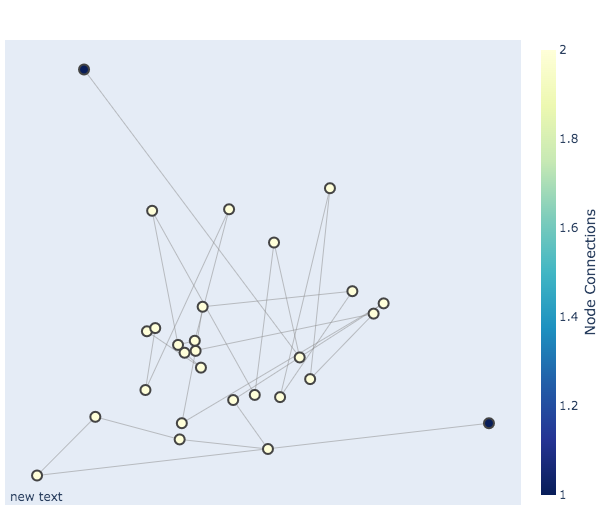}
         \includegraphics[width=\textwidth]{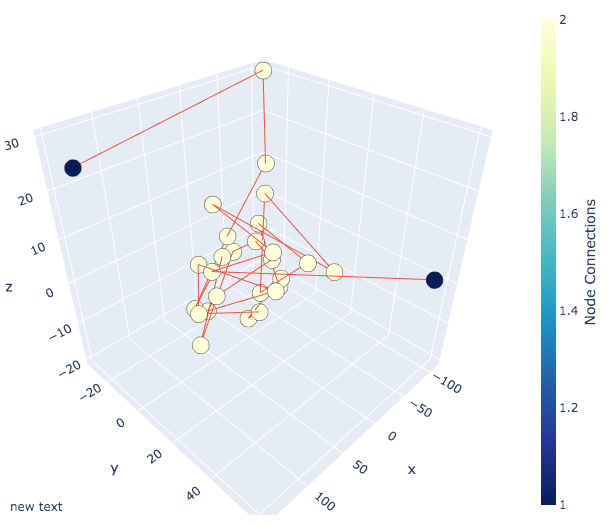}
         \includegraphics[width=\textwidth]{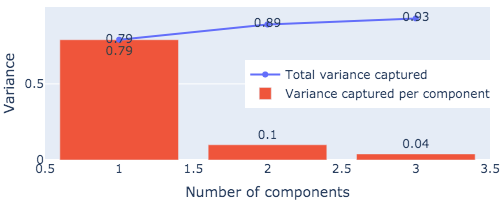}
         \caption{``When I've asked about the role and its possible role in Dragons of Skyrim what's the most anticipated moment that the game takes place in?''}
         \label{fig:9}
    \end{subfigure}
        \begin{subfigure}[b]{0.49\textwidth}
         \includegraphics[width=\textwidth]{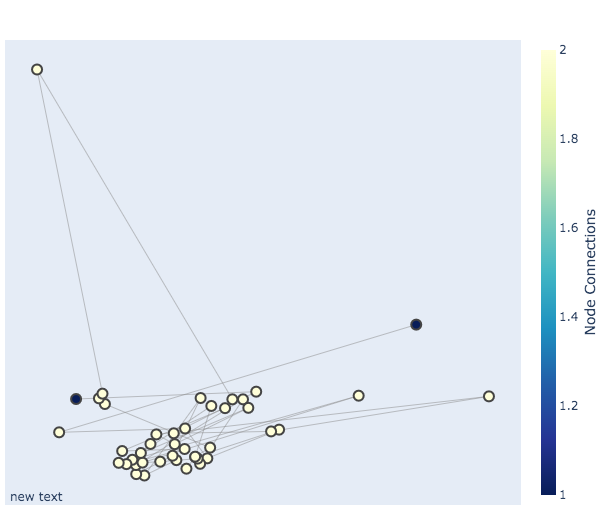}
         \includegraphics[width=\textwidth]{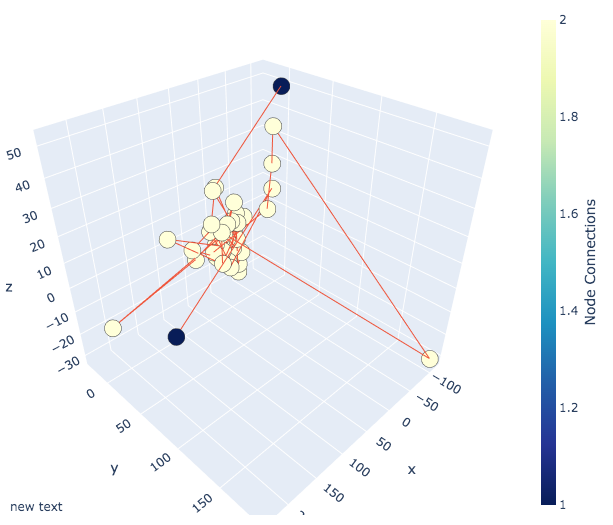}
         \includegraphics[width=\textwidth]{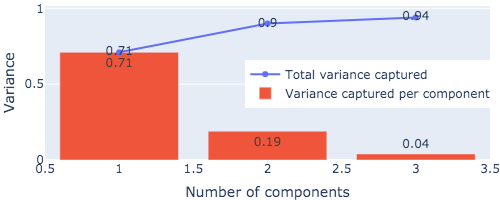}
         \caption{``So I asked if there's something that's coming that's very new: something that's going to be very different than what fans are currently used to watching but it's very different than the rest of the game.''}
         \label{fig:10}
    \end{subfigure}
\end{figure}
\begin{figure}\ContinuedFloat
    \begin{subfigure}[b]{0.49\textwidth}
         \includegraphics[width=\textwidth]{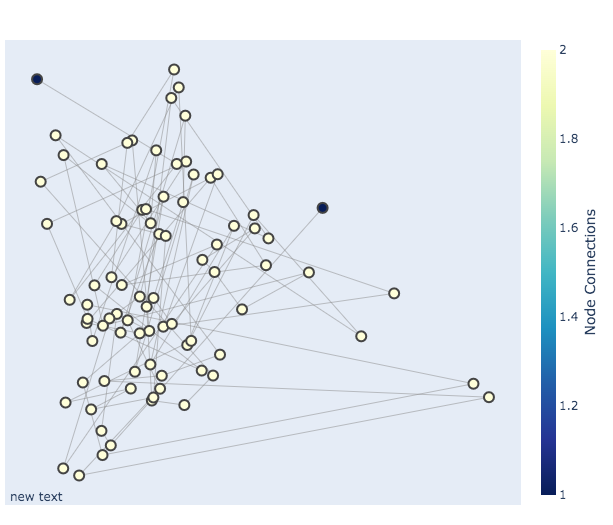}
         \includegraphics[width=\textwidth]{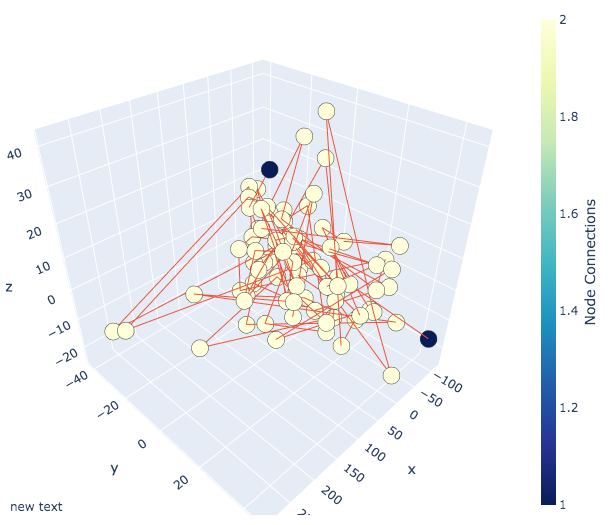}
         \includegraphics[width=\textwidth]{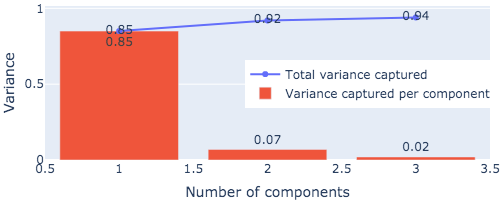}
         \caption{``It's not a traditional RPG. It's not an action RPG. The story is not trying to just take you out into the world and fight off demons and demons and demons and so on. It's not trying to make this 'just a game': but rather a fantasy world based around dragons: dragons being dragon: dragons being dragons: dragons being dragons and to me that's what really made Dragon Age so well.''}
         \label{fig:11}
    \end{subfigure}
        \begin{subfigure}[b]{0.49\textwidth}
         \includegraphics[width=\textwidth]{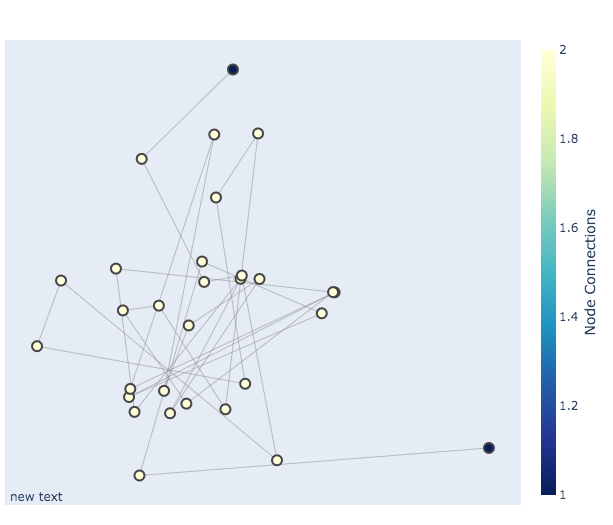}
         \includegraphics[width=\textwidth]{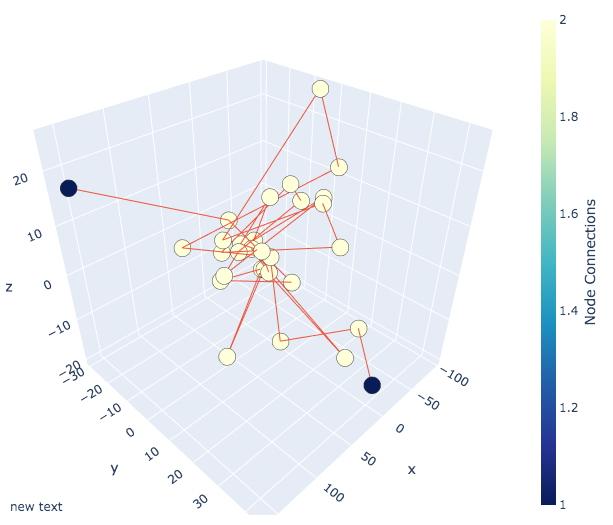}
         \includegraphics[width=\textwidth]{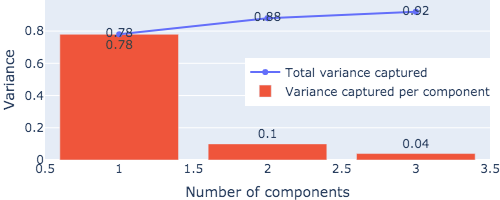}
         \caption{``It's so easy and there are so many different styles and ideas that were going to be introduced in Dragon Age when we were making it: and"''}
         \label{fig:12}
    \end{subfigure}
\end{figure}
\begin{figure}\ContinuedFloat
        \begin{subfigure}[b]{0.49\textwidth}
         \includegraphics[width=\textwidth]{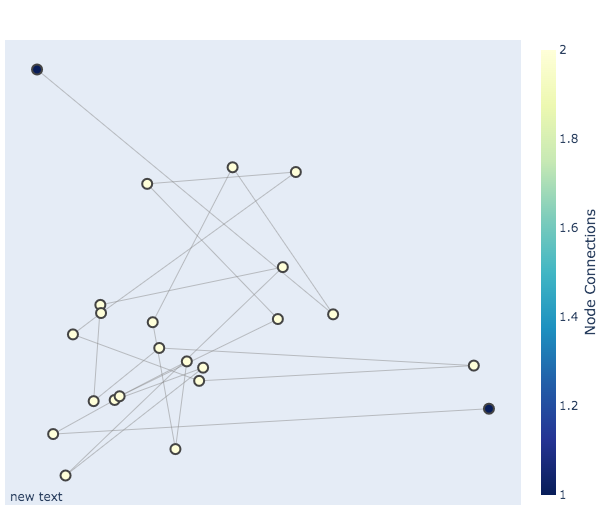}
         \includegraphics[width=\textwidth]{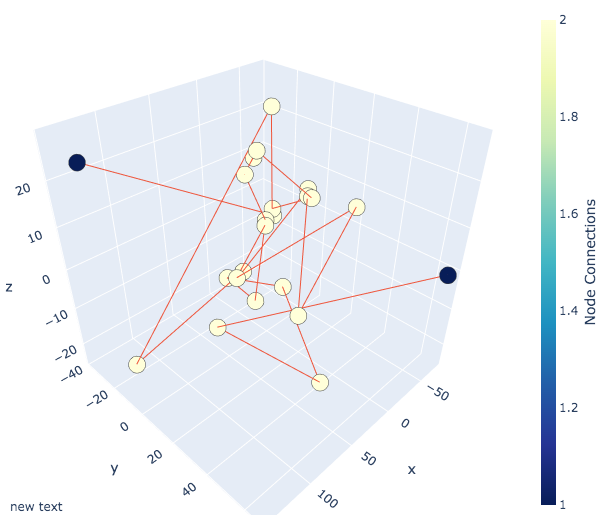}
         \includegraphics[width=\textwidth]{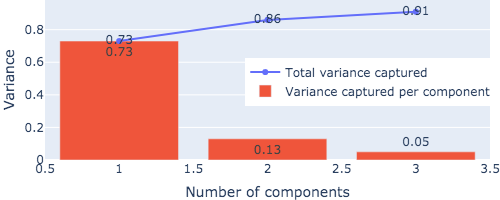}
         \caption{``There are many theories that Donald Trump is a product of his own unprofessional behavior or may even be ""crazy.""''}
         \label{fig:13}
    \end{subfigure}
        \begin{subfigure}[b]{0.49\textwidth}
         \includegraphics[width=\textwidth]{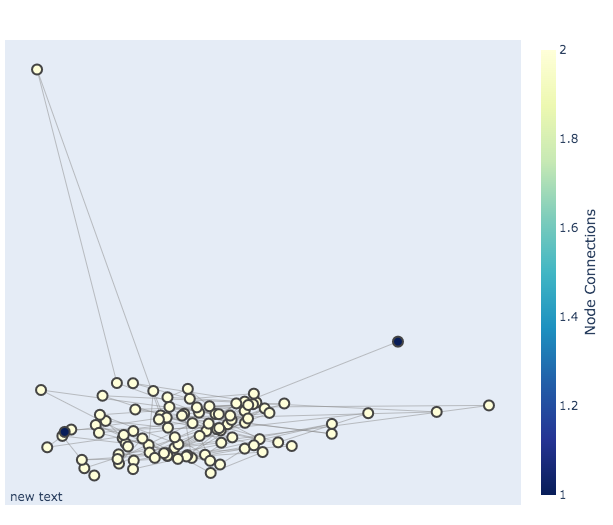}
         \includegraphics[width=\textwidth]{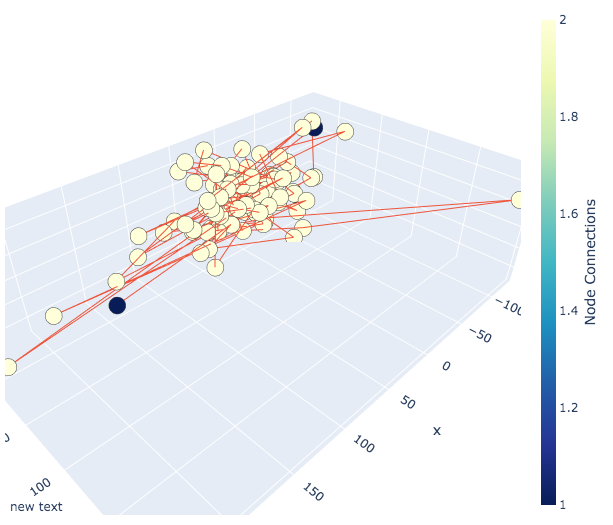}
         \includegraphics[width=\textwidth]{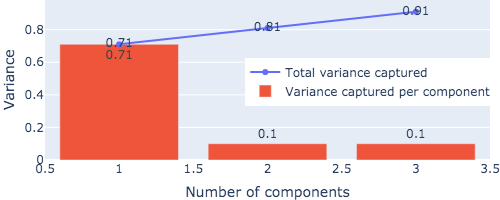}
         \caption{``The last thing you hear about Trump is that he doesn't like women. Trump tweeted that he was ""in fact a feminist"" before taking over the Republican presidential nomination. As for women in a Trump-controlled environment: that may be too much of a stretch: and we have heard no evidence whatsoever about a Trump-induced change in perception of women. But there is evidence: and there is reason to believe: that women could possibly see something real about Trump when he speaks these words.''}
         \label{fig:14}
    \end{subfigure}
\end{figure}
\begin{figure}\ContinuedFloat
        \begin{subfigure}[b]{0.49\textwidth}
         \includegraphics[width=\textwidth]{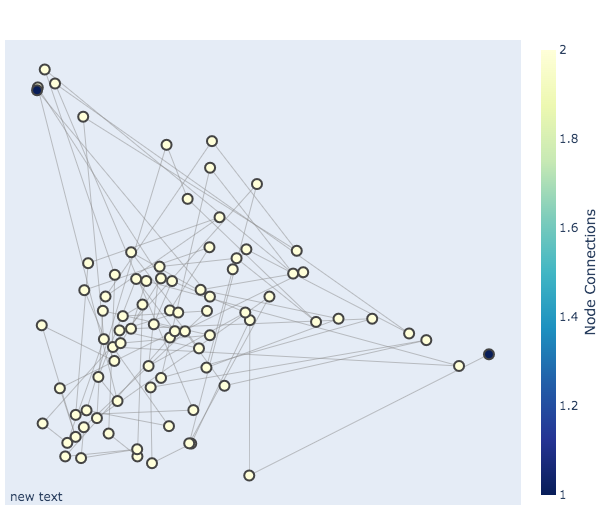}
         \includegraphics[width=\textwidth]{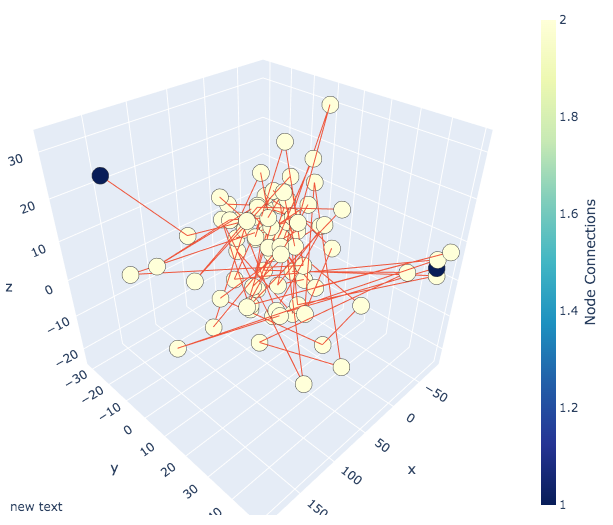}
         \includegraphics[width=\textwidth]{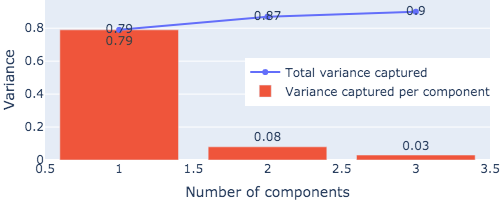}
         \caption{``Even though Hillary Clinton has been accused of making him feel uncomfortable by the Clinton Foundation: Trump still has been accused in countless cases of sexual harassment. He knows what he is doing and he has to stop it from happening to women right now. It's all in the past now. He's probably out of touch with women in the community. He is a sexist who is not doing a good enough job at fixing the country.''}
         \label{fig:15}
    \end{subfigure}
        \begin{subfigure}[b]{0.49\textwidth}
         \includegraphics[width=\textwidth]{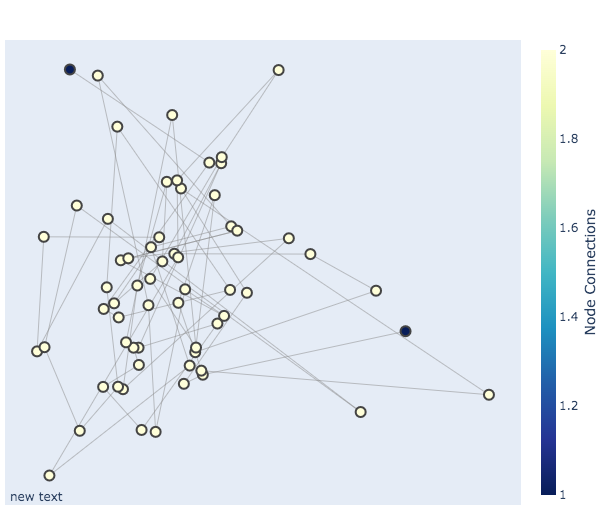}
         \includegraphics[width=\textwidth]{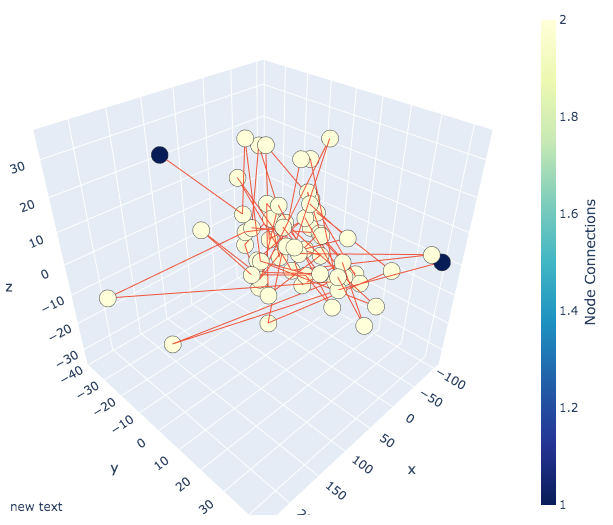}
         \includegraphics[width=\textwidth]{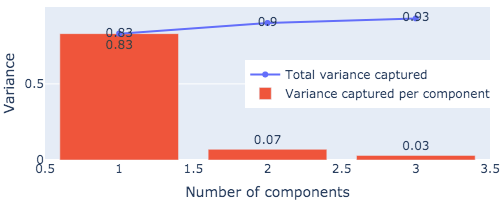}
         \caption{``Trump's words: in addition to the sexism of Donald Trump: may directly suggest that women can see something real in the president-elect when they hear the words. That might seem like a contradiction of history to even the most conservative conservative: but in reality: you can see it all over his own words.''}
         \label{fig:16}
    \end{subfigure}
\end{figure}
\begin{figure}\ContinuedFloat
    \begin{subfigure}[b]{0.49\textwidth}
         \includegraphics[width=\textwidth]{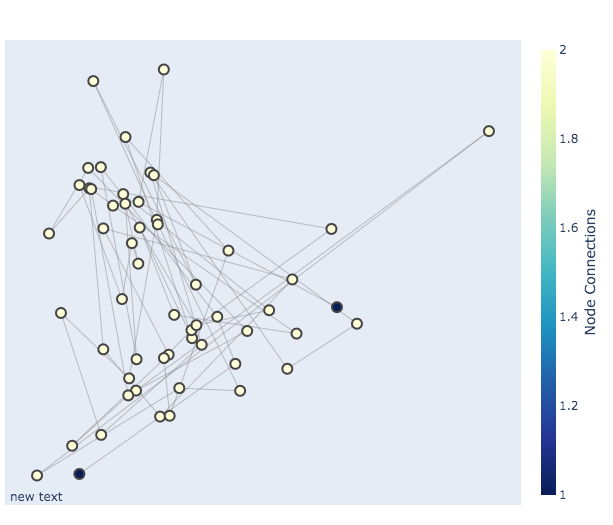}
         \includegraphics[width=\textwidth]{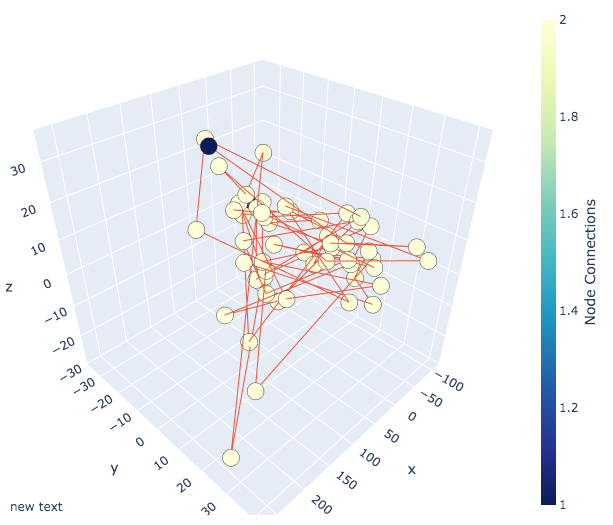}
         \includegraphics[width=\textwidth]{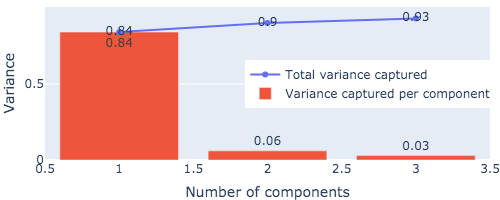}
         \caption{``The most recent presidential debate was conducted under the pretense that Hillary would defeat Donald Trump. However: there were only two men left in the audience. In the following moment: Hillary appeared to mock Trump in one of his most blatant and obnoxious attacks: his most infamous line yet.''}
         \label{fig:17}
    \end{subfigure}
        \begin{subfigure}[b]{0.49\textwidth}
         \includegraphics[width=\textwidth]{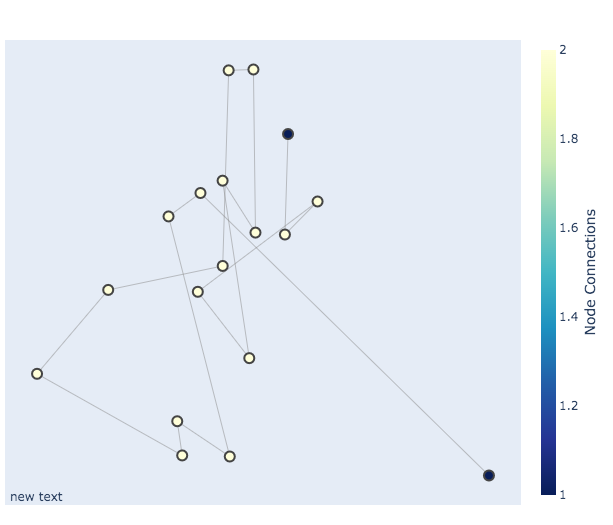}
         \includegraphics[width=\textwidth]{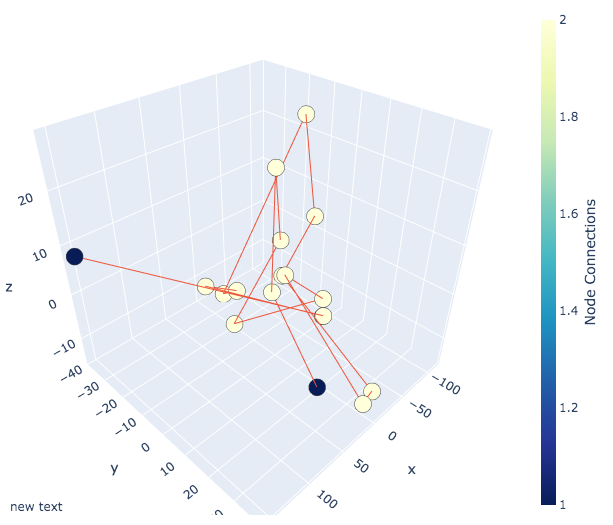}
         \includegraphics[width=\textwidth]{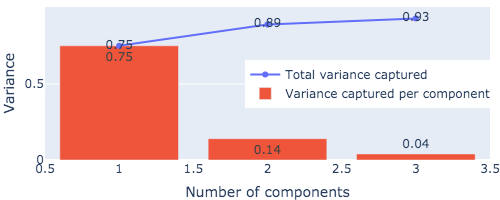}
         \caption{``""So: this is what I'm saying:"" Hillary said with a smirk.''}
         \label{fig:18}
    \end{subfigure}
\end{figure}
\begin{figure}\ContinuedFloat
    \begin{subfigure}[b]{0.49\textwidth}
         \includegraphics[width=\textwidth]{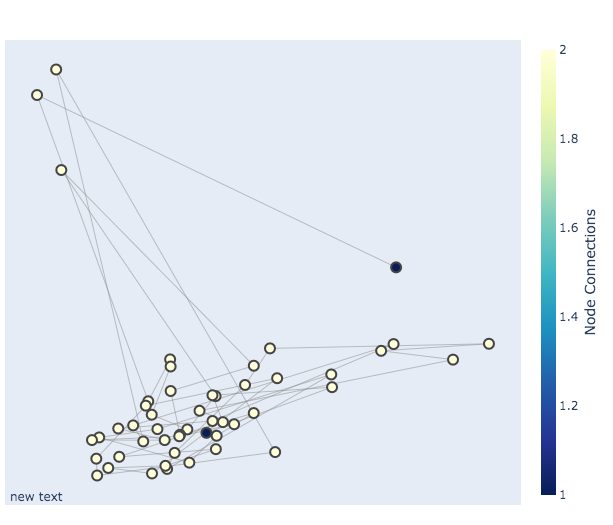}
         \includegraphics[width=\textwidth]{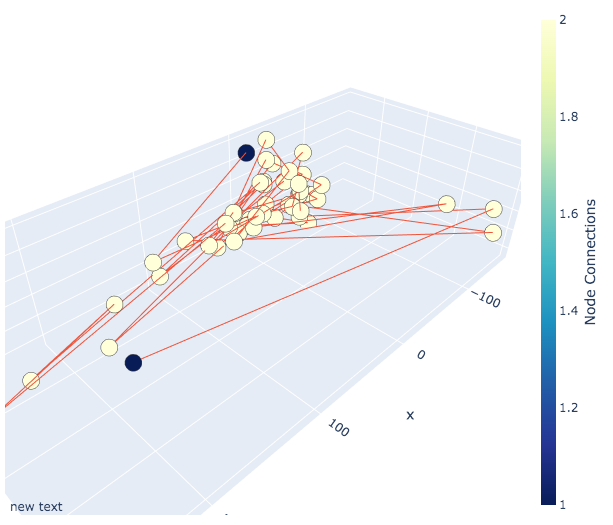}
         \includegraphics[width=\textwidth]{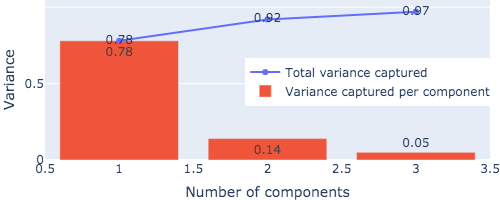}
         \caption{``Hillary wasn't laughing. She was shaking uncontrollably. ""This is what I'm saying: my great-great-great grandmother said. I can't believe it. Why don't you just walk away: and what will you give me?""''}
         \label{fig:19}
    \end{subfigure}
        \begin{subfigure}[b]{0.49\textwidth}
         \includegraphics[width=\textwidth]{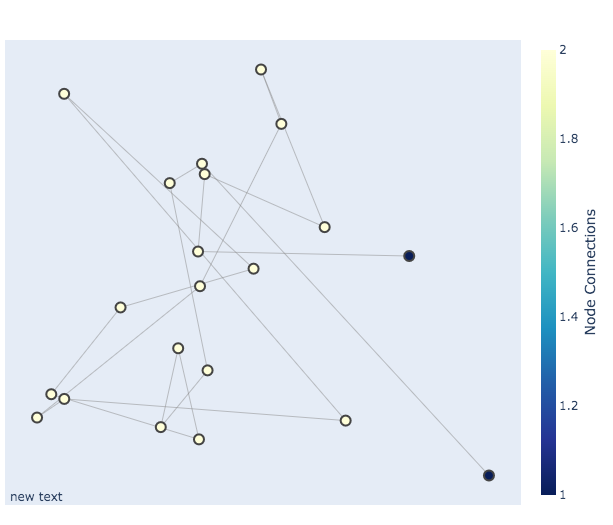}
         \includegraphics[width=\textwidth]{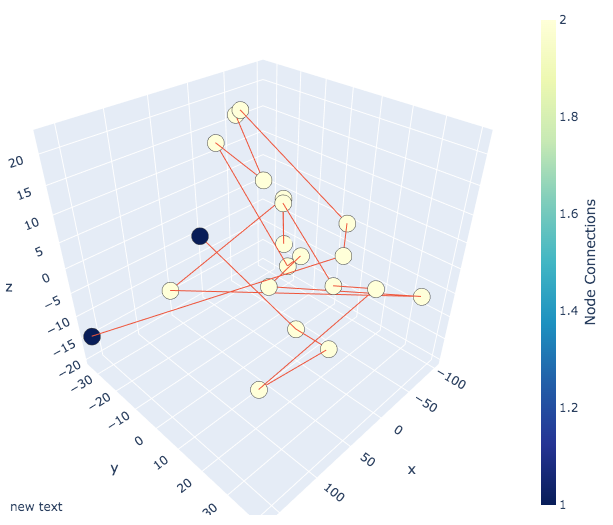}
         \includegraphics[width=\textwidth]{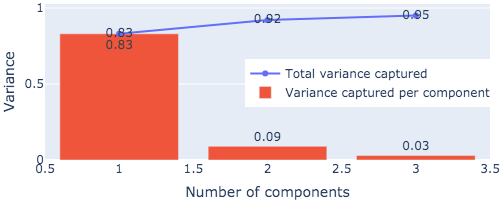}
         \caption{``""So: that's exactly what I'm saying. Don't just talk:"" Donald Trump said.''}
         \label{fig:20}
    \end{subfigure}
\end{figure}
\begin{figure}\ContinuedFloat
        \begin{subfigure}[b]{0.49\textwidth}
         \includegraphics[width=\textwidth]{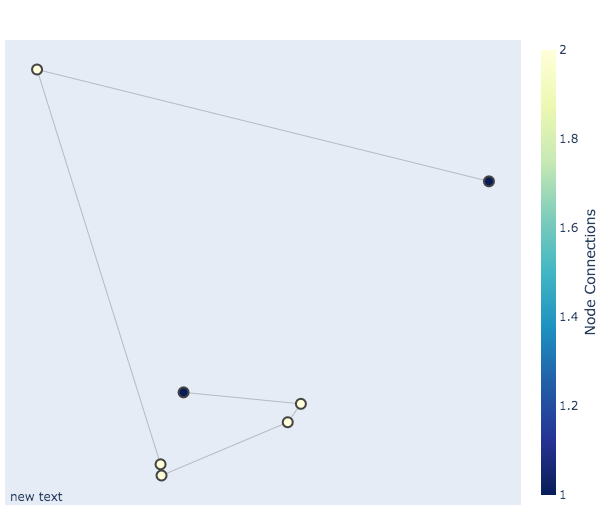}
         \includegraphics[width=\textwidth]{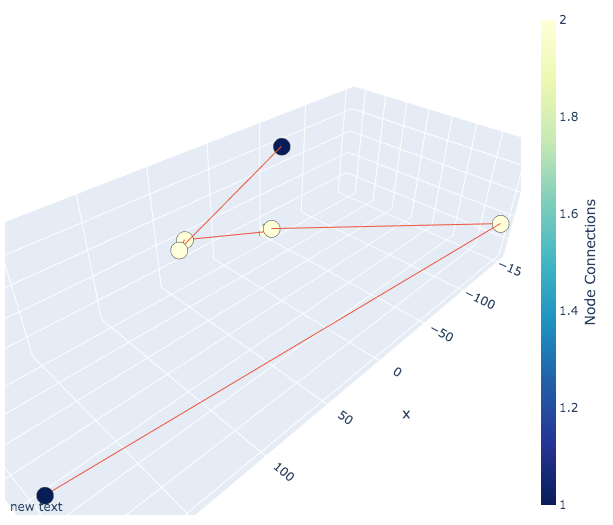}
         \includegraphics[width=\textwidth]{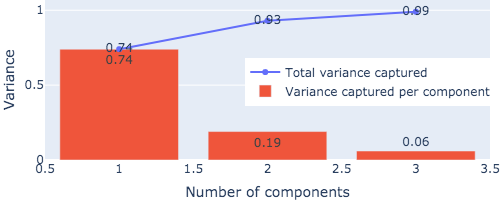}
         \caption{``Hillary didn't really mean that.''}
         \label{fig:21}
    \end{subfigure}
        \begin{subfigure}[b]{0.49\textwidth}
         \includegraphics[width=\textwidth]{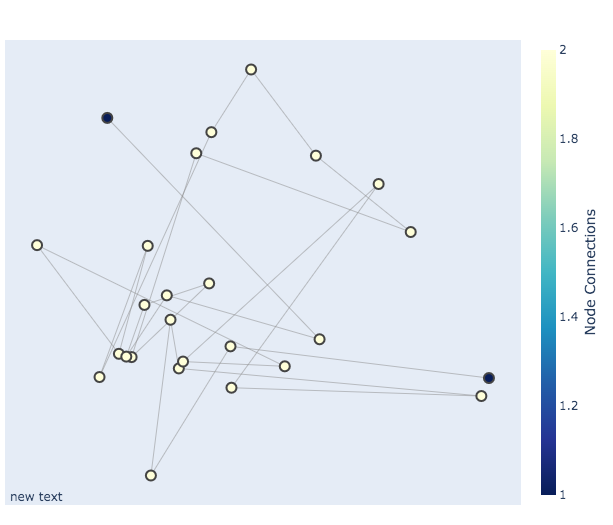}
         \includegraphics[width=\textwidth]{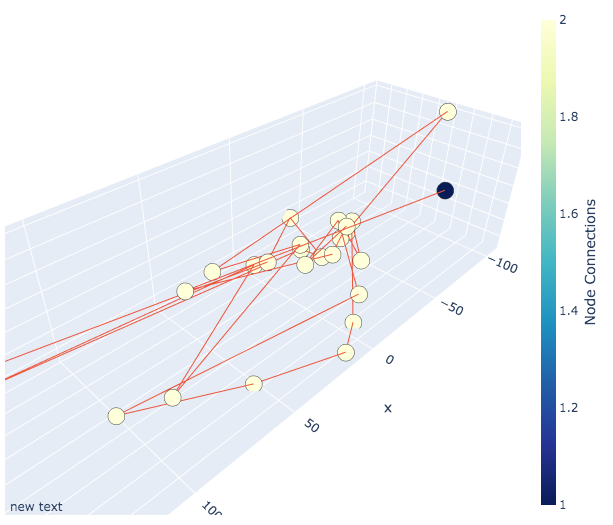}
         \includegraphics[width=\textwidth]{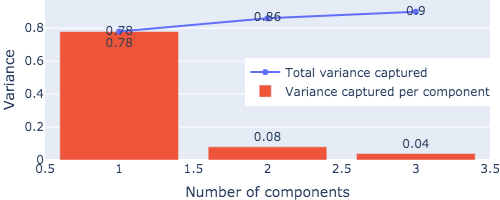}
         \caption{``Donald Trump has long been accused of misogynistic behavior. His most recent ""birther"" comments took a very different tack.''}
         \label{fig:22}
    \end{subfigure}
\end{figure}
\begin{figure}\ContinuedFloat
        \begin{subfigure}[b]{0.49\textwidth}
         \includegraphics[width=\textwidth]{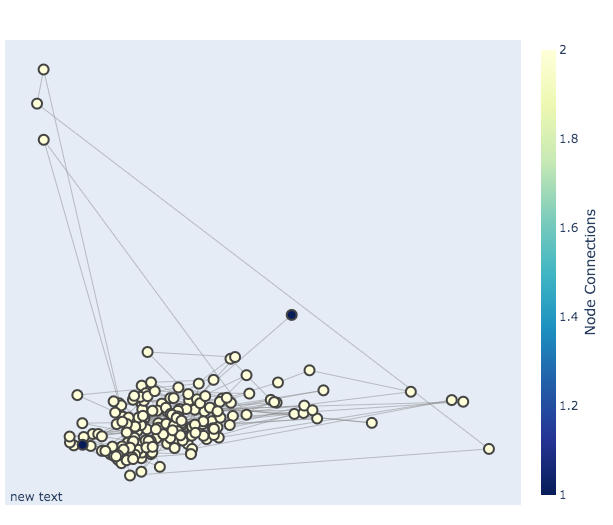}
         \includegraphics[width=\textwidth]{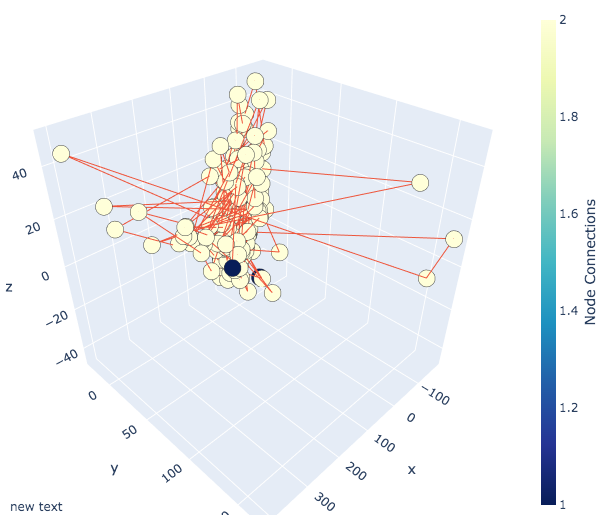}
         \includegraphics[width=\textwidth]{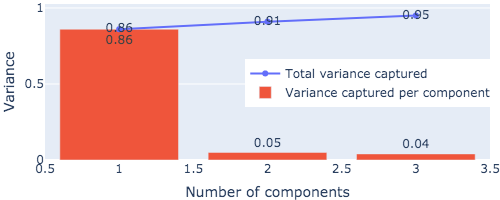}
         \caption{``In my last post: I gave a post on the use of the phrase: ""In response to the increasing role of foreign and domestic forces: we have seen a dramatic increase in the numbers of U.S. military in Afghanistan. The problem now is for the United States to be viewed as the guarantor of stability and stability against an increase in the amount of troops that it has to employ to maintain the peace and stability in Afghanistan."" While I hope that this post: which will be filled with some insight on ""real American involvement in the Taliban government of Helmand:"" will not be a comprehensive and insightful analysis of the situation in the country or at least a critical one is to be expected: I will not be limiting my discussion on the use and justification of the phrase in this post to discussing how ""foreign forces"" is not the primary factor that is often used when the U.S. forces are engaged in combat. That is still the only reason to use the phrase.''}
         \label{fig:23}
    \end{subfigure}
        \begin{subfigure}[b]{0.49\textwidth}
         \includegraphics[width=\textwidth]{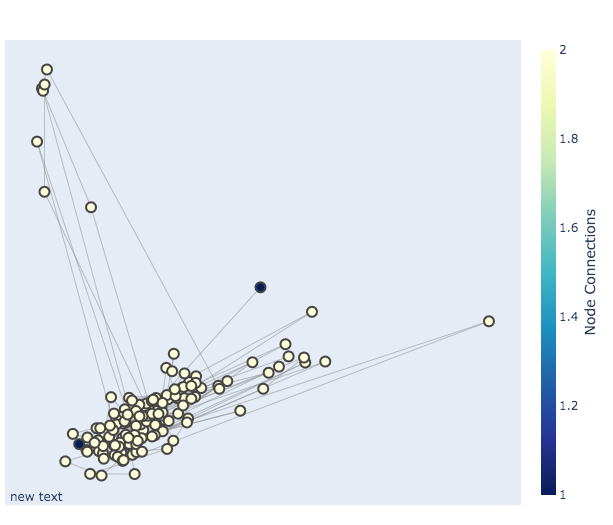}
         \includegraphics[width=\textwidth]{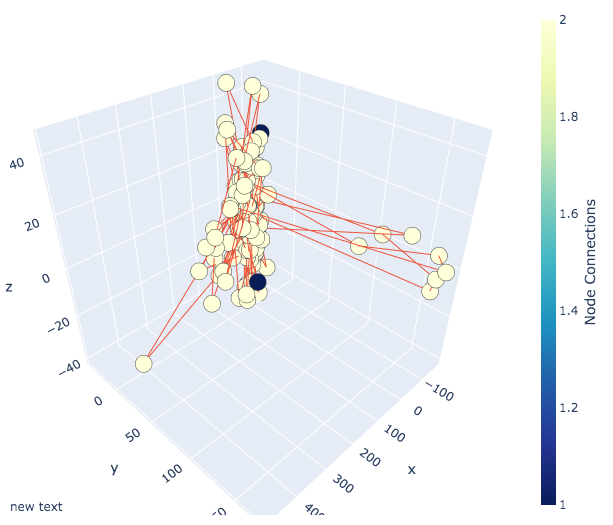}
         \includegraphics[width=\textwidth]{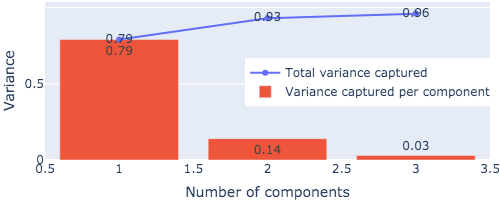}
         \caption{``While not all of my analysis will be focused on specific U.S. military engagements: one thing to be considered is ""international political participation."" The majority of my analysis is driven by a single military strategy that I used to analyze the conflict in Afghanistan—the United States military's involvement in it: and the ongoing role U.S. military forces: such as special operations forces: play. While not all of my analysis comes from a military policy or strategy that is directed at a specific target: I can present specific analyses in response to several key facts (or at least some of the things that will be presented to me in future posts) in relation to how U.S. military forces are involved in the ongoing conflict.''}
         \label{fig:24}
    \end{subfigure}
\end{figure}
\begin{figure}\ContinuedFloat
        \begin{subfigure}[b]{0.49\textwidth}
         \includegraphics[width=\textwidth]{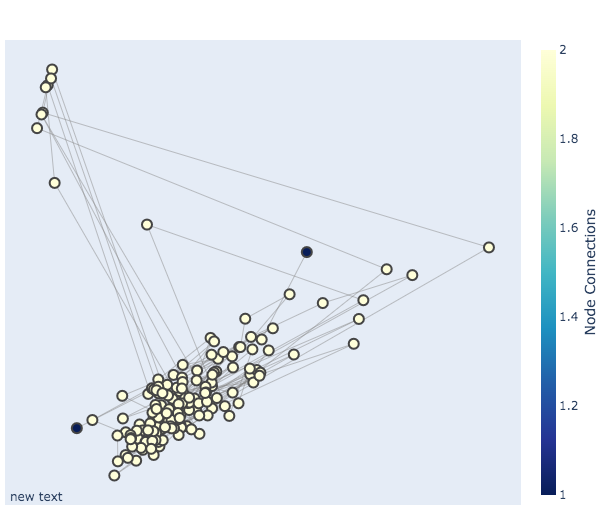}
         \includegraphics[width=\textwidth]{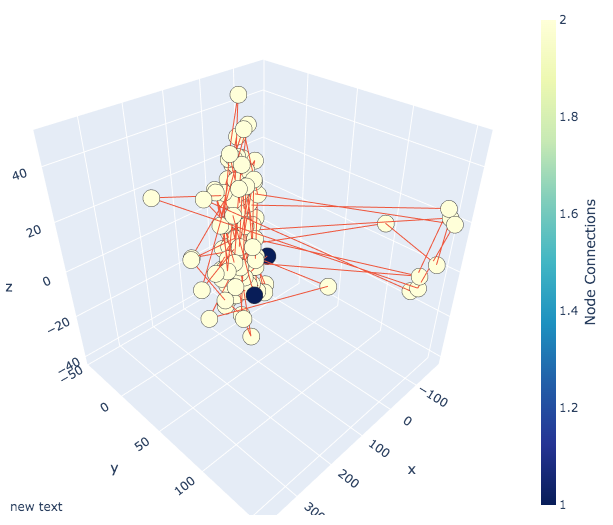}
        \includegraphics[width=\textwidth]{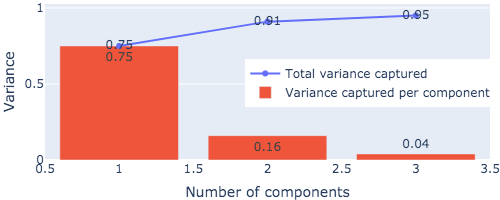}
         \caption{``The problem here is that those who have been following the U.S. military action in Afghanistan for the last few years have been in a position of being surprised by what the U.S. military has been doing: and have seen the military action in and of itself as being in their power to stop the country from falling into the hands of al-Qaida. Many other people are unaware of the huge civilian casualties caused by al-Qaida's operations there: but many of us are shocked by reports from the same source that has been making headlines that the U.S. military has been engaging in drone strikes and airstrikes on tribal territory and targets in northwestern Pakistan. The information that we have come across is that those who think that the U.S. military has engaged in covert war in their country have been wrong before.''}
         \label{fig:25}
    \end{subfigure}
    \begin{subfigure}[b]{0.49\textwidth}
         \includegraphics[width=\textwidth]{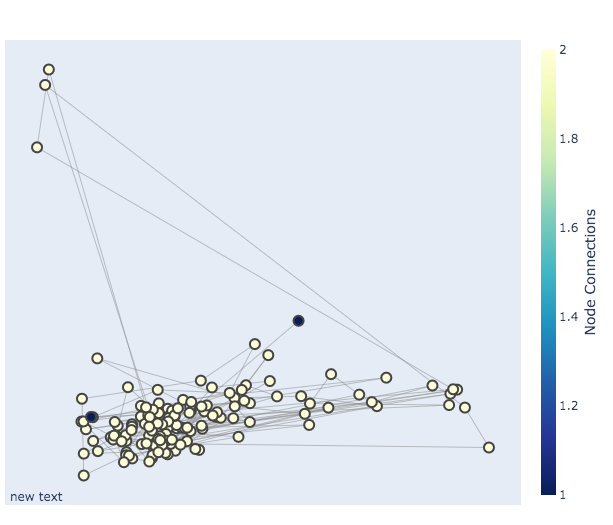}
         \includegraphics[width=\textwidth]{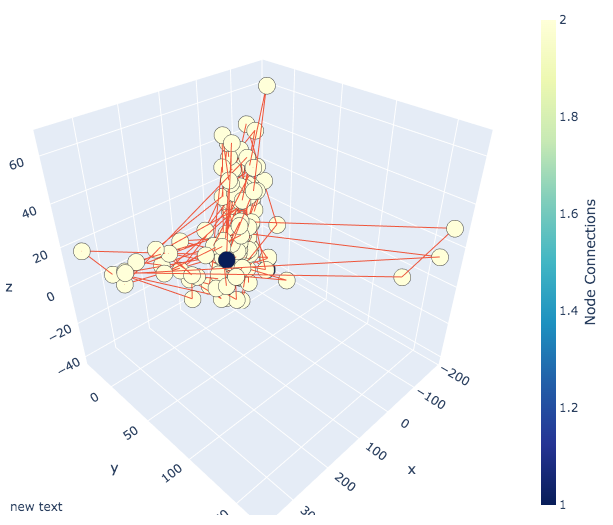}
          \includegraphics[width=\textwidth]{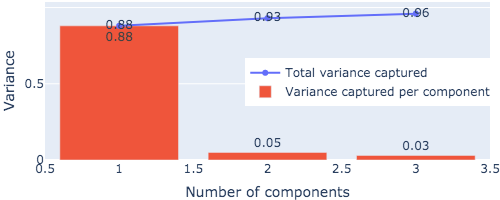}
          \caption{``This is just a sampling of the fact that the U.S. military has engaged in covert wars in the past that we do not take at face value. We have engaged in ""non-combatants"" wars: or ""containment wars:"" in order to prevent or avoid conflict between groups operating in a particular area: which is a conflict based on two different types of government/foreign policy objectives—economic and political. In the past few years: more than 100 countries have joined the U.K. and United States military in ""non-combatants:"" and over 1:1 in all three countries now rely heavily on U.S. bases to keep out al-Qaida.''}
         \label{fig:26}
    \end{subfigure}
\end{figure}
\begin{figure}
    \begin{subfigure}[b]{0.49\textwidth}
         \includegraphics[width=\textwidth]{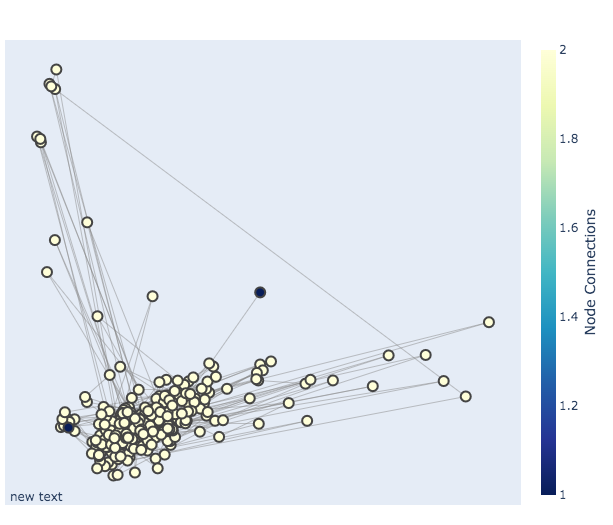}
         \includegraphics[width=\textwidth]{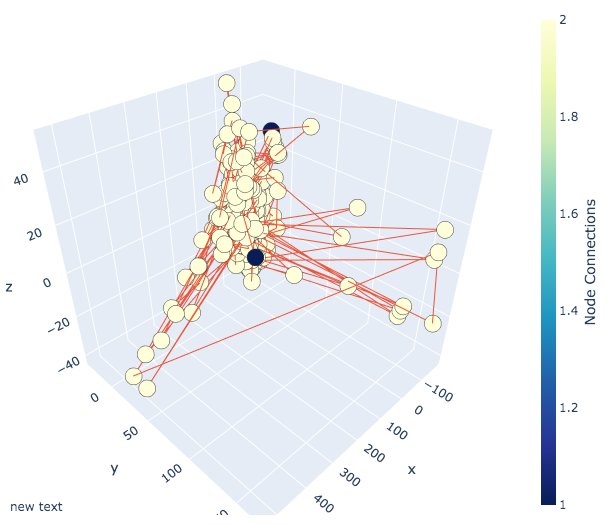}
          \includegraphics[width=\textwidth]{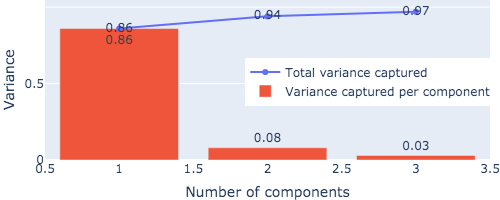}
          \subcaption{``However: there is an important difference in the U.S. military action in the past that I think can be made. It is not surprising that those who have spent time studying for the 2012 Military History books have come to view the US military as being operating mostly in the region: especially in the hands of those within the United States government's leadership. Since 2005 the United States has invaded Kuwait: Afghanistan: and Iraq: in the name of preserving democracy in those countries. That invasion: followed by occupation of Afghanistan: Iraq: and Somalia: and then with more or less full-force occupation of Libya in July of this year: also left the United States with a significant force in Afghanistan: which the United States is now effectively supporting. The Obama administration began bombing Iraq after the invasion: which means that the U.S. army has been in the region for about a year. In fact: the United States has been bombing a third of Iraq's territory and has started bombing and occupying its own territory. That has included a substantial portion of the Kurdish-majority nation of north-eastern Iraq. At the same time: those in power in the United States have been conducting ""non-combatants"" operations which have the purpose of trying to isolate: repel: or even neutralize any threat other than al-Qaida in Iraq as well as any and all threats (including economic) to the interests of international democracy: human rights: and the rule of law.''}
         \label{fig:27}
    \end{subfigure}
        \begin{subfigure}[b]{0.49\textwidth}
         \includegraphics[width=\textwidth]{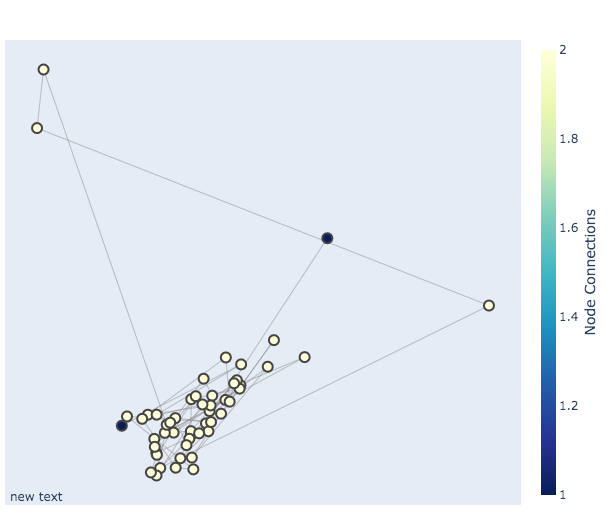}
         \includegraphics[width=\textwidth]{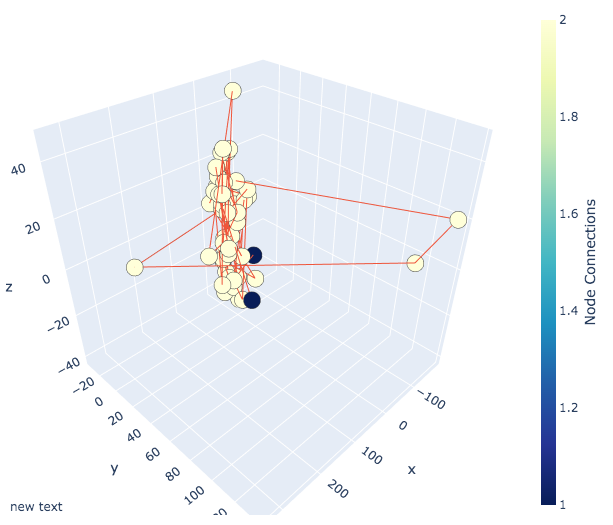}
          \includegraphics[width=\textwidth]{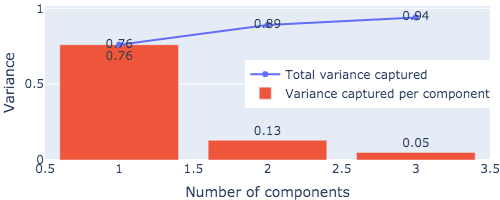}
          \caption{``To get to the point: the United States is providing intelligence support and training to the Taliban on how to carry out the war on terror: not the war on foreign fighters. It's a system that the U.S. has used for over a decade.''}
         \label{fig:28}
    \end{subfigure}
\end{figure}
\begin{figure}\ContinuedFloat
    \begin{subfigure}[b]{0.49\textwidth}
         \includegraphics[width=\textwidth]{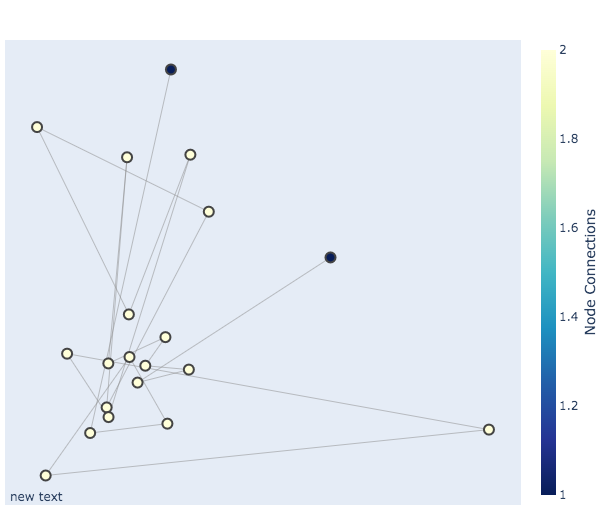}
         \includegraphics[width=\textwidth]{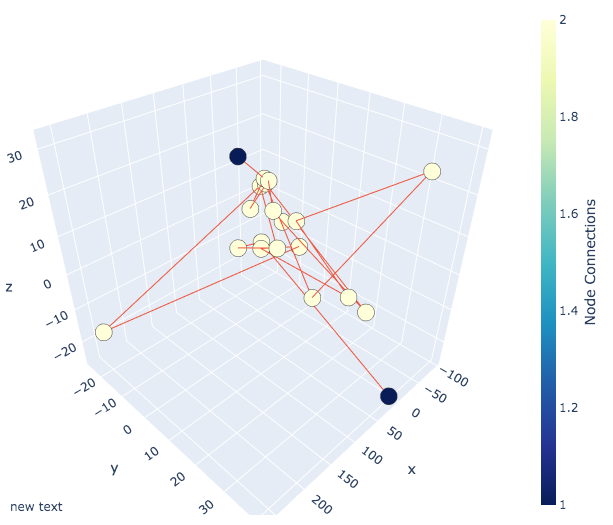}
          \includegraphics[width=\textwidth]{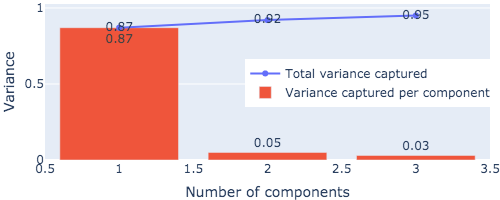}
          \caption{``The situation in Afghanistan today is an ongoing problem. The Obama administration is trying to make sure that"''}
         \label{fig:29}
    \end{subfigure}
    \begin{subfigure}[b]{0.49\textwidth}
         \includegraphics[width=\textwidth]{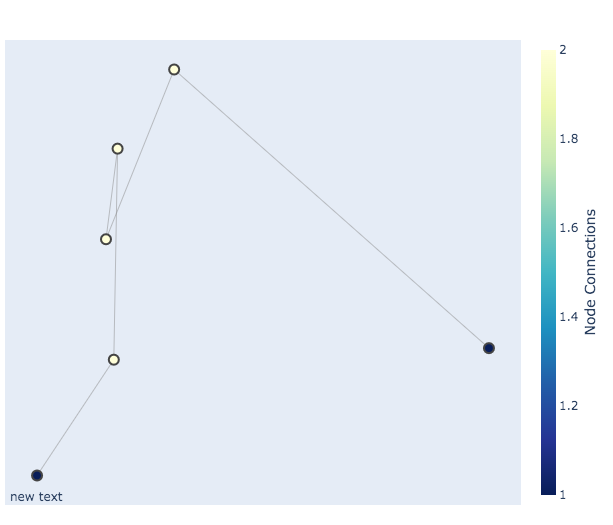}
         \includegraphics[width=\textwidth]{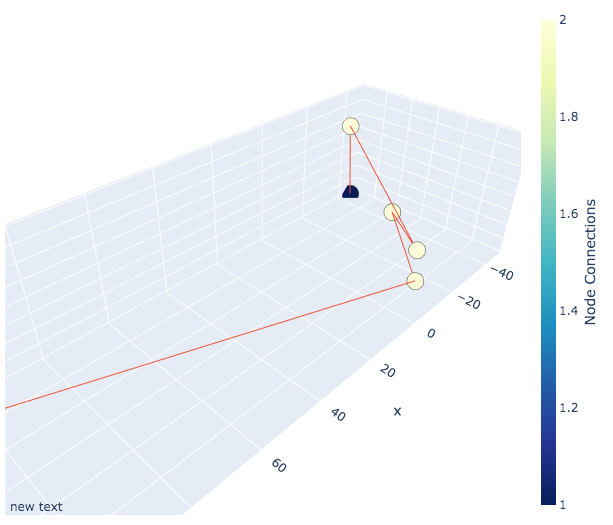}
          \includegraphics[width=\textwidth]{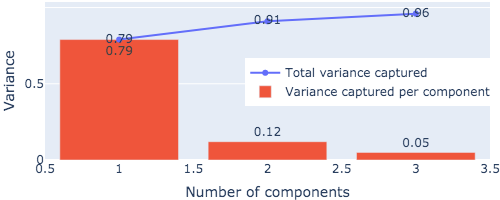}
          \caption{``The KFC Bar \& Grill''}
         \label{fig:30}
    \end{subfigure}
\end{figure}
\begin{figure}\ContinuedFloat
    \begin{subfigure}[b]{0.49\textwidth}
         \includegraphics[width=\textwidth]{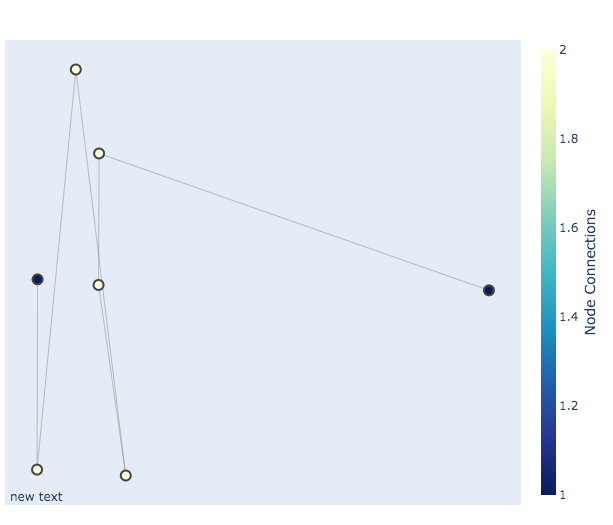}
         \includegraphics[width=\textwidth]{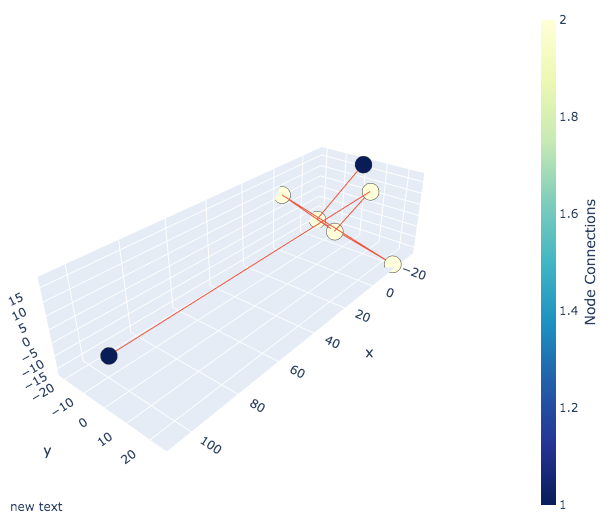}
          \includegraphics[width=\textwidth]{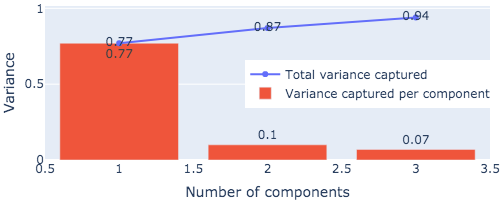}
          \caption{``The Fries \& Dips Bar''}
         \label{fig:31}
    \end{subfigure}
    \begin{subfigure}[b]{0.49\textwidth}
         \includegraphics[width=\textwidth]{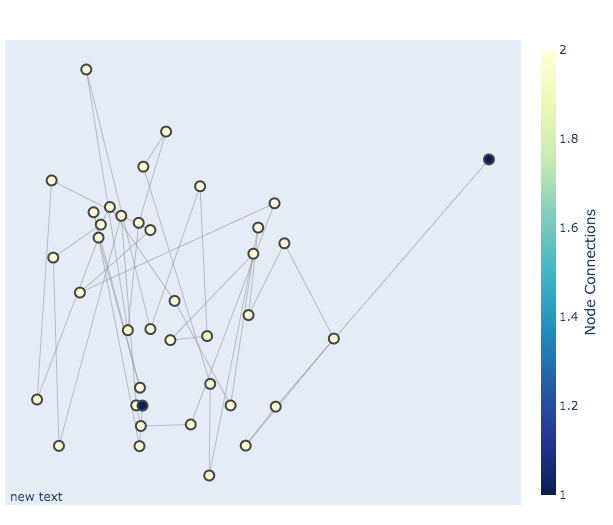}
         \includegraphics[width=\textwidth]{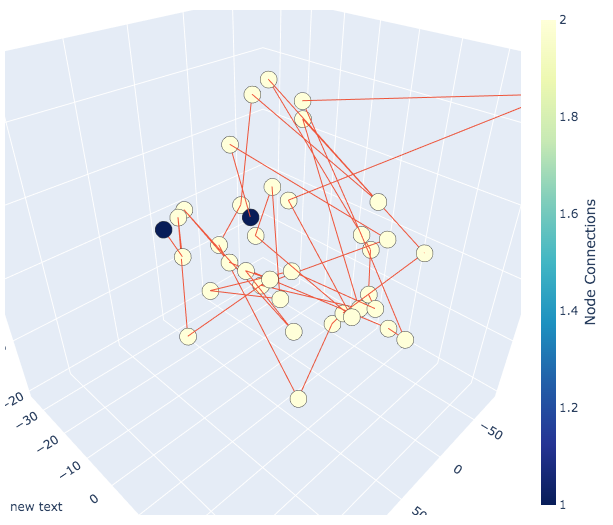}
          \includegraphics[width=\textwidth]{figs/sent_eval/sent2.png}
          \caption{``BarBans.com - The BeerHouse has some great spots to shop for craft beer: too - and is still open for tours. Be warned: it's closed for an extended time.''}
         \label{fig:32}
    \end{subfigure}
\end{figure}
\begin{figure}\ContinuedFloat
    \begin{subfigure}[b]{0.49\textwidth}
         \includegraphics[width=\textwidth]{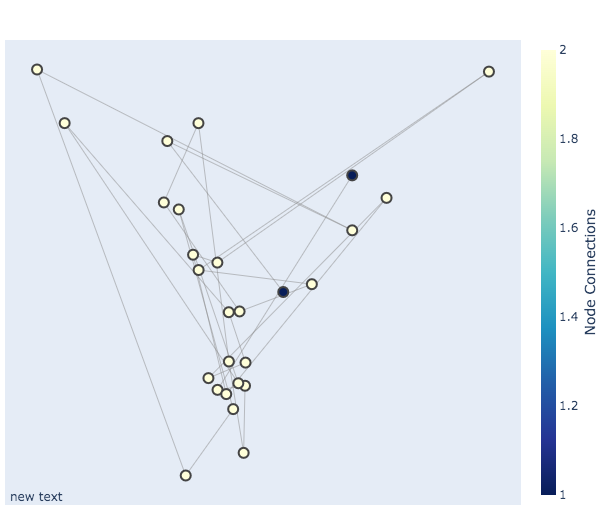}
         \includegraphics[width=\textwidth]{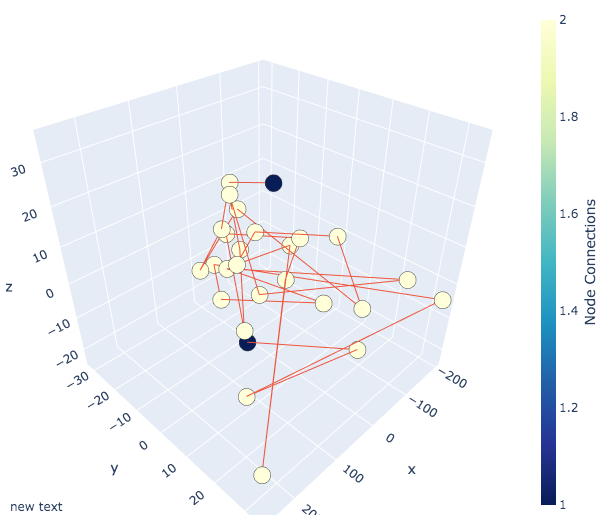}
          \includegraphics[width=\textwidth]{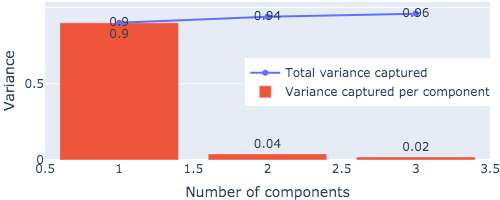}
          \caption{``St. Joe's Pub (1833 N.E. 4th Ave.: Portland). \$7.99: \$9.99''}
         \label{fig:33}
    \end{subfigure}
    \begin{subfigure}[b]{0.49\textwidth}
         \includegraphics[width=\textwidth]{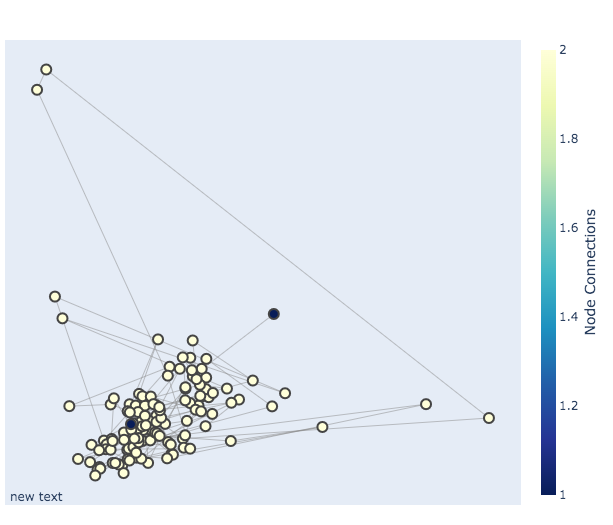}
         \includegraphics[width=\textwidth]{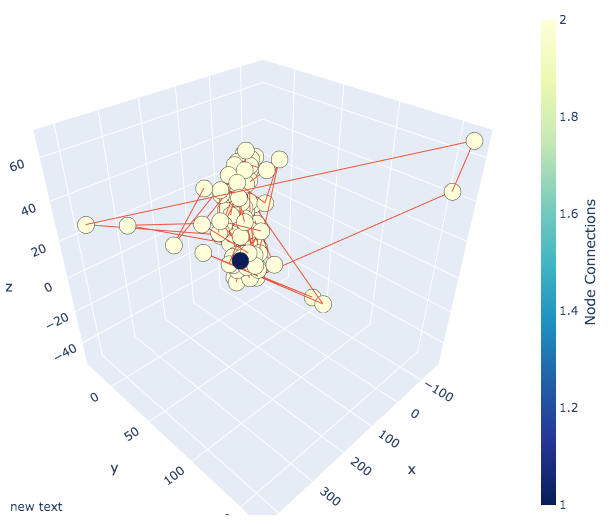}
          \includegraphics[width=\textwidth]{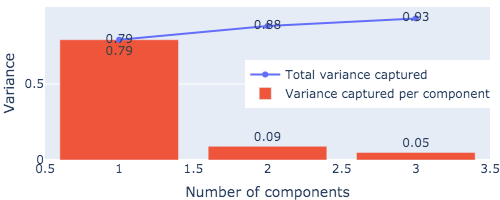}
          \caption{``Brewmaster John H. ""Buck"" Kincaid bought the brewery in 2003 for \$1.75 million. The taproom features the original pub's original menu: but now has a small beer selection: too. The beer house also runs events and beer tappings: and has a full bar. In addition: it's named after the Portland area's most beloved neighborhood band: the Raccoon Raiders. Hops: pints: sodas and bottled beer also all run at 7 p.m. on Saturdays. It also is the hub of the Portland Beer Alliance: a nonprofit association that helps brewers connect with Portland locals in new ways.''}
         \label{fig:34}
    \end{subfigure}
\end{figure}
\begin{figure}\ContinuedFloat
    \begin{subfigure}[b]{0.49\textwidth}
         \includegraphics[width=\textwidth]{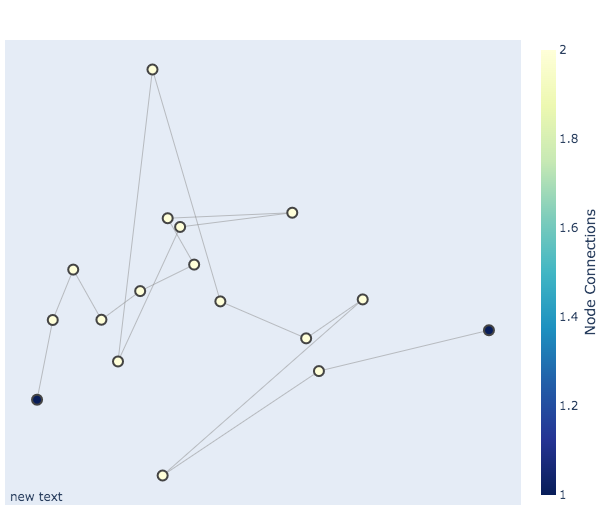}
         \includegraphics[width=\textwidth]{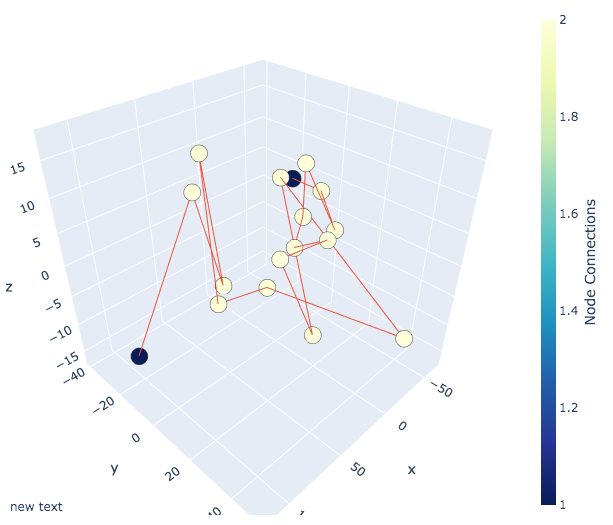}
          \includegraphics[width=\textwidth]{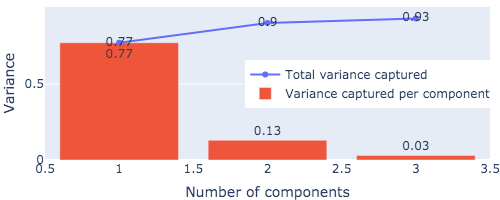}
          \caption{``Beer Co.: one of the biggest beer makers in the world: also opened there.''}
         \label{fig:35}
    \end{subfigure}
    \begin{subfigure}[b]{0.49\textwidth}
         \includegraphics[width=\textwidth]{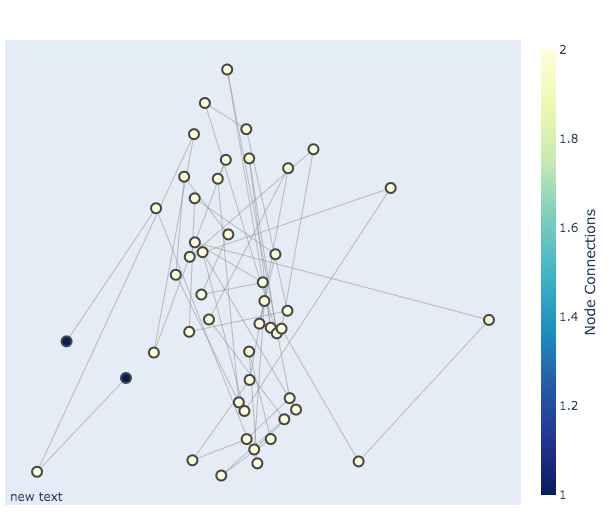}
         \includegraphics[width=\textwidth]{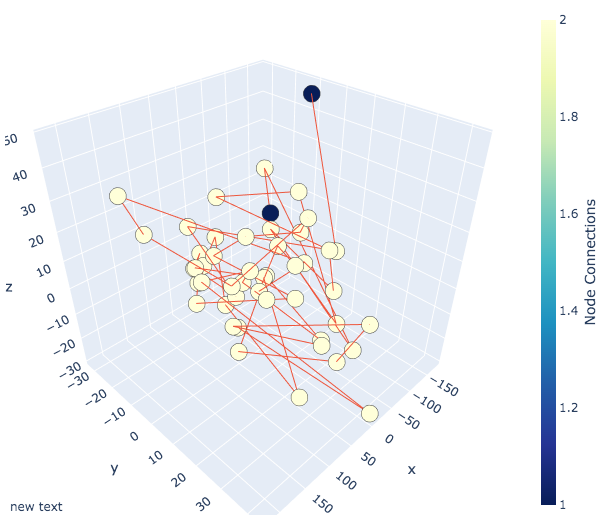}
          \includegraphics[width=\textwidth]{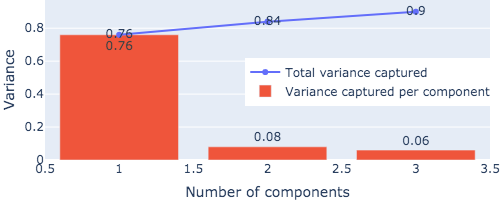}
          \caption{``An unlicensed party member from Russia: who works at the Ukrainian-government health center Kyiv: poses for a photo shortly before the departure of Prime Minister Viktor Yanukovych on March 18: 2014. (Alexei Nikolsky/Reuters)''}
         \label{fig:36}
    \end{subfigure}
\end{figure}
\begin{figure}\ContinuedFloat
    \begin{subfigure}[b]{0.49\textwidth}
         \includegraphics[width=\textwidth]{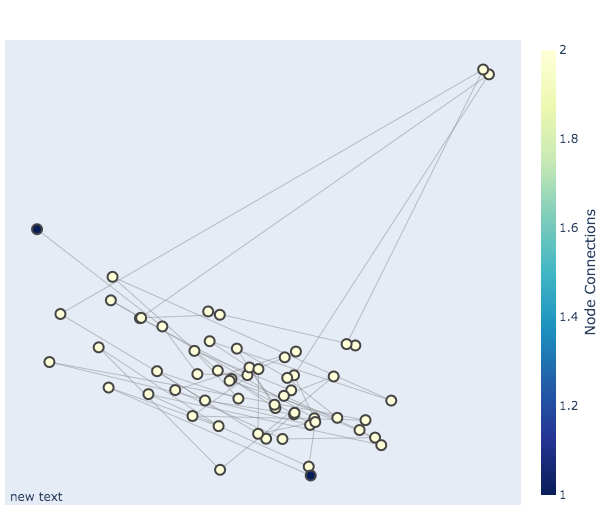}
         \includegraphics[width=\textwidth]{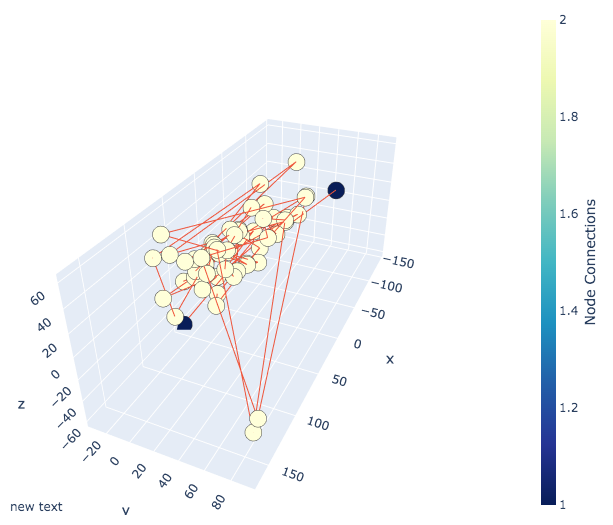}
          \includegraphics[width=\textwidth]{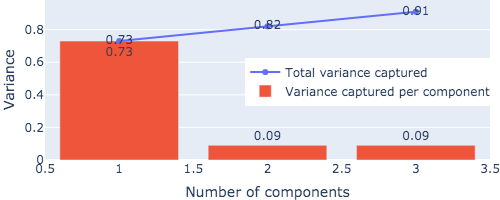}
          \caption{``A week after Russian President Vladimir Putin was sworn in as Ukraine's new prime minister: he signed an executive decree that gave the new premier an additional five months to step down. A week before that: Yanukovych had threatened to end work in Kiev: promising to resign as a ""dangerous man.""''}
         \label{fig:37}
    \end{subfigure}
    \begin{subfigure}[b]{0.49\textwidth}
         \includegraphics[width=\textwidth]{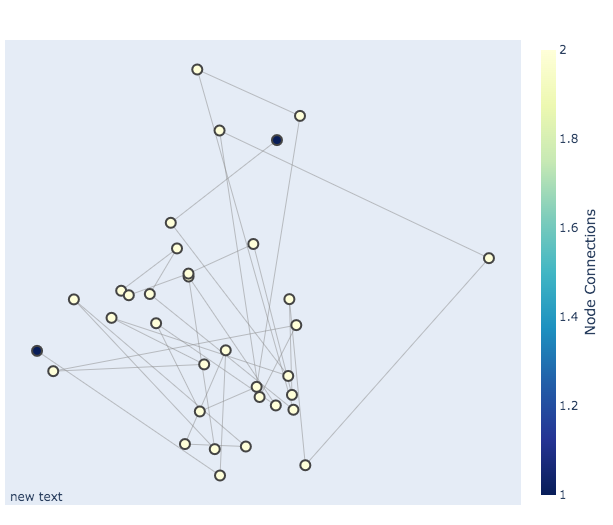}
         \includegraphics[width=\textwidth]{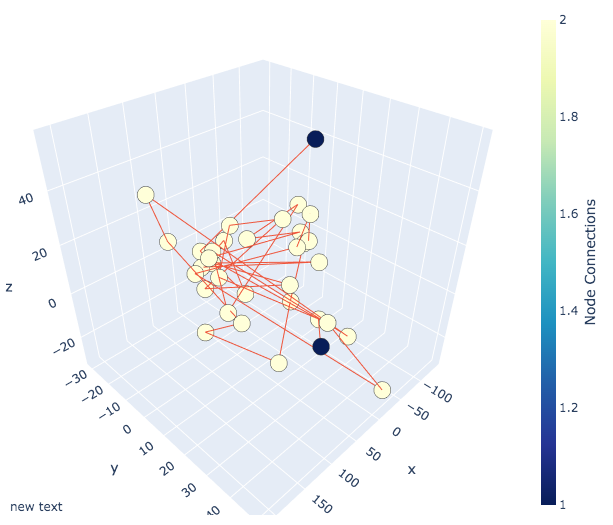}
          \includegraphics[width=\textwidth]{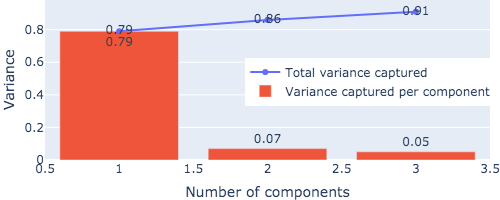}
          \caption{``But the move followed the death of Yanukovych's mentor: the pro-Russian president Viktor Yanukovych himself: who was arrested on April 15 for his role in Yanukovych's death.''}
         \label{fig:38}
    \end{subfigure}
\end{figure}
\begin{figure}\ContinuedFloat
    \begin{subfigure}[b]{0.49\textwidth}
         \includegraphics[width=\textwidth]{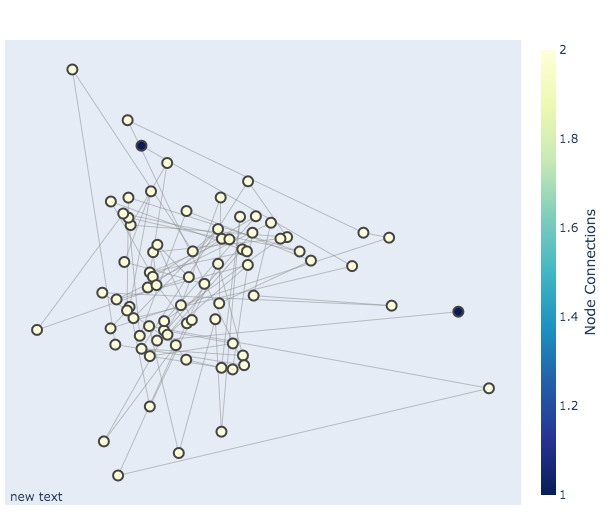}
         \includegraphics[width=\textwidth]{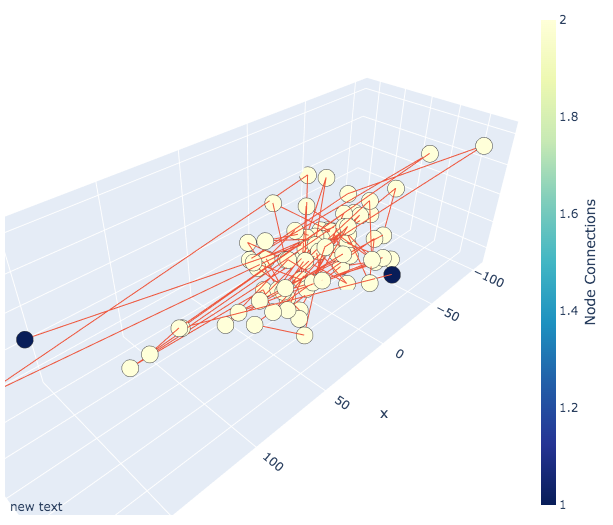}
          \includegraphics[width=\textwidth]{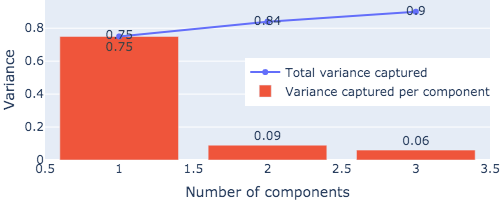}
          \caption{``Now: a new political and legal structure: in which Yanukovych will be able to run a national legislature or run on an ""ordinary"" national legislative committee: and a system of executive committees to determine the direction of Russia's presidency or legislature has been implemented -- in Kiev's capital: in some cases for more than a year -- in the run-up to the Maidan protests last week.''}
         \label{fig:39}
    \end{subfigure}
    \begin{subfigure}[b]{0.49\textwidth}
         \includegraphics[width=\textwidth]{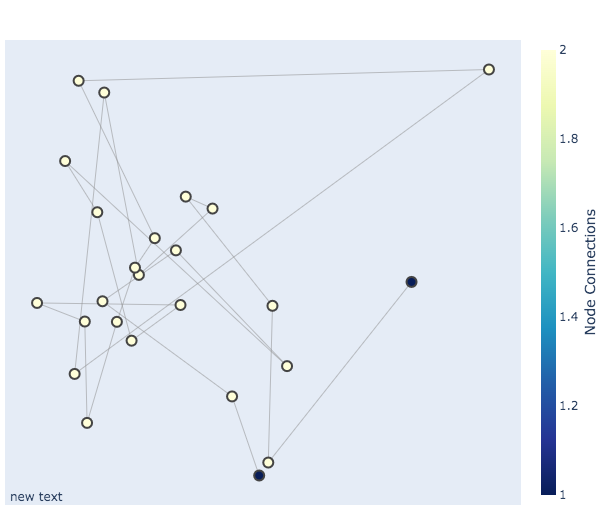}
         \includegraphics[width=\textwidth]{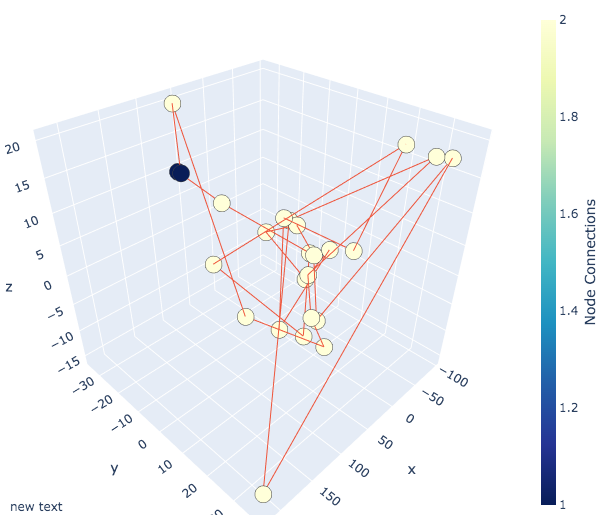}
          \includegraphics[width=\textwidth]{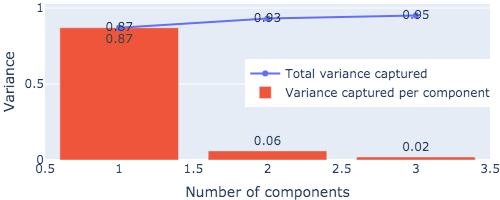}
          \caption{``39:"That means the opposition will be able to control the government: but only if the government will comply with Ukrainian demands."''}
         \label{fig:40}
    \end{subfigure} 
\end{figure}
\begin{figure}\ContinuedFloat
    \begin{subfigure}[b]{0.49\textwidth}
         \includegraphics[width=\textwidth]{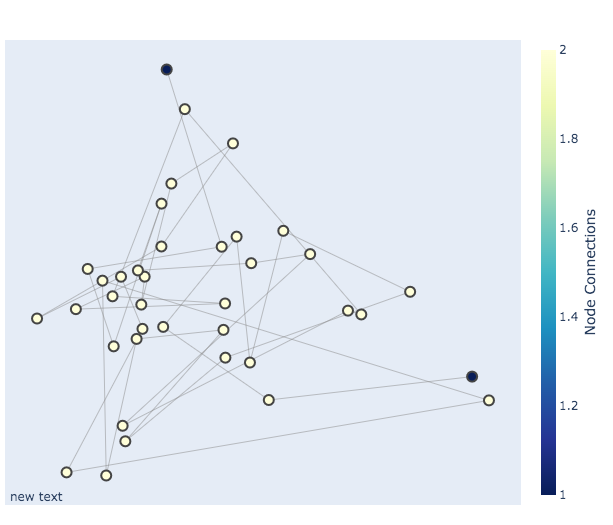}
         \includegraphics[width=\textwidth]{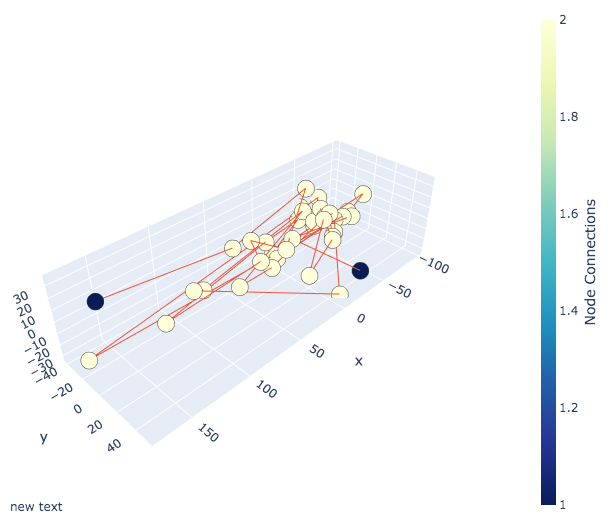}
          \includegraphics[width=\textwidth]{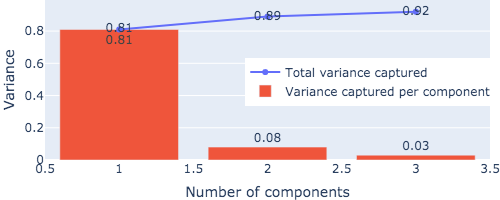}
          \caption{``According to a report on Monday: the authorities in Kiev will be able to keep an additional five ""ordinary"" legislative committees if Yanukovych ""continues to carry out the measures that he promised.""''}
         \label{fig:41}
    \end{subfigure}
    \begin{subfigure}[b]{0.49\textwidth}
         \includegraphics[width=\textwidth]{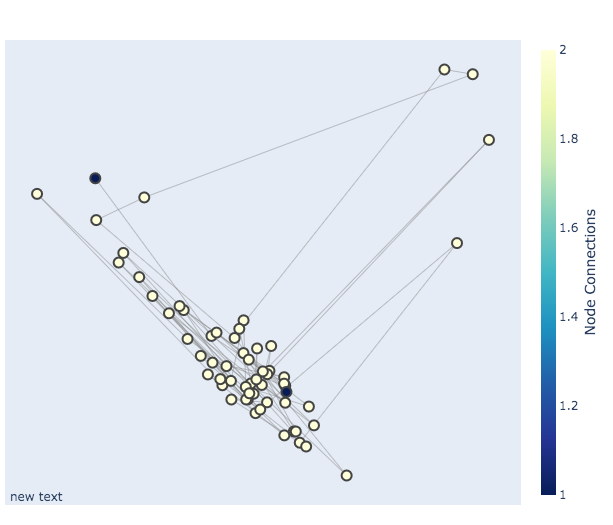}
         \includegraphics[width=\textwidth]{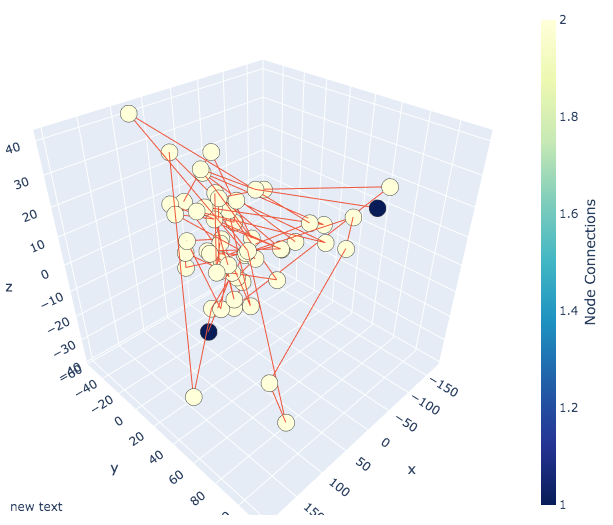}
          \includegraphics[width=\textwidth]{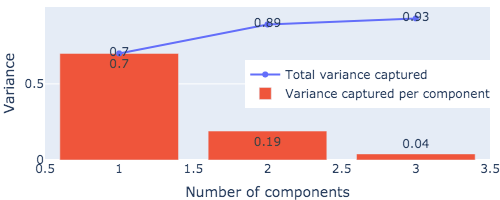}
          \caption{``The new order: which is in stark contrast to the one set in 2009 in a deal struck by the United States and the European Union: was supposed to make it easier for Ukraine and Russia to work side-by-side on policy issues such as Ukraine's status as having to leave the EU.''}
         \label{fig:42}
    \end{subfigure}
\end{figure}
\begin{figure}\ContinuedFloat
    \begin{subfigure}[b]{0.49\textwidth}
         \includegraphics[width=\textwidth]{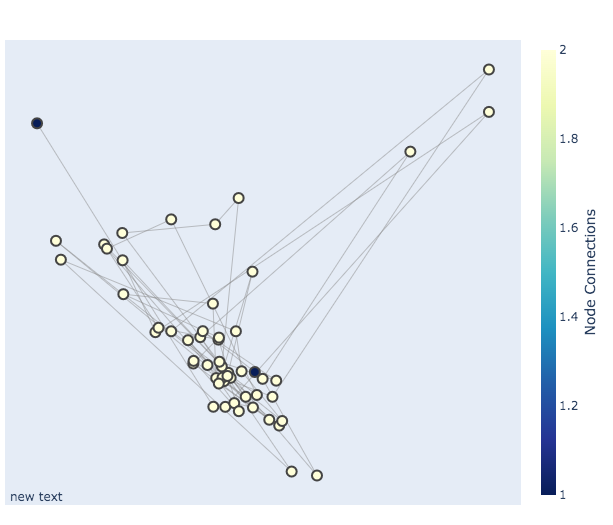}
         \includegraphics[width=\textwidth]{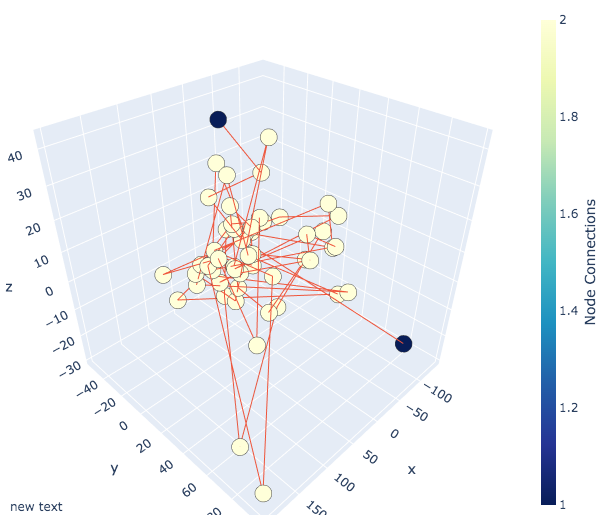}
         \includegraphics[width=\textwidth]{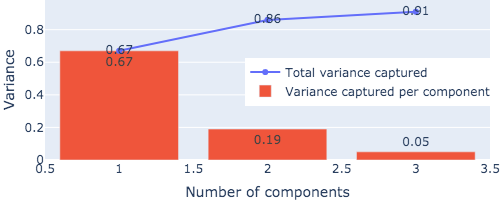}
         \caption{``But in the wake of Yanukovych's death: new powers have come into effect from the Russian military in the wake of his ouster: and Ukrainian President Petro Poroshenko has signed an order instructing the country's national defense forces to intervene militarily to keep it alive.''}
         \label{fig:43}
    \end{subfigure}
    \begin{subfigure}[b]{0.49\textwidth}
         \includegraphics[width=\textwidth]{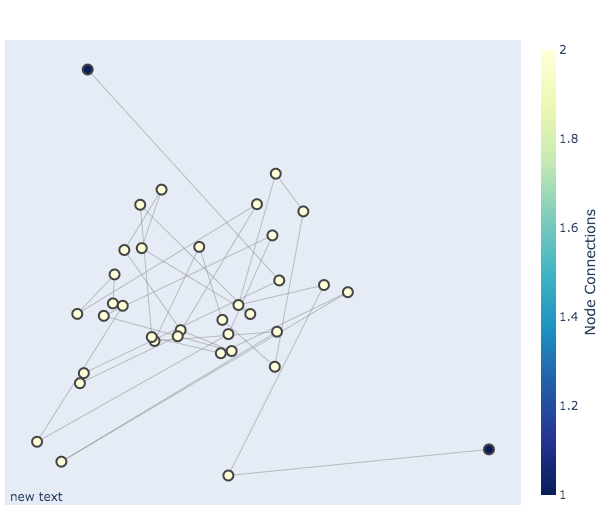}
         \includegraphics[width=\textwidth]{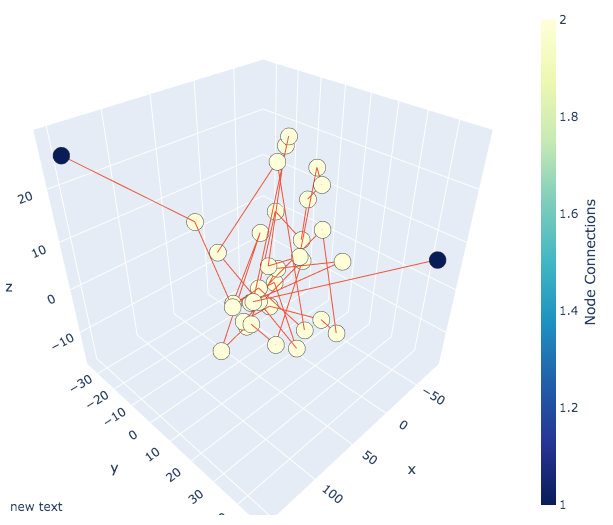}
          \includegraphics[width=\textwidth]{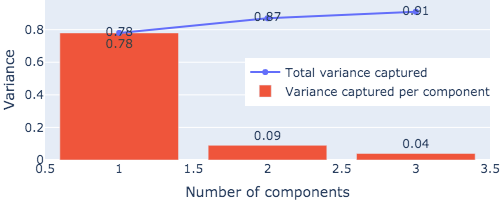}
          \caption{``The current president: Petro Poroshenko: has been forced to deal with the situation in Kiev: and it appears his goal is to move the political process closer to his own agenda.''}
         \label{fig:44}
    \end{subfigure}
\end{figure}
\begin{figure}\ContinuedFloat
    \begin{subfigure}[b]{0.49\textwidth}
         \includegraphics[width=\textwidth]{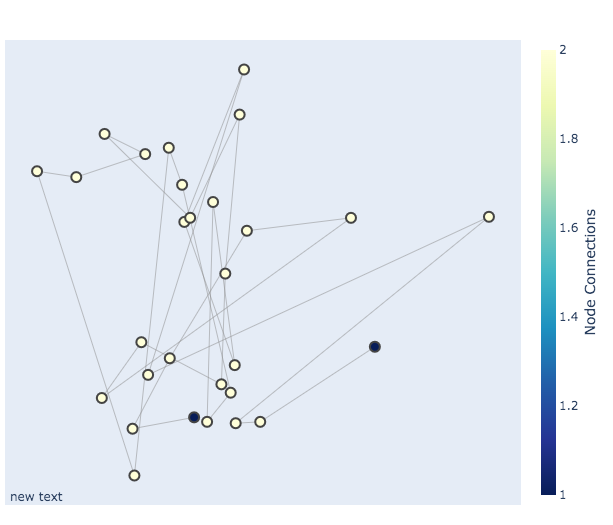}
         \includegraphics[width=\textwidth]{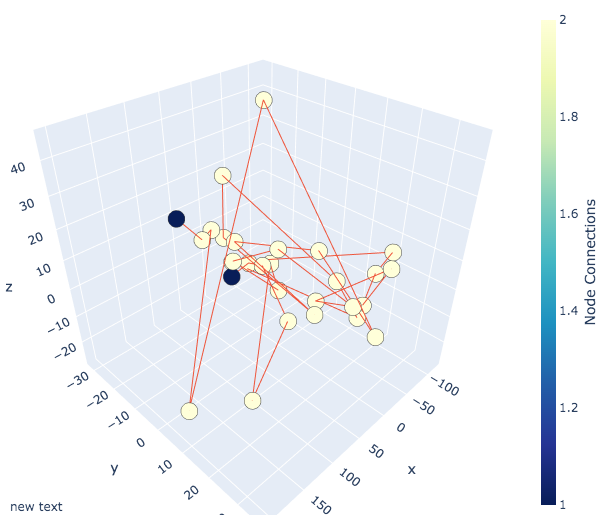}
          \includegraphics[width=\textwidth]{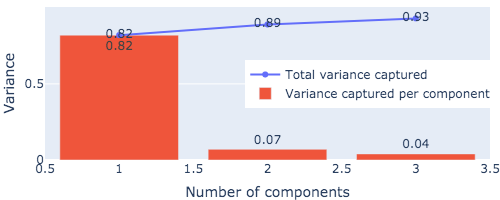}
          \caption{``[N.Y.'s new president: Petro Poroshenko: ""I am doing the best I can"" are we talking about? ... ]''}
         \label{fig:45}
    \end{subfigure}
    \begin{subfigure}[b]{0.49\textwidth}
         \includegraphics[width=\textwidth]{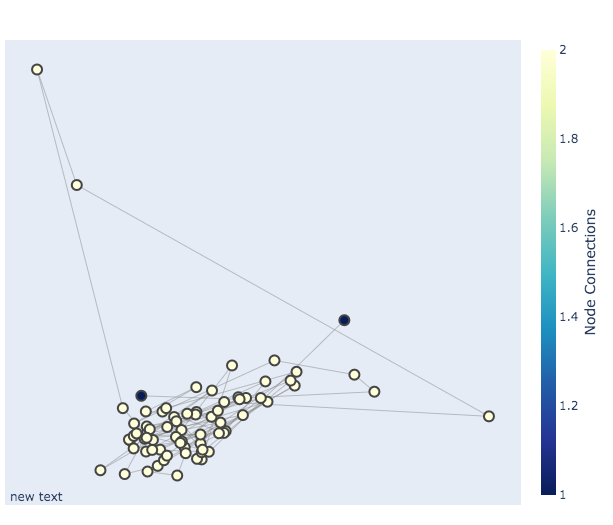}
         \includegraphics[width=\textwidth]{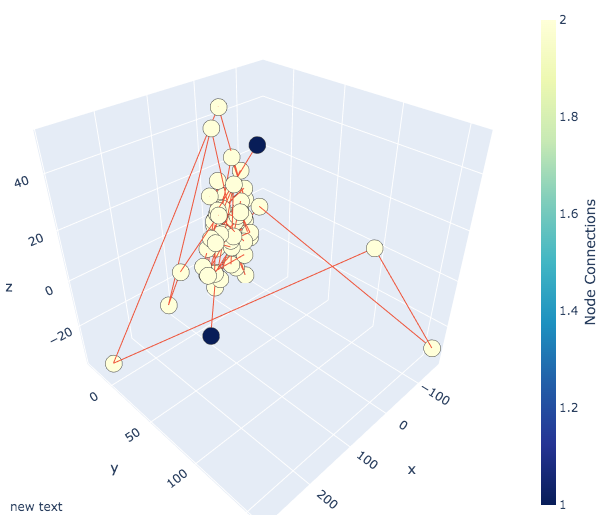}
          \includegraphics[width=\textwidth]{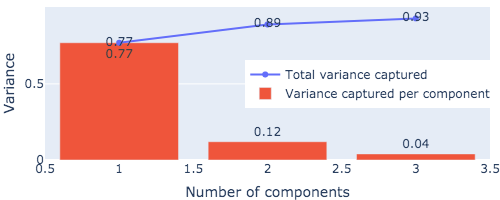}
          \caption{``And while Poroshenko's decree may provide some cover for a U.S.-led military intervention that will be required to keep the state's borders safe from attack: the main obstacle to a U.S.-led war in the region: and perhaps the first step in the next step in his agenda: is the potential military involvement of Russian forces in the war.''}
         \label{fig:46}
    \end{subfigure}
\end{figure}
\begin{figure}\ContinuedFloat
    \begin{subfigure}[b]{0.49\textwidth}
         \includegraphics[width=\textwidth]{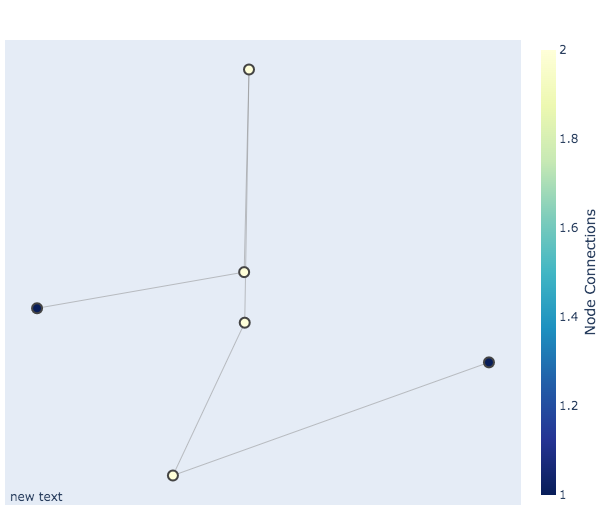}
         \includegraphics[width=\textwidth]{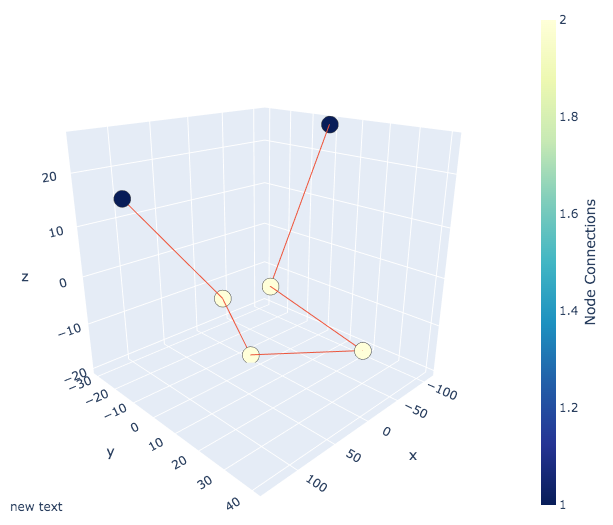}
         \includegraphics[width=\textwidth]{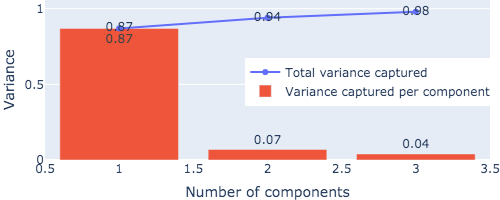}
         \caption{``This is no big deal.''}
          \label{fig:47}
    \end{subfigure}
    \begin{subfigure}[b]{0.49\textwidth}
         \includegraphics[width=\textwidth]{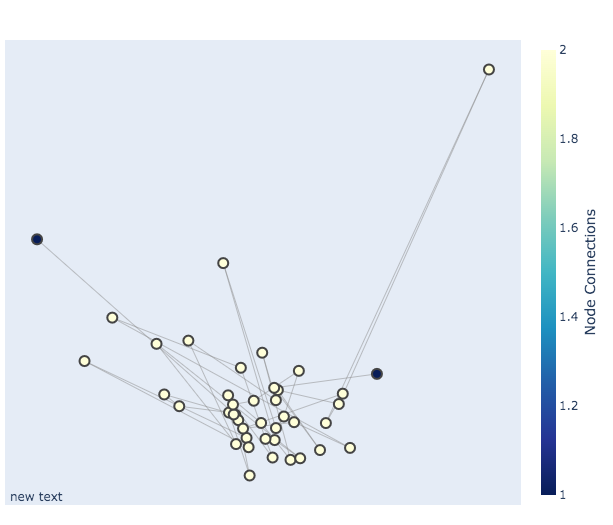}
         \includegraphics[width=\textwidth]{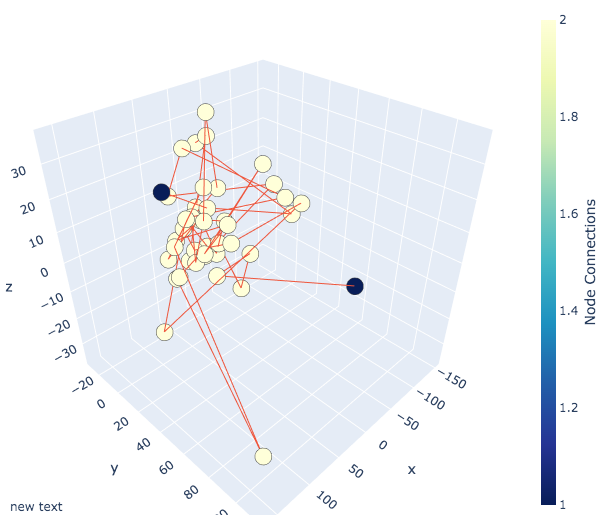}
         \includegraphics[width=\textwidth]{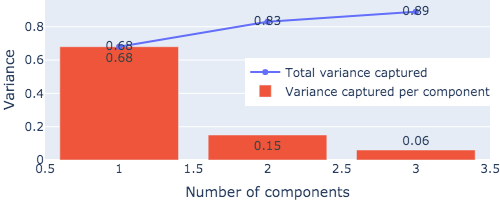}
         \caption{``But the problem may be that: for now: the new order is a huge leap for Ukraine: where Poroshenko's new powers mean he has little time to push through a complex set of legislation and regulations.''}
         \label{fig:48}
    \end{subfigure}
\end{figure}
\begin{figure}\ContinuedFloat
    \begin{subfigure}[b]{0.49\textwidth}
         \includegraphics[width=\textwidth]{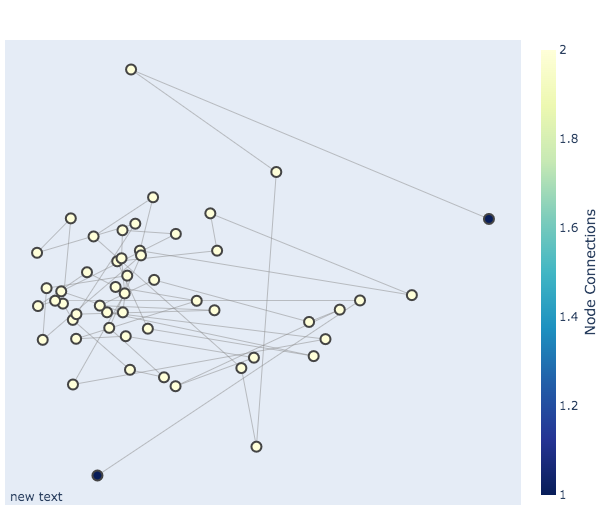}
         \includegraphics[width=\textwidth]{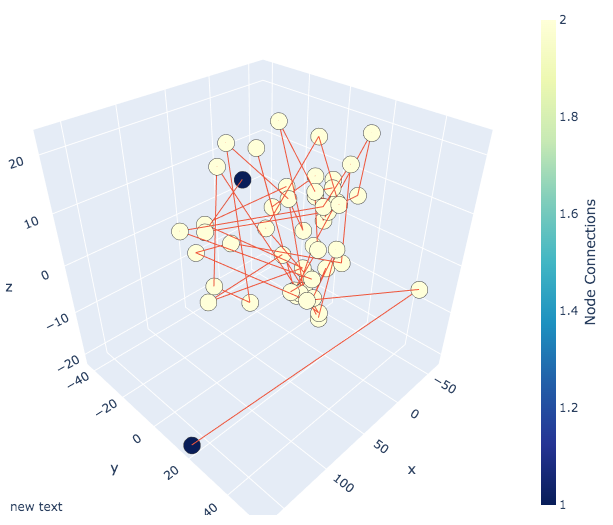}
         \includegraphics[width=\textwidth]{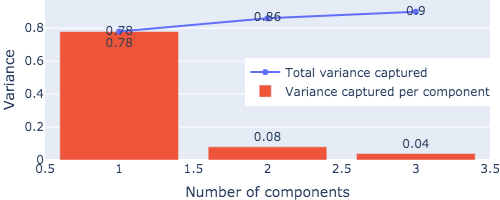}
         \caption{``While many of Poroshenko's initiatives are focused on keeping the situation in Kiev stable -- the Ukrainian police and its security service are still patrolling the country: and their forces are fighting insurgents in the southeastern sector of the country -- his approach is actually not that different.''}
         \label{fig:49}
    \end{subfigure}
    \begin{subfigure}[b]{0.49\textwidth}
         \includegraphics[width=\textwidth]{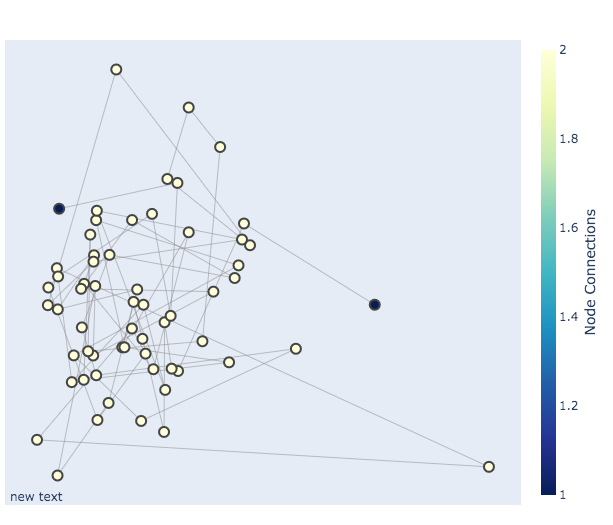}
         \includegraphics[width=\textwidth]{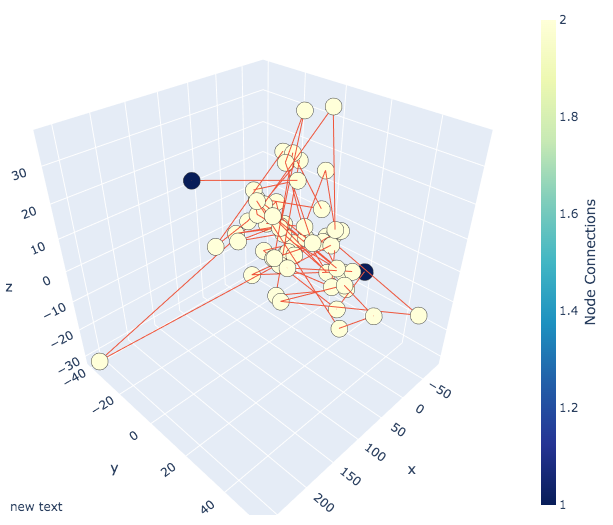}
         \includegraphics[width=\textwidth]{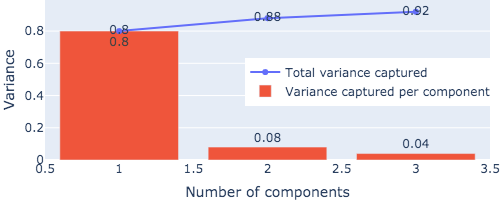}
         \caption{``Poroshenko's move may be a symbolic step toward a new political system that will be more democratic: not less: and more accountable to the people: many of whom he now calls ""honest Ukrainians."" But for Kiev: it means they are beginning to wonder whether they are being manipulated by Russia:''}
         \label{fig:50}
    \end{subfigure} 
\end{figure}
\begin{figure}\ContinuedFloat
    \begin{subfigure}[b]{0.49\textwidth}
         \includegraphics[width=\textwidth]{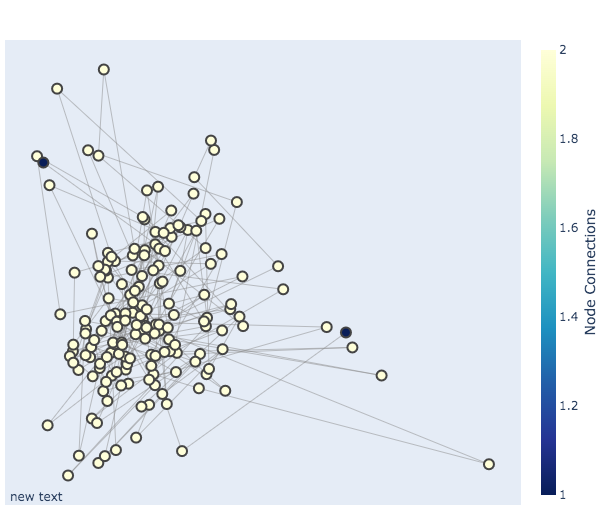}
         \includegraphics[width=\textwidth]{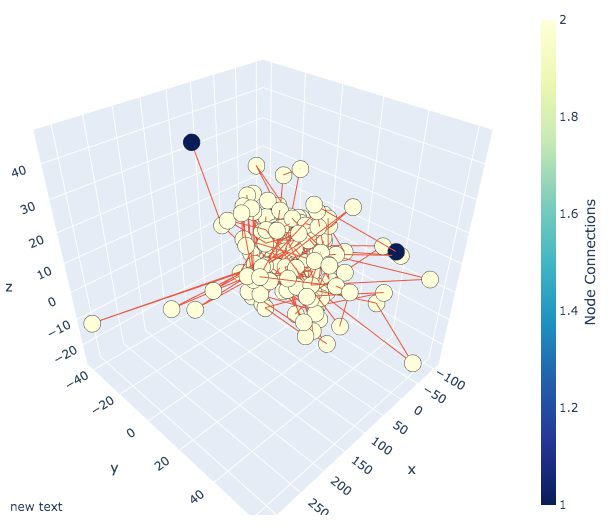}
         \includegraphics[width=\textwidth]{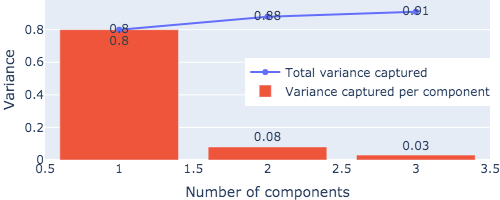}
         \caption{``There is no way of knowing who does this to me: for sure -- and I mean it completely and in both capitals. My question is: can I have a conversation about the legitimacy of this arrangement that is actually just as good as the arrangement we agreed on: that will protect those of us who do not share our values: and that will also be good for the country? ... The government in Kiev has done two things by decree: It has brought in new legislative committee: which is based on the principle that the majority of votes from the electorate is needed for an election. It has also created some parliamentary committee which will be based on a concept that we had in 2013: called the ""Kremlin Committee to Protect Elections."" It has already introduced legislation to do the same thing. And there are very concrete measures within it: to get better security and more accountability.''}
         \label{fig:51}
    \end{subfigure}
    \begin{subfigure}[b]{0.49\textwidth}
         \includegraphics[width=\textwidth]{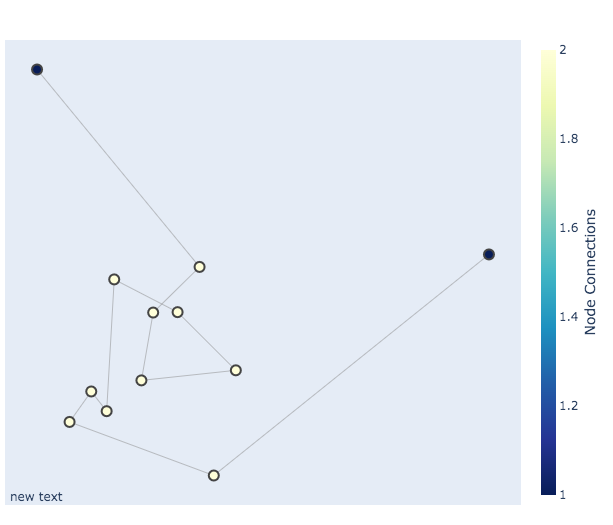}
         \includegraphics[width=\textwidth]{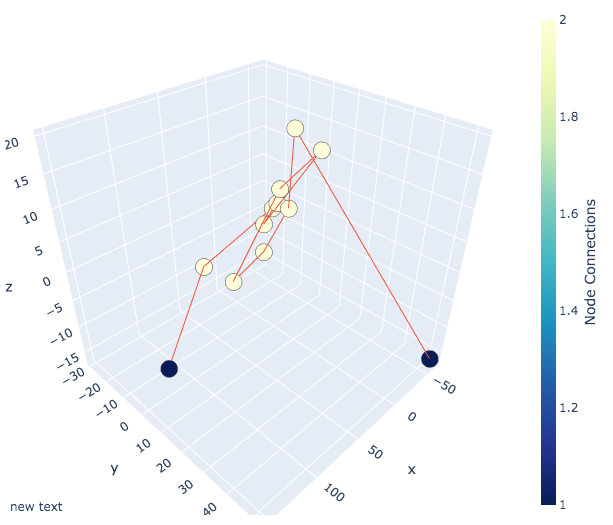}
         \includegraphics[width=\textwidth]{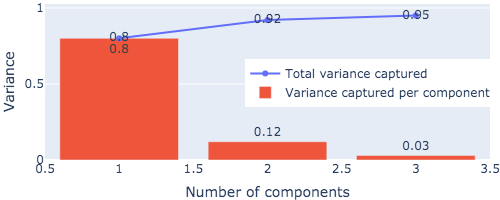}
         \caption{``But that was not the main priority of the new government.''}
         \label{fig:52}
    \end{subfigure}
\end{figure}
\begin{figure}
    \begin{subfigure}[b]{0.49\textwidth}
         \includegraphics[width=\textwidth]{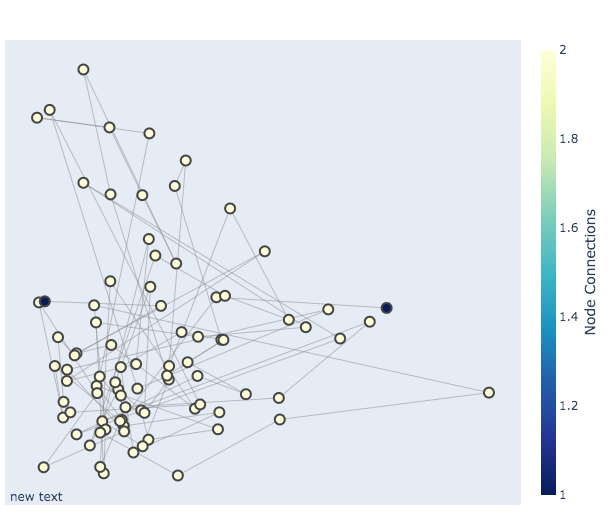}
         \includegraphics[width=\textwidth]{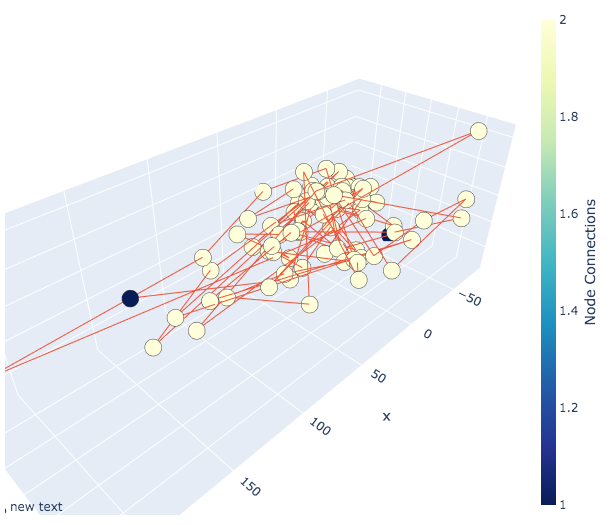}
         \includegraphics[width=\textwidth]{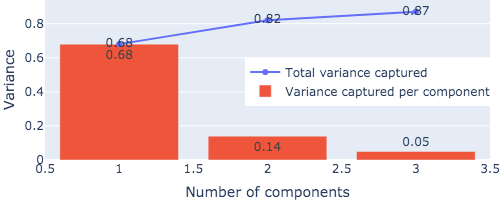}
         \caption{``The change was announced by the new administration's press secretary: Yulia Tymoshenko -- not only was the new new prime minister: but the head of the new Komsomolskaya Pravda: also named as the secretary-general: who will be a member of both the Kiev and the Donbass political parties. That leaves a few key issues that have been at the heart of the current Ukrainian political situation.''}
         \label{fig:53}
    \end{subfigure}
    \begin{subfigure}[b]{0.49\textwidth}
         \includegraphics[width=\textwidth]{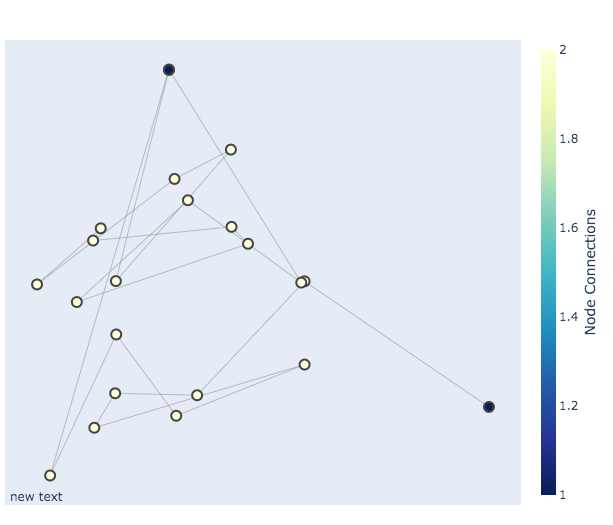}
         \includegraphics[width=\textwidth]{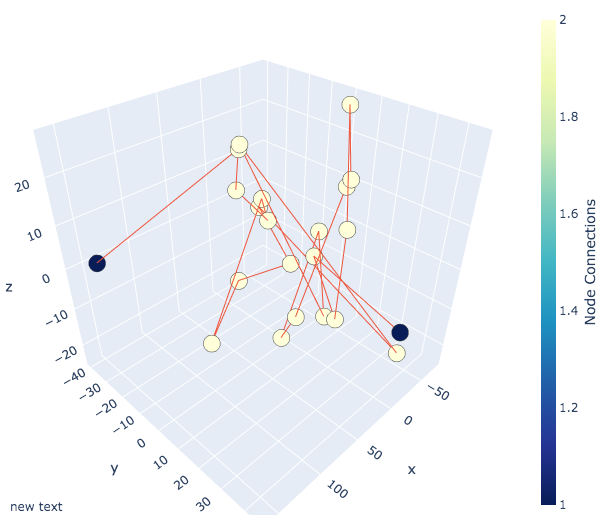}
         \includegraphics[width=\textwidth]{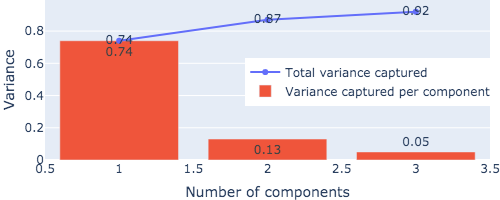}
         \caption{``Poroshenko has already vowed that he will ""promote and protect"" the new political system: which"''}
         \label{fig:54}
    \end{subfigure}
\end{figure}
\begin{figure}\ContinuedFloat
    \begin{subfigure}[b]{0.49\textwidth}
         \includegraphics[width=\textwidth]{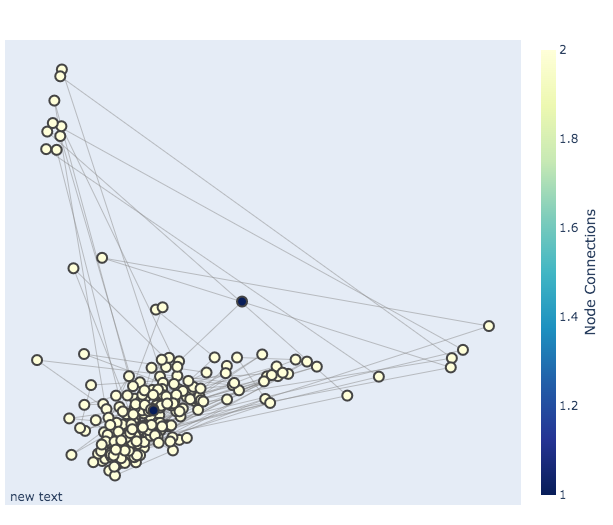}
         \includegraphics[width=\textwidth]{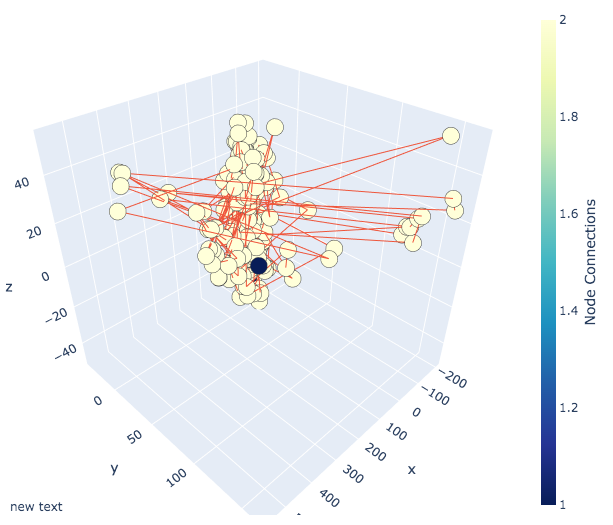}
         \includegraphics[width=\textwidth]{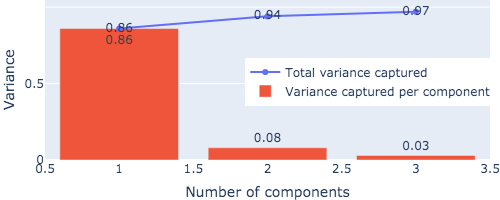}
         \caption{``8:"K.P. is the founder of the K.P. Network: a decentralized marketplace for social media companies. K.P. launched the first decentralised online stock exchange in May of 2012. Its first platform: K.P. Market: was created in September of this year and has since gained widespread support and acceptance. The founder is the founder: with K.P. founder Arvind Subramanian as co-founder and CEO . Before joining K.P. on May 31: 2008: K.P. was a private investment company and managed a hedge fund with financial backing from a couple of private equity firms. He also acted as a private equity investor and CEO from 1999-2006 while he was still at K.P. He worked at BNSF until 2004-2005: after which he joined BNP Paribas. Since 2008: he has been part of the K.P. community on Twitter as well as on the BNSF platform. As of April 2016: he is on Twitter's board of directors and is currently ranked among the top twenty companies in the world by Forbes (the world's most-read website: according to a study released in 2013 by the Forbes Magazine) and Forbes in May 2015."''}
         \label{fig:55}
    \end{subfigure}
    \begin{subfigure}[b]{0.49\textwidth}
         \includegraphics[width=\textwidth]{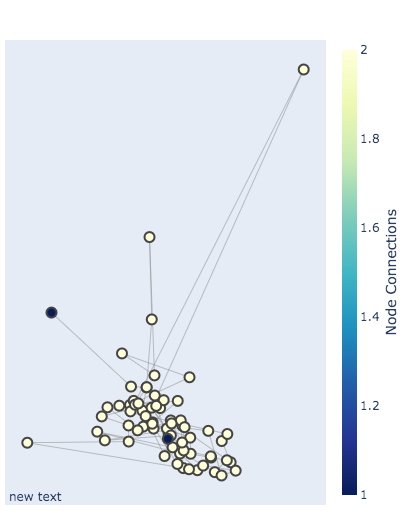}
         \includegraphics[width=\textwidth]{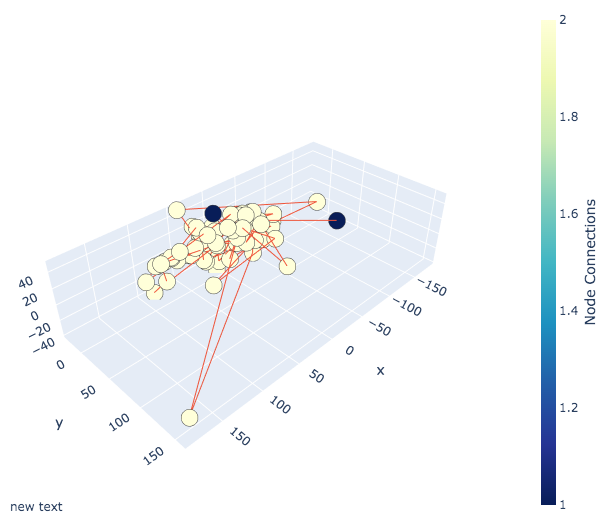}
         \includegraphics[width=\textwidth]{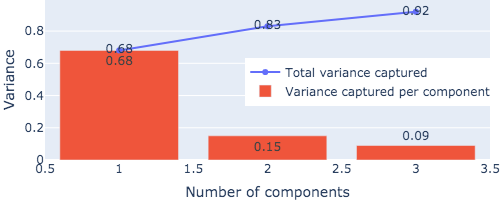}
         \caption{``Riot Squad Roush is currently working the situation in the city as part of C4A: providing security for businesses who've been affected by the protest. In an interview here: Riot said: ""The Riot Squad is now operating the emergency check area and we have secured the main entrance to the main site with riot shields.""''}
         \label{fig:56}
    \end{subfigure}
\end{figure}
\begin{figure}\ContinuedFloat
    \begin{subfigure}[b]{0.49\textwidth}
         \includegraphics[width=\textwidth]{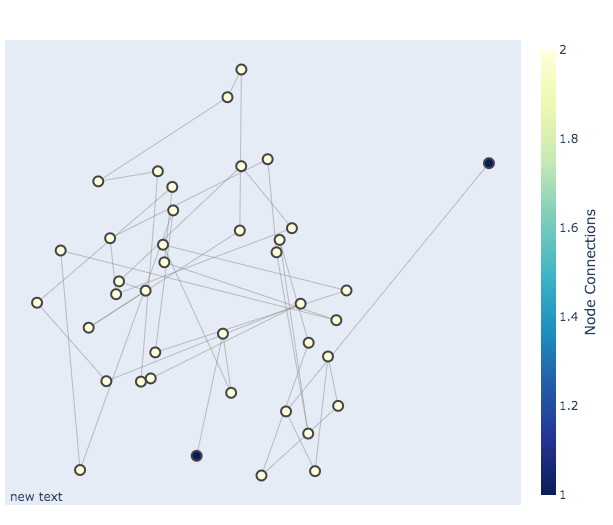}
         \includegraphics[width=\textwidth]{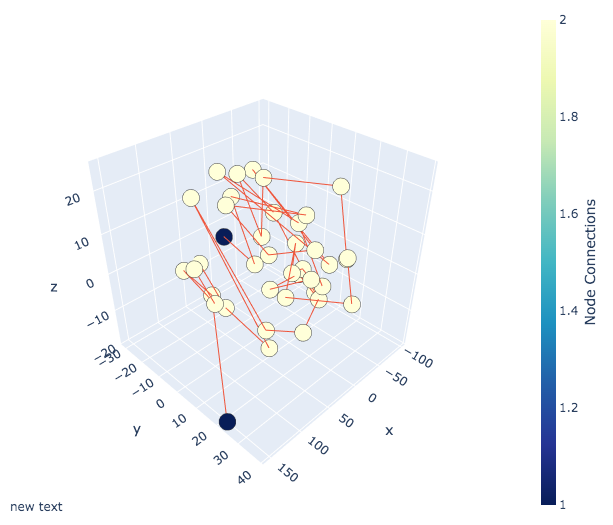}
         \includegraphics[width=\textwidth]{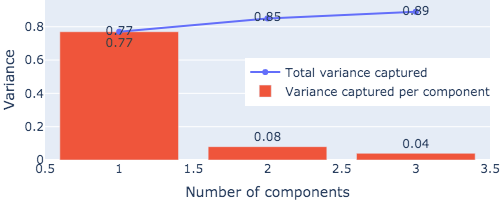}
         \caption{``11:"Bishop Dallmann of North Texas said she was inspired when her husband was asked about whether women had no rights: and she called for men to be given equal access to abortion services.''}
         \label{fig:57}
    \end{subfigure}
    \begin{subfigure}[b]{0.49\textwidth}
         \includegraphics[width=\textwidth]{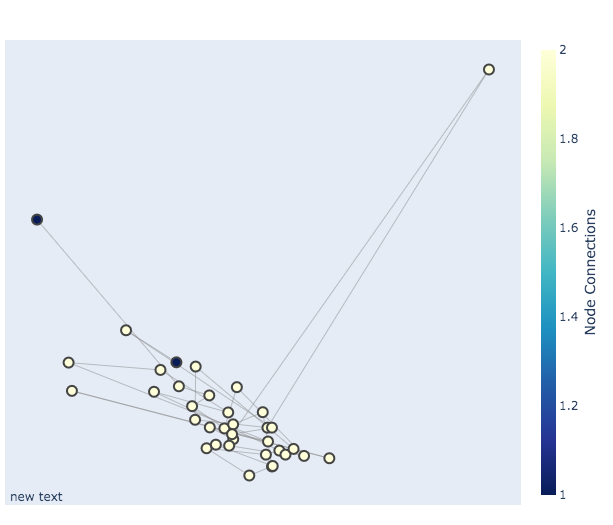}
         \includegraphics[width=\textwidth]{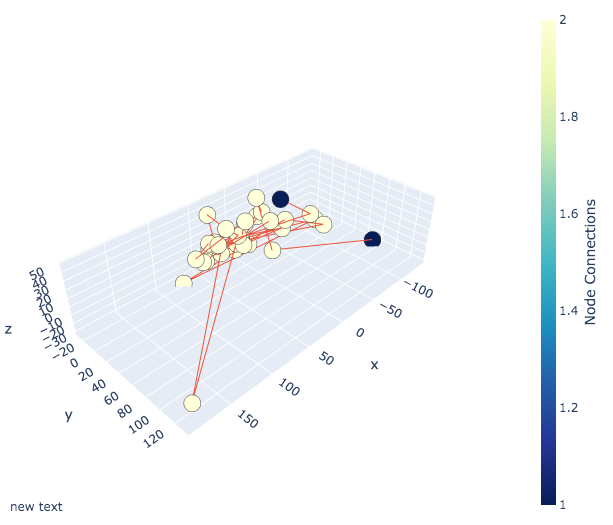}
         \includegraphics[width=\textwidth]{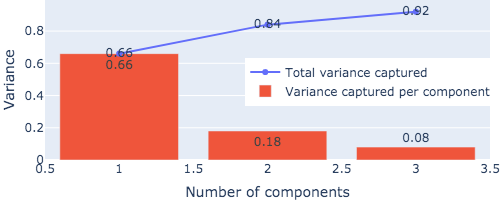}
         \caption{``She said ""all the men who are on the other side would like to give them the abortion: even if that means giving away their own organs in the name of their faith.""''}
         \label{fig:58}
    \end{subfigure}
\end{figure}
\begin{figure}\ContinuedFloat
    \begin{subfigure}[b]{0.49\textwidth}
         \includegraphics[width=\textwidth]{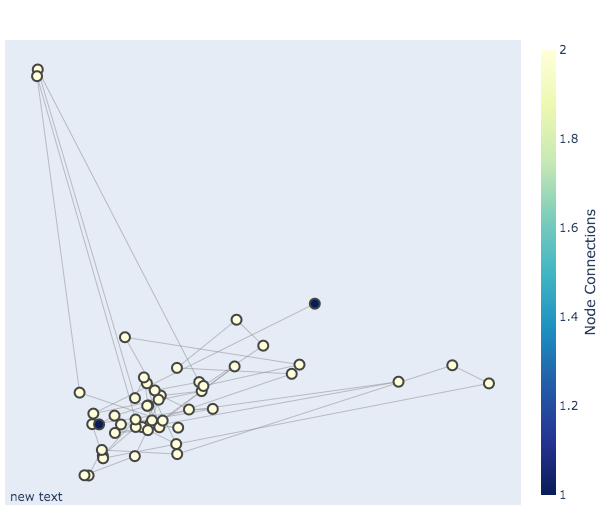}
         \includegraphics[width=\textwidth]{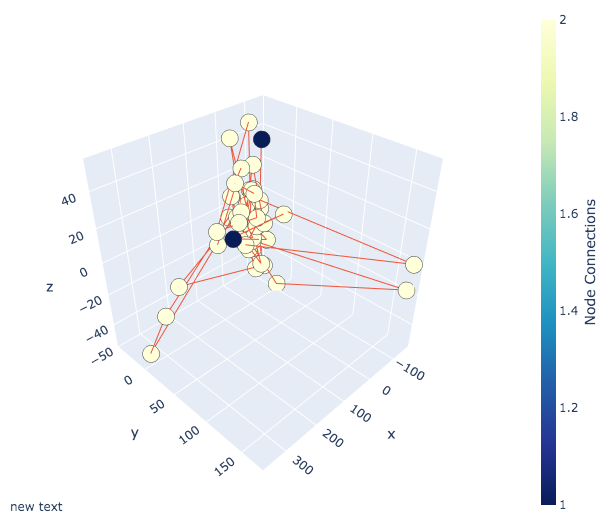}
         \includegraphics[width=\textwidth]{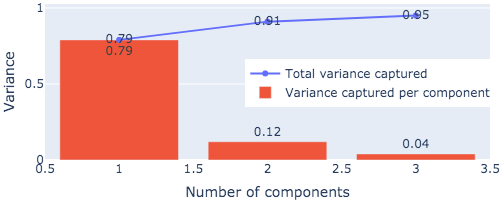}
         \caption{``At least 12 states allow women to withhold their rights to elect and provide abortion services: according to the National Alliance for Pregnant Women. But it doesn't make sense for these men to be allowed access if women are only given reproductive health care.''}
         \label{fig:59}
    \end{subfigure}
    \begin{subfigure}[b]{0.49\textwidth}
         \includegraphics[width=\textwidth]{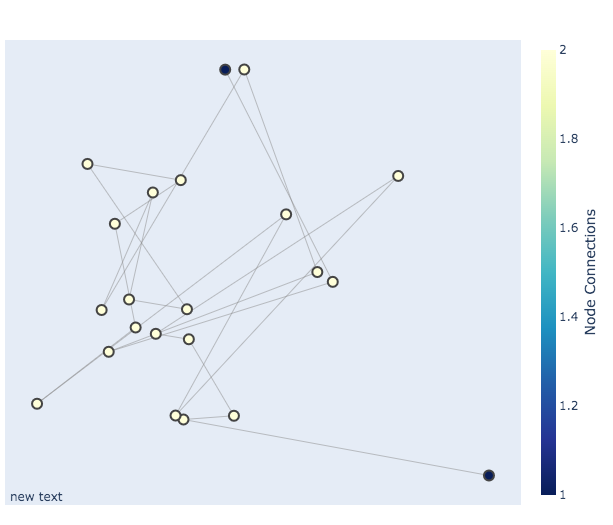}
         \includegraphics[width=\textwidth]{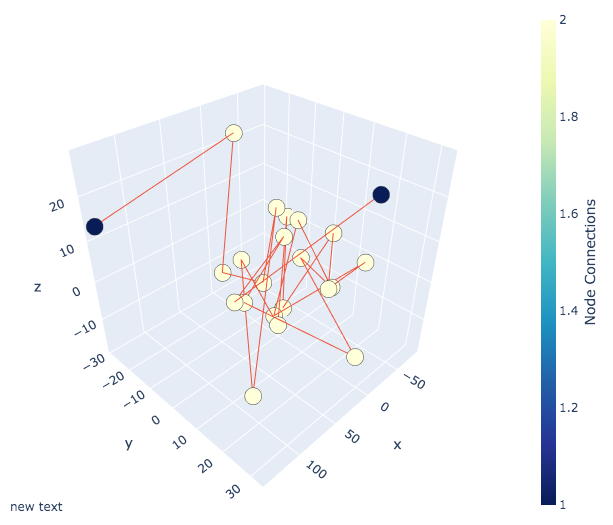}
         \includegraphics[width=\textwidth]{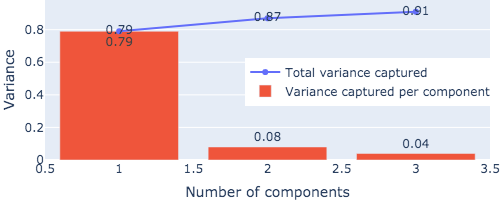}
         \caption{``The group advocates for the right of women to have abortions: and believes the law is inhumane and discriminatory.''}
         \label{fig:60}
    \end{subfigure} 
\end{figure}
\begin{figure}\ContinuedFloat
    \begin{subfigure}[b]{0.49\textwidth}
         \includegraphics[width=\textwidth]{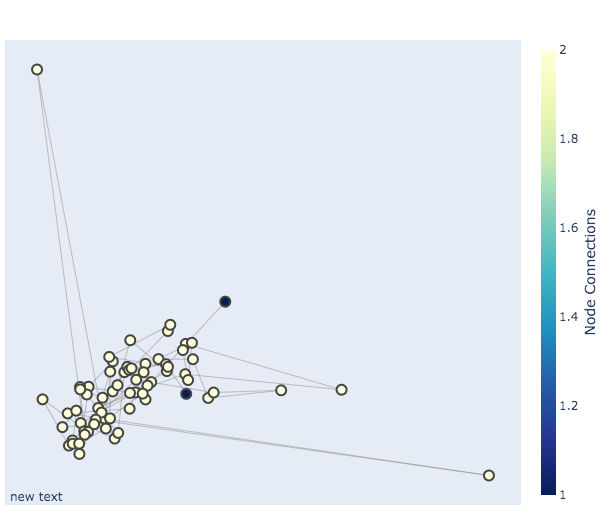}
         \includegraphics[width=\textwidth]{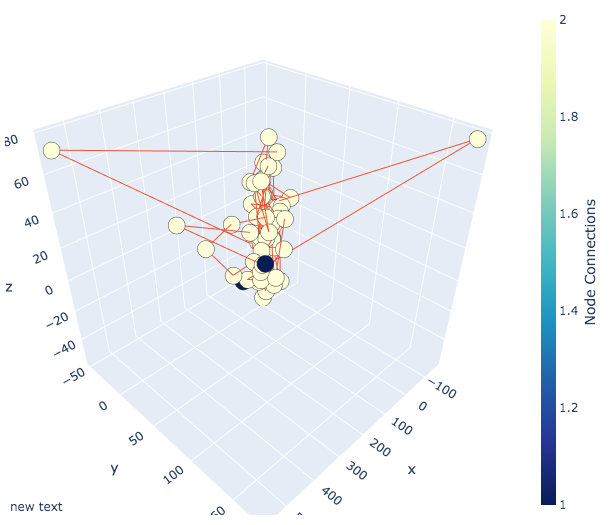}
         \includegraphics[width=\textwidth]{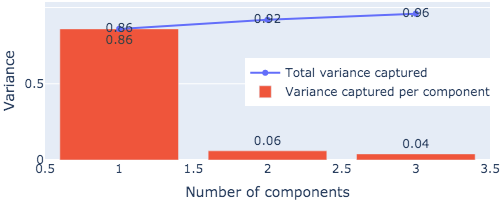}
         \caption{``""It only makes sense for women to have these health services: who aren't legally able to do them: and I hope that other states see it as something that goes against the fundamental values that the state holds such and such:"" said Dr. Ann B. Seibel: president of the Catholic Alliance for Pregnant Women.''}
         \label{fig:61}
    \end{subfigure}
    \begin{subfigure}[b]{0.49\textwidth}
         \includegraphics[width=\textwidth]{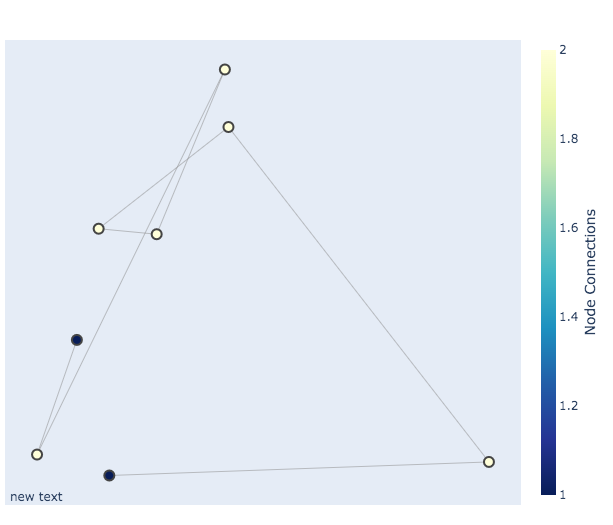}
         \includegraphics[width=\textwidth]{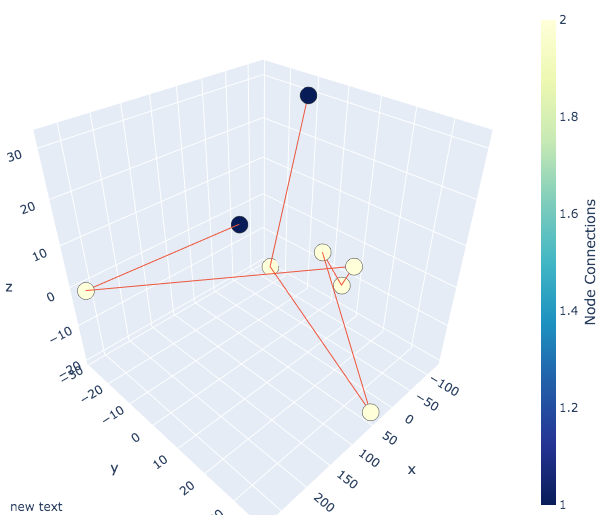}
         \includegraphics[width=\textwidth]{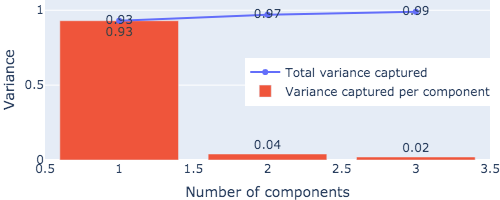}
         \caption{``The Associated Press contributed to this report."''}
         \label{fig:62}
    \end{subfigure}
\end{figure}
\begin{figure}\ContinuedFloat
    \begin{subfigure}[b]{0.49\textwidth}
         \includegraphics[width=\textwidth]{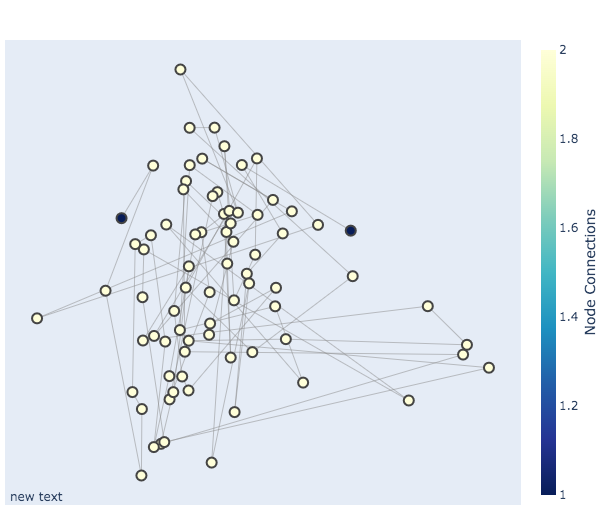}
         \includegraphics[width=\textwidth]{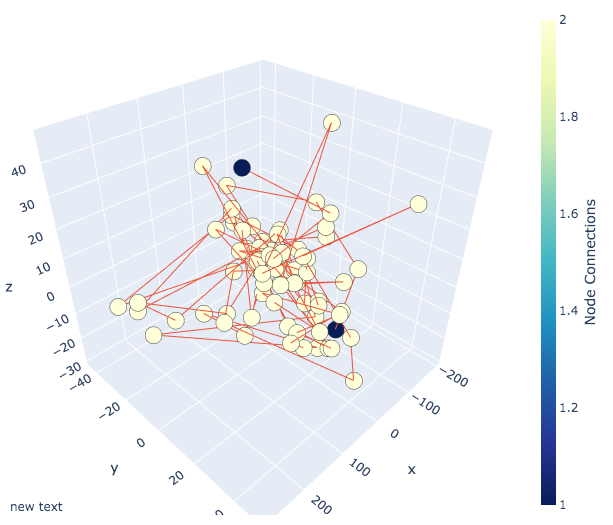}
         \includegraphics[width=\textwidth]{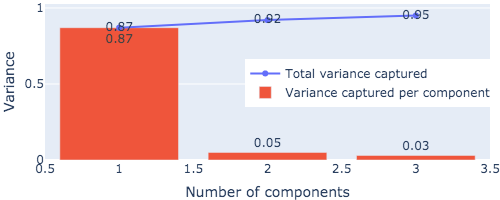}
         \caption{``13:"The world's third greatest mountain lion: a rare breed that ranges in size from 20:600 to 40:000 m: belongs to a subspecies of lion: and is the largest in the European Union. Its ability to detect danger is due largely to its ability to run. However: because of its long hair: it can take a beating and can occasionally run away.''}
         \label{fig:63}
    \end{subfigure}
    \begin{subfigure}[b]{0.49\textwidth}
         \includegraphics[width=\textwidth]{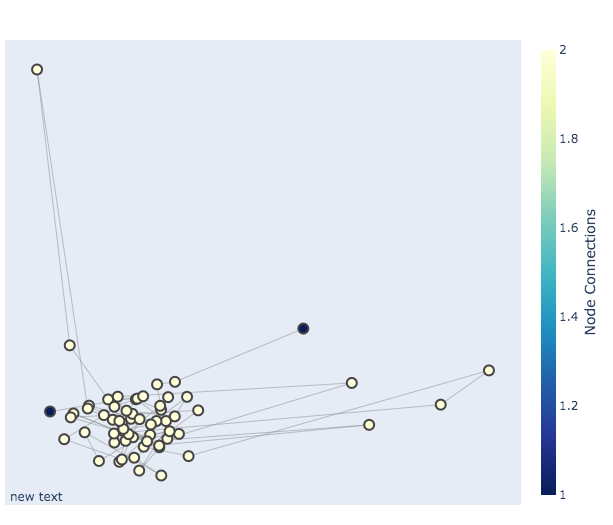}
         \includegraphics[width=\textwidth]{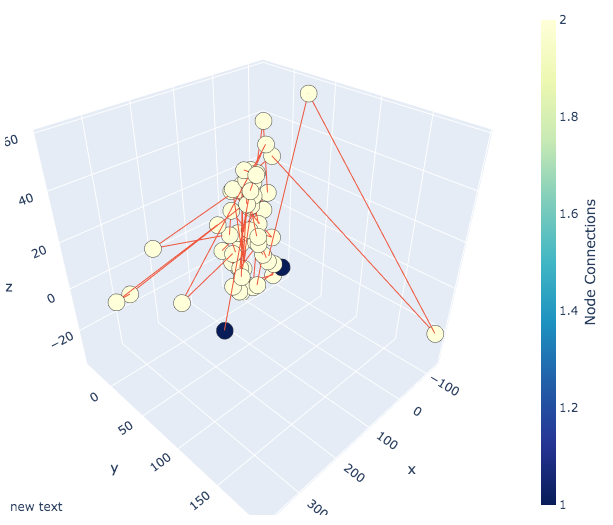}
         \includegraphics[width=\textwidth]{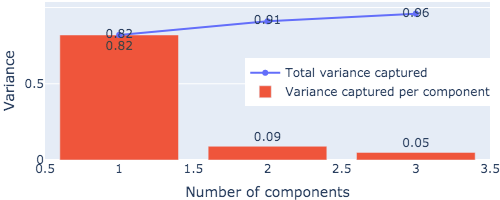}
         \caption{``But these animals have a weakness that is not a problem for humans - the ability to fight. If any lion encounters or bites another lion: it immediately dies: usually with fatal injury. This is due largely to the fact that lions generally fight with other animals because predators don't go in the same direction.''}
         \label{fig:64}
    \end{subfigure}
\end{figure}
\begin{figure}\ContinuedFloat
    \begin{subfigure}[b]{0.49\textwidth}
         \includegraphics[width=\textwidth]{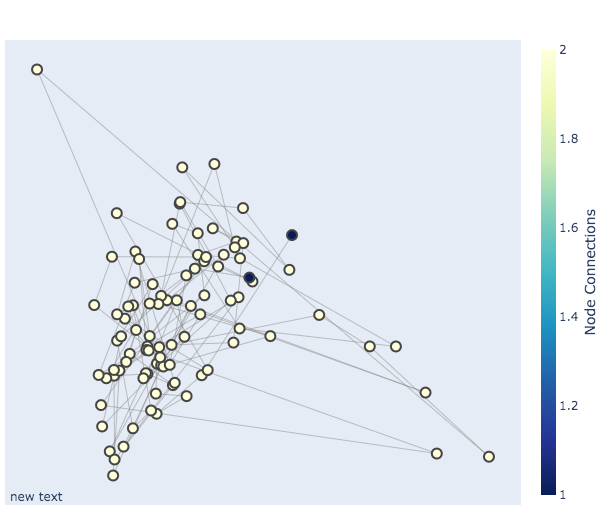}
         \includegraphics[width=\textwidth]{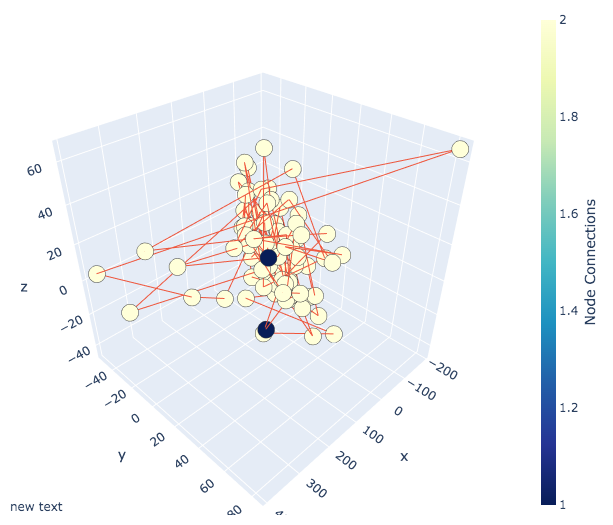}
         \includegraphics[width=\textwidth]{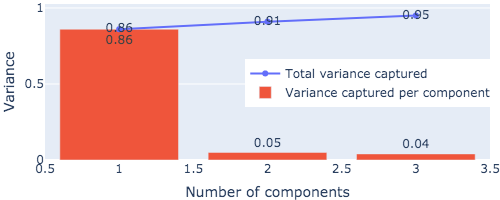}
         \caption{``""These animals do survive the whole ordeal: but only if they are able to adapt and survive within those confines:"" said Dr. William Nesbit from the New Zealand Department of Animal Research and the U.K.'s University of KwaZulu-Natal. ""Unfortunately: the lions may not be good at finding and catching prey because they are too slow to deal with threats at large - sometimes it could take several months for one to break free.""''}
         \label{fig:65}
    \end{subfigure}
    \begin{subfigure}[b]{0.49\textwidth}
         \includegraphics[width=\textwidth]{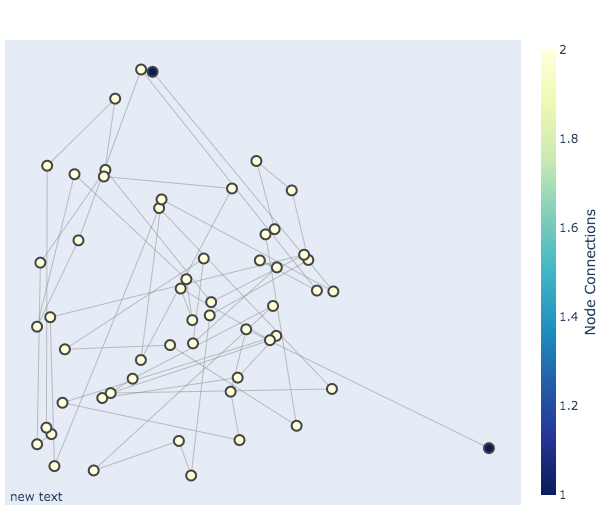}
         \includegraphics[width=\textwidth]{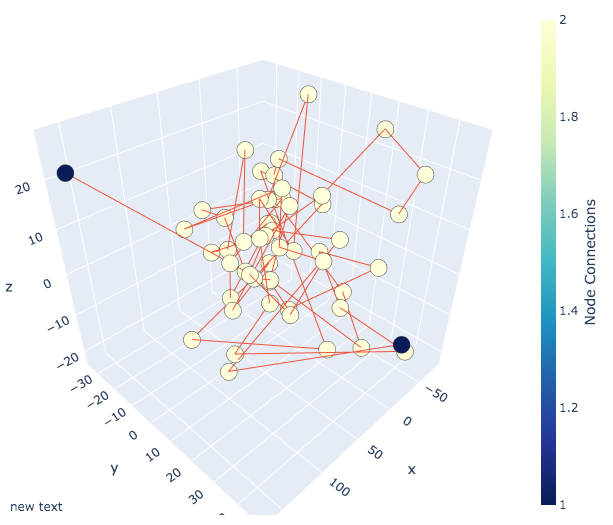}
         \includegraphics[width=\textwidth]{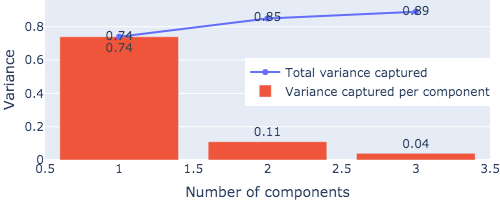}
         \caption{``""Their survival rate is one of the only means to defend against predators:"" Nesbit and his colleagues reported in Nature Communications. ""But even to survive in such conditions as this: they are hardy: and they will not be a bad pick for hunters."""''}
         \label{fig:66}
    \end{subfigure}
\end{figure}
\begin{figure}\ContinuedFloat
    \begin{subfigure}[b]{0.49\textwidth}
         \includegraphics[width=\textwidth]{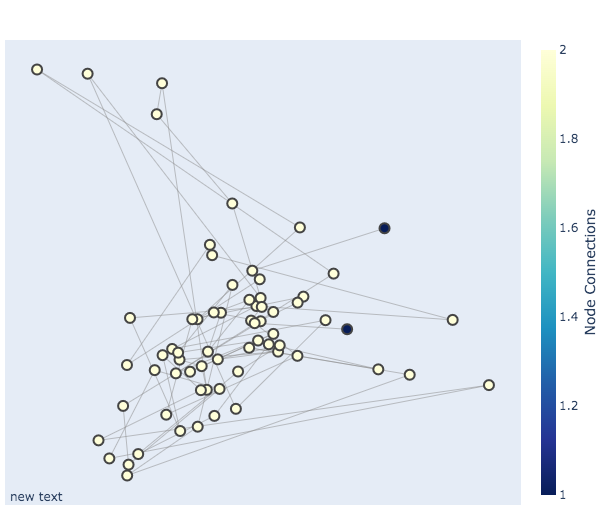}
         \includegraphics[width=\textwidth]{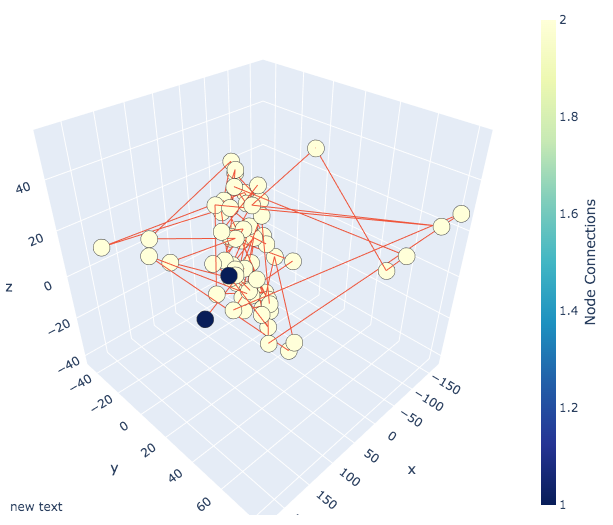}
         \includegraphics[width=\textwidth]{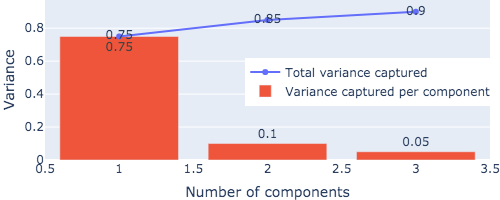}
         \caption{``RALEIGH: N.C. — North Carolina Gov. Pat McCrory on Thursday signed bills into law to allow people across the state to purchase health insurance on the state's open-and-shut network: as state law enforcement agencies investigate those who try to purchase insurance without a driver's license or valid driver's license.''}
         \label{fig:67}
    \end{subfigure}
    \begin{subfigure}[b]{0.49\textwidth}
         \includegraphics[width=\textwidth]{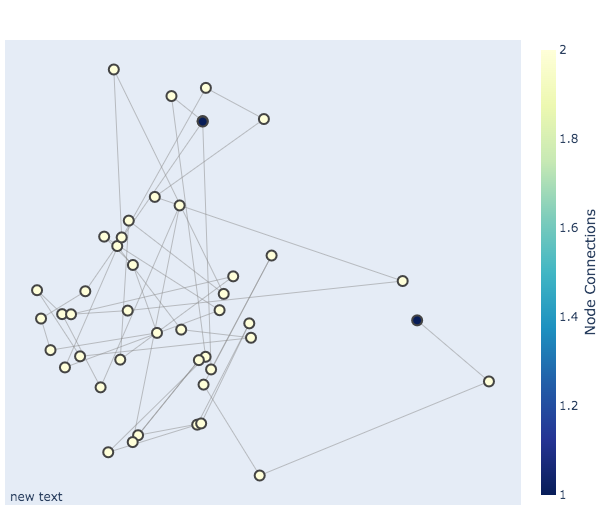}
         \includegraphics[width=\textwidth]{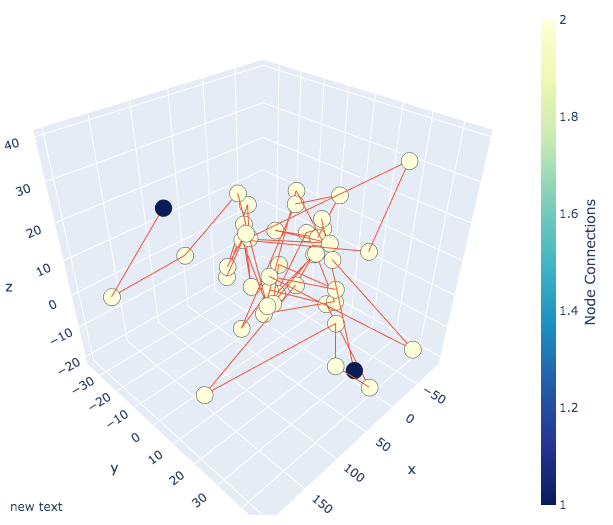}
         \includegraphics[width=\textwidth]{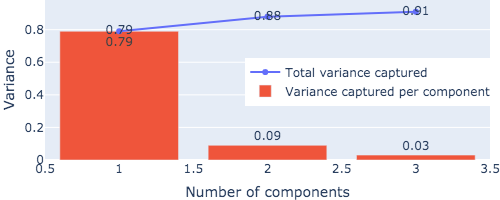}
         \caption{``In response to questions: the McCrory administration announced Thursday that the state's health exchanges were being closed because of an ""unexpected increase in violent crime and unsafe driving."" It also released an update on the new regulations.''}
         \label{fig:68}
    \end{subfigure}
\end{figure}
\begin{figure}\ContinuedFloat
    \begin{subfigure}[b]{0.49\textwidth}
         \includegraphics[width=\textwidth]{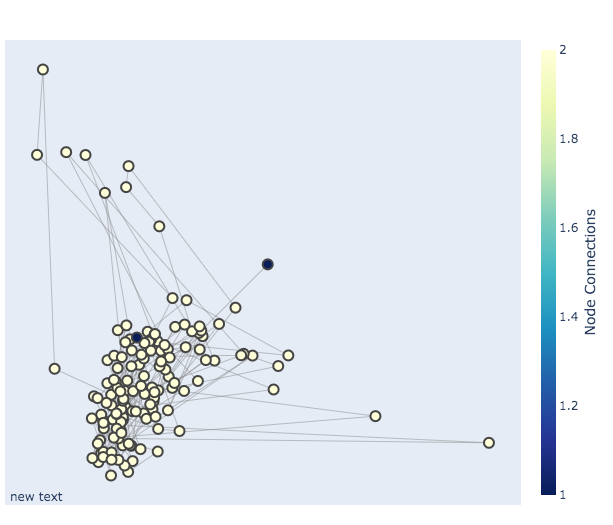}
         \includegraphics[width=\textwidth]{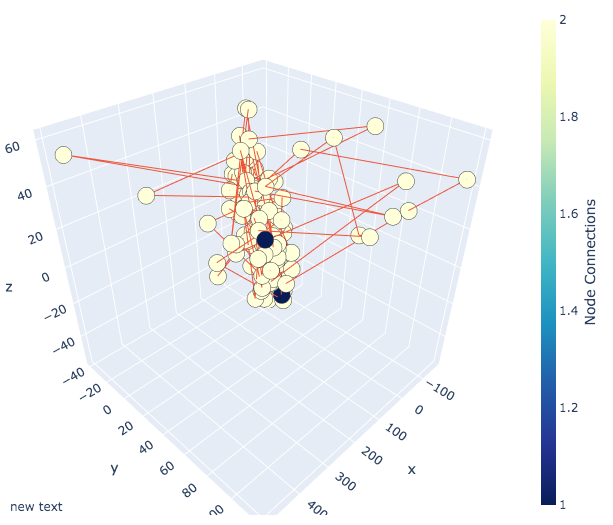}
         \includegraphics[width=\textwidth]{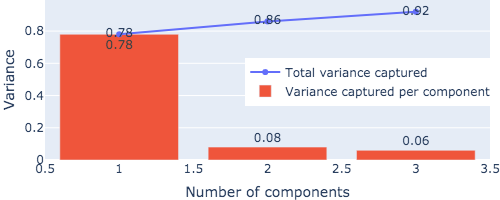}
         \caption{``""We're working closely with law enforcement efforts to try to increase safety across this state:"" said North Carolina Gov. Pat McCrory in a statement. ""We are committed to ensuring that our employees and our state's residents are making a responsible and responsible choice when it comes to accessing a health care plan in the state of North Carolina."" The new regulations came into effect on Jan. 1: 2016: but will only take effect for ""emergency circumstances:"" meaning the federal government cannot seize data from private corporations. A state spokeswoman refused to comment on whether the law would also require that NC drivers and insurance companies are notified of ""unexpected spikes in incidents of violent crime and unsafe driving.""''}
         \label{fig:69}
    \end{subfigure}
    \begin{subfigure}[b]{0.49\textwidth}
         \includegraphics[width=\textwidth]{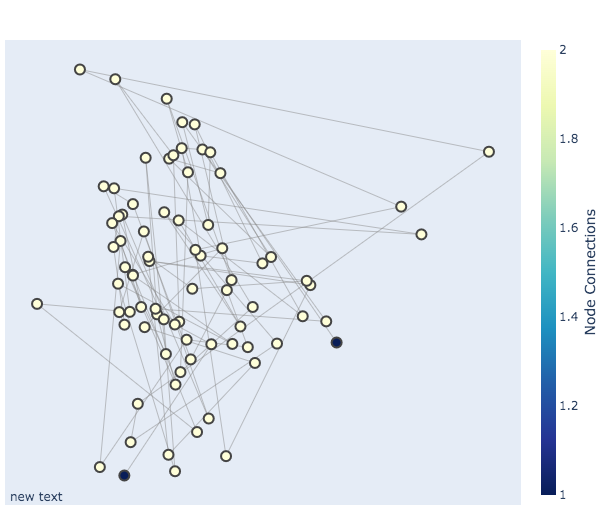}
         \includegraphics[width=\textwidth]{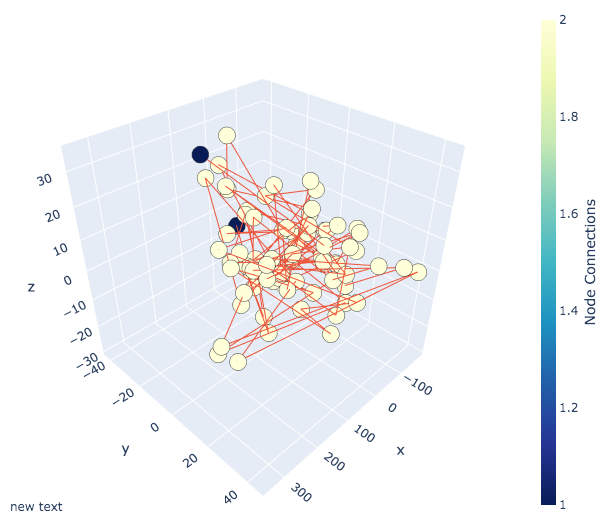}
         \includegraphics[width=\textwidth]{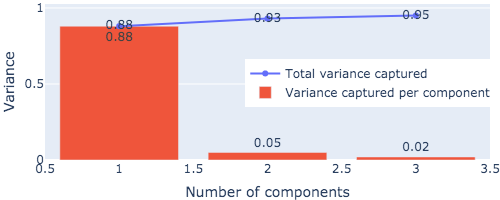}
         \caption{``North Carolina is one of six states in the country that require insurance companies to provide medical data to all their customers: regardless of whether it results in accidents or not. In order to qualify for tax credits or subsidies: the NC General Assembly created a national network of health insurers that provides coverage across North Carolina. The state's health insurance exchange network already offers access to more than 16 million people nationwide.''}
         \label{fig:70}
    \end{subfigure}
\end{figure}
\begin{figure}\ContinuedFloat
    \begin{subfigure}[b]{0.49\textwidth}
         \includegraphics[width=\textwidth]{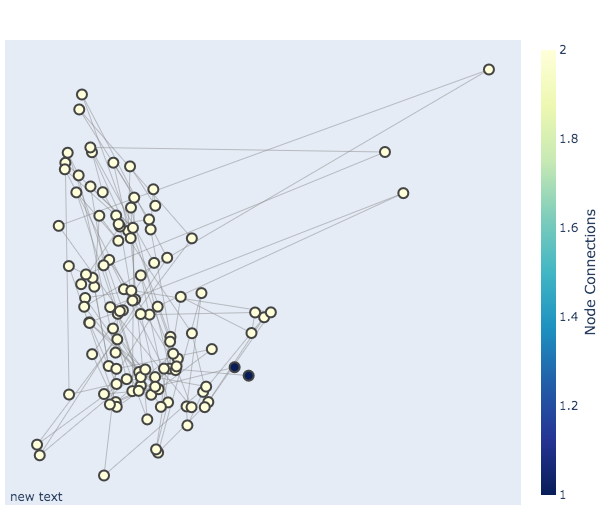}
         \includegraphics[width=\textwidth]{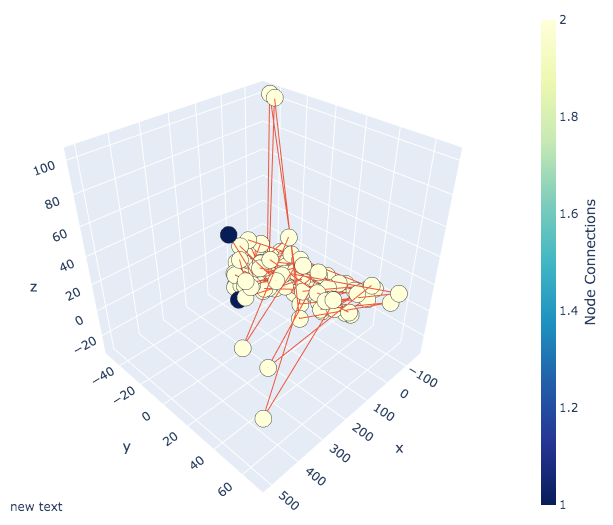}
         \includegraphics[width=\textwidth]{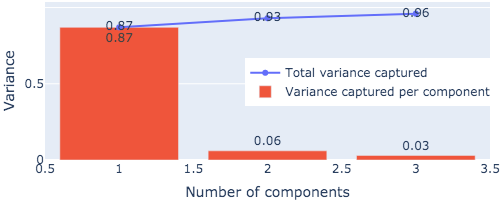}
         \caption{``""It's an important step forward in making sure that these people have insurance as soon as they sign up. We believe that the governor can make this important change while still taking steps to ensure that they have health insurance as soon as they make that decision:"" said Richard N. Anderson: CEO and chief executive officer of Public Citizen's North Carolina Department of Insurance. ""As Governor North Carolina: I am proud to introduce two bills that will create greater public confidence in our government's ability to manage the federal government's health care: reducing health insurance premiums even more.""''}
         \label{fig:71}
    \end{subfigure}
    \begin{subfigure}[b]{0.49\textwidth}
         \includegraphics[width=\textwidth]{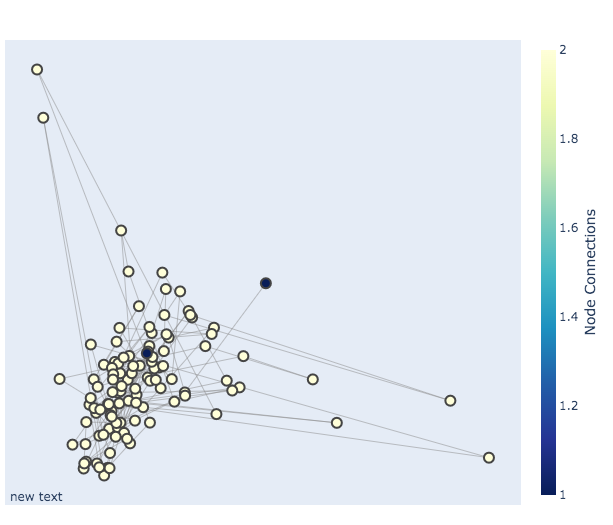}
         \includegraphics[width=\textwidth]{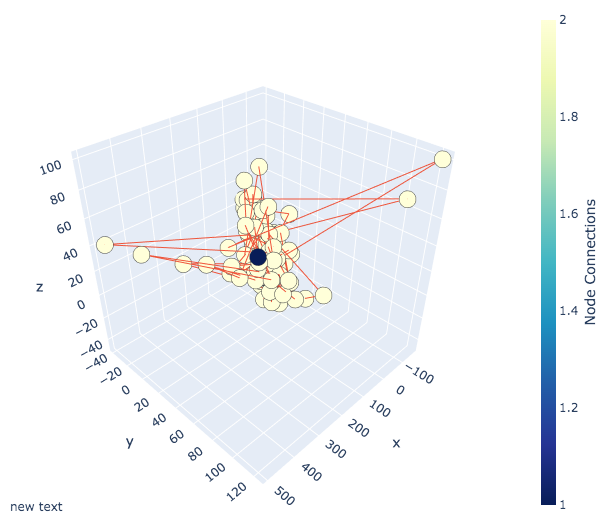}
         \includegraphics[width=\textwidth]{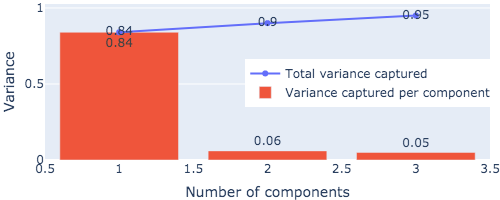}
         \caption{``According to state figures the federal government holds roughly 17.5 million people in health insurance and does not account for a handful of states as a whole. As part of the law: McCrory said it would require federal money to go to state-based health insurance exchanges: which he said would result in ""increases in uninsured persons: resulting in thousands of people who are denied coverage by the government."" The federal grant will not be matched to the states because they still have their own government-run exchanges: N.C. News reported.''}
         \label{fig:72}
    \end{subfigure}
\end{figure}
\begin{figure}\ContinuedFloat
    \begin{subfigure}[b]{0.49\textwidth}
         \includegraphics[width=\textwidth]{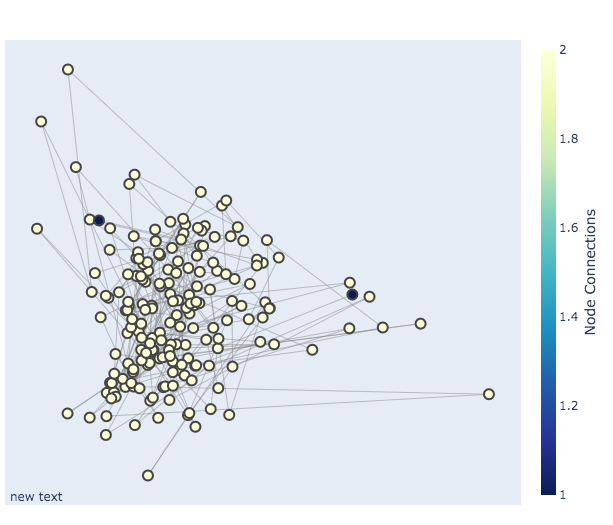}
         \includegraphics[width=\textwidth]{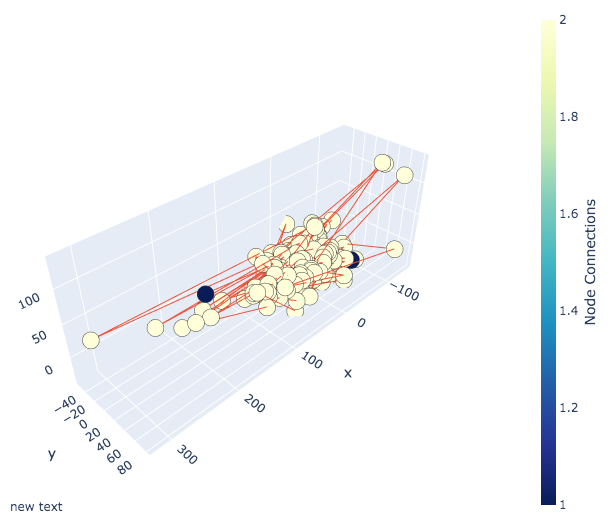}
         \includegraphics[width=\textwidth]{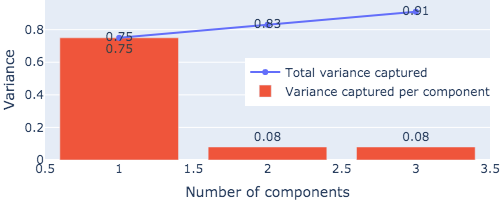}
         \caption{``The federal law requires states to provide consumers with information about their insurance when an insurance policy is required in their state: such as the type of person enrolled: whether they have access to health care or what type of medical coverage they have. Other information that is included in the individual form is included in the policy: such as physician reports and required documentation to make informed decisions about what kind of care he or she plans to have. All state-based health exchanges provide an ""incomplete"" list of health care policies and have no information on whether consumers are on a pre-existing condition or have an ""emergency illness."" While all states provide complete information: states can limit how long it may take to make a new health plan through insurance regulators. In some states: insurers have to offer an additional health plan for the person without a health card or pre-existing condition: a policy that can cost up to \$6:000 on average per year.''}
         \label{fig:73}
    \end{subfigure}
    \begin{subfigure}[b]{0.49\textwidth}
         \includegraphics[width=\textwidth]{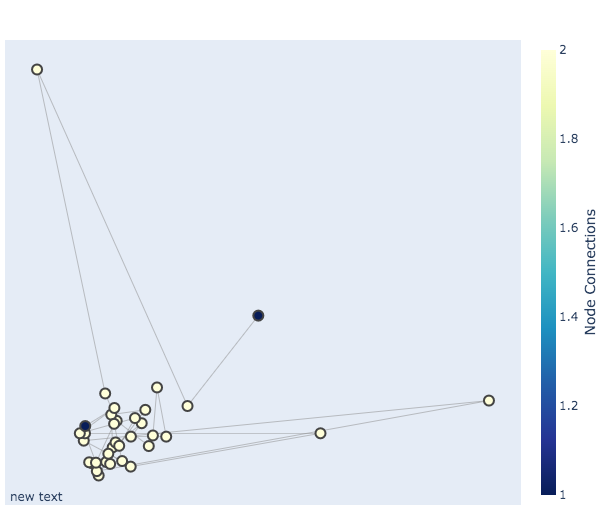}
         \includegraphics[width=\textwidth]{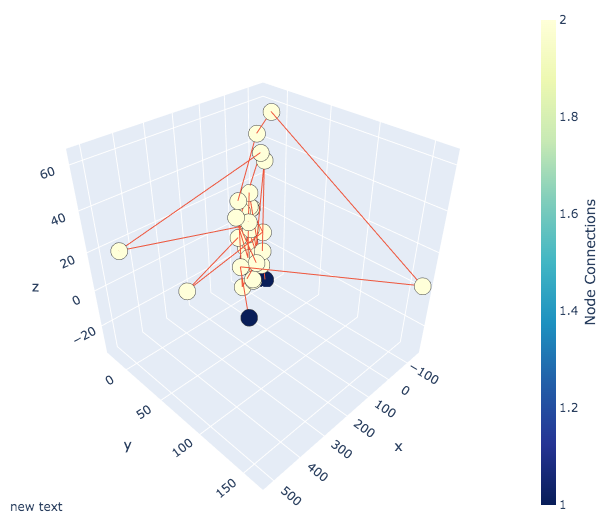}
         \includegraphics[width=\textwidth]{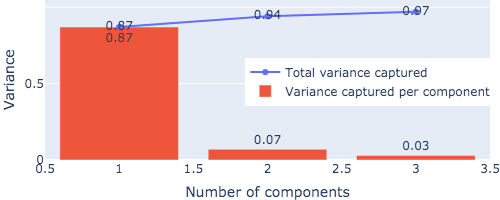}
         \caption{``State plans don't always provide full details of an individual's insurance coverage: but the federal government is generally able to provide coverage to people based on the insurance policies they have.''}
         \label{fig:74}
    \end{subfigure}
\end{figure}
\begin{figure}\ContinuedFloat
    \begin{subfigure}[b]{0.49\textwidth}
         \includegraphics[width=\textwidth]{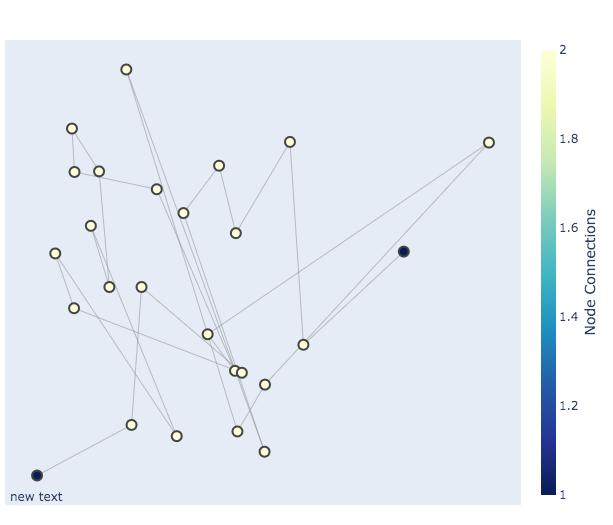}
         \includegraphics[width=\textwidth]{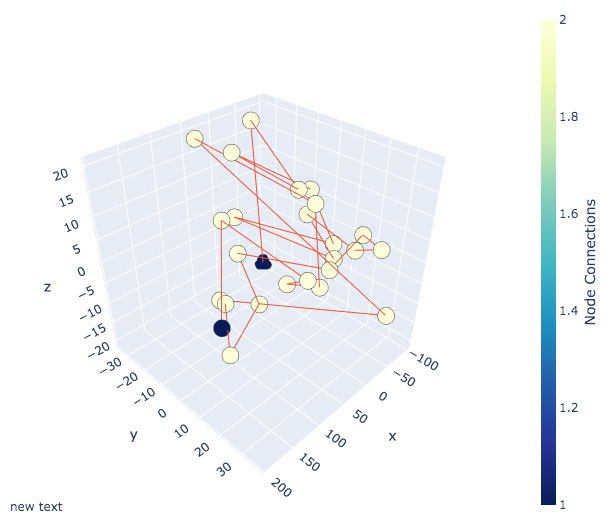}
         \includegraphics[width=\textwidth]{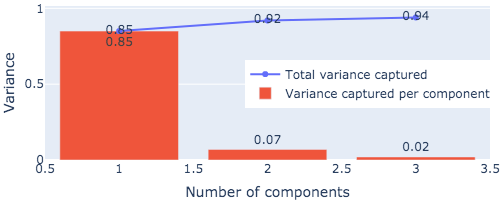}
         \caption{``In North Carolina: the state-led N.C. Health Centers are the only federally run health insurance exchanges run by citizens.''}
         \label{fig:75}
    \end{subfigure}
    \begin{subfigure}[b]{0.49\textwidth}
         \includegraphics[width=\textwidth]{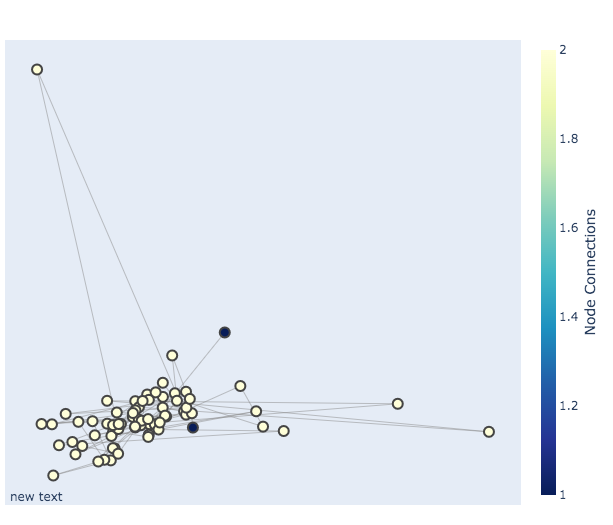}
         \includegraphics[width=\textwidth]{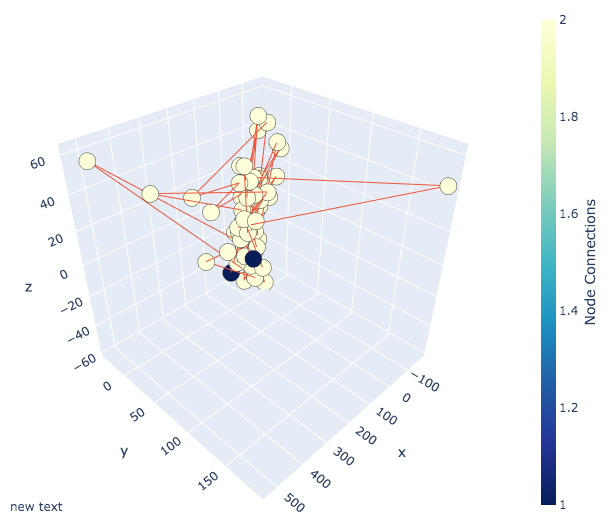}
         \includegraphics[width=\textwidth]{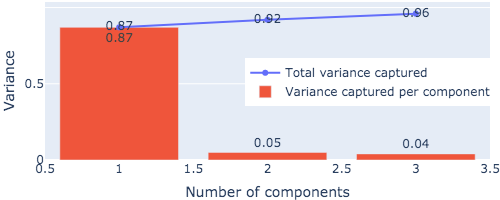}
         \caption{``""It should not be difficult for state officials to decide who is and who isn't covered under Obamacare:"" said John Meehan: N.C. State and federal spokeswoman. ""When you have a system like this it's not difficult for people to decide which insurance plans to buy: which coverage to buy and what type of coverage to buy.""''}
         \label{fig:76}
    \end{subfigure}
\end{figure}
\begin{figure}\ContinuedFloat
    \begin{subfigure}[b]{0.49\textwidth}
         \includegraphics[width=\textwidth]{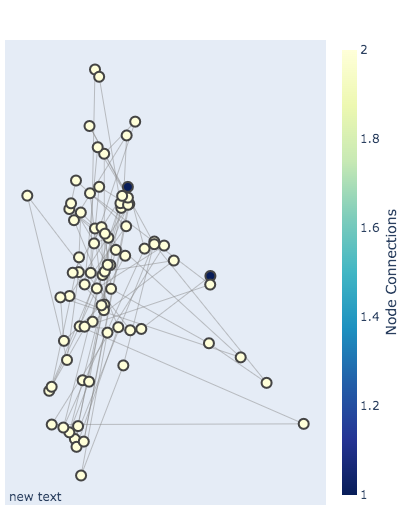}
         \includegraphics[width=\textwidth]{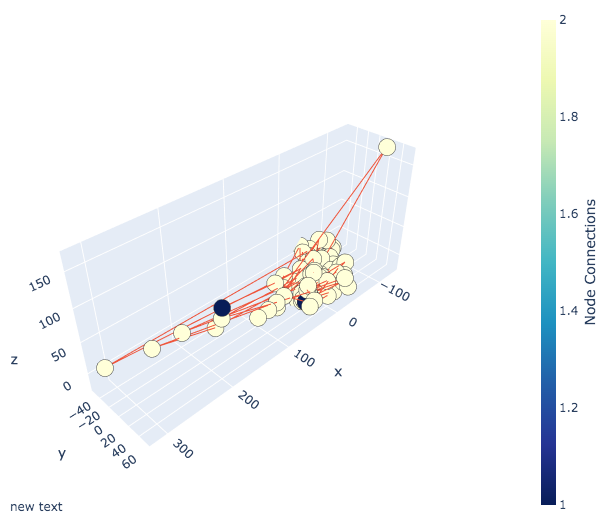}
         \includegraphics[width=\textwidth]{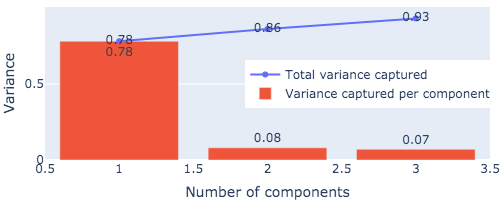}
         \caption{``The new law will allow states to offer federal-run coverage in a way that will not include people without insurance who already have a health plan. The McCrory administration plans to sign the legislation into law by the end of the year: while its secretary will act to create new laws to address specific insurance policies for state and local residents affected by the new legislation to help resolve any changes they make.''}
         \label{fig:77}
    \end{subfigure}
    \begin{subfigure}[b]{0.49\textwidth}
         \includegraphics[width=\textwidth]{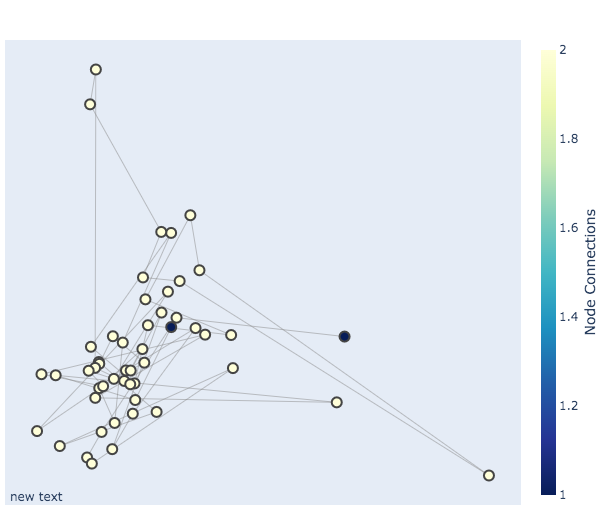}
         \includegraphics[width=\textwidth]{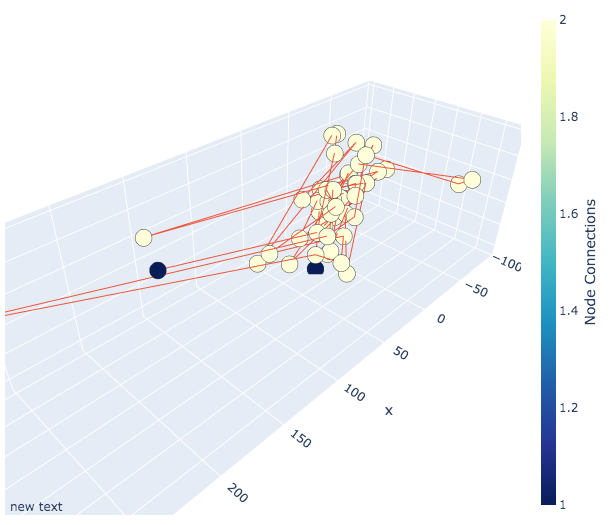}
         \includegraphics[width=\textwidth]{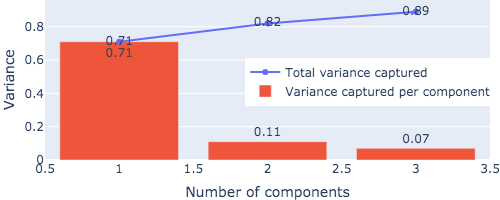}
         \caption{``Other state officials will also need to update their existing policies with the law to avoid confusion over what an individual can buy and what state-based coverage can include. N.C. Governor Pat McCrory said the state is in the process of developing"''}
         \label{fig:78}
    \end{subfigure}
\end{figure}
\begin{figure}
    \begin{subfigure}[b]{0.49\textwidth}
         \includegraphics[width=\textwidth]{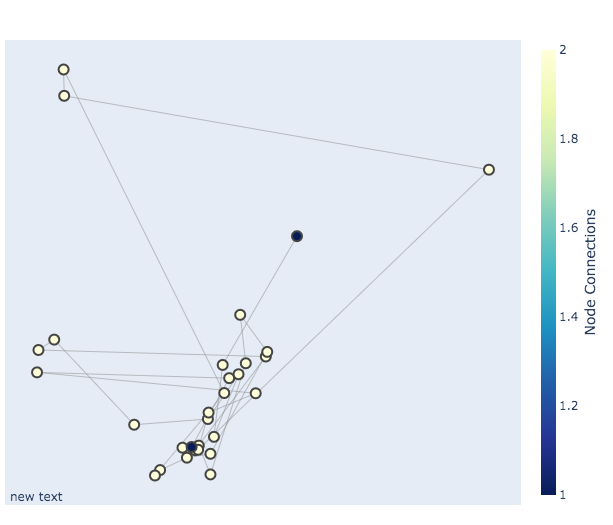}
         \includegraphics[width=\textwidth]{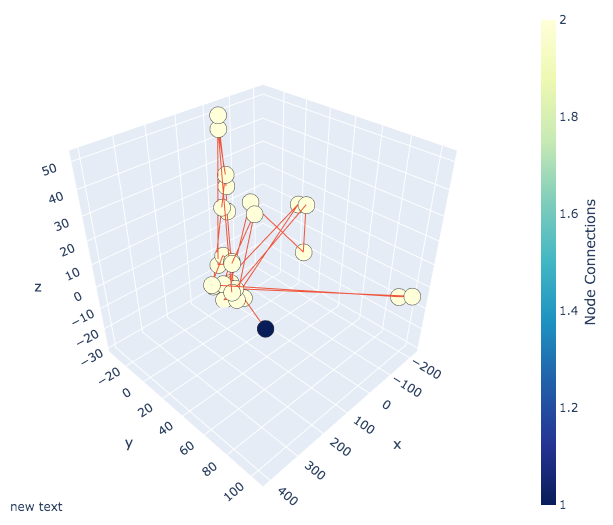}
         \includegraphics[width=\textwidth]{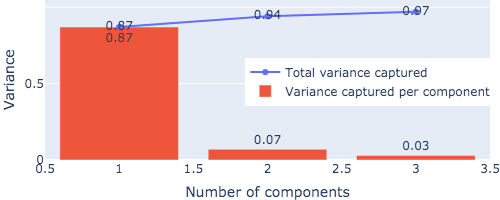}
         \caption{``16:"HBO said on Sept. 2 its television show ""Game of Thrones"" would premiere at 9 p.m. Thursday: March 11.''}
         \label{fig:79}
    \end{subfigure}
    \begin{subfigure}[b]{0.49\textwidth}
         \includegraphics[width=\textwidth]{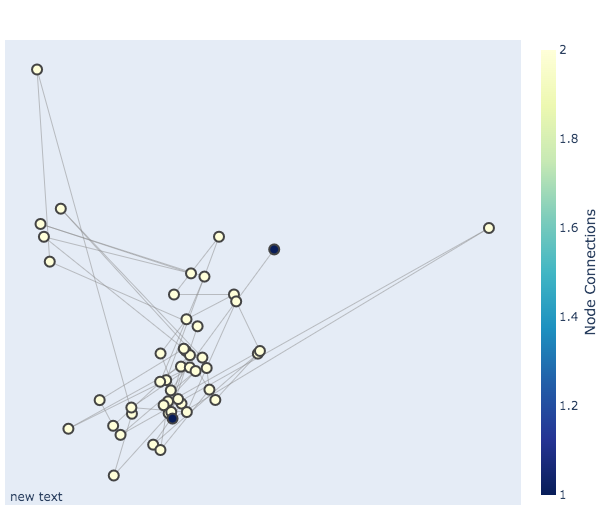}
         \includegraphics[width=\textwidth]{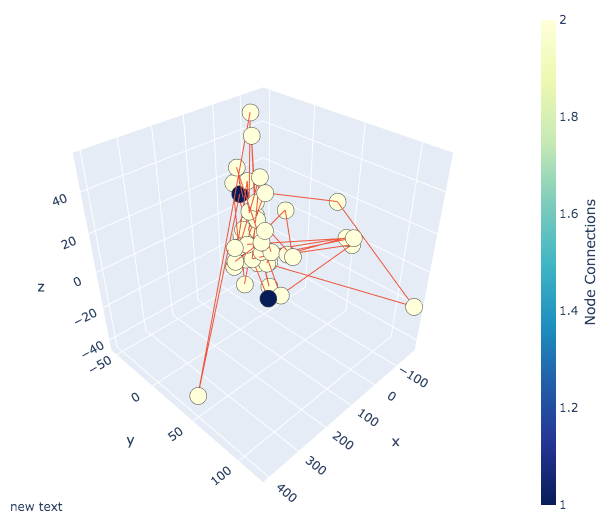}
         \includegraphics[width=\textwidth]{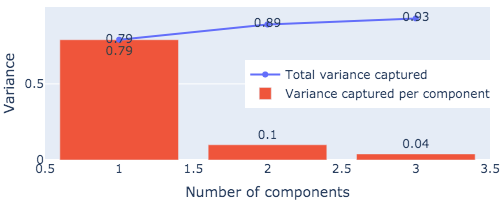}
         \caption{``The series: which centers on Cersei Lannister: was founded by producer Peter Dinklage in 2007 after Dinklage was appointed general in the George R. R. Martin-written ""Star Wars"" series.[1]''}
         \label{fig:80}
    \end{subfigure} 
\end{figure}
\begin{figure}\ContinuedFloat
    \begin{subfigure}[b]{0.49\textwidth}
         \includegraphics[width=\textwidth]{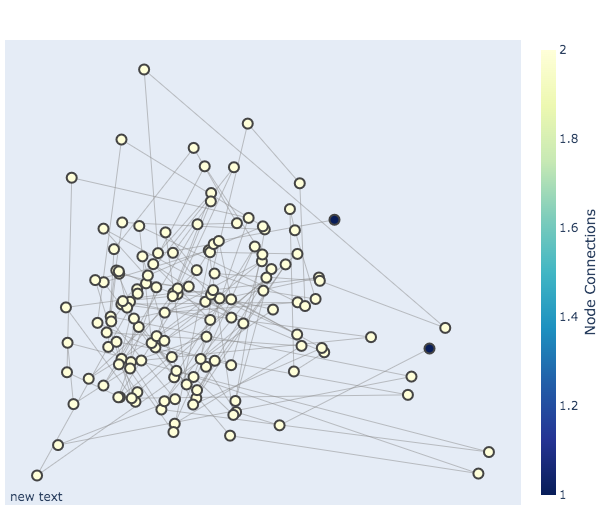}
         \includegraphics[width=\textwidth]{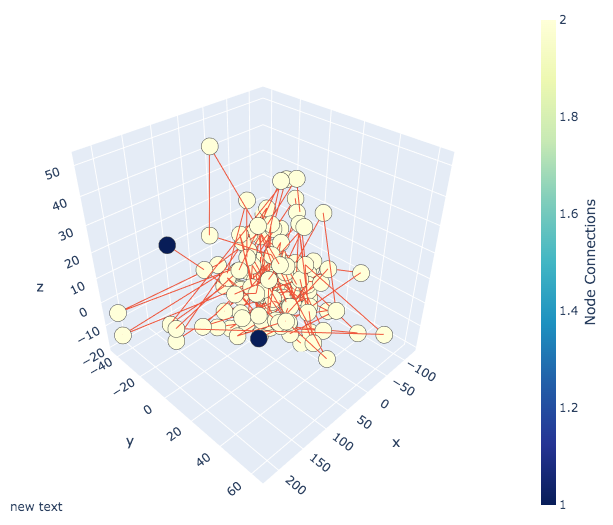}
         \includegraphics[width=\textwidth]{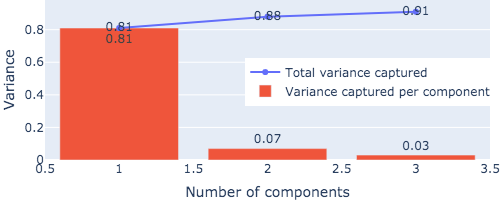}
         \caption{``""This is a truly remarkable thing to happen this week. But there is so much we love about the show. We know Jon Snow has been an integral part of the show's success for decades and we're proud that his legacy will continue to help the show grow and expand from now until it's a live reality show and HBO is making a lot of the fans want to get caught up in this next season:"" HBO president Michael Lombardo said in a statement. ""The show's success this week has given the fans a very unique opportunity to have a moment of their own watching a show at peak performance in its current form – and with every move we make: there's more to come.""''}
         \label{fig:81}
    \end{subfigure}
    \begin{subfigure}[b]{0.49\textwidth}
         \includegraphics[width=\textwidth]{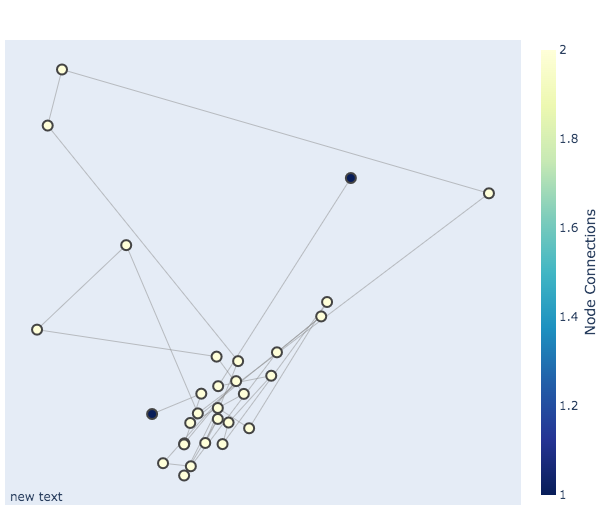}
         \includegraphics[width=\textwidth]{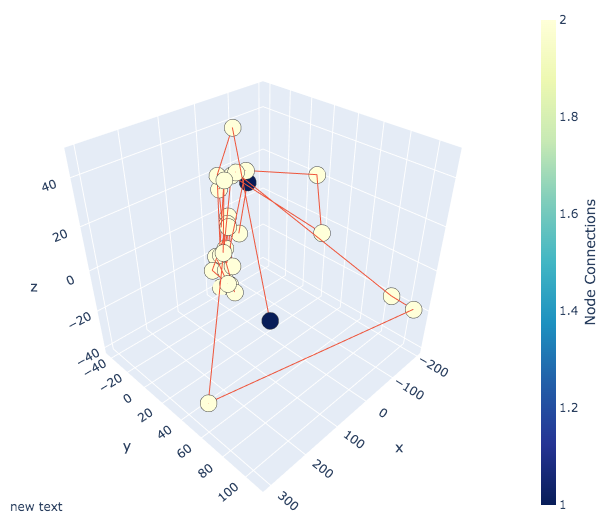}
         \includegraphics[width=\textwidth]{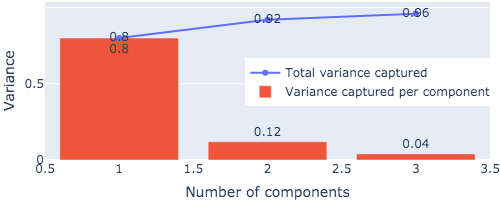}
         \caption{``81:It is also fitting that HBO would start at 6 a.m. Sept. 2 with a six-hour episode in its broadcast lineup.''}
         \label{fig:82}
    \end{subfigure}
\end{figure}
\begin{figure}\ContinuedFloat
    \begin{subfigure}[b]{0.49\textwidth}
         \includegraphics[width=\textwidth]{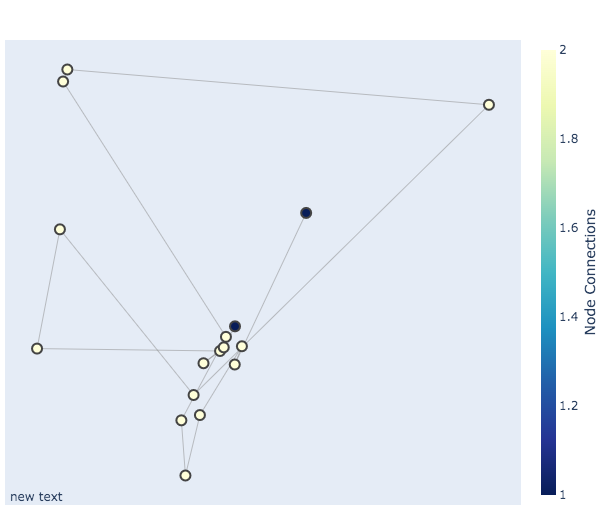}
         \includegraphics[width=\textwidth]{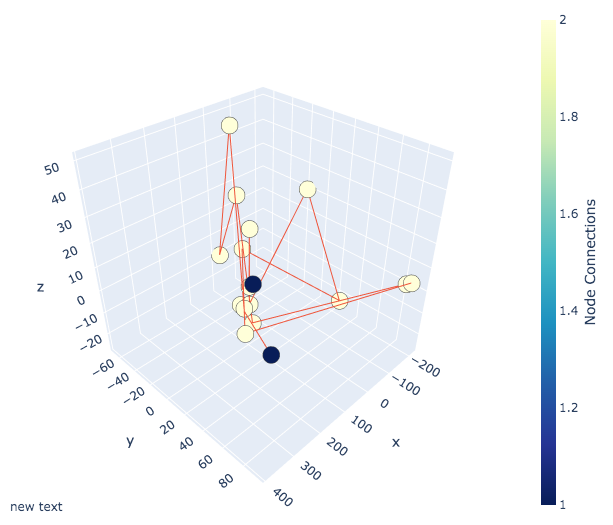}
         \includegraphics[width=\textwidth]{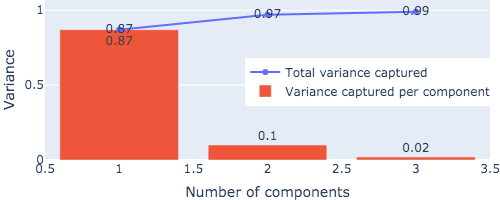}
         \caption{``Dunkirk will air at 7 p.m. Sept. 4 on HBO.''}
         \label{fig:83}
    \end{subfigure}
    \begin{subfigure}[b]{0.49\textwidth}
         \includegraphics[width=\textwidth]{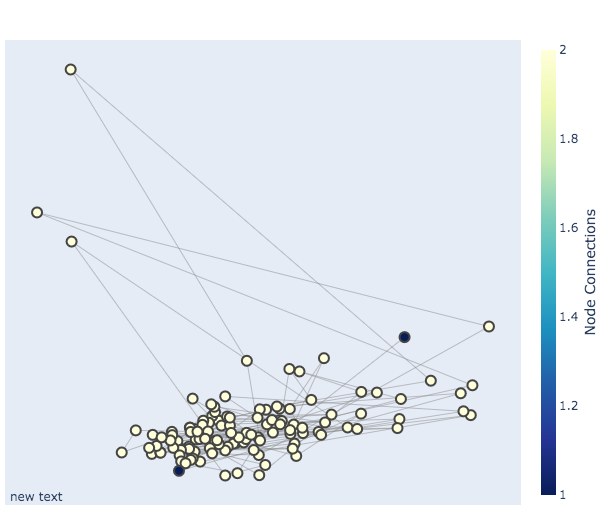}
         \includegraphics[width=\textwidth]{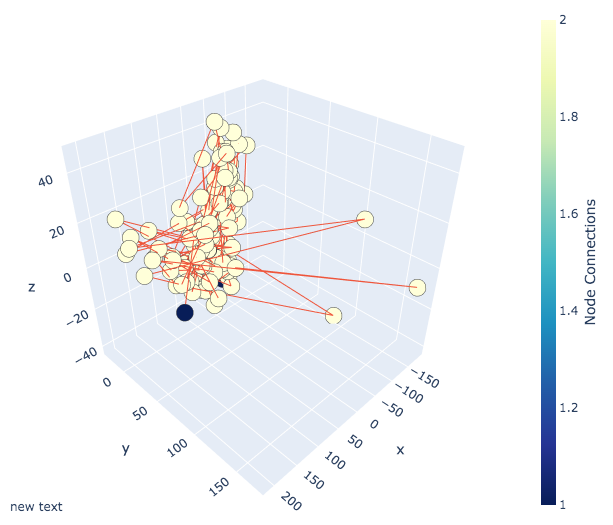}
         \includegraphics[width=\textwidth]{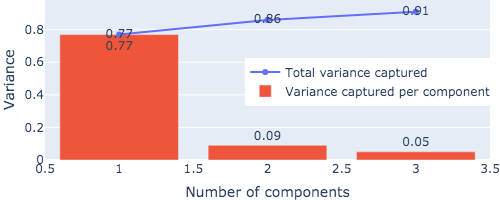}
         \caption{``""The Walking Dead"" began its 10 week run at a time when TV fans knew there was no ""Game of Thrones"" on the air: while some were concerned that it would end at the midpoint of the season. However: the show won the No. 1 spot and ""Game of Thrones"" will now fall one-third or four-fifths of the way down in the rankings: which will come as some fans were wondering about what was going on with the show's top 10 shows: such as ""The Walking Dead"": ""The Americans: Deadwood"": and ""Mad Men.""''}
         \label{fig:84}
    \end{subfigure}
\end{figure}
\begin{figure}\ContinuedFloat
    \begin{subfigure}[b]{0.49\textwidth}
         \includegraphics[width=\textwidth]{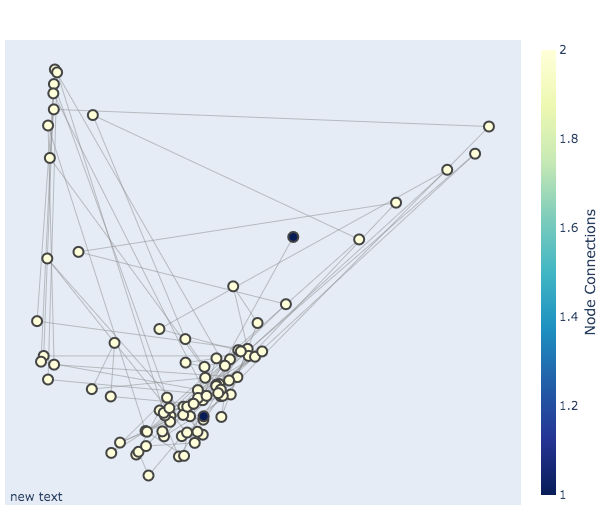}
         \includegraphics[width=\textwidth]{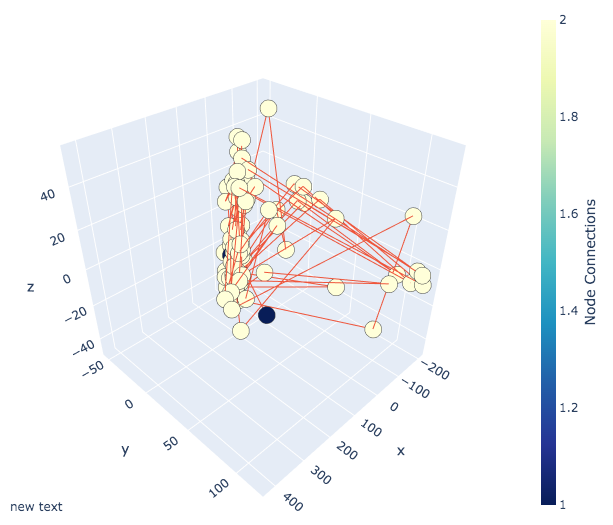}
         \includegraphics[width=\textwidth]{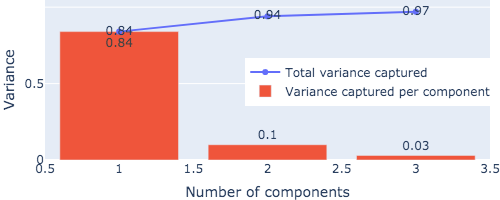}
         \caption{``Other top 10 shows included: ""Game of Thrones"" rose seven to No. 9 ""Game of Thrones"" fell to No. 8 (which is still an up-or-down decision based on ratings) ""House of Cards"" dropped four to No. 9 ""Game of Thrones"": ranked No. 6 on TBS and No. 10 in all three broadcast networks and ""The Walking Dead"" ranked No. 10 in all three networks.''}
         \label{fig:85}
    \end{subfigure}
    \begin{subfigure}[b]{0.49\textwidth}
         \includegraphics[width=\textwidth]{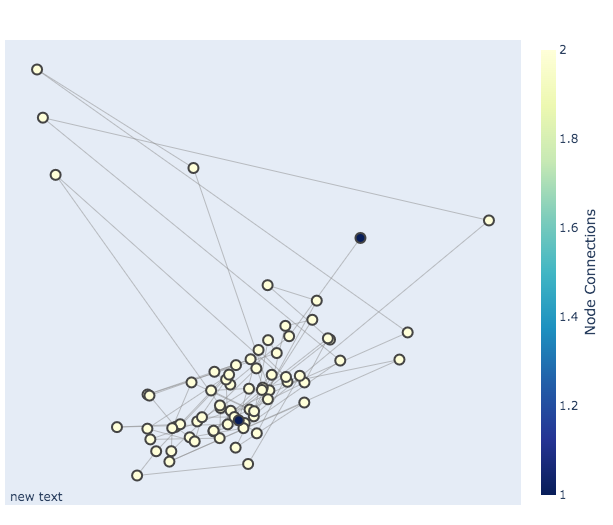}
         \includegraphics[width=\textwidth]{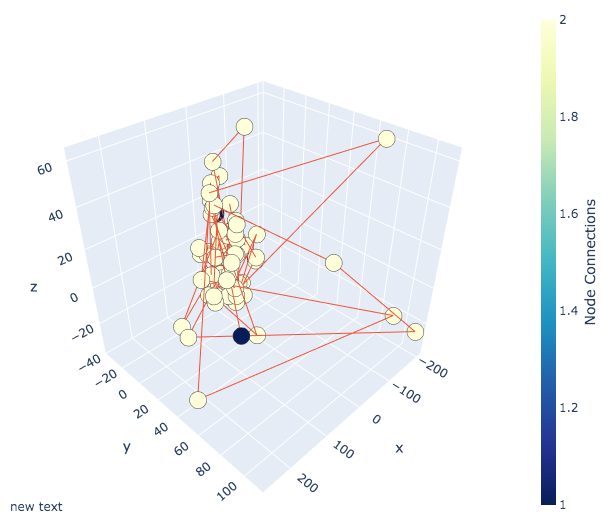}
         \includegraphics[width=\textwidth]{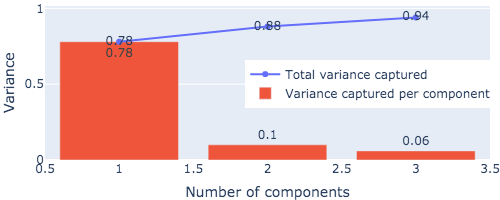}
         \caption{``This week: HBO premiered season two of ""Game of Thrones."" ""Game of Thrones"" has now been renewed for a sixth season: making ""Game of Thrones"" the third most-watched animated series in television history. Season 3 marked the 20th time in three seasons that two series had been renewed and the fourth most-watched animated show ever.''}
         \label{fig:86}
    \end{subfigure}
\end{figure}
\begin{figure}\ContinuedFloat
    \begin{subfigure}[b]{0.49\textwidth}
         \includegraphics[width=\textwidth]{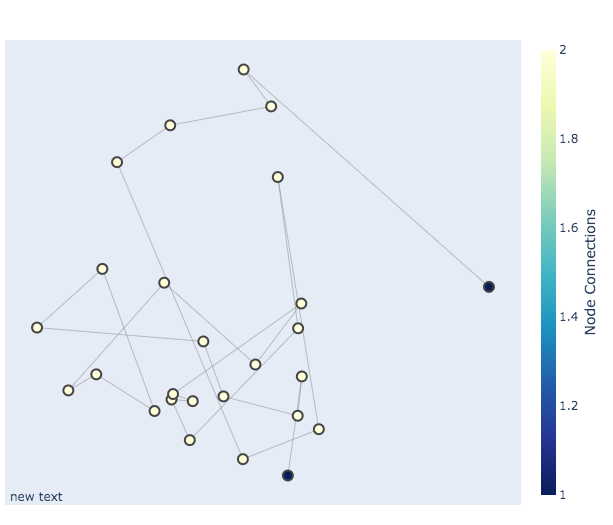}
         \includegraphics[width=\textwidth]{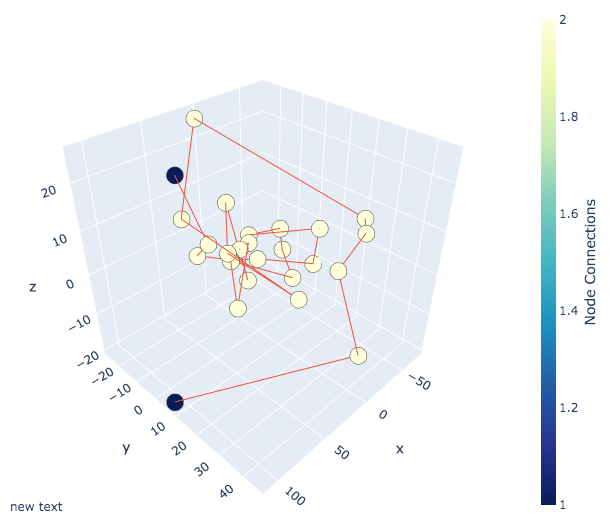}
         \includegraphics[width=\textwidth]{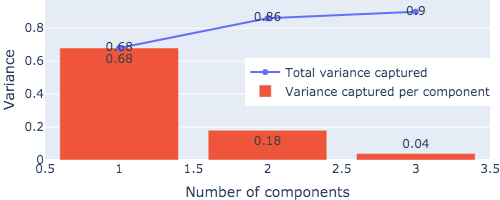}
         \caption{``For more on the upcoming ""Game of Thrones"" season five broadcast schedule: including new drama specials for season five: click here."''}
         \label{fig:87}
    \end{subfigure}
    \begin{subfigure}[b]{0.49\textwidth}
         \includegraphics[width=\textwidth]{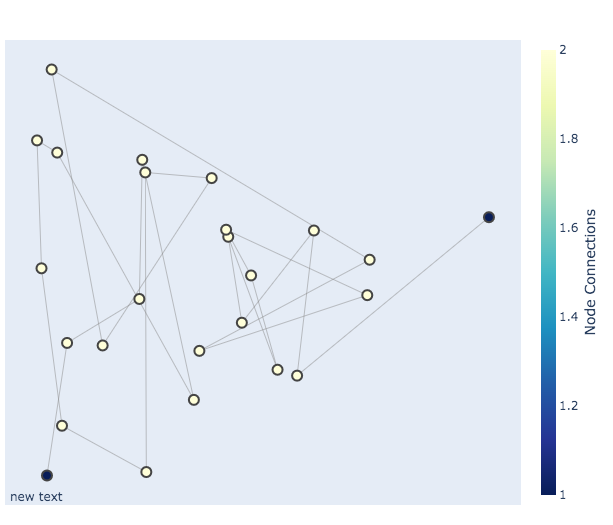}
         \includegraphics[width=\textwidth]{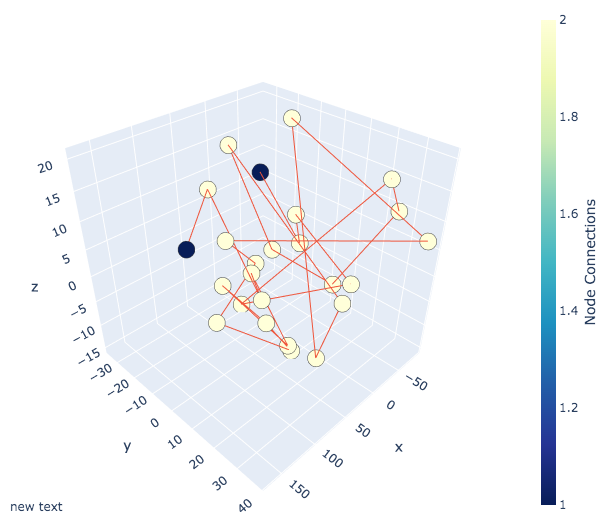}
         \includegraphics[width=\textwidth]{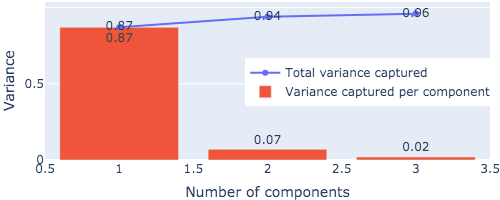}
         \caption{``Dangerous weather: cold weather: etc. don't happen during the year. And they never in the past were.''}
         \label{fig:88}
    \end{subfigure}
\end{figure}
\begin{figure}\ContinuedFloat
    \begin{subfigure}[b]{0.49\textwidth}
         \includegraphics[width=\textwidth]{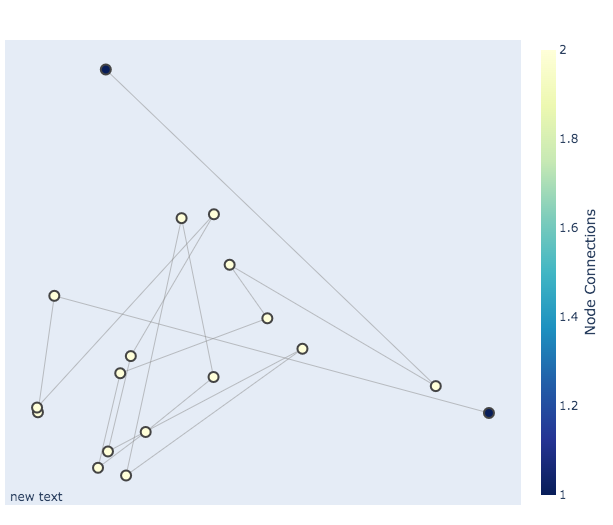}
         \includegraphics[width=\textwidth]{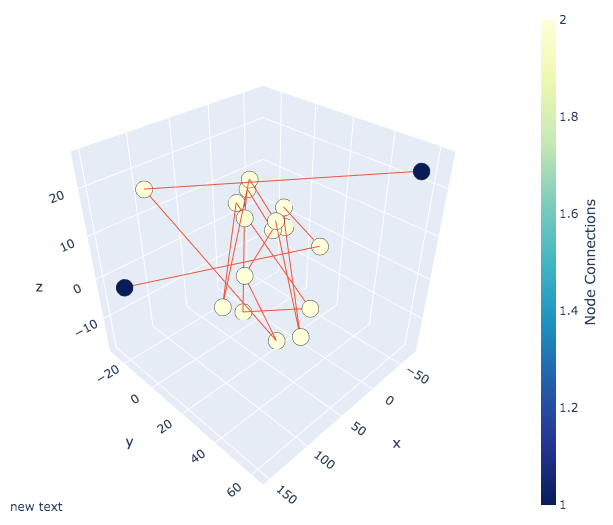}
         \includegraphics[width=\textwidth]{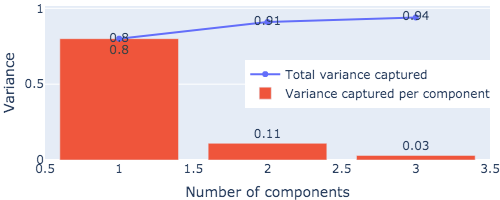}
         \caption{``""We're getting down to the business of weather forecasting for the coming three months."""''}
         \label{fig:89}
    \end{subfigure}
    \begin{subfigure}[b]{0.49\textwidth}
         \includegraphics[width=\textwidth]{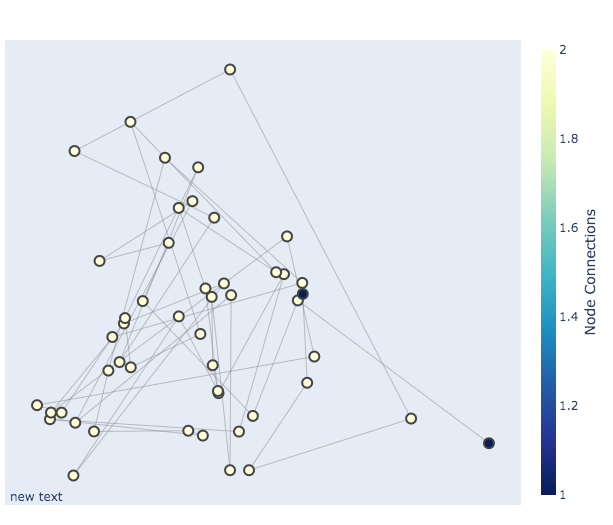}
         \includegraphics[width=\textwidth]{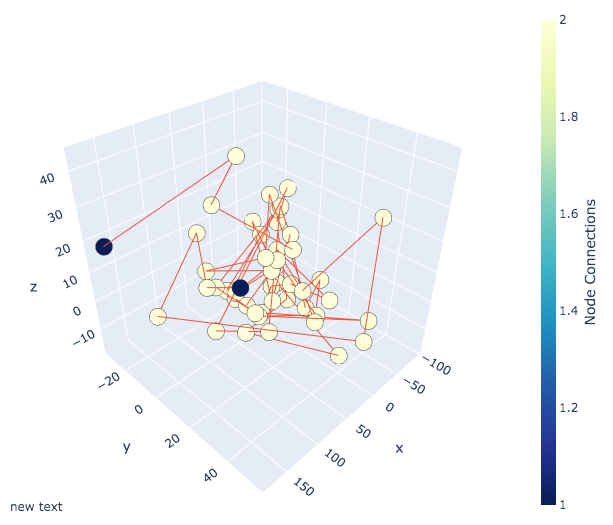}
         \includegraphics[width=\textwidth]{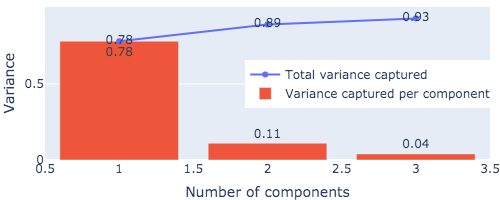}
         \caption{``Kanye West: So: as people look at me with this face like they see that I'm all set and I just want me to come out and sing with you every day: this is the look to me. (Photo from YouTube)''}
         \label{fig:90}
    \end{subfigure}
\end{figure}
\begin{figure}\ContinuedFloat
    \begin{subfigure}[b]{0.49\textwidth}
         \includegraphics[width=\textwidth]{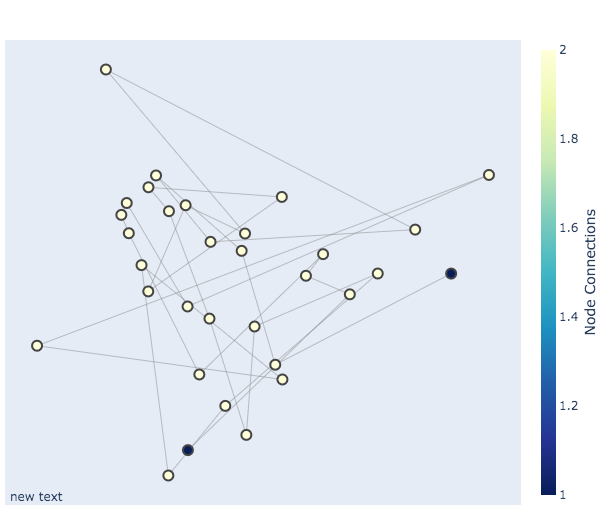}
         \includegraphics[width=\textwidth]{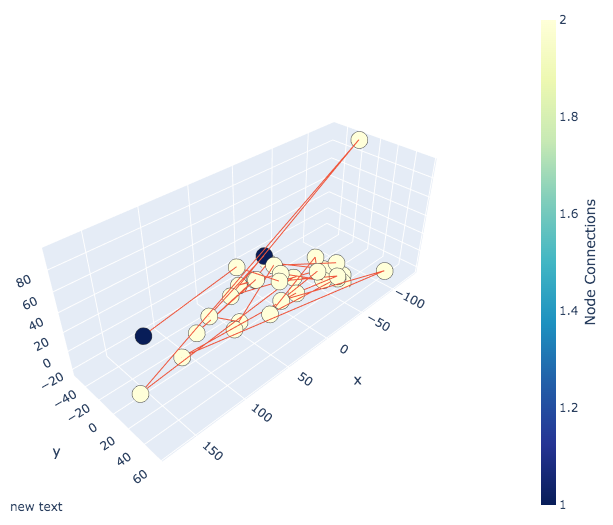}
         \includegraphics[width=\textwidth]{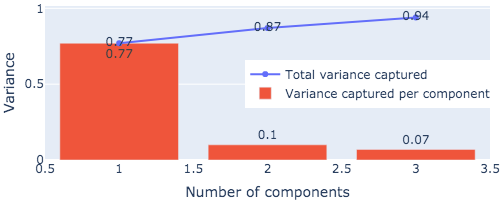}
         \caption{``Kanye West has no plans to give a performance during his upcoming album ""Living Colour:"" but ""Living Colour"" is his favorite song of all time.''}
         \label{fig:91}
    \end{subfigure}
    \begin{subfigure}[b]{0.49\textwidth}
         \includegraphics[width=\textwidth]{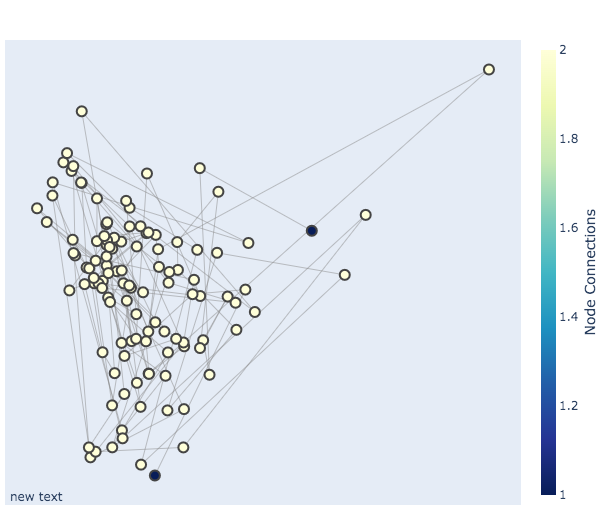}
         \includegraphics[width=\textwidth]{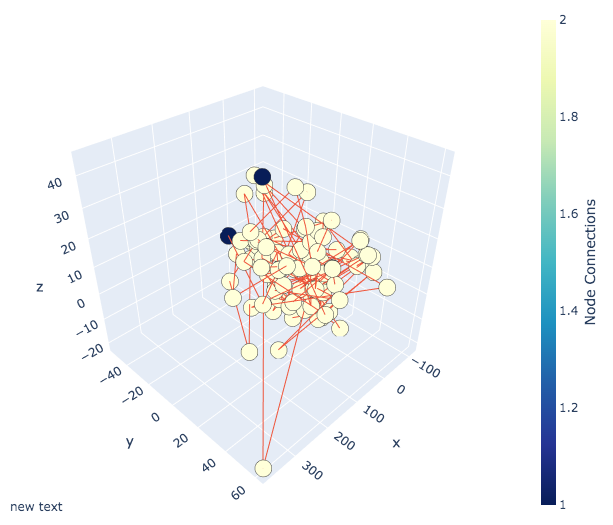}
         \includegraphics[width=\textwidth]{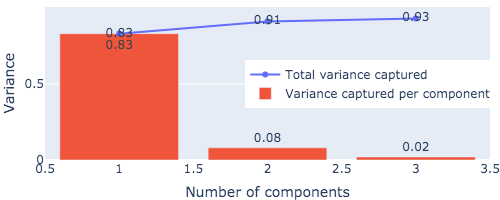}
         \caption{``After performing ""Living Colour"" alongside the rapper: West said: ""That just really got my mind racing because I've been playing with this and this is not an art I want to see play. I want to make something big out of: just as a child: playing with my mom. I think that's the best place I got this chance when I was just starting out. (Laughs) And she had her own dream. To play so much on my stage: and I hope she gets to hear that tonight. That's where all the fun stuff comes in.""''}
         \label{fig:92}
    \end{subfigure}
\end{figure}
\begin{figure}\ContinuedFloat
    \begin{subfigure}[b]{0.49\textwidth}
         \includegraphics[width=\textwidth]{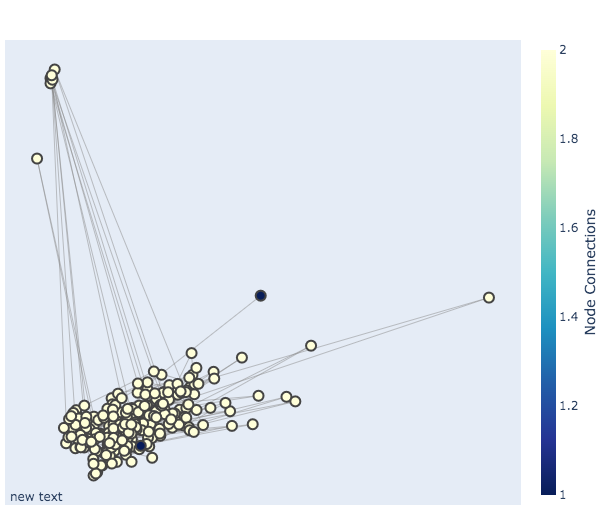}
         \includegraphics[width=\textwidth]{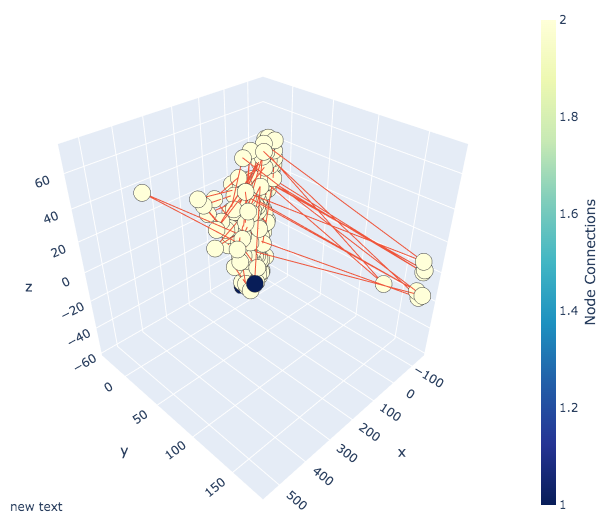}
         \includegraphics[width=\textwidth]{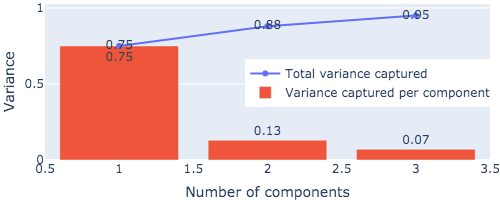}
         \caption{``West also confirmed that his upcoming solo album won't feature an ""A-list producer"" like he had earlier this year. ""But that doesn't mean I won't play with a producer. It probably won't come from any other musician or anybody. Just because I'm doing this not because I want to be famous doesn't mean I have to. I like doing other things. That's the best thing about it because you're working with an artist that you're working in partnership with: and when you've done something: you're making something special. It's not like you're singing: I've done it. It's a collaboration. That's really what was so fun. Not just because I do this. It's what we do. So: in the same way: it's just because you get out and do something that you've never done before on your own. I haven't done it before: you do it yourself: but because of all the success: I've got all the confidence you need right now. And it doesn't matter who the next step is. When you see this album: I haven't even had the time for the time to actually be there.""''}
         \label{fig:93}
    \end{subfigure}
    \begin{subfigure}[b]{0.49\textwidth}
         \includegraphics[width=\textwidth]{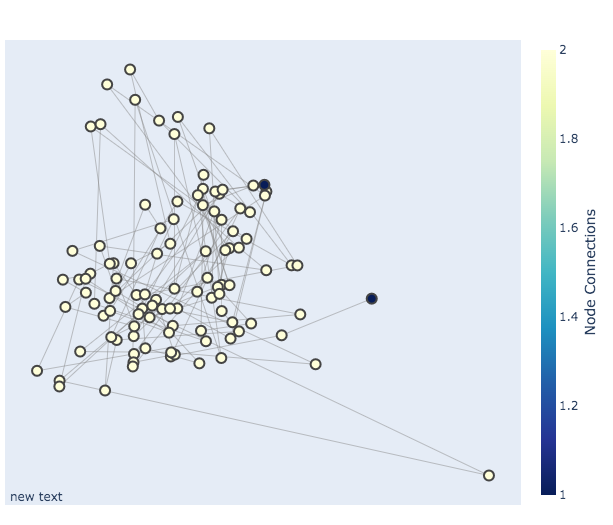}
         \includegraphics[width=\textwidth]{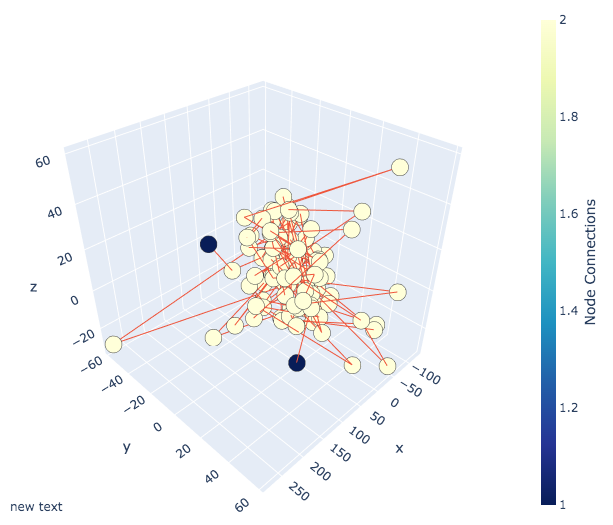}
         \includegraphics[width=\textwidth]{figs/sent_eval/sent94.png}
         \caption{``For now: West is still performing a video in honor of the pop superstar's birthday: at which point he'll be dancing with his former bandmates for a crowd of 3:000 people at the Hyatt Regency and Las Vegas music venue: PGA Miami. ""The video is coming out in three months. So: I'm already doing a lot for the next three months. And if I'm going to record it again: well it's good to be in my groove:"" he told Rolling Stone in March.''}
         \label{fig:94}
    \end{subfigure}
\end{figure}
\begin{figure}\ContinuedFloat
    \begin{subfigure}[b]{0.49\textwidth}
         \includegraphics[width=\textwidth]{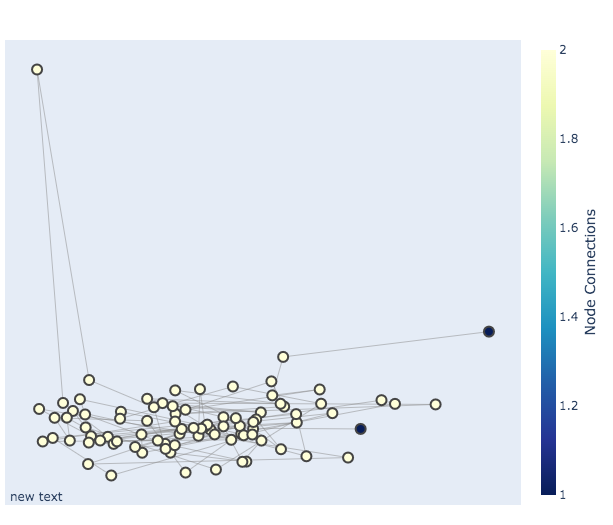}
         \includegraphics[width=\textwidth]{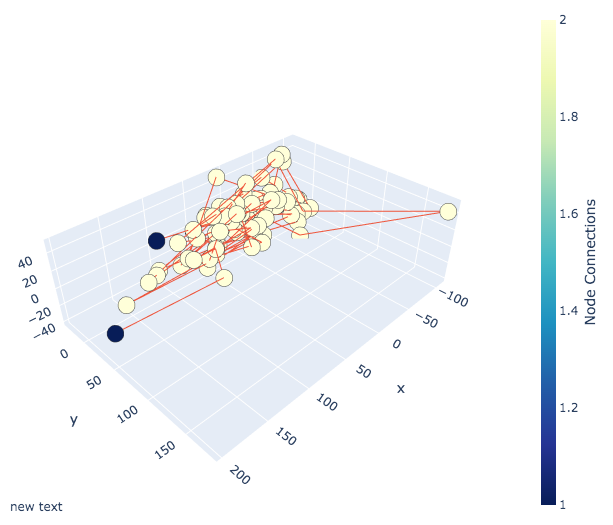}
         \includegraphics[width=\textwidth]{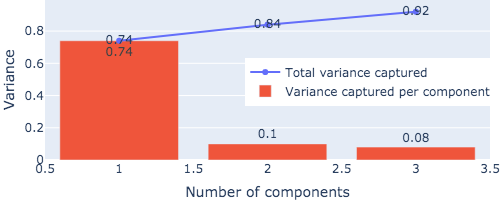}
         \caption{``West told the magazine in November that he plans on releasing a new album in 2016: ""I'll play one: then I'll have one and a half albums on my iPod at the same time. I don't know how long anything will go. I mean: it's hard work at two different different music websites for the time being. But we're gonna go for it and see what happens and be great to have it.""''}
         \label{fig:95}
    \end{subfigure}
    \begin{subfigure}[b]{0.49\textwidth}
         \includegraphics[width=\textwidth]{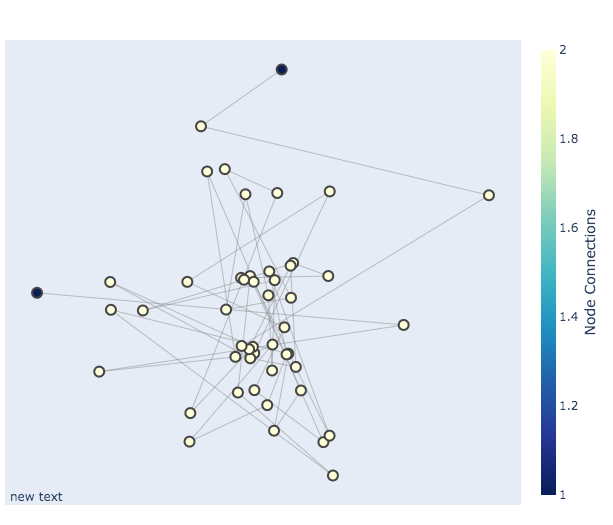}
         \includegraphics[width=\textwidth]{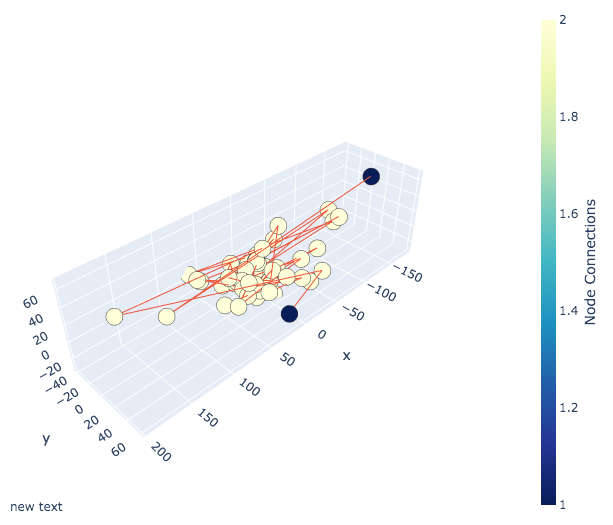}
         \includegraphics[width=\textwidth]{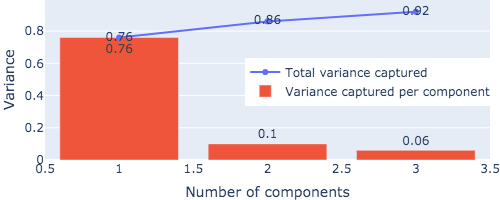}
         \caption{``More than anything: West is making an effort to keep working to keep himself active. When asked about how the band might do for a new album: West had these thoughts in the press conference announcing this week that his solo release on February 16.''}
         \label{fig:96}
    \end{subfigure}
\end{figure}
\begin{figure}\ContinuedFloat
    \begin{subfigure}[b]{0.49\textwidth}
         \includegraphics[width=\textwidth]{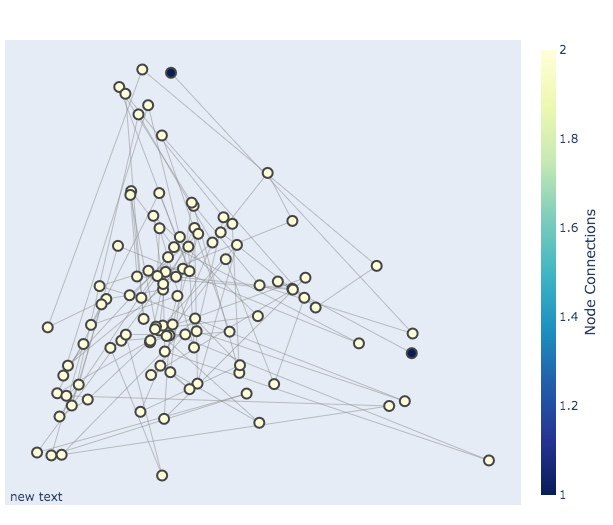}
         \includegraphics[width=\textwidth]{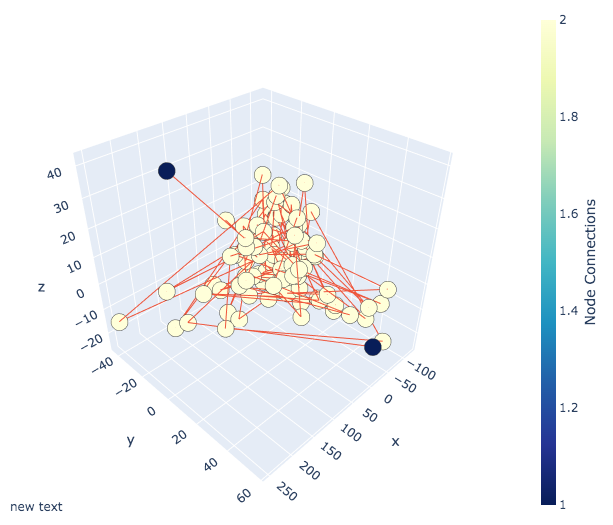}
         \includegraphics[width=\textwidth]{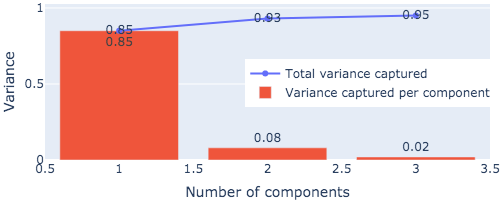}
         \caption{``Oh man: I got a lot of really good advice: he explained. "I'm working hard on it because the record label was a bit pissed at me. So: I guess we'll see. I'm gonna have to look at it for a couple more weeks. But: for now: I just want to do what I do. I'm just working really hard on it. It's not gonna be like a two songs: a lot of records: it's gonna be like two music websites.''}
         \label{fig:97}
    \end{subfigure}
    \begin{subfigure}[b]{0.49\textwidth}
         \includegraphics[width=\textwidth]{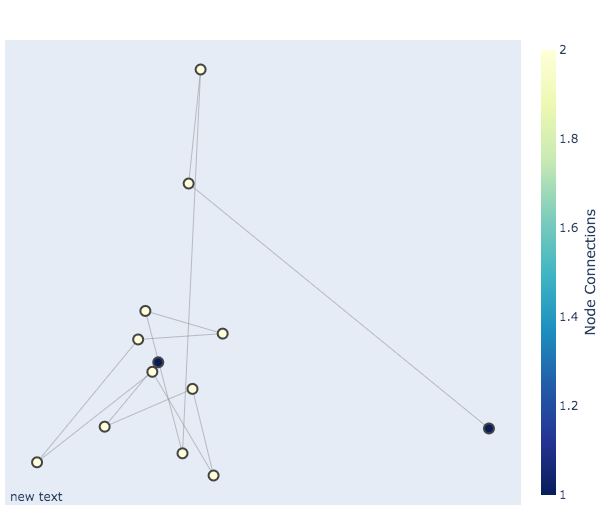}
         \includegraphics[width=\textwidth]{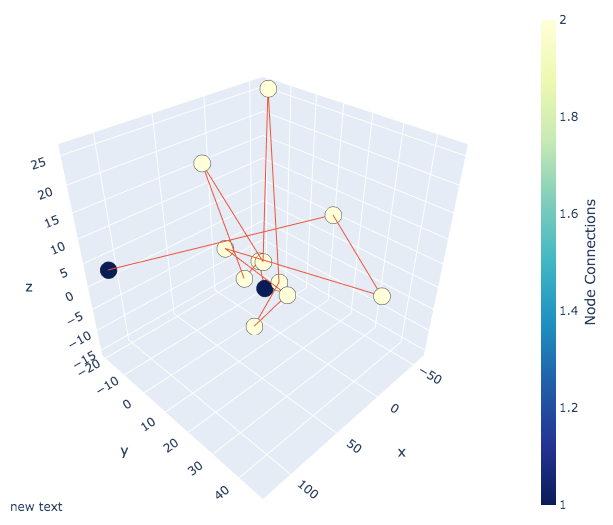}
         \includegraphics[width=\textwidth]{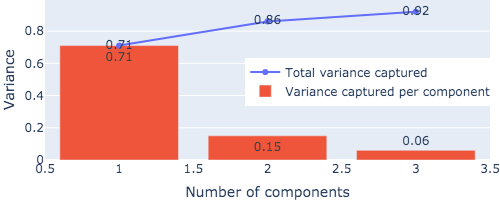}
         \caption{``23:(See "Stuff to Know About It")''}
         \label{fig:98}
    \end{subfigure}
\end{figure}
\begin{figure}\ContinuedFloat
    \begin{subfigure}[b]{0.49\textwidth}
         \includegraphics[width=\textwidth]{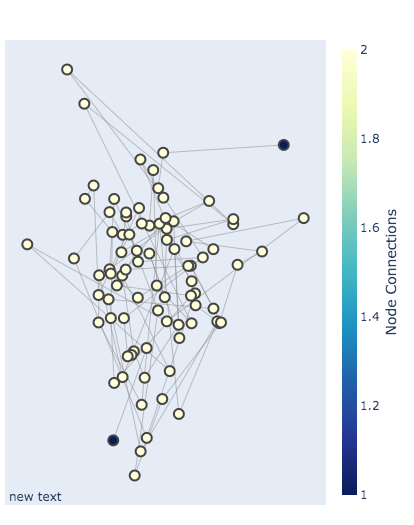}
         \includegraphics[width=\textwidth]{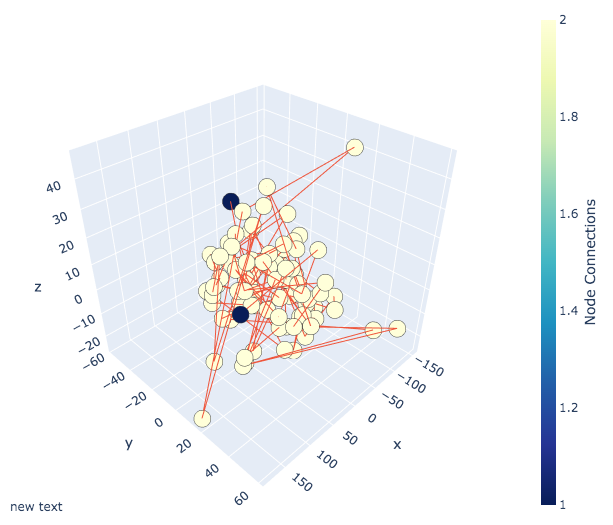}
         \includegraphics[width=\textwidth]{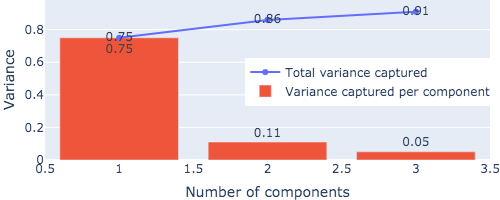}
         \caption{``The National Archives of Virginia has some good tidings for you if you decide to buy a copy—and it sounds like the government might just have it all figured out in the next few weeks. In a letter dated Monday: Richard C. Corman: assistant secretary for state for domestic affairs: has a new look at ""Stuff to Know About It:"" with some interesting details about the National Archives' digital services.''}
         \label{fig:99}
    \end{subfigure}
    \begin{subfigure}[b]{0.49\textwidth}
         \includegraphics[width=\textwidth]{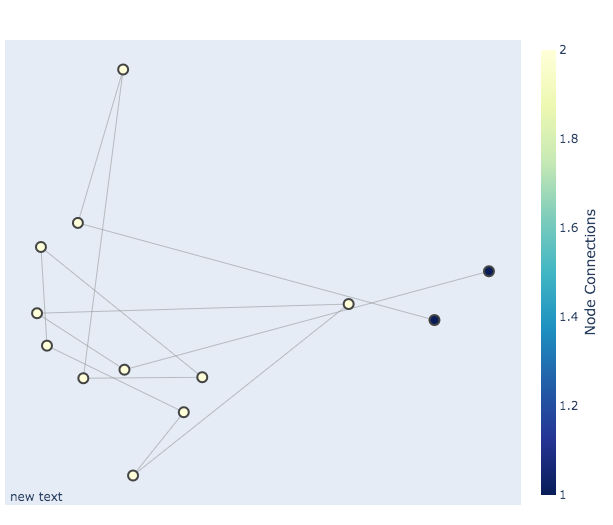}
         \includegraphics[width=\textwidth]{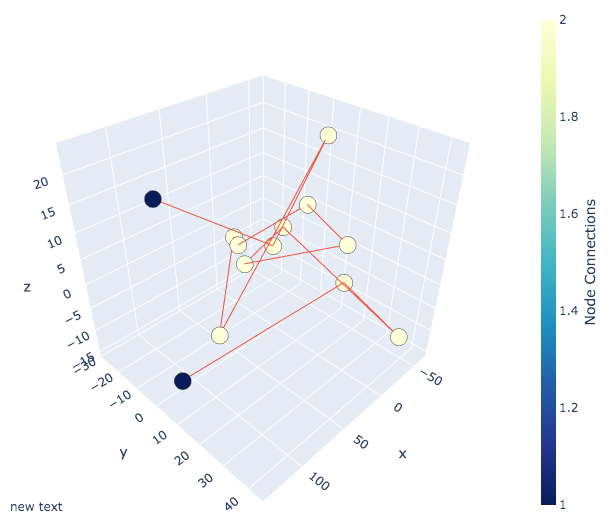}
         \includegraphics[width=\textwidth]{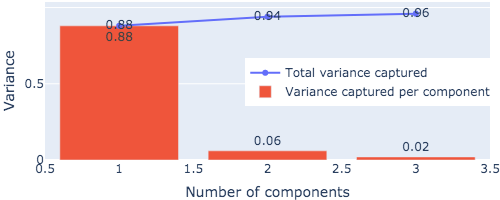}
         \caption{``Here's a rundown of some interesting notes from the press release:''}
         \label{fig:100}
    \end{subfigure}
\caption{Visualization of sentence vector as a directed graph in $2$-dimensional and $3$-dimensional spaces.}
\label{fig:graph}
\end{figure}

\end{document}